\documentclass[runningheads]{llncs}

\usepackage{eccv}

\usepackage{eccvabbrv}

\usepackage{tabularx}
\usepackage{graphicx}
\usepackage{booktabs}
\usepackage{tikz}
\usepackage{amssymb}
\usepackage{amsmath}

\usepackage{multirow}

\usepackage{longtable}
\usepackage{xcolor}
\usepackage{colortbl}
\usepackage{subcaption}

\usepackage{array} 
\usepackage{multirow}
\definecolor{headerblue}{RGB}{30, 80, 160}
\definecolor{rowgray}{RGB}{240, 242, 245}

\usepackage[accsupp]{axessibility}  %

\usepackage[pagebackref,breaklinks,colorlinks,citecolor=eccvblue]{hyperref}

\usepackage{orcidlink}

\newcommand{\method}{LoRA$^2$\xspace}

\begin{document}

\title{Not All Layers Are Created Equal: Adaptive LoRA Ranks for Personalized Image Generation}

\titlerunning{Adaptive LoRA Ranks for Personalized Image Generation}

\author{Donald Shenaj\inst{1}\orcidlink{0000-0002-6501-9437},
Federico Errica\inst{2}\orcidlink{0000-0001-5181-2904},
Antonio Carta\inst{1}\orcidlink{0000-0002-0003-2323}}

\authorrunning{D. Shenaj, F. Errica, A. Carta}

\institute{University of Pisa \and NEC Laboratories Europe}
\maketitle

\begin{center}
    \centering
    \includegraphics[width=0.9\linewidth]{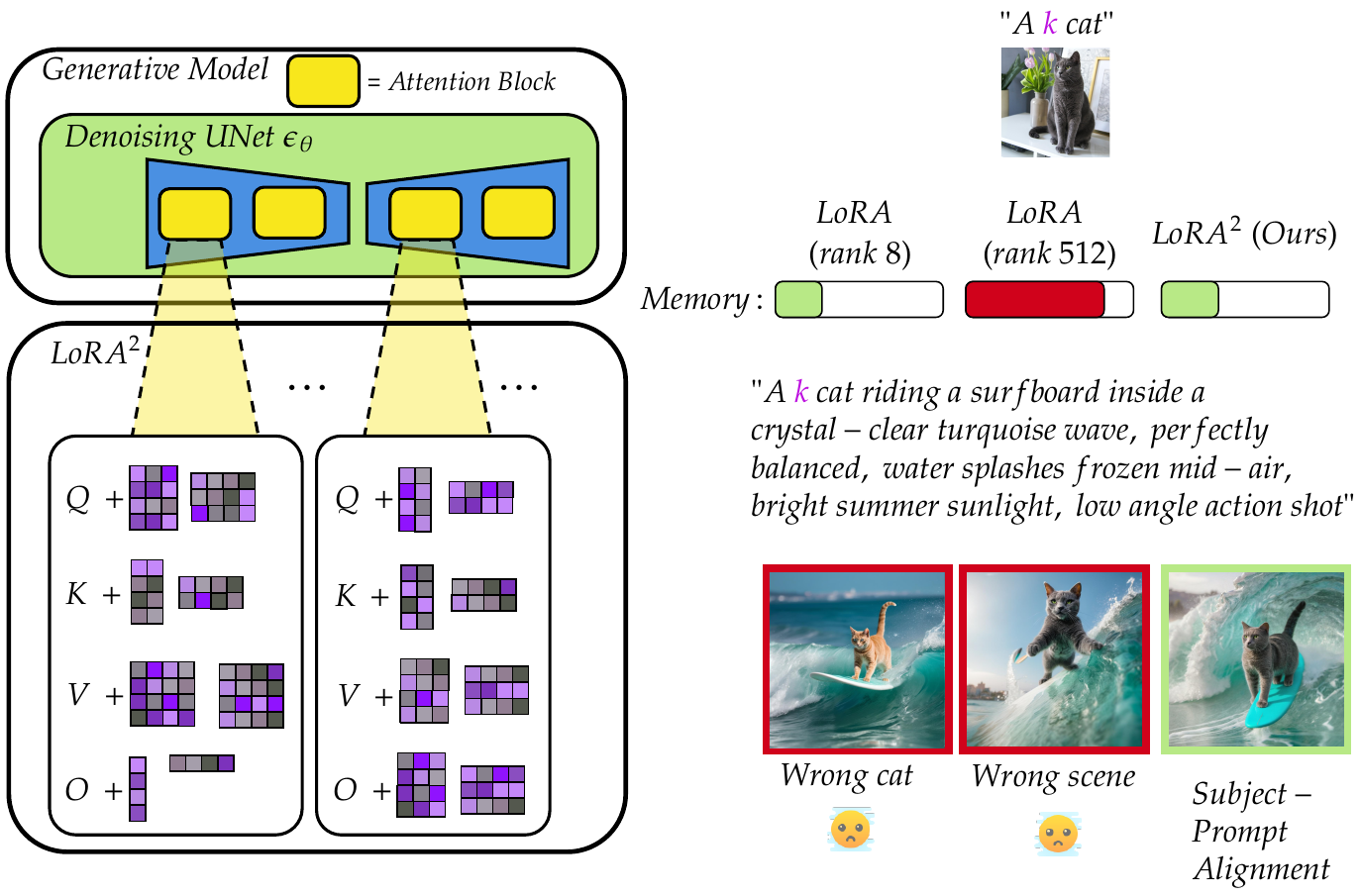}
    \captionof{figure}{(Left) In \method{}, each LoRA component is rank-adaptive and task-dependent. (Right) \method{} achieves better subject-prompt alignment and memory consumption.}
\label{fig:teaser}
\end{center}

\begin{abstract}
Low Rank Adaptation (LoRA) is the de facto fine-tuning strategy to generate personalized images from pre-trained diffusion models. Choosing a good rank is extremely critical, since it trades off performance and memory consumption, but today the decision is often left to the community's consensus, regardless of the personalized subject's complexity. The reason is evident: the cost of selecting a good rank for each LoRA component is combinatorial, so we opt for practical shortcuts such as fixing the same rank for all components.
In this paper, we take a first step to overcome this challenge. Inspired by variational methods that learn an adaptive width of neural networks, we let the ranks of each layer freely adapt during fine-tuning on a subject. We achieve it by imposing an ordering of importance on the rank's positions, effectively encouraging the creation of higher ranks when strictly needed. Qualitatively and quantitatively, our approach, \method, achieves a competitive trade-off between DINO, CLIP-I, and CLIP-T across 29 subjects while requiring much less memory and lower rank than high rank LoRA versions. Code: \url{https://github.com/donaldssh/NotAllLayersAreCreatedEqual}.
\end{abstract}

\section{Introduction}
\label{sec:introduction}

Personalized diffusion models \cite{ruiz2023dreambooth, gal2022image, kumari2023multi} are a popular application where a pretrained text-to-image generative model is finetuned to generate new subjects or styles with a few sample images.
Online repositories such as Civitai \cite{civitai} and HuggingFace \cite{huggingface} host thousands of personalized diffusion models trained to capture specific subjects or artistic styles. Most of these models are obtained via Low-Rank Adaptation (LoRA)\cite{hu2022lora}, a parameter-efficient fine-tuning technique that injects low-rank updates into pretrained diffusion backbones.

A successful personalized model should satisfy \textbf{three key objectives:} \textbf{(1)} high-quality generation of the desired subject or style, \textbf{(2)} strong fidelity to the textual prompt, and \textbf{(3)} low memory footprint (\cref{fig:teaser}).

In practice, these objectives are tightly coupled with the choice of the LoRA rank. Current practice adopts a simple heuristic: a fixed rank is selected and used uniformly across all LoRA components and all subjects. While this strategy provides reasonable average performance, it \textbf{severely restricts flexibility} for various reasons. First, the optimal rank depends on the subject; complex subjects may require higher ranks to capture fine-grained appearance variations, whereas simpler subjects can be modeled with substantially lower ranks. Second, the optimal ranks vary across layers and architectures; many layers may need small ranks while others would require higher capacities. A globally fixed rank prevents layer-wise specialization, resulting in a higher memory footprint without any performance benefits (\cref{fig:teaser}).

The reason for choosing such heuristic, regardless of the subject and layer, is the combinatorial explosion of a full layer-wise and subject-specific hyperparameter search. In this paper, we propose \method, a novel approach that adapts LoRA ranks during fine-tuning. Inspired by adaptive-width methods based on variational inference, \method encourages an ordering over the rank indices of each LoRA component, effectively pushing it to achieve the minimal effective rank necessary for the task. This structured parameterization enables high image quality with reduced memory usage compared to a global LoRA rank.

Experimental results demonstrate that \method achieves a better trade-off between subject fidelity, text alignment, and memory consumption compared to fixed-rank LoRA baselines. Across 29 personalized subjects and two diffusion backbones (SDXL and KOALA), our method improves this trade-off over fixed-rank configurations with similar or higher memory usage. For example, models with rank 512 achieve strong subject fidelity but require up to 2.8 GB of parameters, whereas \method attains comparable scores with only 0.40 GB, illustrating the efficiency of adaptive learning of the LoRA ranks.

Our analysis also reveals that optimal ranks vary significantly across subjects and layers, confirming that a globally fixed rank is inherently suboptimal. The adaptive behavior enables the model to allocate capacity where it is most beneficial while minimizing unnecessary parameters. Finally, ablation studies further show that regularizing both the rank parameters and LoRA weights allows \method to produce compact models with minimal degradation in generation quality.

\section{Related Work}
\label{sec:related-work}

\subsection{Personalization in Diffusion Models}

Diffusion models~\cite{ho2020denoising, rombach2022high, song2021denoising} have achieved remarkable success in image synthesis due to their strong representation capacity and compatibility with multi-modal conditioning, particularly text guidance. Their ability to generate high-fidelity and diverse images has made them the dominant paradigm for text-to-image generation. 

Beyond generic generation, recent advances have improved the adaptability of diffusion models through personalization techniques that tailor a pretrained backbone to specific subjects or styles while preserving creative flexibility. Methods such as DreamBooth~\cite{ruiz2023dreambooth}, Textual Inversion~\cite{gal2022image}, and StyleDrop~\cite{styledrop} adapt a base model using a small set of reference images, allowing it to generate new renditions of a particular object, person, or artistic style across diverse contexts.

More recently, Low-Rank Adaptation (LoRA)~\cite{hu2022lora} has emerged as a parameter-efficient alternative for personalization. Instead of fully fine-tuning model weights, LoRA introduces low-rank update matrices that significantly reduce the number of trainable parameters while maintaining generation quality. This design enables efficient training, lightweight storage, and modular deployment, allowing users to maintain separate personalization modules for individual subjects. The compact size of LoRA adapters further facilitates sharing and reuse through public model repositories, making it a widely adopted approach for subject-driven conditioning in diffusion models.

\subsection{Adaptive Architectures}
The term adaptive architectures refers to all those methods that dynamically modify the computational graph of a machine learning model. Early works in this space are constructive approaches that progressively increase a model's capacity, for instance cascade correlation \cite{fahlman_cascade_1989}. Firefly network descent \cite{wu_firefly_2020} relies on an auxiliary objective function to expand both width and depth at fixed intervals. Other methods grow networks by either duplicating or splitting units in a continual learning setting \cite{yoon_lifelong_2018}, or by periodically creating identical offsprings of neurons \cite{wu_splitting_2019}. More recently, \cite{mitchell_self_2023} proposed natural gradient–based heuristics to grow or shrink layers in MLPs and CNNs.

Contrary to growing methods, pruning \cite{blalock_state_2020} and distillation \cite{hinton_distilling_2015} aim to reduce network size, typically trading off performance for efficiency. Pruning methods remove connections \cite{mishra_accelerating_2021} or entire neurons \cite{valerio_dynamic_2022,dufort_maxwell_2024}, including dynamic approaches that apply hard or soft masks during training \cite{guo_dynamic_2016,he2018soft}. Distillation instead transfers knowledge from a larger model to a smaller one \cite{gou_knowledge_2021}.

Adaptive Width Neural Networks (AWNs) \cite{errica2026adaptive} take a different and simpler perspective by learning layer width directly through gradient descent within a single training loop. Instead of relying on explicit growth rules or splitting heuristics, AWNs introduce a continuous, monotonically decreasing importance distribution over neurons, allowing the model to smoothly \textit{expand or contract} its effective width during optimization. This formulation enables structured truncation and dynamic capacity adaptation without separate architectural interventions.

\subsection{Adaptive LoRA}
The literature on learning adaptive LoRA ranks tends to be more developed in the NLP domain. AdaLoRA \cite{zhang2023adaptive} computes an importance score based on the gradients and adds a soft orthogonality constraint. DoRA \cite{liu2024dora} improves the importance measure of AdaLoRA by making it more robust to noise and sparse gradients at convergence. ARD-LoRA \cite{shinwari2025ard} introduces a scaling factor that controls the rank and it is learned by optimizing a meta-objective. To the best of our knowledge, the effectiveness of adaptive LoRA has not been validated for personalized diffusion models, possibly because these techniques do not trivially transfer to computer vision models.
    
Empirical findings in the literature show benefits in adapting the rank of specific components, often found via an extensive manual search. \cite{bidermanLoRALearnsLess2024} shows that LoRA has less adaptation and less forgetting in LLM post-training. MLPs drive most of the performance of LoRAs, while attention layers can be excluded. \cite{li2024faster} finds that in during finetuning, the encoder features stay relatively constant, whereas the decoder features exhibit substantial variations across different time-steps. B-LoRA\cite{frenkel2024implicit} showed that certain blocks in the SDXL UNet are more responsible for content, and some are more responsible for style. The same approach has been used by UnZipLoRA~\cite{liu2025unziplora} to achieve subject-style separation. Overall, these results motivate our exploration of adaptive rank methods.

\section{Method}
\label{sec:method}
The idea behind our approach is to impose, for each LoRA, an adaptive ordering of importance across the rank dimension of LoRA weight matrices. Such orderings, learned via backpropagation as any other parameter, are used to determine the adaptive rank of each LoRA. Before introducing our method, however, we provide a refresher on LoRA and the variational framework for adaptive width neural networks of \cite{errica2026adaptive}, which we frame to our needs.

\subsection{LoRA Refresher}
Low Rank Adaptation (LoRA)\cite{hu2022lora} is a Parameter-Efficient Fine-Tuning (PEFT) technique designed to adapt large pre-trained models, including diffusion models, without the need to update all model parameters. This is achieved by introducing low-rank weights alongside those of a frozen model's component $\ell$. Specifically, given a frozen weight matrix $W^*_\ell \in \mathbb{R}^{m \times n}$, LoRA updates only a residual weight $\Delta W_\ell \in \mathbb{R}^{m \times n}$, which is computed as two low learnable rank matrices $B_\ell \in \mathbb{R}^{m \times r}$ and 
$A_\ell \in \mathbb{R}^{r \times n}$, with rank $r \ll \min(m,n)$. The choice of the rank $r$ naturally induces a trade-off between flexibility and efficiency, and in the literature it is typically set to the same value for all the model's components. For each component $\ell$, the final adapted weights can be represented as:
\begin{equation}
W'_\ell = W^*_\ell + \Delta W_\ell = W_\ell + B_\ell A_\ell.
\label{eq:lora}
\end{equation}

\subsection{Adaptive Rank Variational Framework}
Given a dataset of $N$ \textit{i.i.d.}\ samples, with generic $i$-th input $x_i$ and output $y_i$, a typical learning objective is maximizing the log-likelihood of the data
\begin{align}
    \log p(Y|X) = \log \prod_{i=1}^N p(y_i| x_i)=\sum_{i=1}^N \log p(y_i| x_i).
    \label{eq:objective}
\end{align}
where $p(y_i|x_i)$ is a probabilistic model, properly defined for each use case.

To formalize learning of a \textit{possibly infinite} rank for each LoRA component $\ell \in [1,L]$ of our image-generation model, we first consider a continuous random variable $\lambda_\ell$ that controls the \textit{finite} choice of the rank for component $\ell$, in a way that we will describe later. In addition, we introduce an infinite set of random variable $\boldsymbol{\theta}_{\ell r}, r \in [1,\infty]$, where $r$ can be thought as a ``rank index'' meaning that, as the rank increases from $r$ to $r+1$, a new set of weights has to be introduced in LoRA -- effectively expanding matrices $\boldsymbol{B}$ and $\boldsymbol{A}$ -- and these new weights will be associated with the multidimensional random variable $\boldsymbol{\theta}_{\ell r+1}$. For notational convenience, we define $\boldsymbol{\theta}_\ell=\left\{\boldsymbol{\theta}_{\ell r}\right\}_{r=1}^{\infty}$, $\boldsymbol{\theta}=\left\{\boldsymbol{\theta}_{\ell}\right\}_{\ell=1}^{L}$ and $\boldsymbol{\lambda}=\left\{\lambda_{\ell}\right\}_{\ell=1}^{L}$. Under these assumptions, we can write $p(Y| X) = \int p(Y,\boldsymbol{\theta},\boldsymbol{\lambda} | X)d\boldsymbol{\theta}d\boldsymbol{\lambda}$, which is unfortunately intractable. Therefore, we apply the same variational approach of \cite{errica2026adaptive}, which we refer to for the full details, with the only conceptual distinction that $r$ here refers to a rank index instead of a neuron index. 

To maximize an intractable Eq. \ref{eq:objective}, we can instead work with the evidence lower bound (ELBO):
\begin{align}
    & \log p(Y|X)  \geq
    \mathbb{E}_{q(\boldsymbol{\lambda},\boldsymbol{\theta})}\left[\log \frac{p(Y,\boldsymbol{\lambda},\boldsymbol{\theta} | X)}{q(\boldsymbol{\lambda},\boldsymbol{\theta})} \right],
    \label{eq:elbo}
\end{align}
where we make the following assumptions about the joint distribution $p(Y,\boldsymbol{\lambda},\boldsymbol{\theta} | X)$ of the generative model and the associated variational distribution $q(\boldsymbol{\lambda},\boldsymbol{\theta})$:  

\begin{align}
    &p(Y, \boldsymbol{\lambda}, \boldsymbol{\theta} | X) = \prod_{i=1}^N p(y_i, \boldsymbol{\lambda}, \boldsymbol{\theta} | x_i) && p(y_i, \boldsymbol{\lambda}, \boldsymbol{\theta} | x_i) = p(y_i | \boldsymbol{\lambda}, \boldsymbol{\theta}, x_i) p(\boldsymbol{\lambda}) p(\boldsymbol{\theta}) \\
    & p(\boldsymbol{\lambda})=\prod^{L}_{\ell=1}p(\lambda_\ell)=\prod^{L}_{\ell=1}\mathcal{N}(\lambda_{\ell}; \mu^\lambda_\ell, \sigma^\lambda_\ell)  &&  p(\boldsymbol{\theta}) = \prod^{L}_{\ell=1}\prod^{\infty}_{r=1}p(\theta_{\ell r})    
\end{align}
\begin{align}    
    & p(\theta_{\ell r})=\mathcal{N}(\theta_{\ell r}; \mathbf{0}, \text{diag}(\sigma^{\theta}_{\ell})) && p(y_i | \boldsymbol{\lambda}, \boldsymbol{\theta}, x_i) = \text{LoRA Neural Net}\label{eq:neunet} \\
    & q(\boldsymbol{\lambda},\boldsymbol{\theta})=q(\boldsymbol{\lambda})q(\boldsymbol{\theta} | \boldsymbol{\lambda}) && q(\boldsymbol{\lambda})=\prod_{\ell=1}^L q(\lambda_\ell)=\prod_{\ell=1}^L\mathcal{N}(\lambda_\ell; \nu_\ell, 1) \\
    & q(\boldsymbol{\theta} | \boldsymbol{\lambda})=\prod_{\ell=1}^L \prod_{r=1}^{D_\ell}q(\theta_{\ell r})\prod_{r'=D_\ell+1}^{\infty}p(\theta_{\ell r'}) && q(\theta_{\ell r}) = \mathcal{N}(\theta_{\ell r}; \rho_{\ell r}, \mathbf{I}).
\end{align}
Here, $\mu_\ell^\lambda, \sigma_\ell^\lambda, \sigma_\ell^\theta$ represent hyper-parameters controlling our prior assumptions about ideal ranks and ideal value of the LoRA weights, whereas $\nu_\ell, \rho_{\ell r}$ are learnable variational parameters that control the effective LoRA rank and LoRA weights at component $\ell$, respectively. In particular, $D_\ell$ represents the \textit{finite} rank used for LoRA at component $\ell$, and it is computed as the quantile function of a discretized exponential $f_\ell(x;\nu_\ell) = (1-e^{-\nu_\ell(x+1)}) - (1-e^{-\nu_\ell x})$, evaluated at $0.9$. In other words, the effective rank $D_\ell$ at component $\ell$ \textbf{is determined via a continuous parameter} $\nu_\ell$ that acts as a proxy for the ideal rank and can be easily learned.

The final probabilistic objective reduces to
\begin{align}
    & \sum_\ell^L\log\frac{p(\nu_\ell;\mu_\ell^\lambda,\sigma_\ell^\lambda)}{q(\nu_\ell;\nu_\ell)} + \sum_\ell^L\sum_{r=1}^{D_\ell}\log \frac{p(\rho_{\ell r};\sigma_\ell^\theta)}{q(\rho_{\ell r};\rho_{\ell r})} + \sum_{i=1}^N \log p(y_i | \boldsymbol{\nu}, \boldsymbol{\rho}, x_i),
    \label{eq:objective-final}
\end{align}
which is essentially composed of an \textit{optional} regularization term for the desired rank, an \textit{optional} regularization over the LoRA weights, and a \textbf{mandatory} loss term associated with the fine-tuning task. This loss can be optimized via standard backpropagation: as $\nu_\ell$ changes, we dynamically recompute the rank of each LoRA component $\ell$, effectively introducing or cutting parameters on the fly. This means that, in principle, the model's size can change during training.

\subsection{Adaptive Rank LoRA}
\label{subsec:adaptive-lora}
To learn an effective LoRA rank per LoRA component $\ell$, we must incorporate the discretized exponential $f_\ell(x;\nu_\ell)$ into $\Delta W_\ell$, in a way that reflects how the variational framework of the previous section determines the effective rank $D_\ell$. For this reason, we remind that the role of the discretized exponential is to assign a decreasing ordering of importance to each rank index, meaning that we would like the last columns of $B_\ell$ to be less important than the former ones (or, equivalently, the last rows of $A_\ell$). This way, changes to the first rank indices will have a greater effect on performances, while we can safely increase the rank index without impacting $\Delta W_\ell$ too much.

\begin{figure}[t]
\centering
\resizebox{0.8\textwidth}{!}{\input{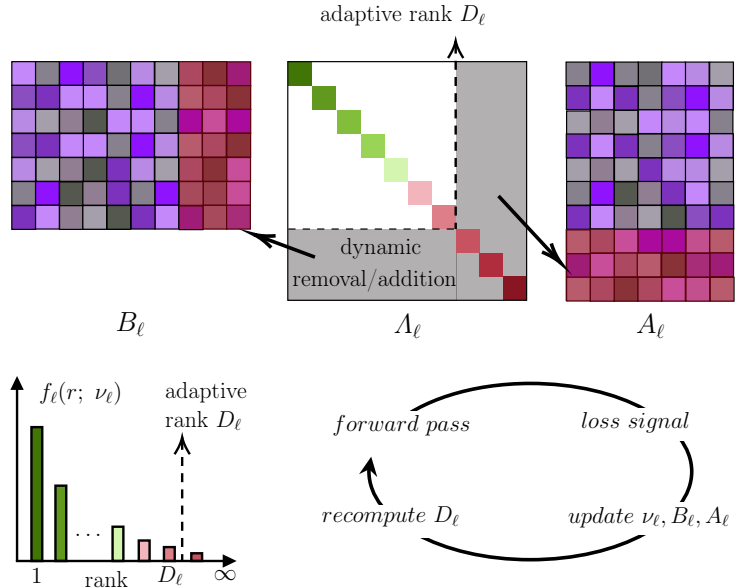}}
\caption{\method{} works by dynamically determining an adaptive rank $D_\ell$ for each LoRA component by truncating an exponential distribution $f_\ell(r;\nu_\ell)$, parametrized by a learnable $\nu_\ell$. This makes the rank dependent on the component and the task.}
\label{fig:method}
\end{figure}

For this reason, we formally consider $p(y_i | \boldsymbol{\nu}, \boldsymbol{\rho}, x_i)$ as a generic neural network and construct each LoRA component as follows:
\begin{align}
    \Delta W_\ell = B_\ell \Lambda_\ell A_\ell, &&\Lambda_\ell=diag\left(f(1; \nu_\ell),\dots,f(D_\ell; \nu_\ell)\right) 
\end{align}

This approach is extremely easy to implement and can grow/shrink dynamically during training; in the case of a growing $D_\ell$, as new rank dimensions are added we randomly initialize the new weights of $B_\ell$ and $A_\ell$. The approach is visually represented in \cref{fig:method}.

\subsubsection{Weight Initialization.} The rescaling generated by $\Lambda_\ell$ has an effect on convergence speedup, since it affects the gradients. To counteract this effect, we apply a ``rescaled'' Kaiming initialization; in particular, we initialize $A_\ell$ weights from a Gaussian distribution with standard deviation $\frac{\sqrt{2}}{\sqrt{\sum_{j=1}^{D_{\ell}} f^2_{\ell}(j)}}$. Instead, $B_\ell$ is initialized as a zero matrix following \cite{hu2022lora}.

\subsubsection{Implicit Space Search.} The main conceptual advantage of \method{} is that it replaces the search over a very large number of different LoRA architectures. In principle, finetuning $S$ subjects while trying $K$ different ranks for a network with $L$ components amounts to training $SK^L$ different architectural configurations, way beyond any practical application even for small values of $K$ and $L$. Instead, continuous optimization of $\boldsymbol{\nu}$ allows to softly introduce new ranks when needed and truncate those that are not necessary any longer, \textbf{all in a single training run}. Therefore, despite the introduction of (optional) regularization hyper-parameters, we argue that our approach makes the search over a huge amount of LoRA architectures much more feasible than before.

\subsubsection{Training Loss.} 
We finetune the LoRA modules using a combination of three losses, which are related in spirit to the ones of Equation \ref{eq:objective-final} in the variational framework. The main reconstruction loss is
\begin{equation}
\mathcal{L}_{\mathrm{MSE}} = \frac{1}{N} \sum_{i=1}^{N}  \|\hat{\epsilon}_i - \epsilon_i\|^2,
\end{equation}
where $\hat{\epsilon}_i$ is the model prediction, $\epsilon_i$ the target noise, , and $N$ the batch size.

We regularize the adaptive LoRA rank rates to remain close to a target:
\begin{equation}
\mathcal{L}_{\mathrm{reg}} = \sum_{\ell \in [1,\dots,L]} \left| \nu_\ell - \nu_{\mathrm{target}} \right|,
\quad
\nu_{\mathrm{target}} = -\frac{\log(1-q)}{r_\mathrm{target}},
\end{equation}
with $q$ being the quantile and $r_\mathrm{target}$ the rank we would like to push the LoRA components towards.
To encourage more selective and confident cross-token alignments, we minimize the entropy of the cross-attention maps:
\begin{equation}
\mathcal{L}_{\mathrm{entropy}}
=
- \frac{1}{|\mathcal{C}|}
\sum_{\ell \in \mathcal{C}}
\mathbb{E}_{p_\ell}
\left[
\log p_\ell
\right],
\end{equation}
where $\mathcal{C}$ denotes the set of components over which the cross-attention is computed, and $p_\ell$ represents the softmax-normalized attention map at component $\ell$. The overall loss, therefore, can be written as:
\begin{equation}
\mathcal{L}_{\mathrm{total}} = \mathcal{L}_{\mathrm{MSE}} + \lambda_r \mathcal{L}_{\mathrm{reg}} + \lambda_e \mathcal{L}_{\mathrm{entropy}},
\end{equation}
with $\lambda_r$ and $\lambda_e$ weighting factors.
k
\section{Experiments}
\label{sec:experiments}

We use SDXL \cite{podell2024sdxl} and KOALA-700m \cite{lee2024koala} as backbones for our experiments. On SDXL, we use 50 inference steps \cite{shah2024ziplora,shenaj2025lora}; on KOALA-700m, 25 \cite{koala700m_hf}.
To learn personalized subjects, we employ LoRA finetuning using the DreamBooth protocol \cite{ruiz2023dreambooth}. Our experiments are conducted on a set of 30 subjects sourced from
\cite{ruiz2023dreambooth}.
We select one random subject (vase) for hyper-parameter tuning, and then test on the remaining 29 subjects.
For each subject, we explore LoRA models of different capacities, with ranks  $\in \{8, 16, 32, 64, 128$$, 256, 512\}$. In \method experiments, the hyper-parameter tuning process selected 500 training steps for SDXL and 800 steps for KOALA. We fixed the learning rate of the Adam optimizer to $5e^{-5}$ and fixed weights $\lambda_r=\lambda_e=1e^{-4}$. For LoRA, we use 1000 training steps as in \cite{shah2024ziplora, shenaj2025lora}.
For each subject, we collect 10 prompts (please refer to the supplementary material) and then generate 5 images per prompt. We then compute the DINO, CLIP-I, and CLIP-T scores, comparing the features of each generated image with the features of the original subject image or the features of the prompt. To aggregate the score, we average the score of each subject across each generation in a prompt, and then across all prompts. In this way, we have a single score for each subject, and we average them across all subjects.

\section{Results}
\label{sec:results}

\begin{figure}[!h]
\centering
\renewcommand{\arraystretch}{1.2}
\setlength{\tabcolsep}{2pt}
\resizebox{\linewidth}{!}{%
\begin{tabular}{
    >{\centering\arraybackslash}m{2cm}
    >{\centering\arraybackslash}m{2.5cm} 
    >{\centering\arraybackslash}m{2.5cm} 
    >{\centering\arraybackslash}m{2.5cm} 
    >{\centering\arraybackslash}m{2.5cm} 
    >{\centering\arraybackslash}m{2.5cm} 
}
\includegraphics[width=2cm]{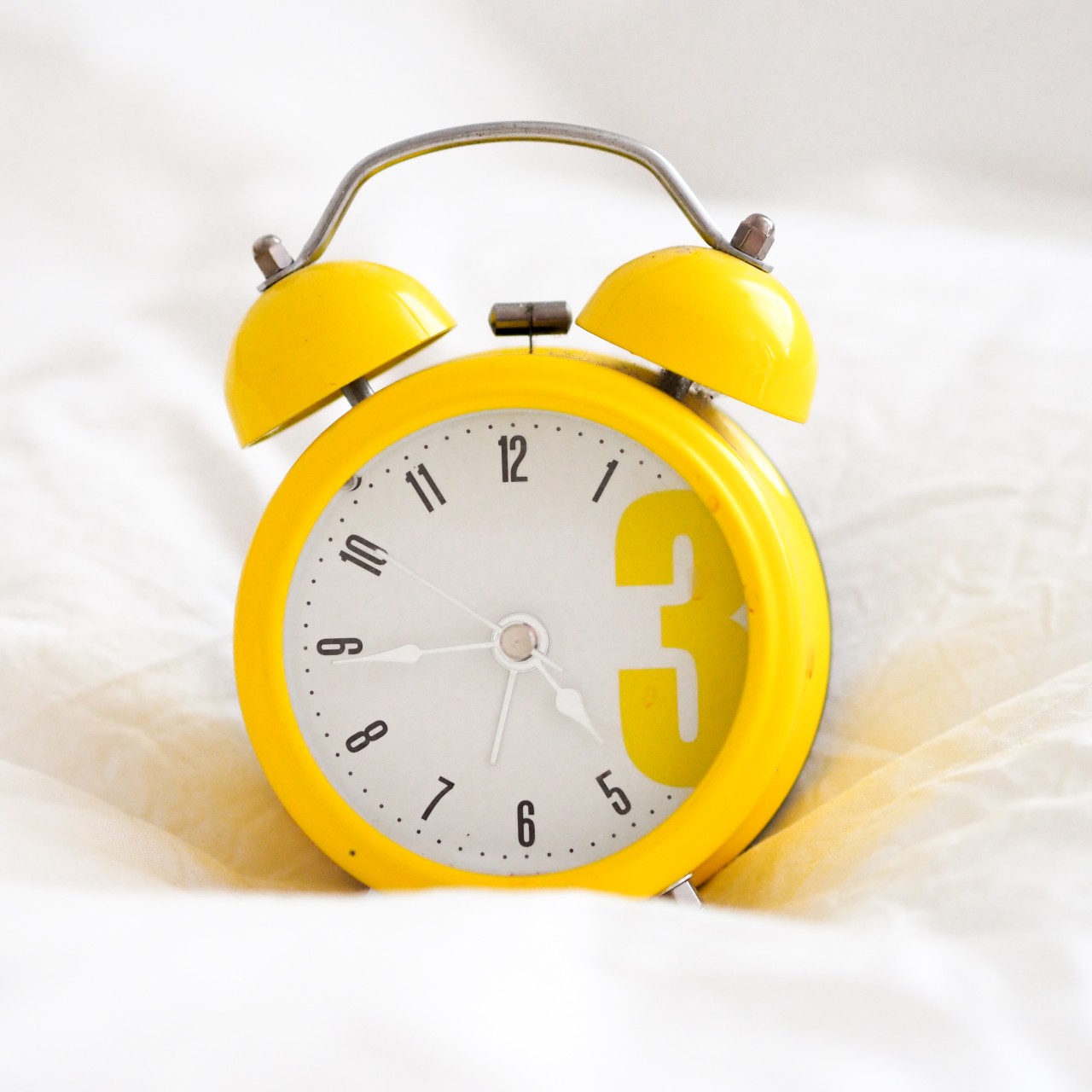}
& \textbf{``a \textcolor{red}{k} clock next to a cup of coffee on a kitchen counter''} 
& \textbf{``a \textcolor{red}{k} clock placed on pink silk fabric''} 
& \textbf{``a \textcolor{red}{k} clock on a mossy rock in a forest''} 
& \textbf{``a \textcolor{red}{k} clock with a city skyline in the background''} 
& \textbf{``a \textcolor{red}{k} clock in the snow under warm sunlight''} \\[2mm]

\scriptsize Rank 8
& \includegraphics[width=2.3cm]{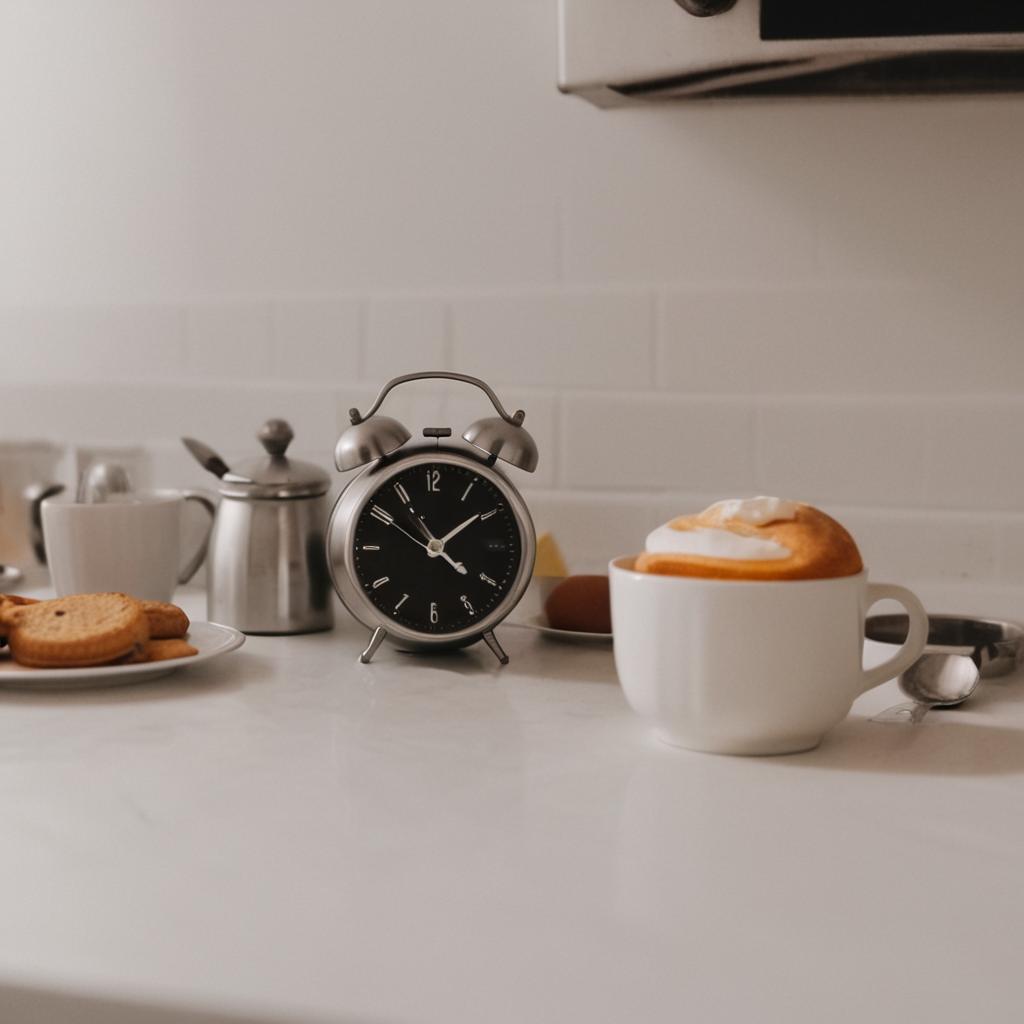}
& \includegraphics[width=2.3cm]{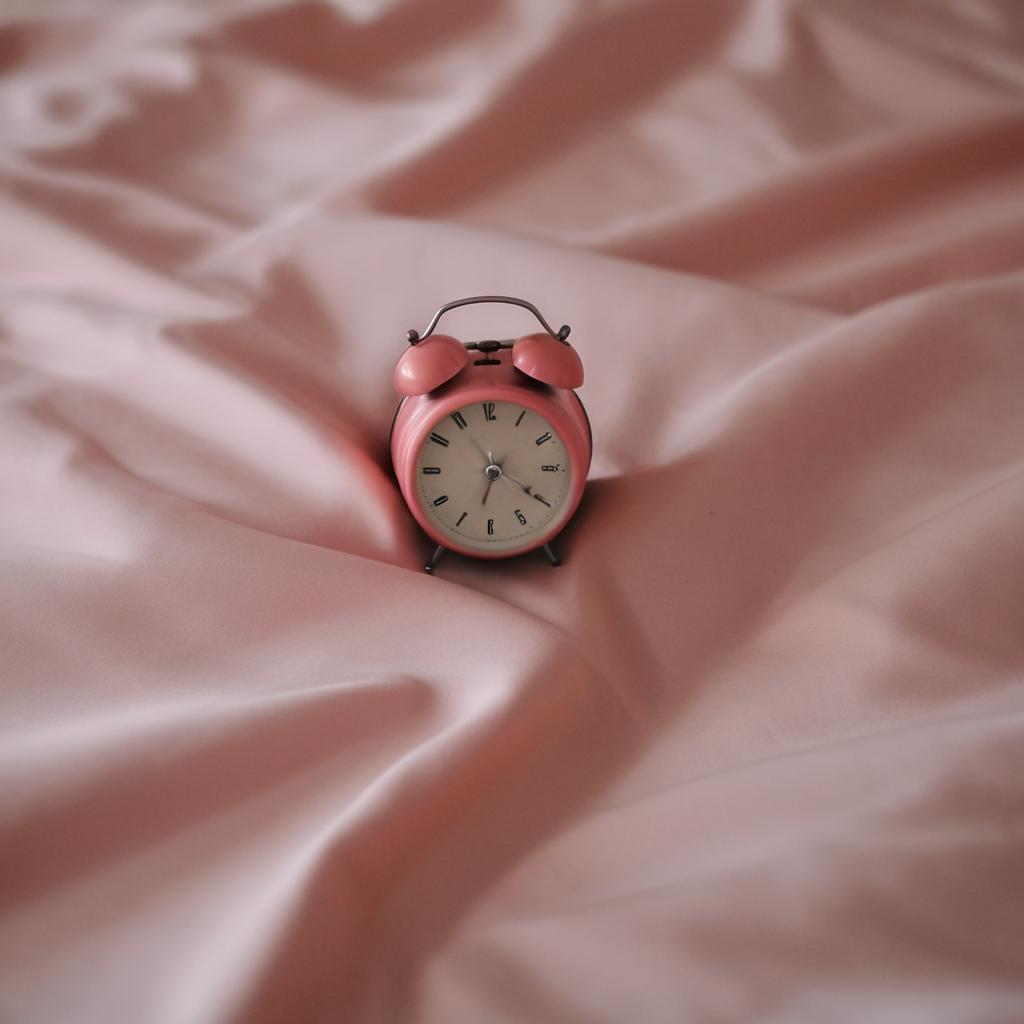}
& \includegraphics[width=2.3cm]{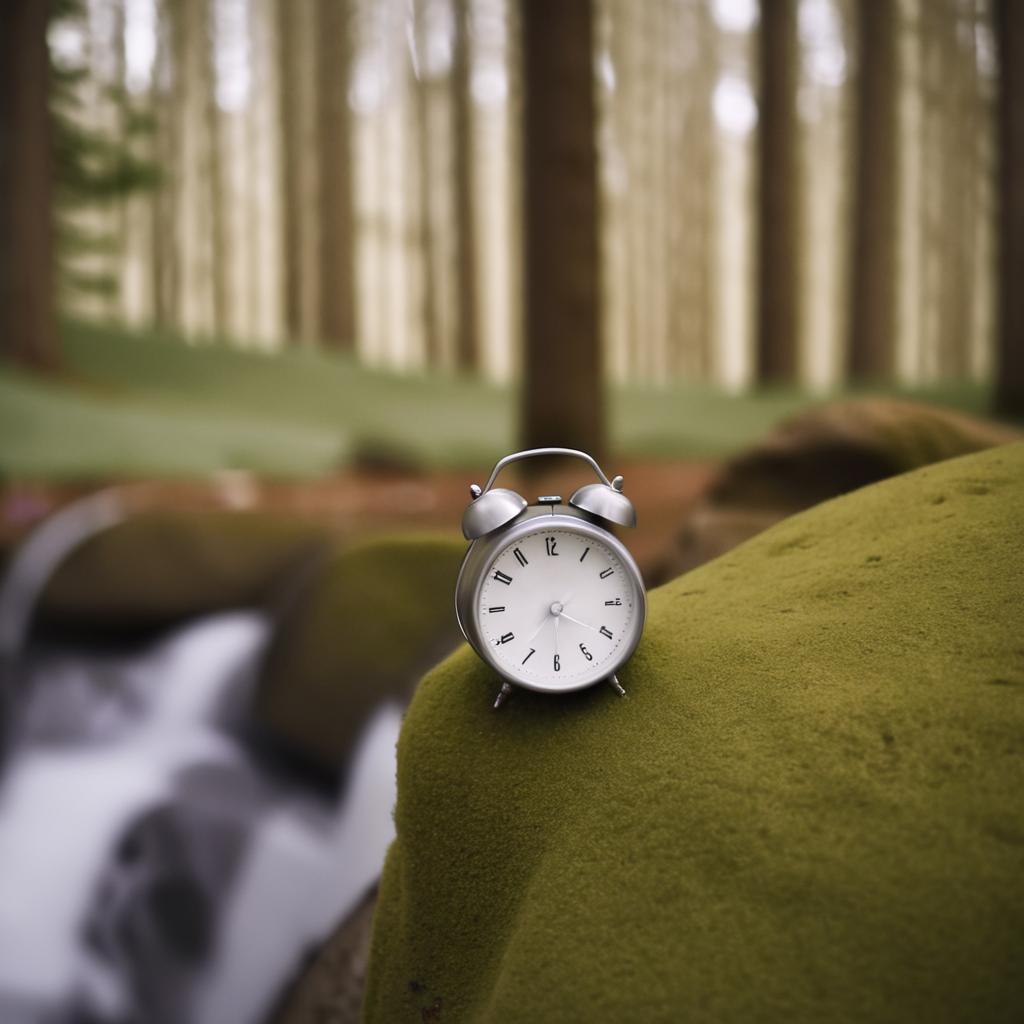}
& \includegraphics[width=2.3cm]{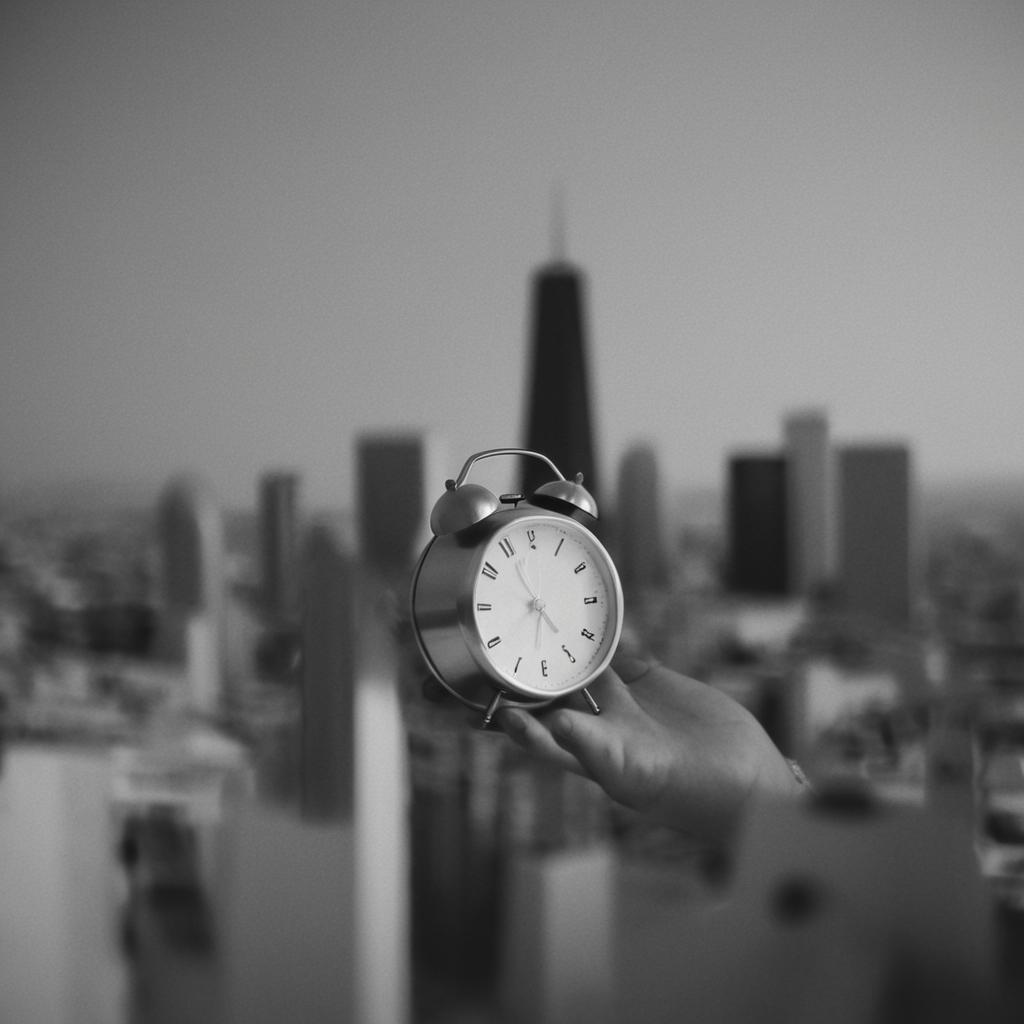}
& \includegraphics[width=2.3cm]{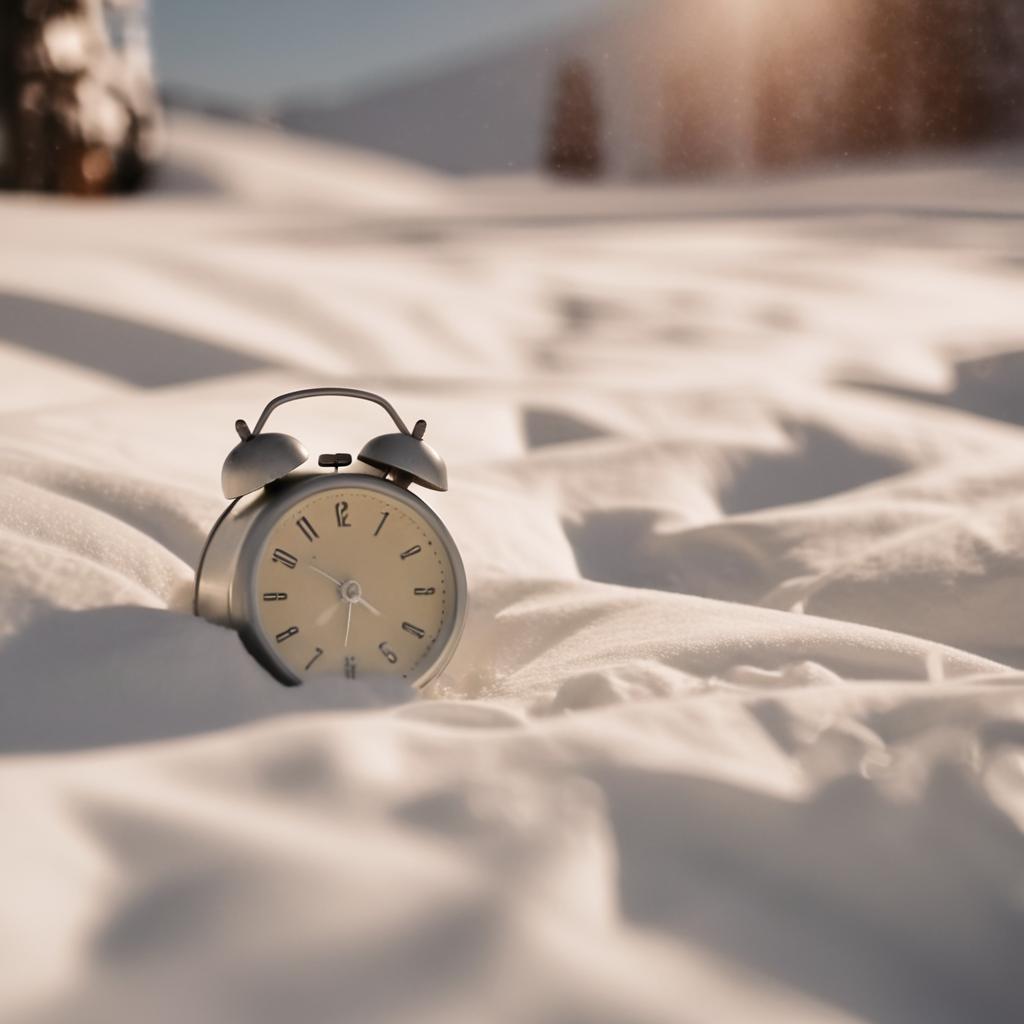}
\\[1mm]

\scriptsize Rank 64
& \includegraphics[width=2.3cm]{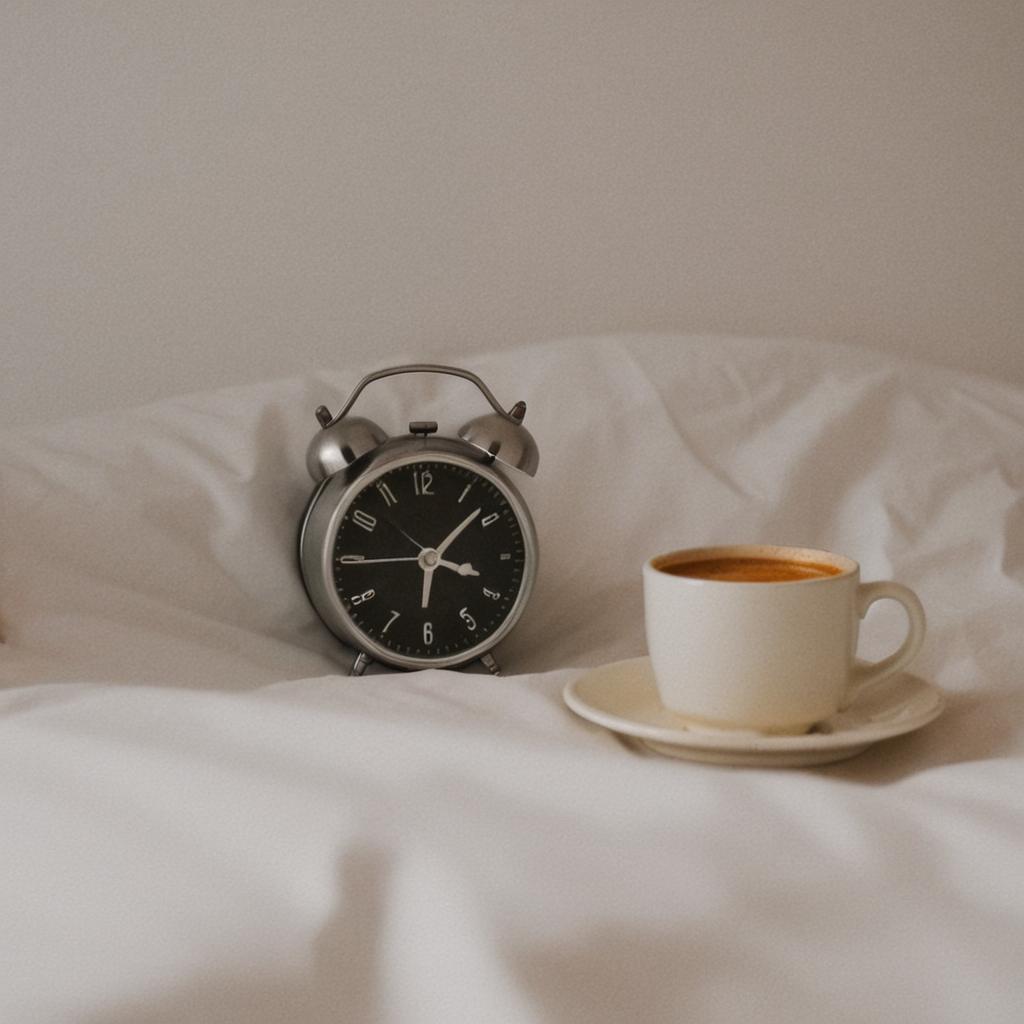}
& \includegraphics[width=2.3cm]{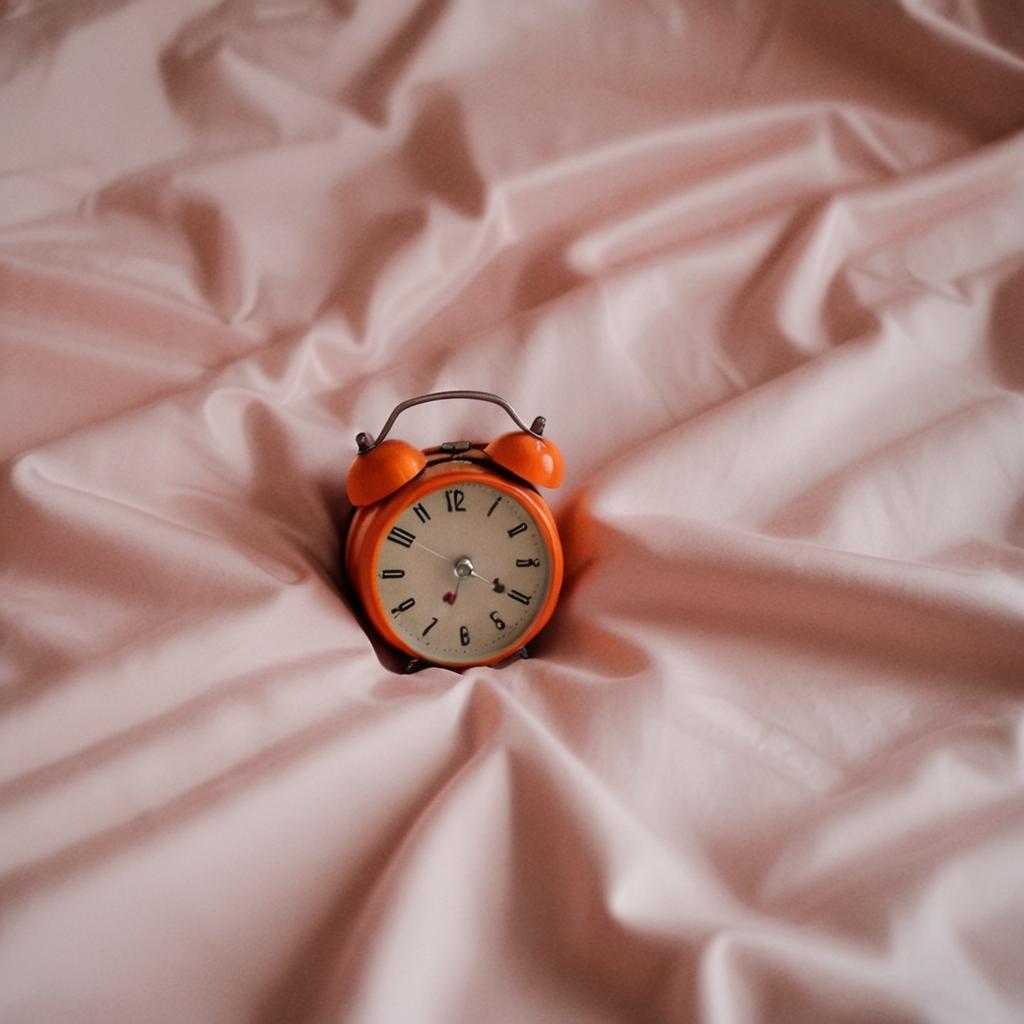}
& \includegraphics[width=2.3cm]{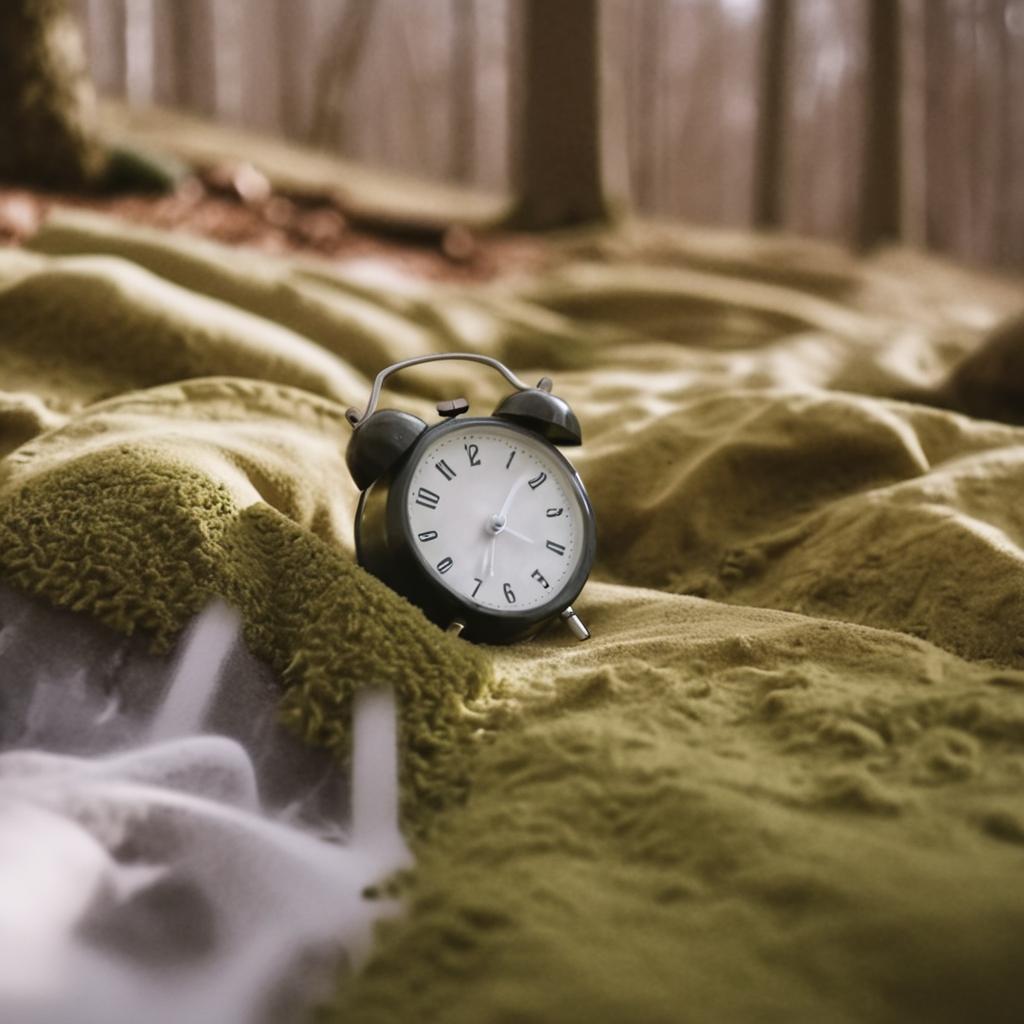}
& \includegraphics[width=2.3cm]{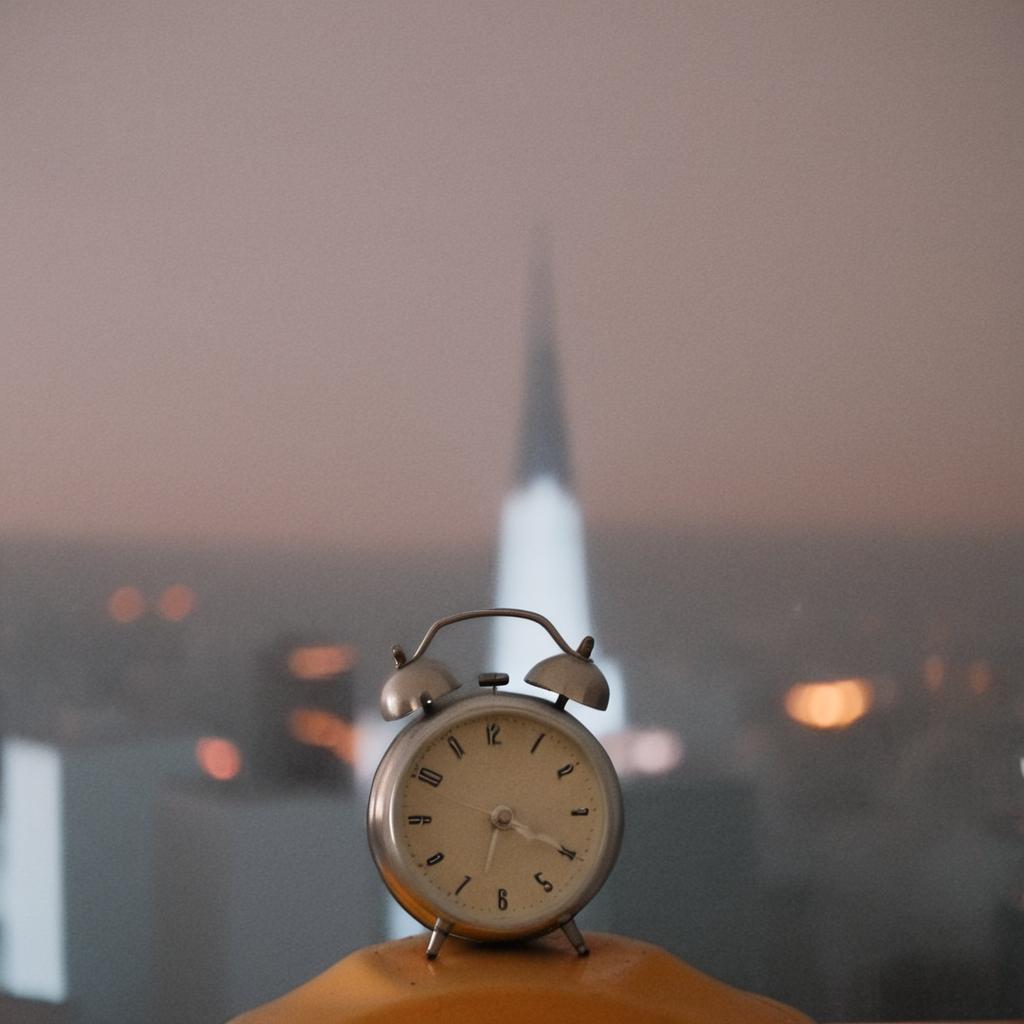}
& \includegraphics[width=2.3cm]{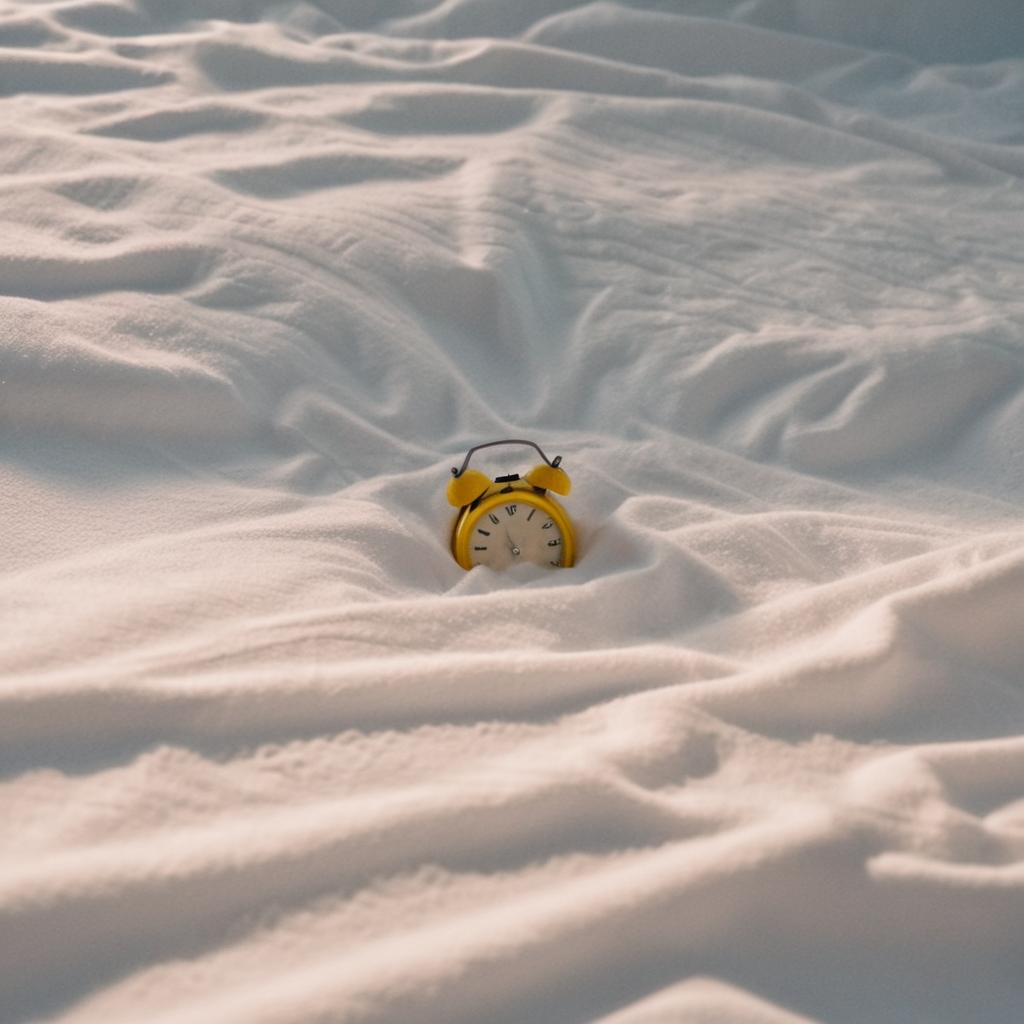}
\\[1mm]

\scriptsize Rank 512
& \includegraphics[width=2.3cm]{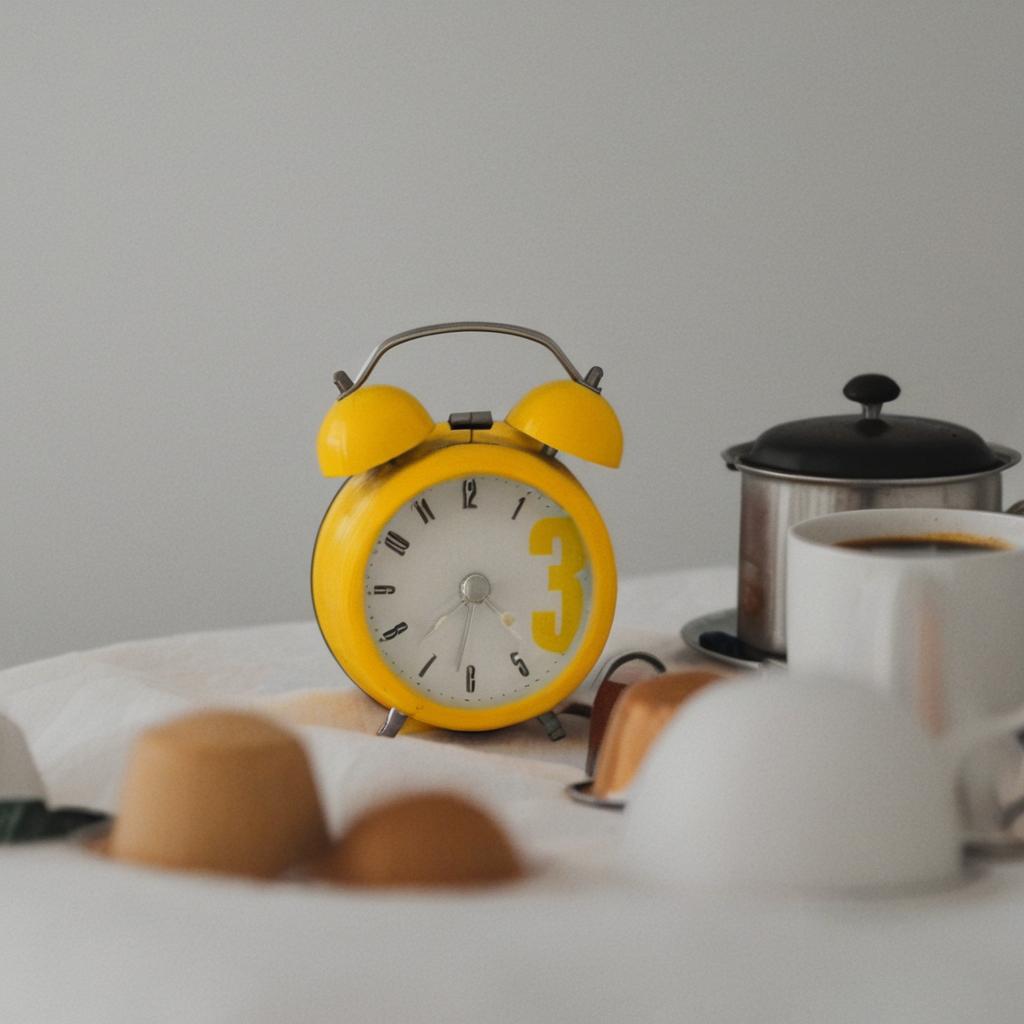}
& \includegraphics[width=2.3cm]{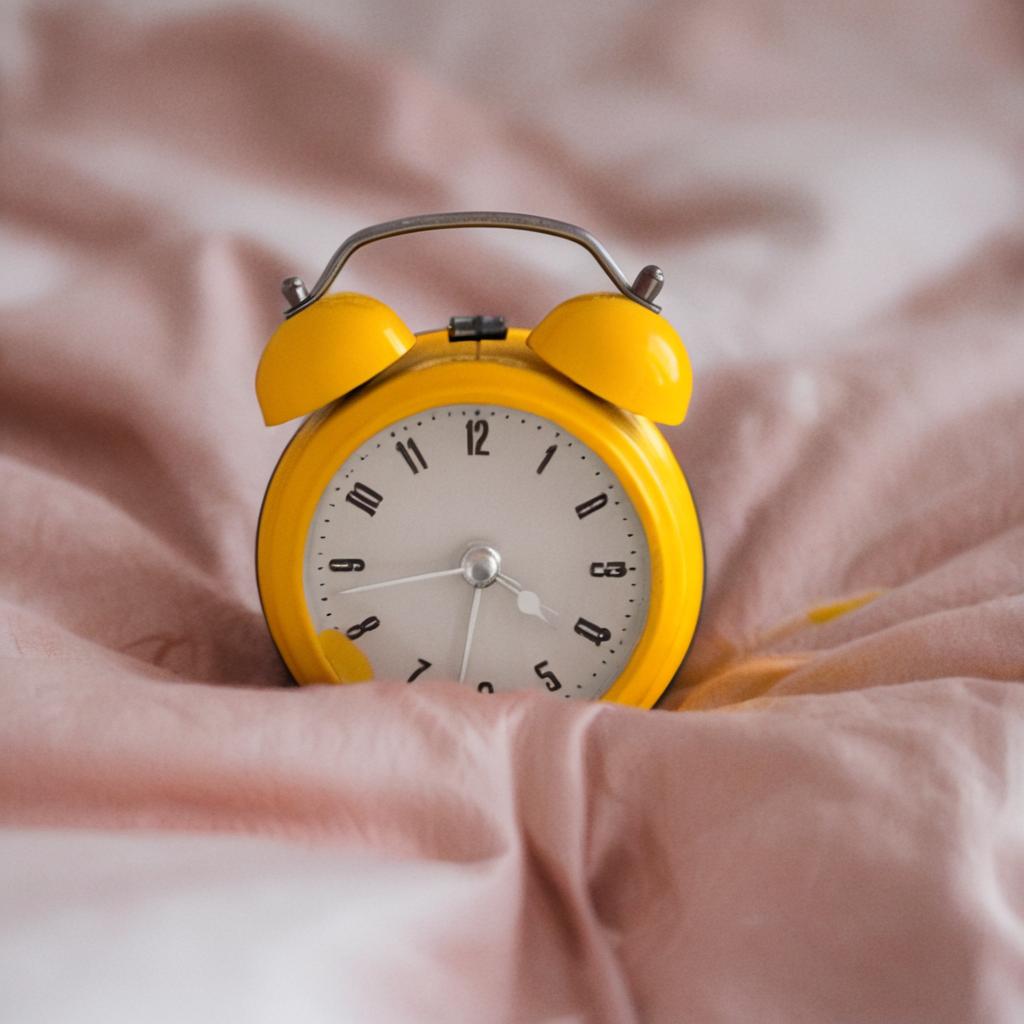}
& \includegraphics[width=2.3cm]{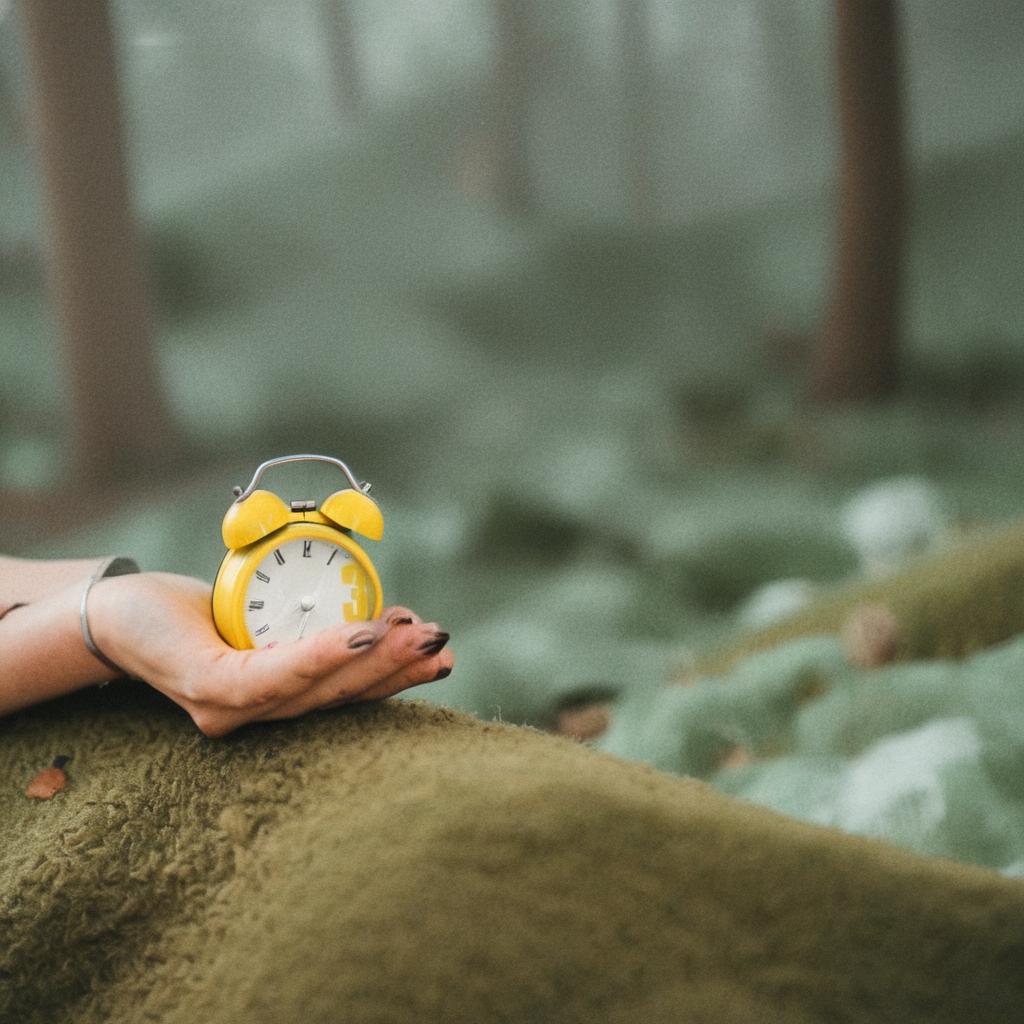}
& \includegraphics[width=2.3cm]{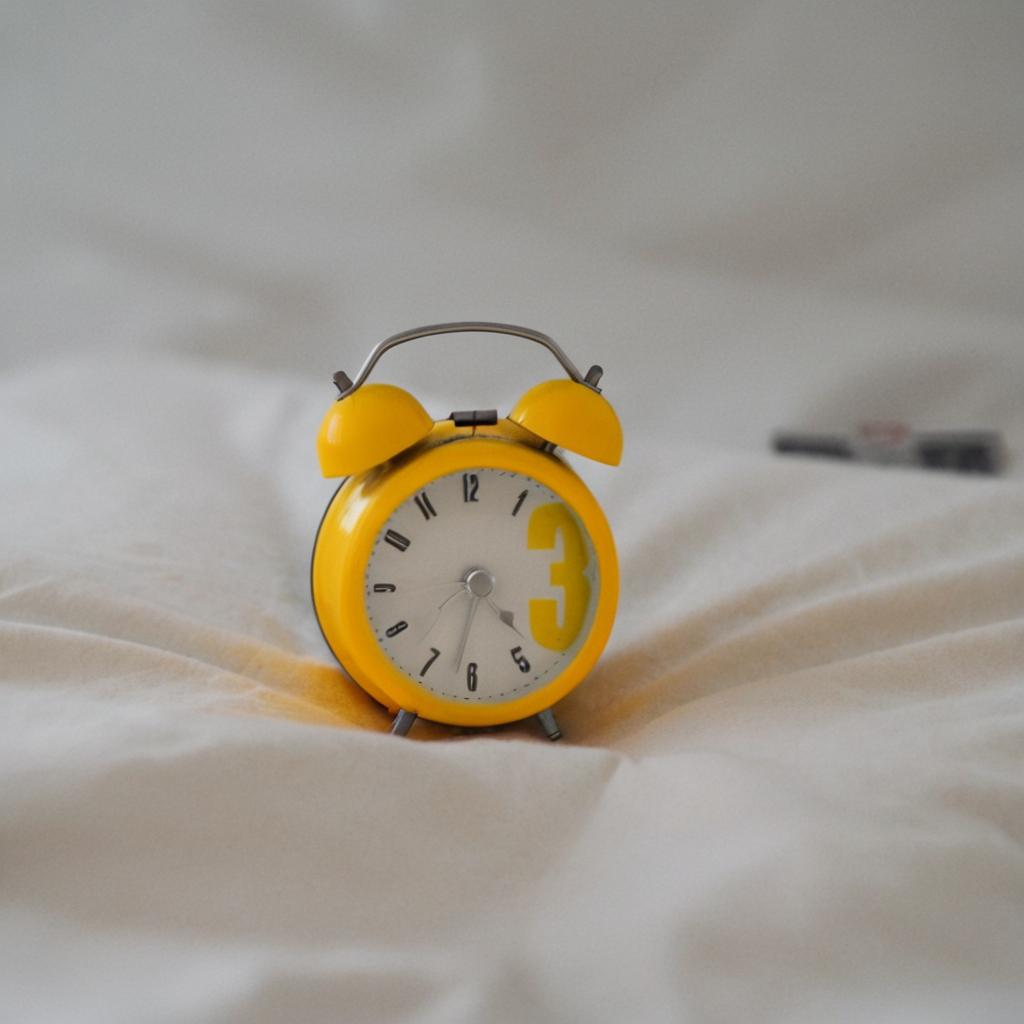}
& \includegraphics[width=2.3cm]{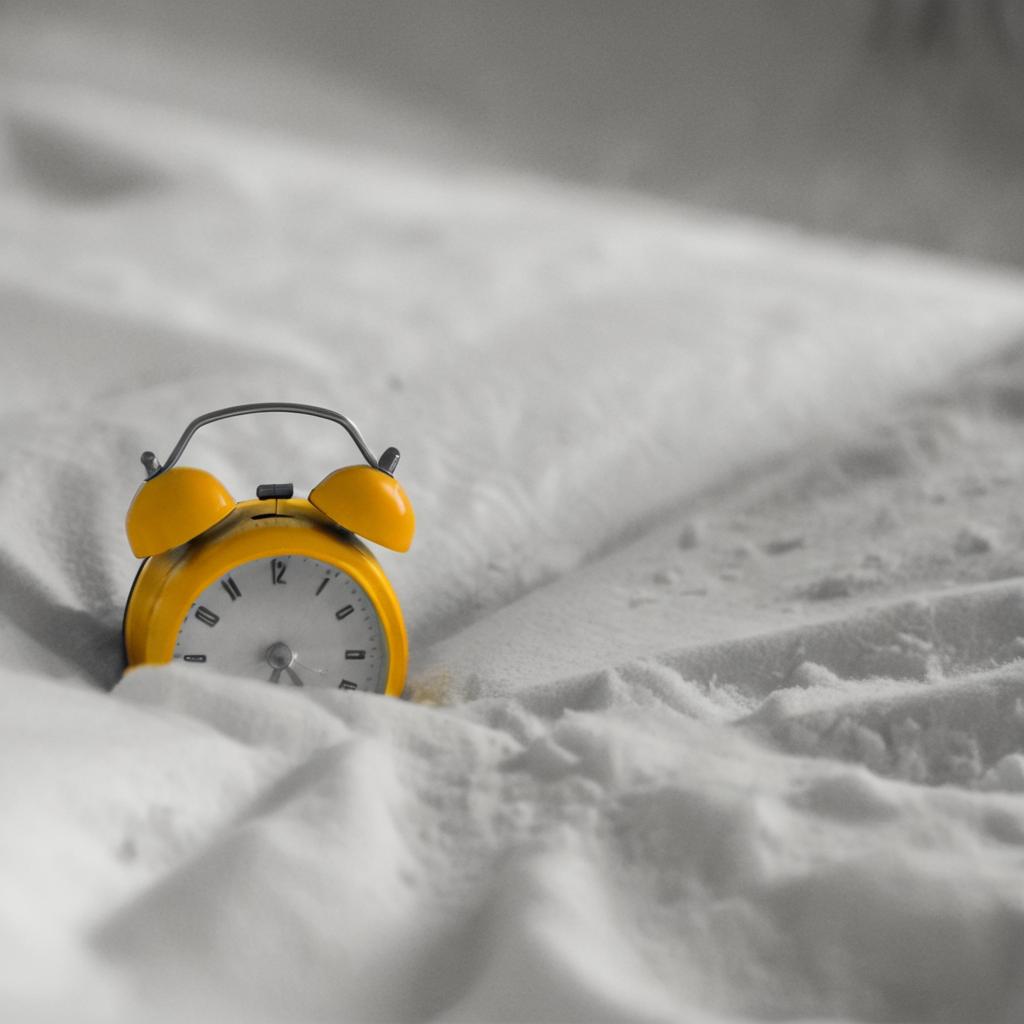}
\\[1mm]

\scriptsize \method
& \includegraphics[width=2.3cm]{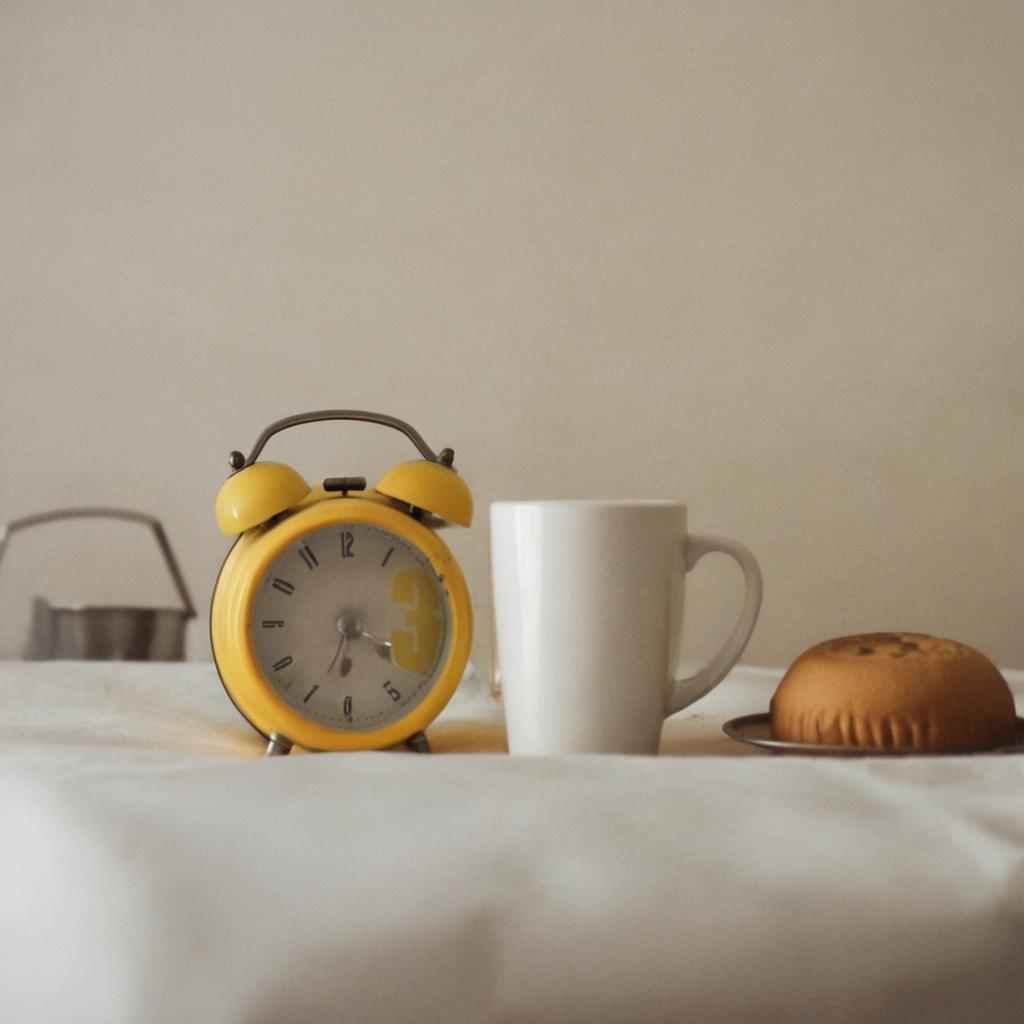}
& \includegraphics[width=2.3cm]{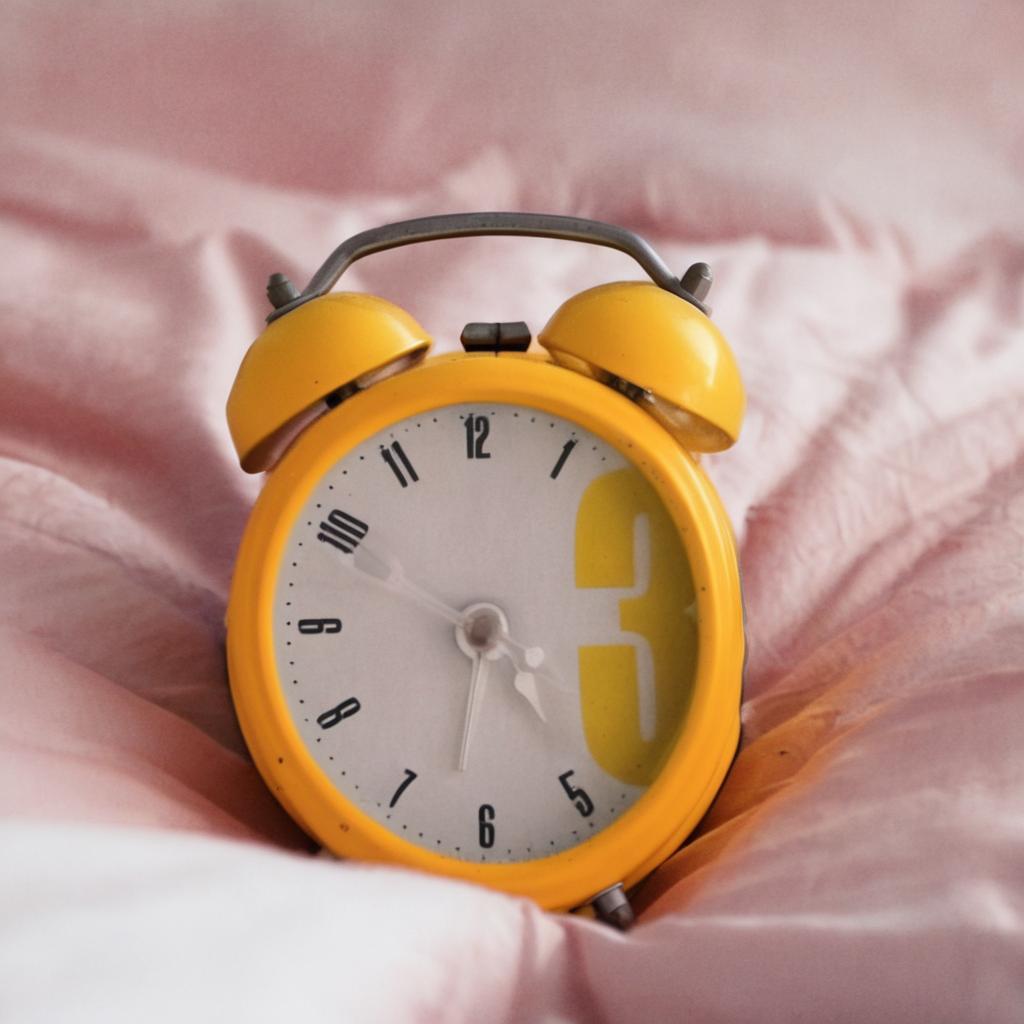}
& \includegraphics[width=2.3cm]{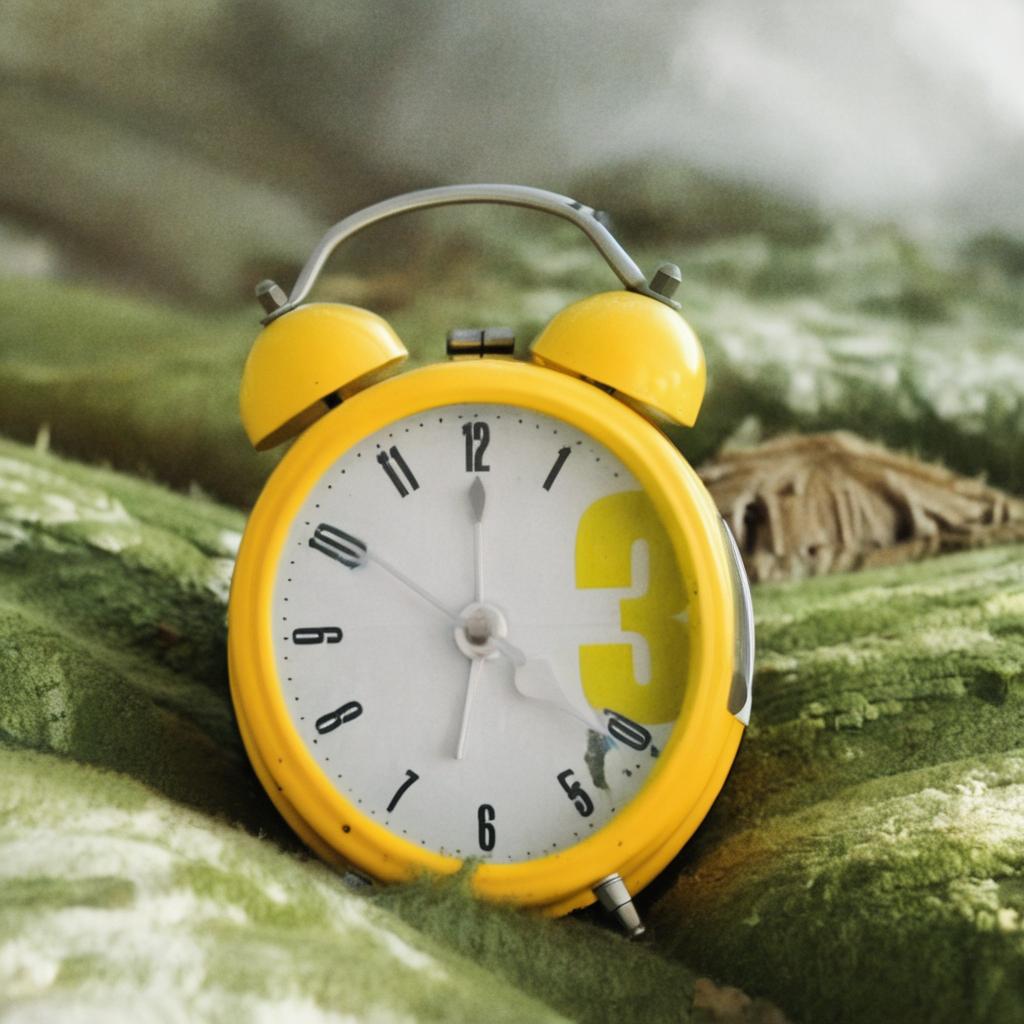}
& \includegraphics[width=2.3cm]{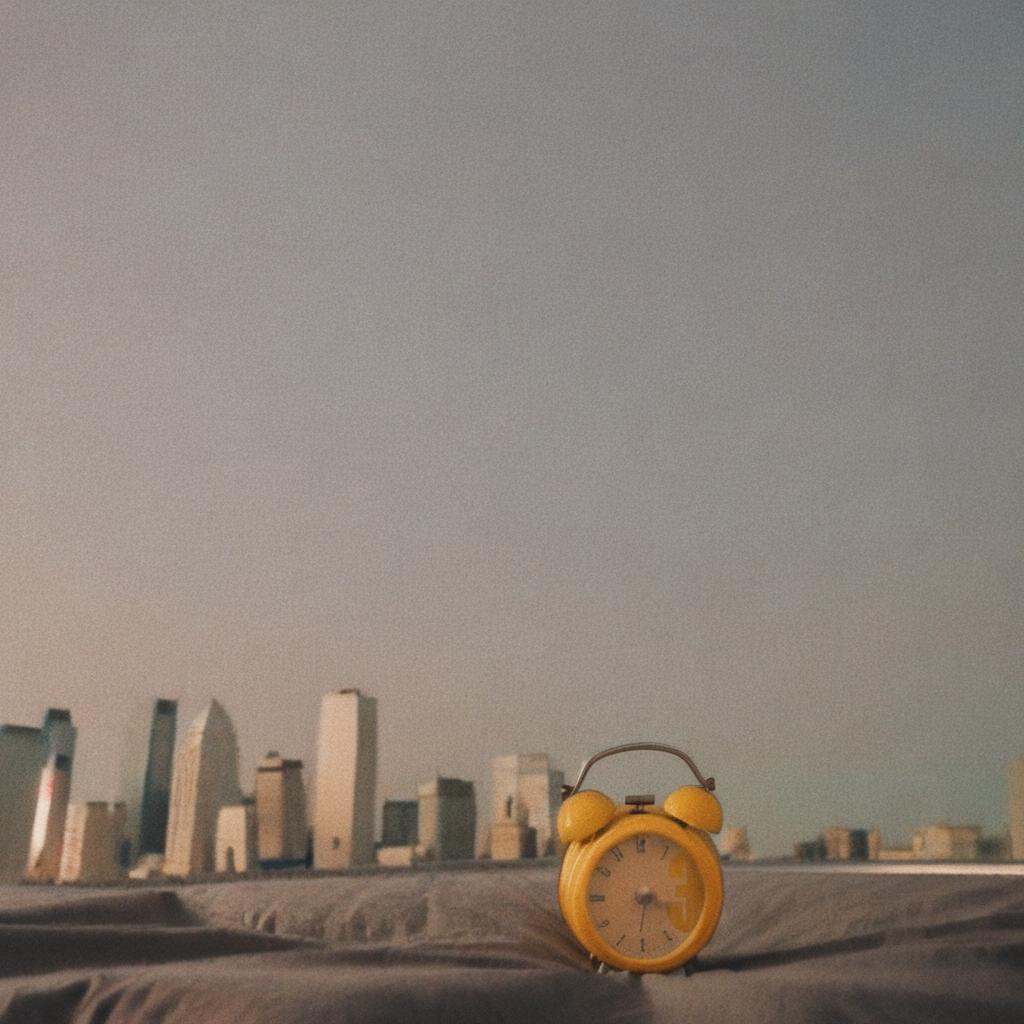}
& \includegraphics[width=2.3cm]{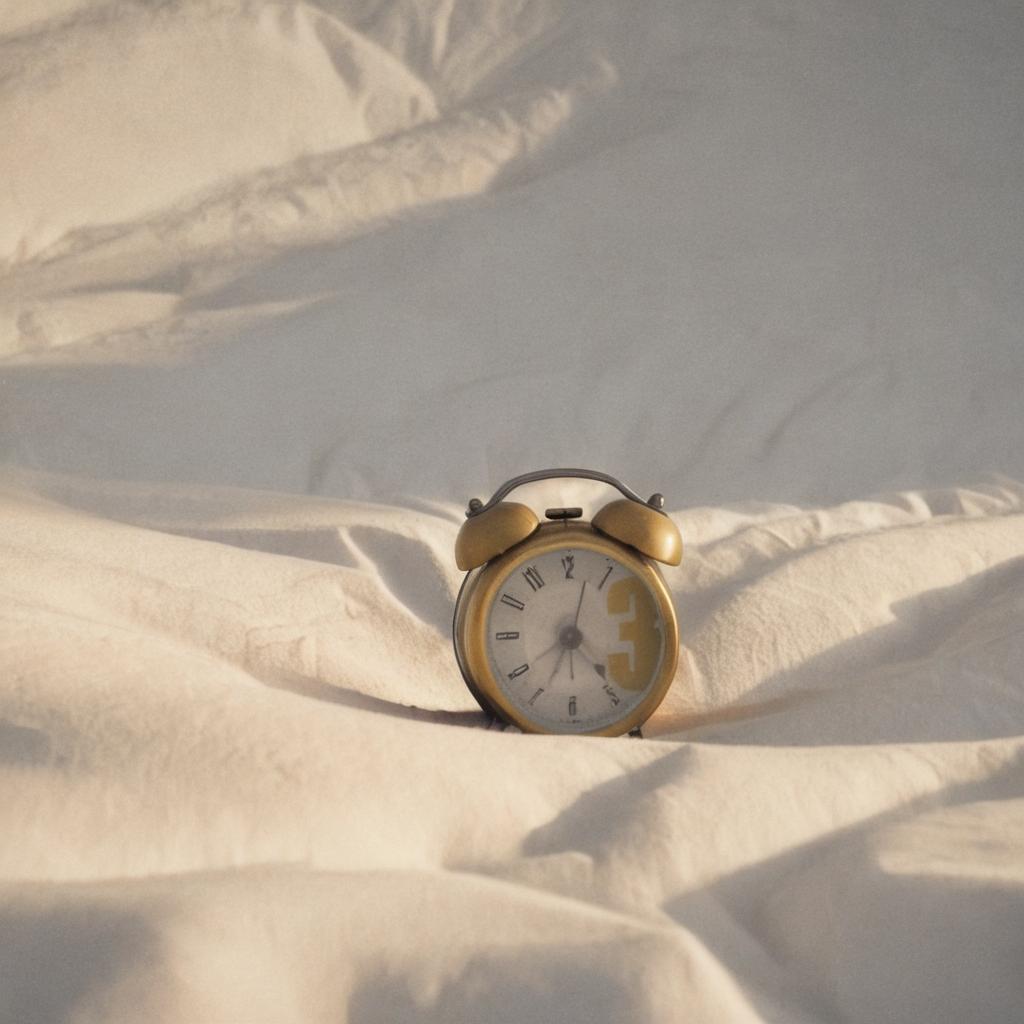}
\\
\end{tabular}%
}
\caption{Images generated using SDXL backbone for the ``clock" subject.  The original subject is present on the top left. 
}
\label{fig:qualitative_comparison_SDXL}
\end{figure}

\subsection{Qualitative Results}

\begin{figure}[!h]
\centering
\renewcommand{\arraystretch}{1.2}
\setlength{\tabcolsep}{2pt}
\resizebox{\linewidth}{!}{%
\begin{tabular}{
    >{\centering\arraybackslash}m{2cm}
    >{\centering\arraybackslash}m{2.5cm} 
    >{\centering\arraybackslash}m{2.5cm} 
    >{\centering\arraybackslash}m{2.5cm} 
    >{\centering\arraybackslash}m{2.5cm} 
    >{\centering\arraybackslash}m{2.5cm} 
}
\includegraphics[width=2cm]{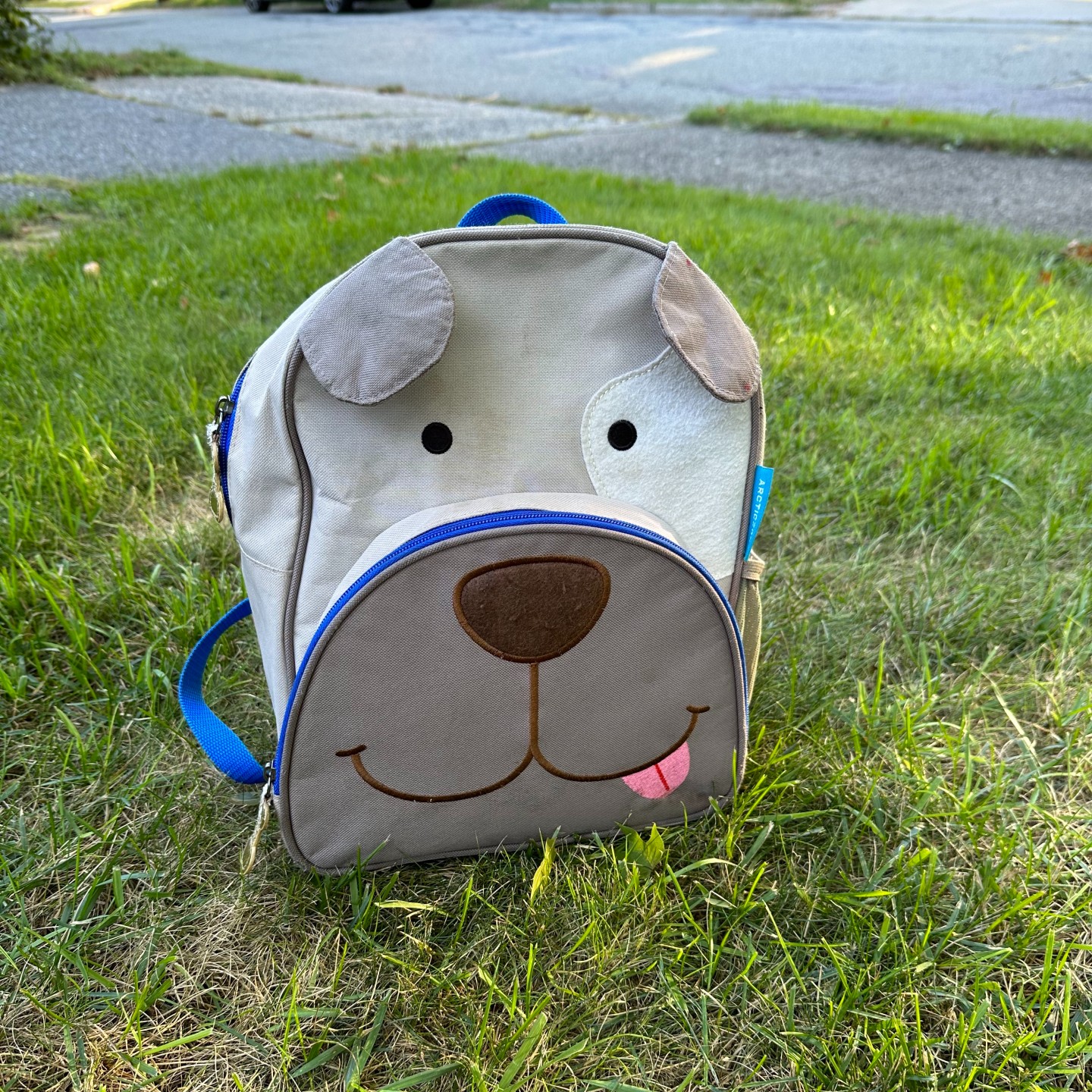}
& \textbf{``a \textcolor{red}{k} backpack on a cobblestone street after rain''} 
& \textbf{``a \textcolor{red}{k} backpack on a glass table with reflections''} 
& \textbf{``a \textcolor{red}{k} backpack with mountains and mist in the background''} 
& \textbf{``a \textcolor{red}{k} backpack floating in crystal clear water''} 
& \textbf{``a \textcolor{red}{k} backpack surrounded by neon lights''} \\[2mm]

\scriptsize Rank 8
& \includegraphics[width=2.3cm]{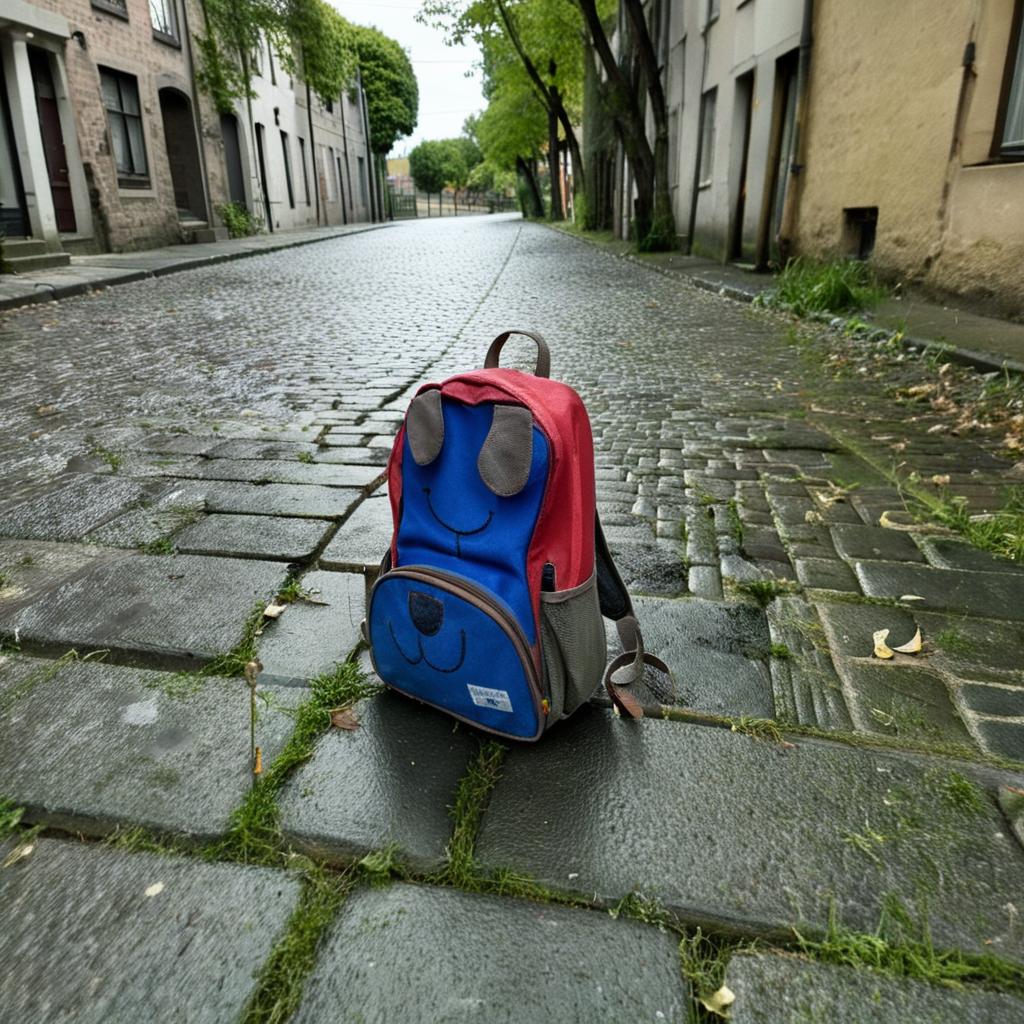}
& \includegraphics[width=2.3cm]{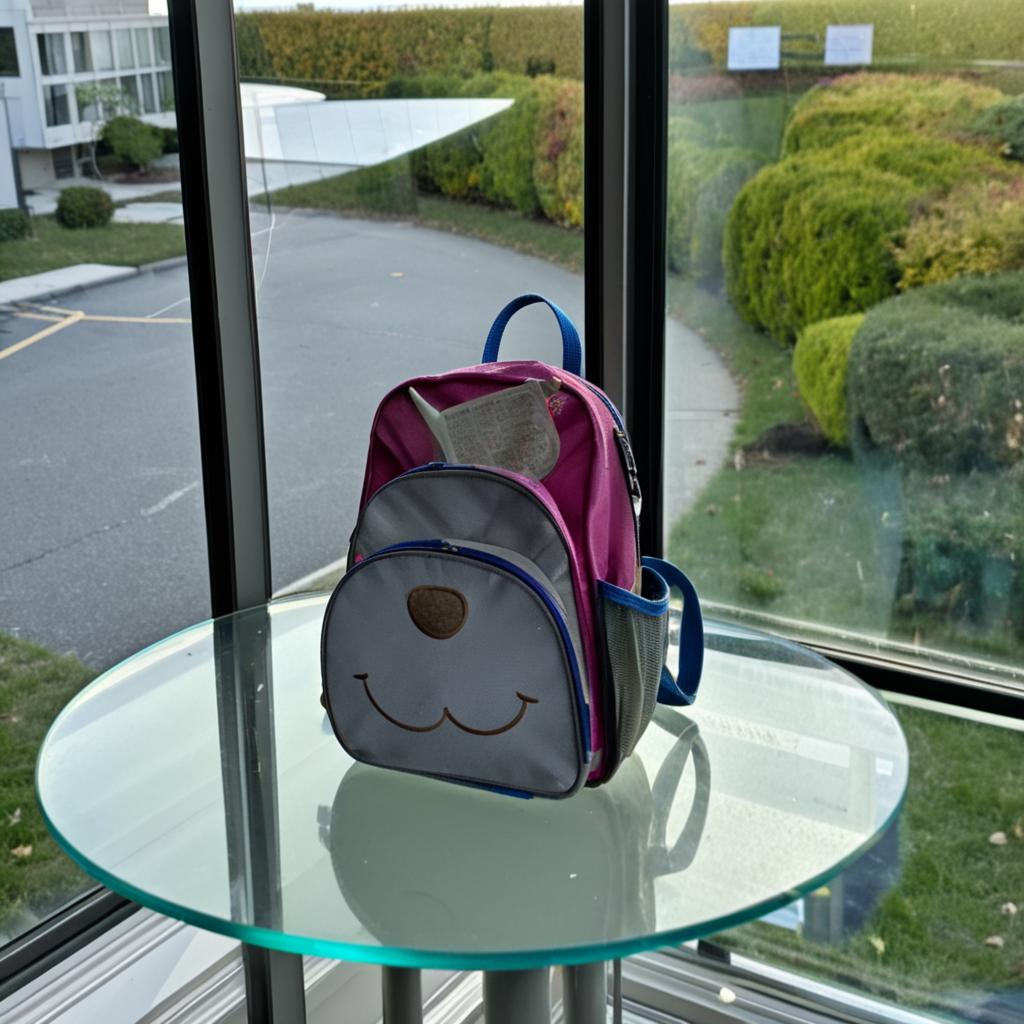}
& \includegraphics[width=2.3cm]{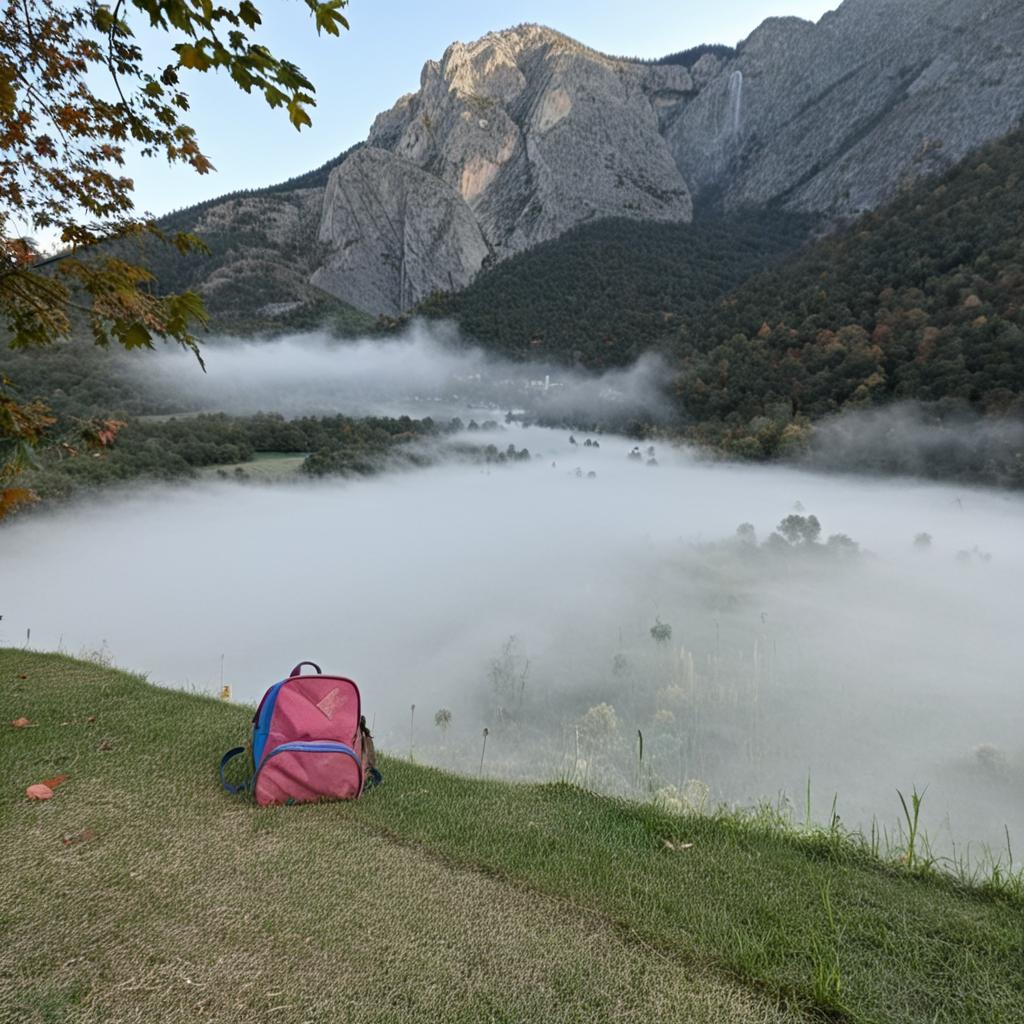}
& \includegraphics[width=2.3cm]{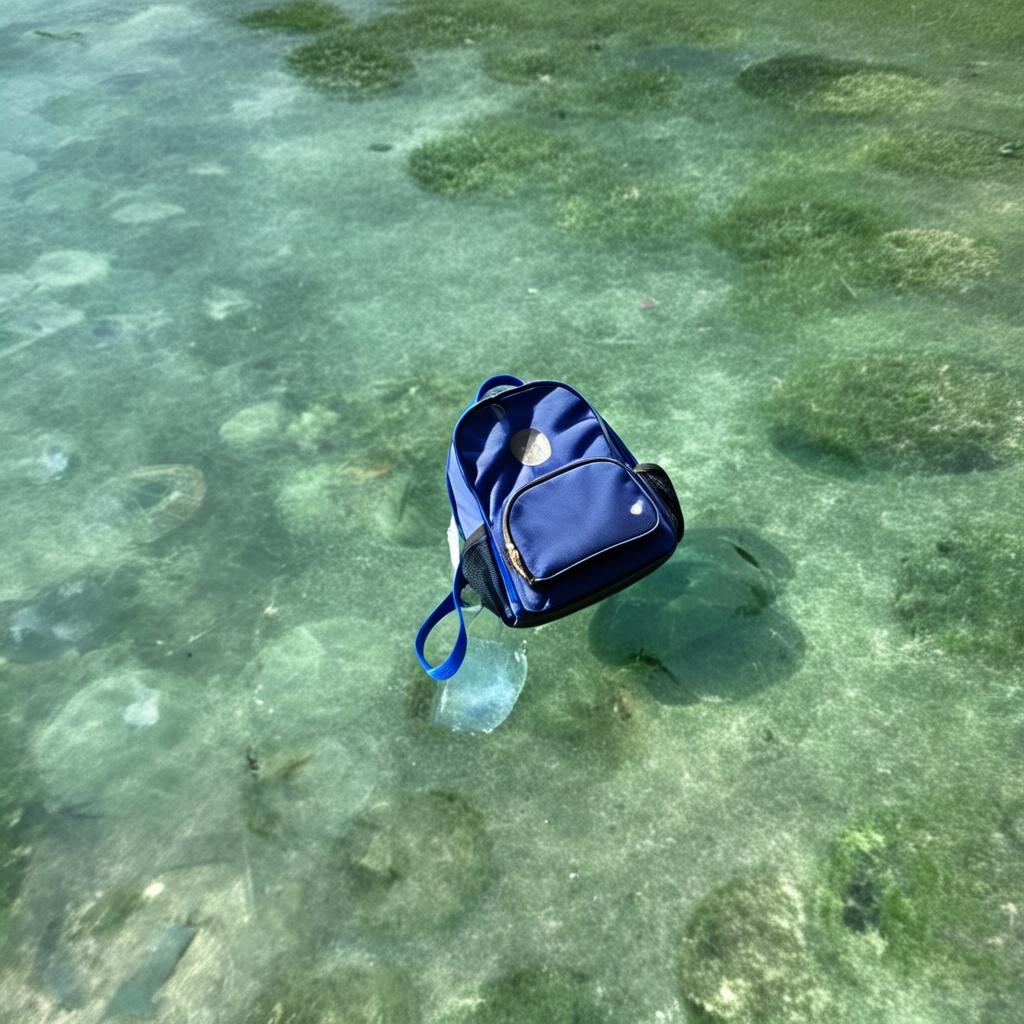}
& \includegraphics[width=2.3cm]{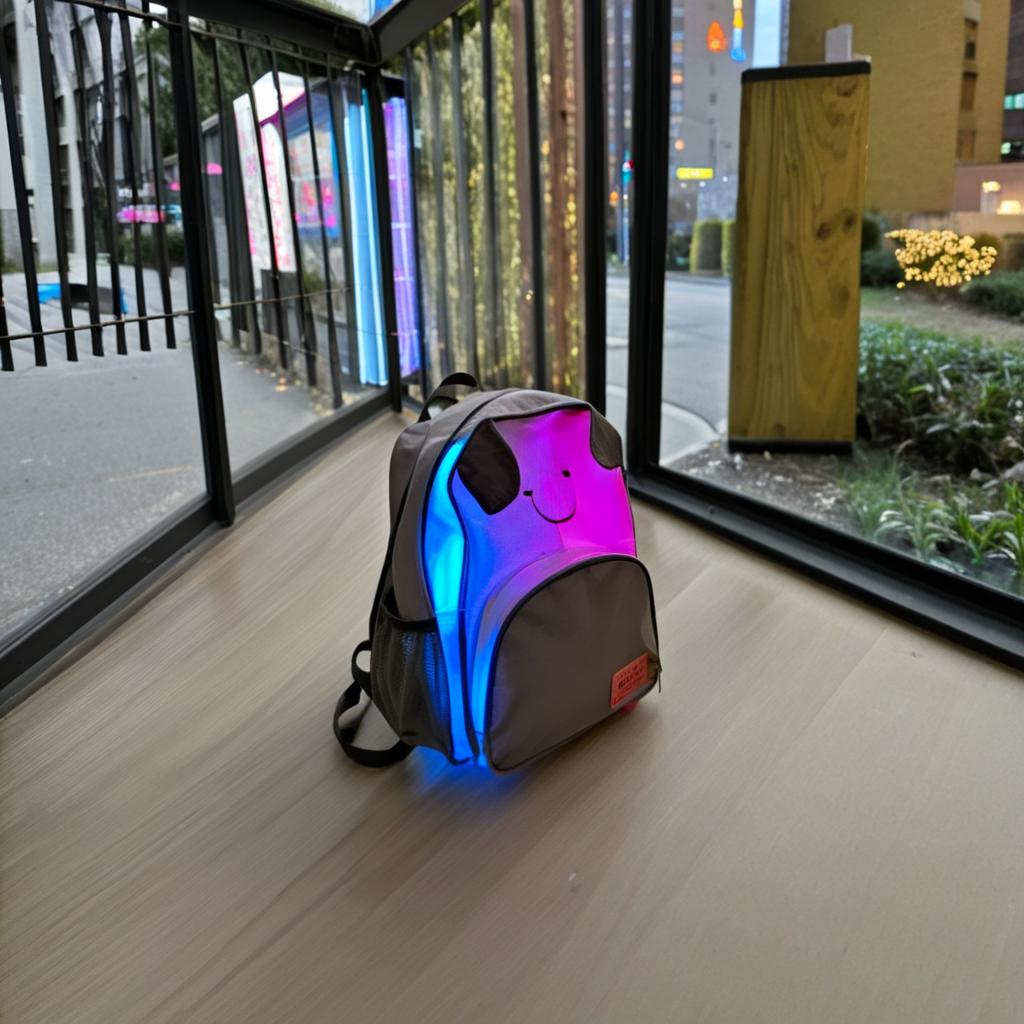}
\\[1mm]

\scriptsize Rank 64
& \includegraphics[width=2.3cm]{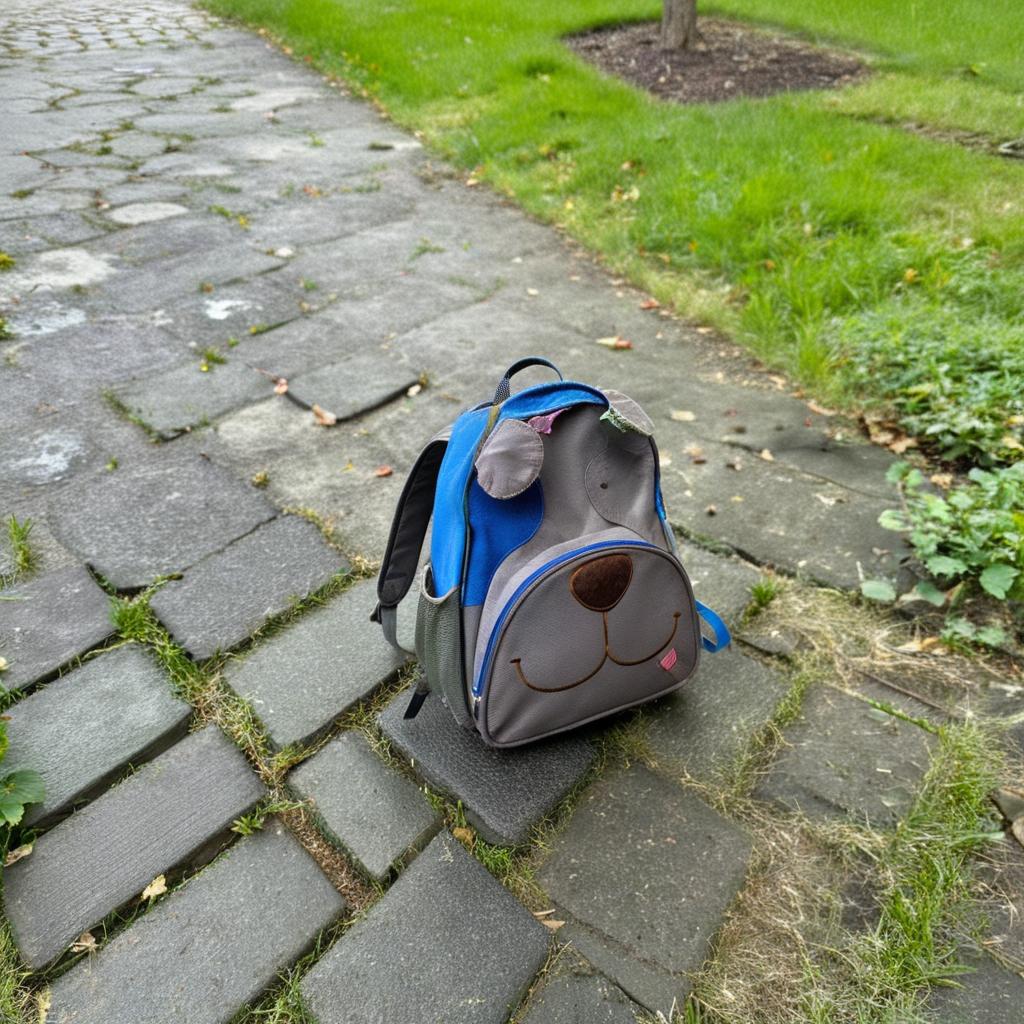}
& \includegraphics[width=2.3cm]{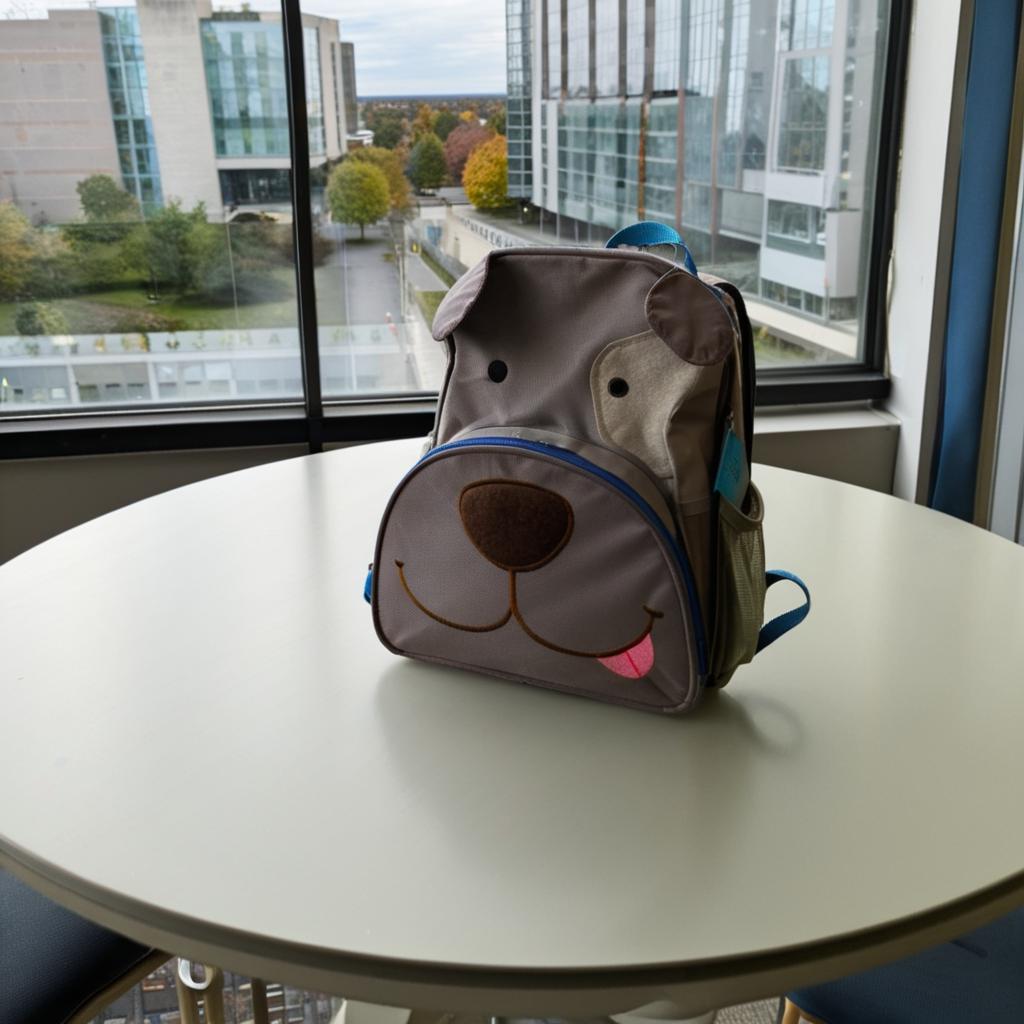}
& \includegraphics[width=2.3cm]{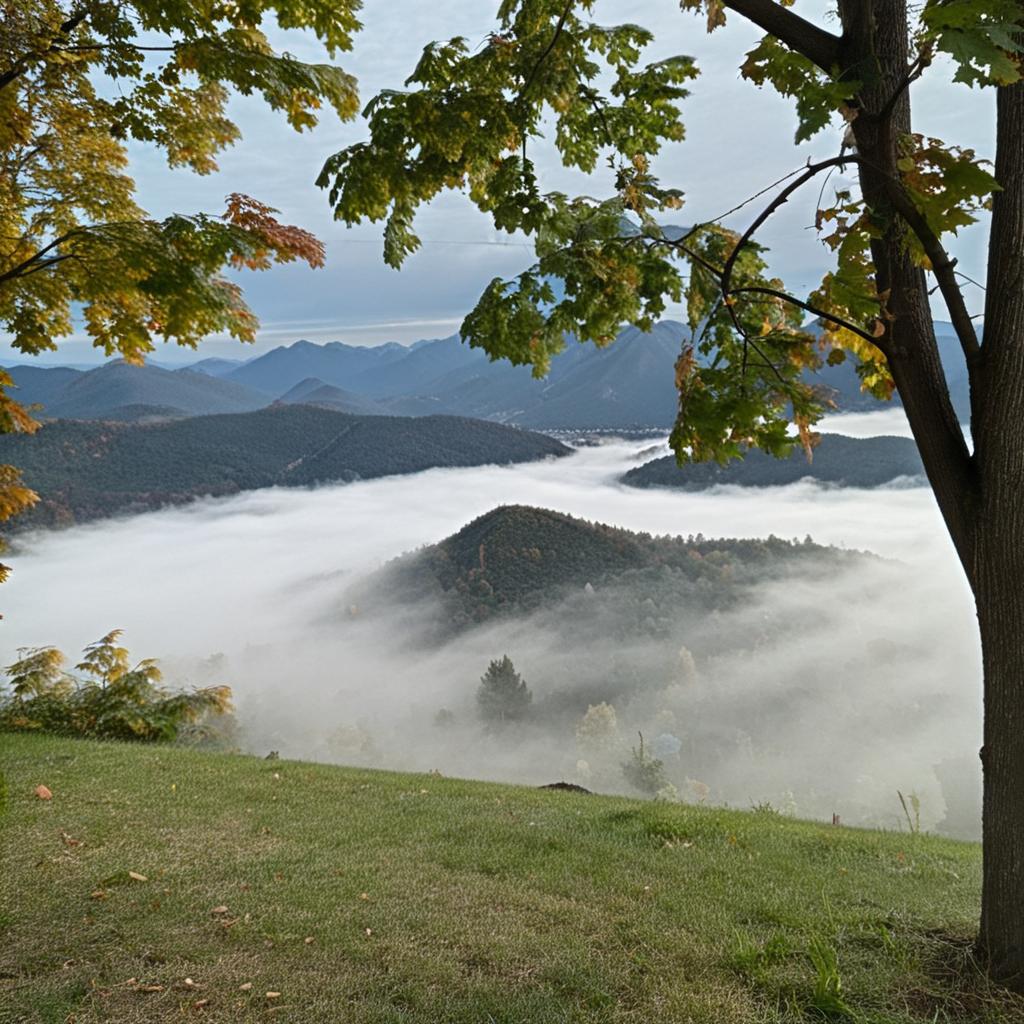}
& \includegraphics[width=2.3cm]{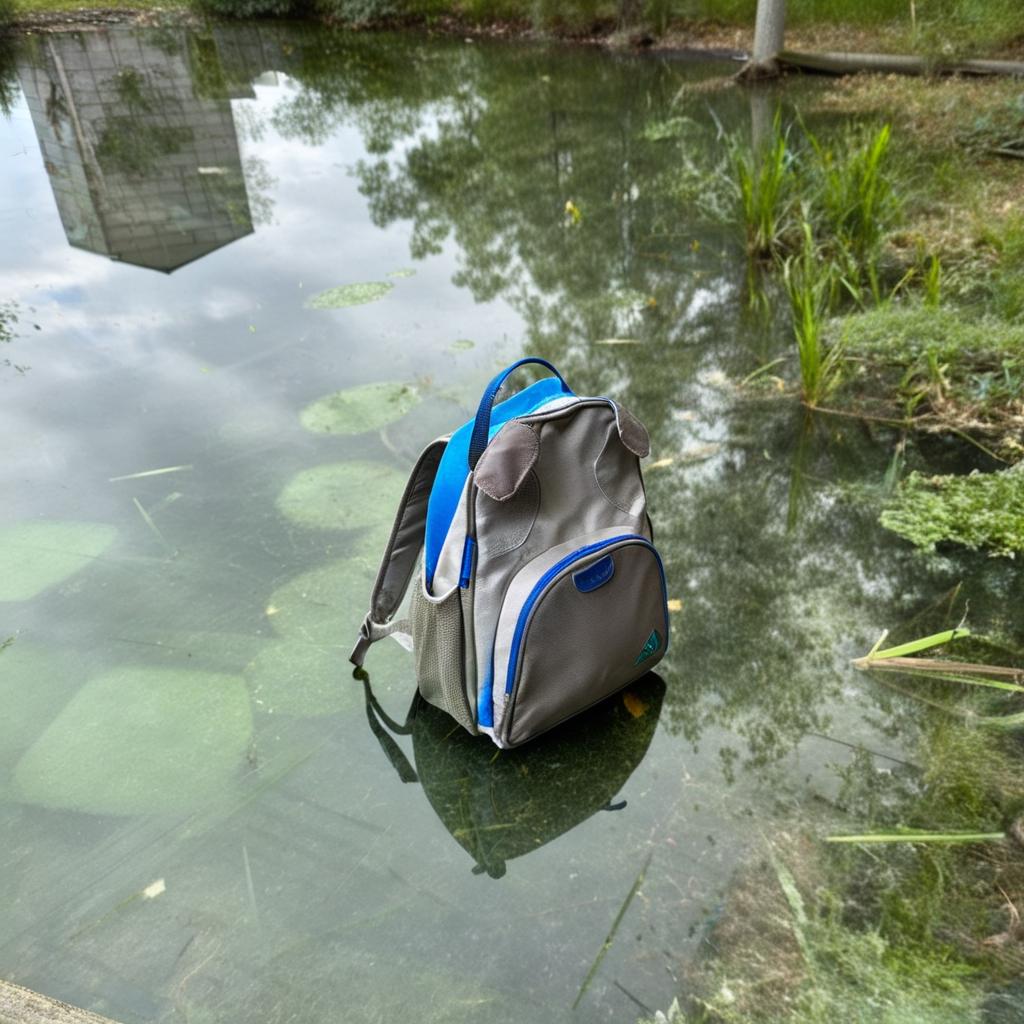}
& \includegraphics[width=2.3cm]{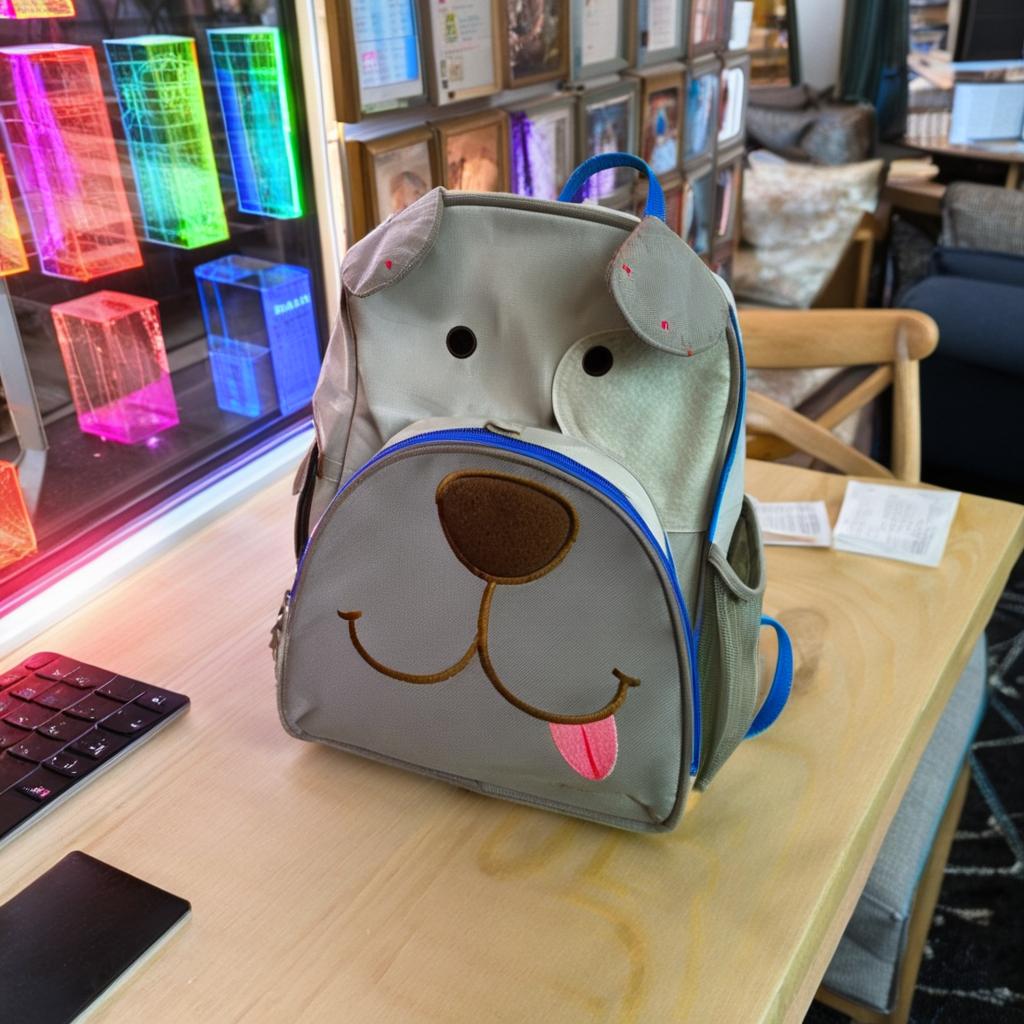}
\\[1mm]

\scriptsize Rank 512
& \includegraphics[width=2.3cm]{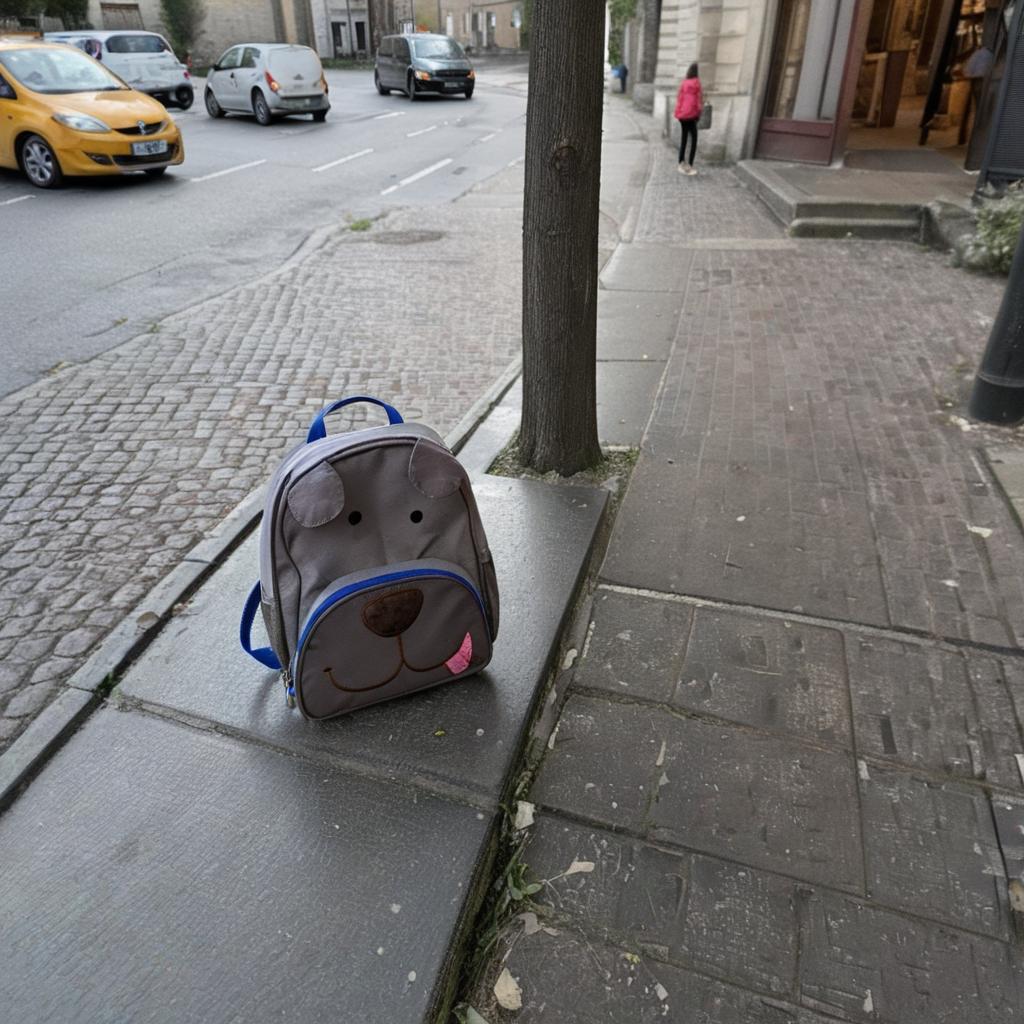}
& \includegraphics[width=2.3cm]{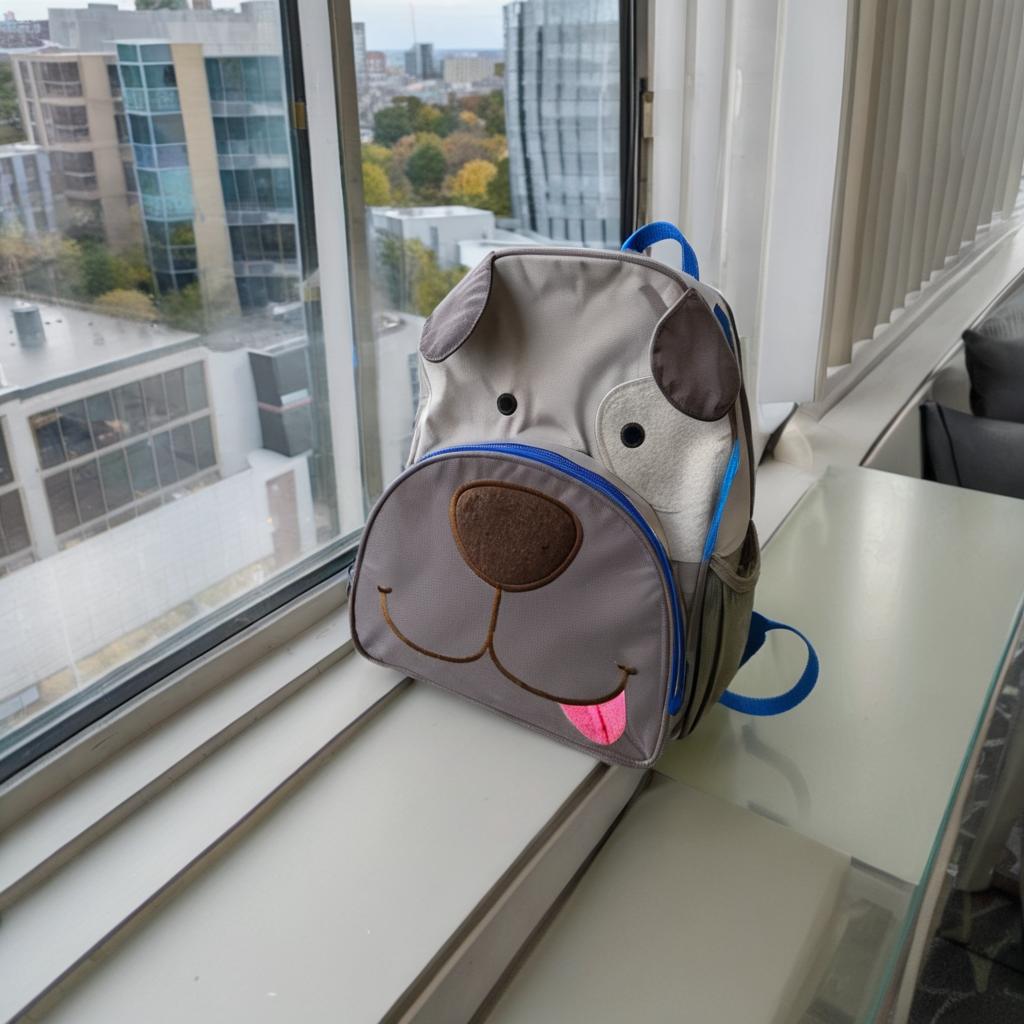}
& \includegraphics[width=2.3cm]{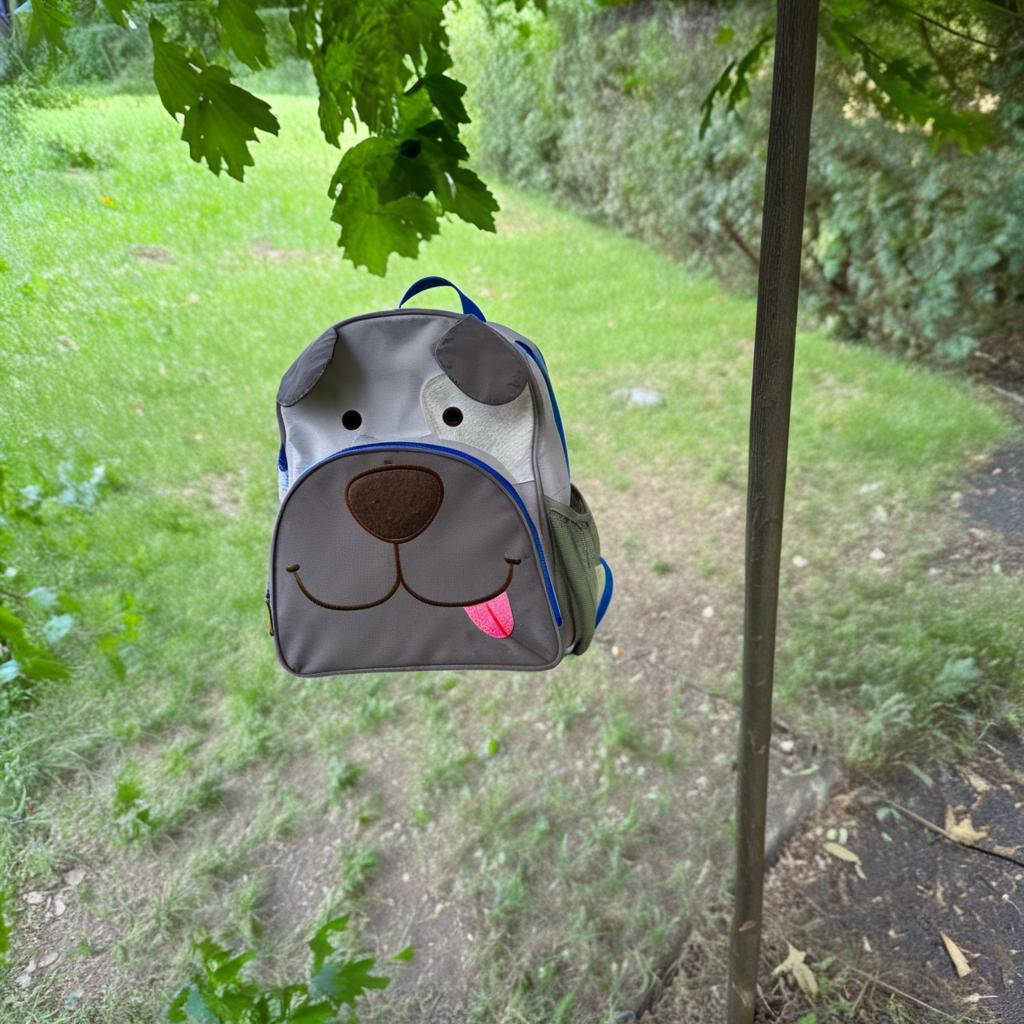}
& \includegraphics[width=2.3cm]{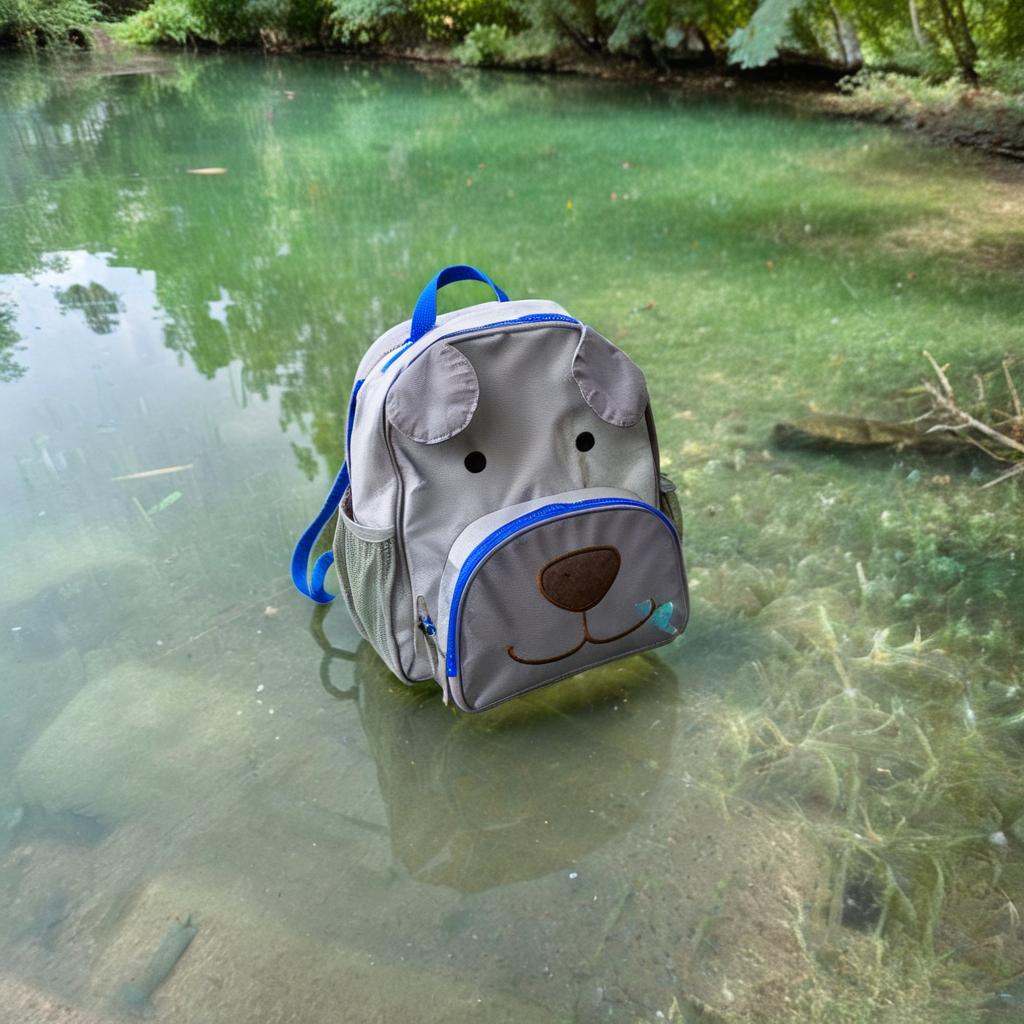}
& \includegraphics[width=2.3cm]{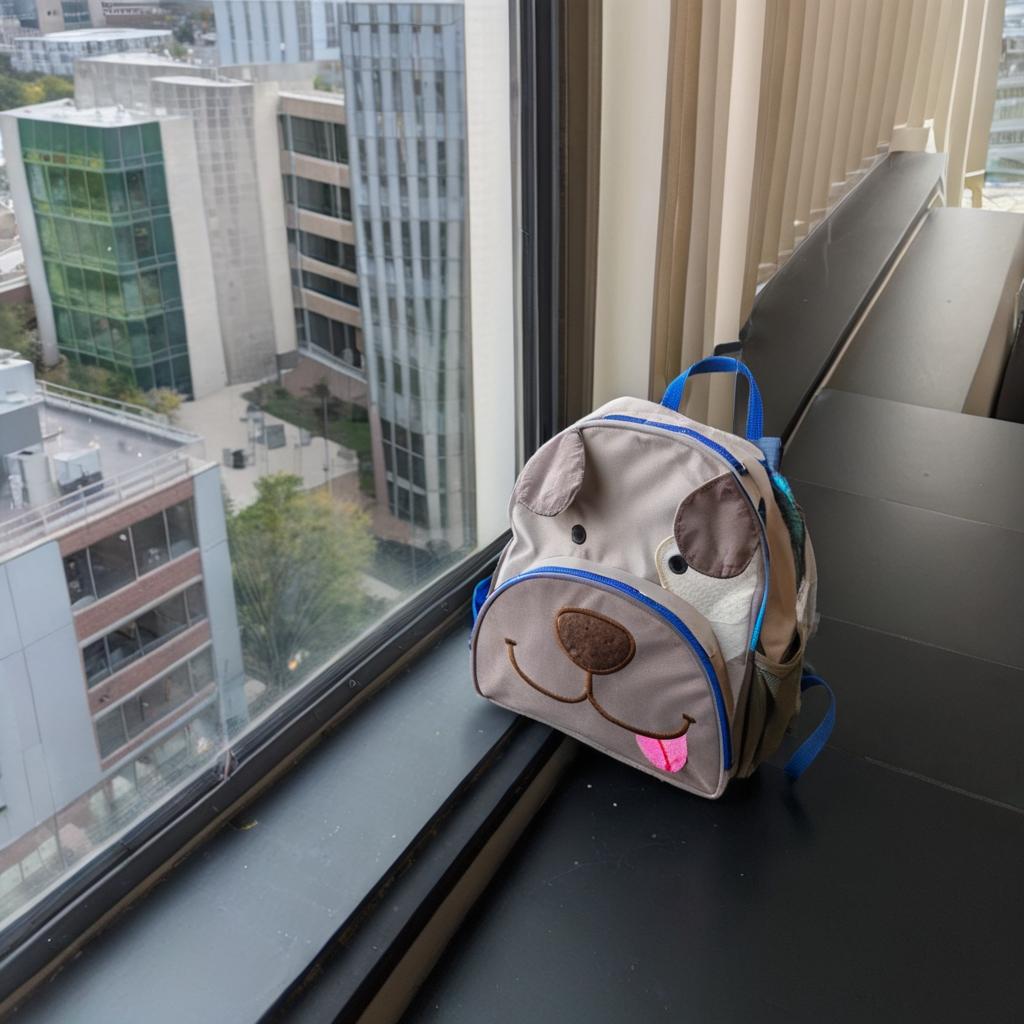}
\\[1mm]

\scriptsize \method
& \includegraphics[width=2.3cm]{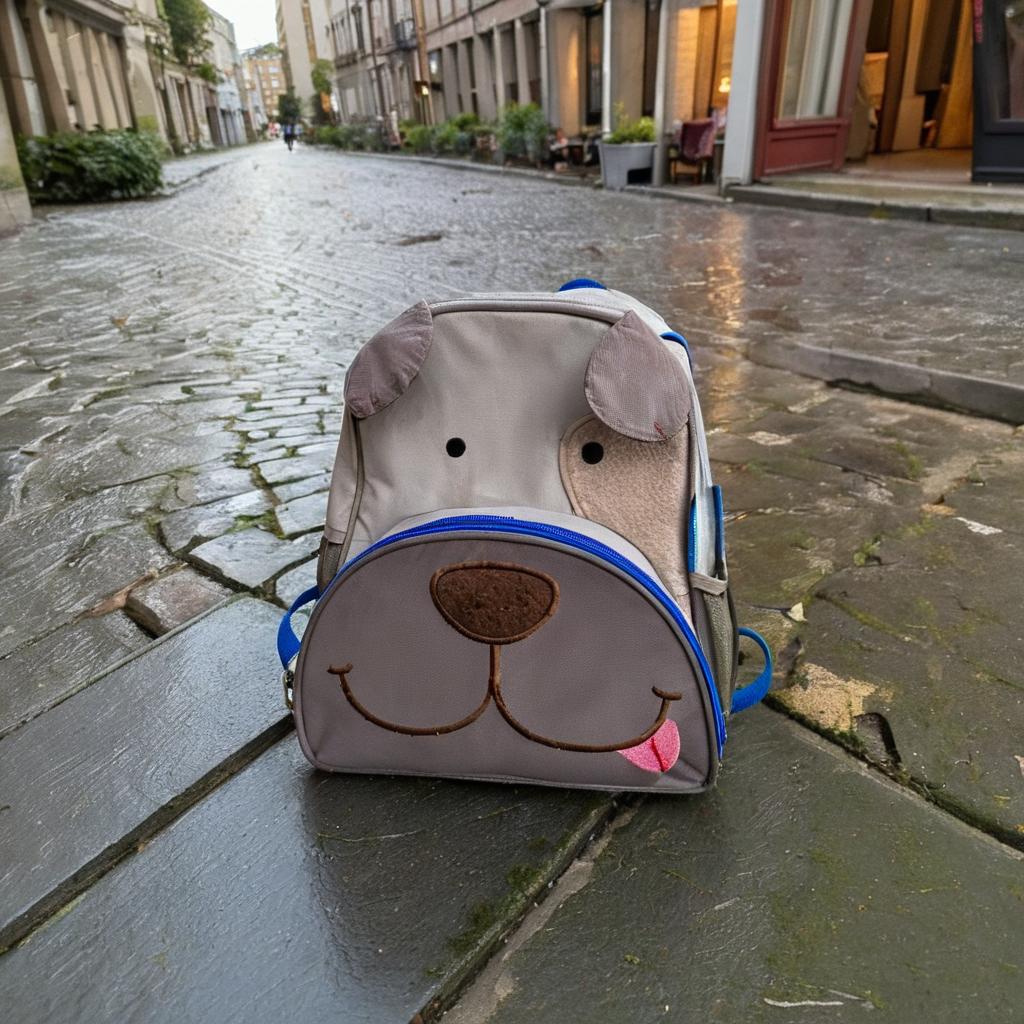}
& \includegraphics[width=2.3cm]{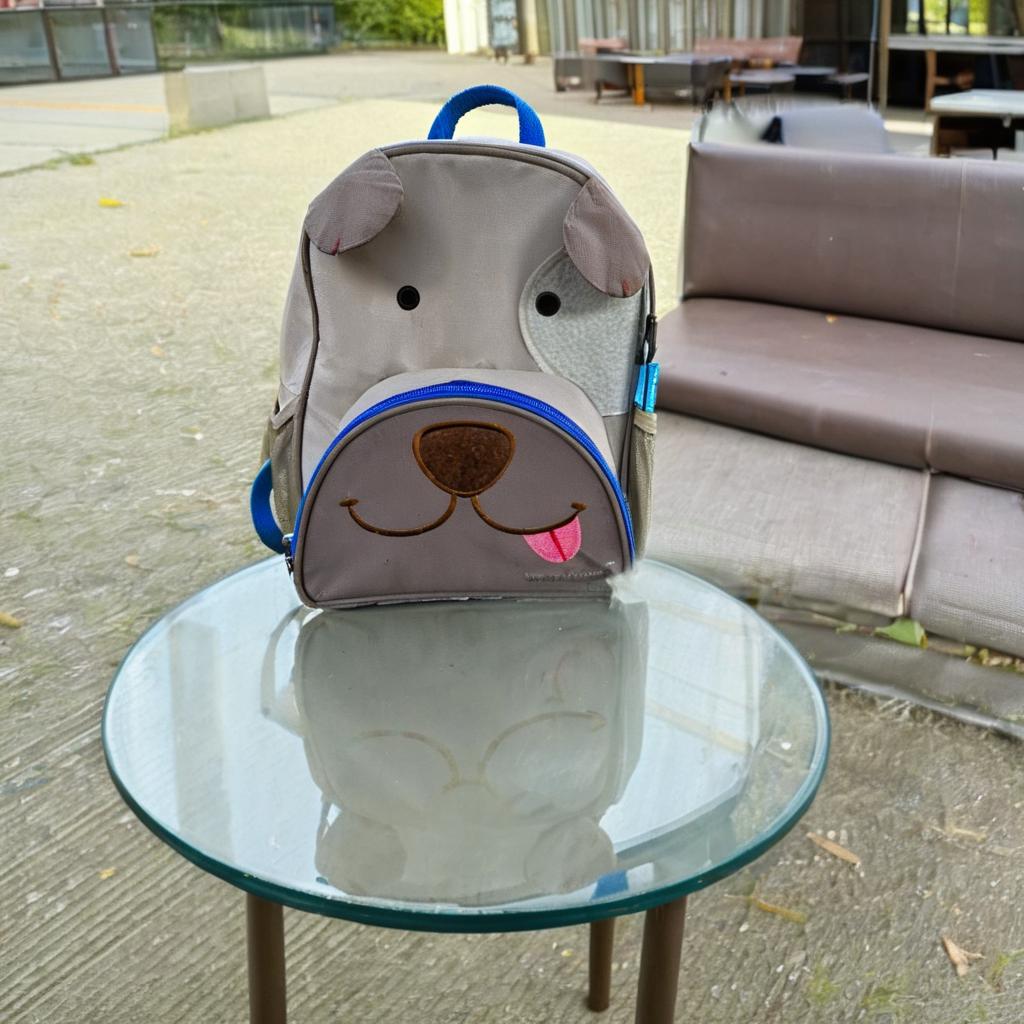}
& \includegraphics[width=2.3cm]{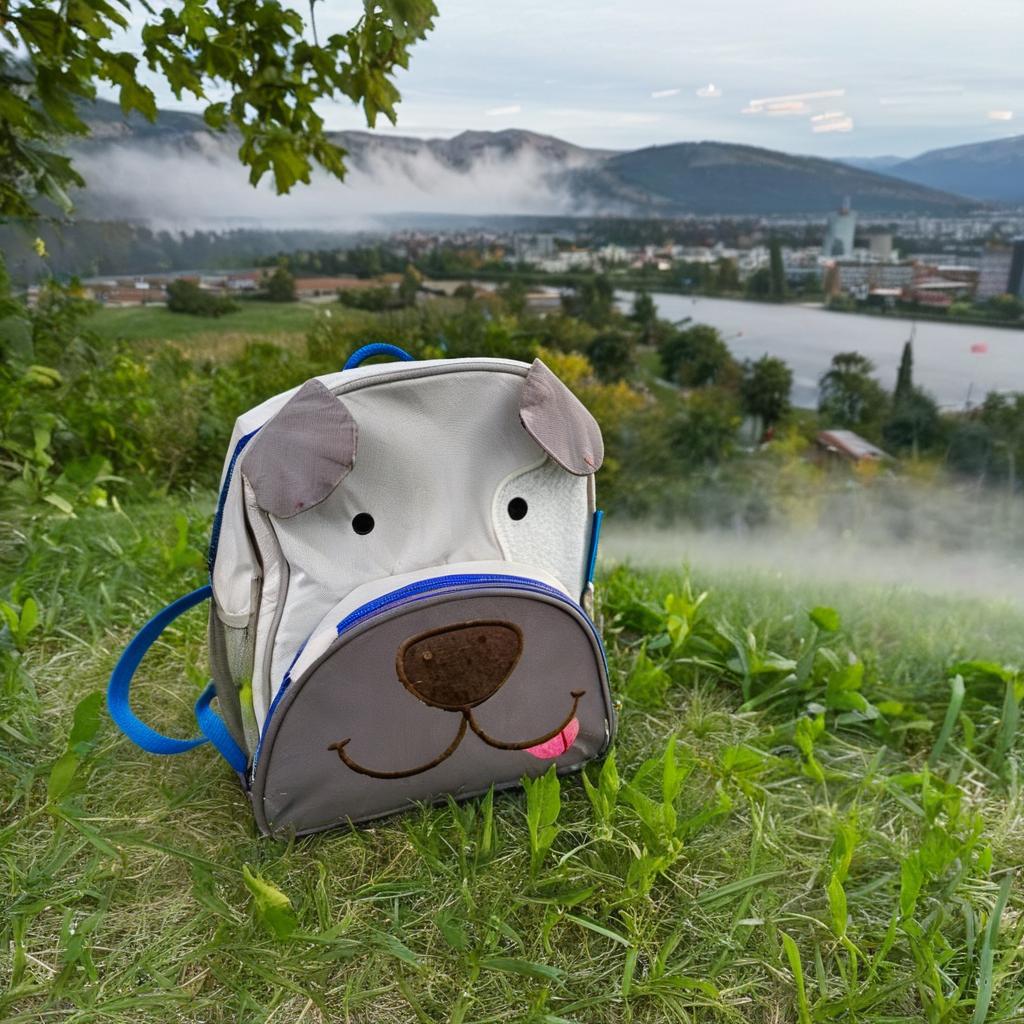}
& \includegraphics[width=2.3cm]{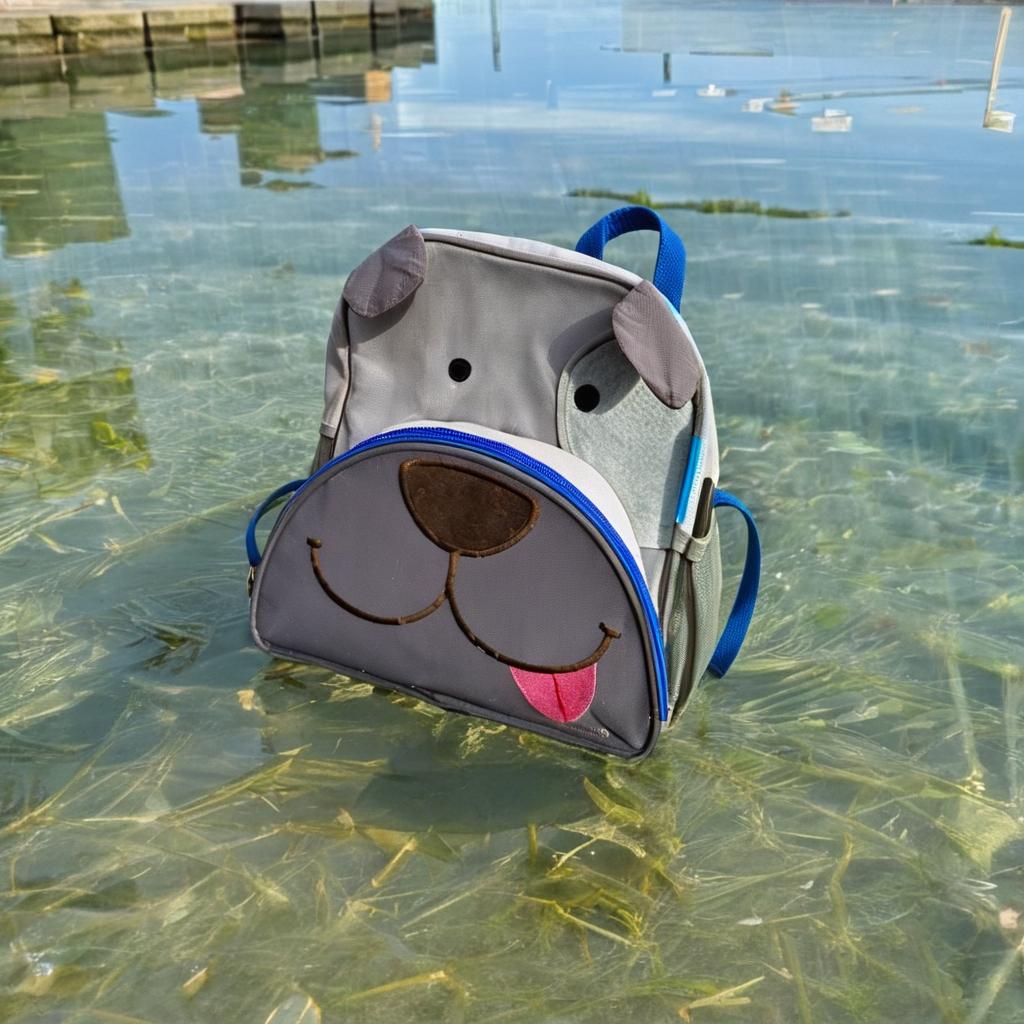}
& \includegraphics[width=2.3cm]{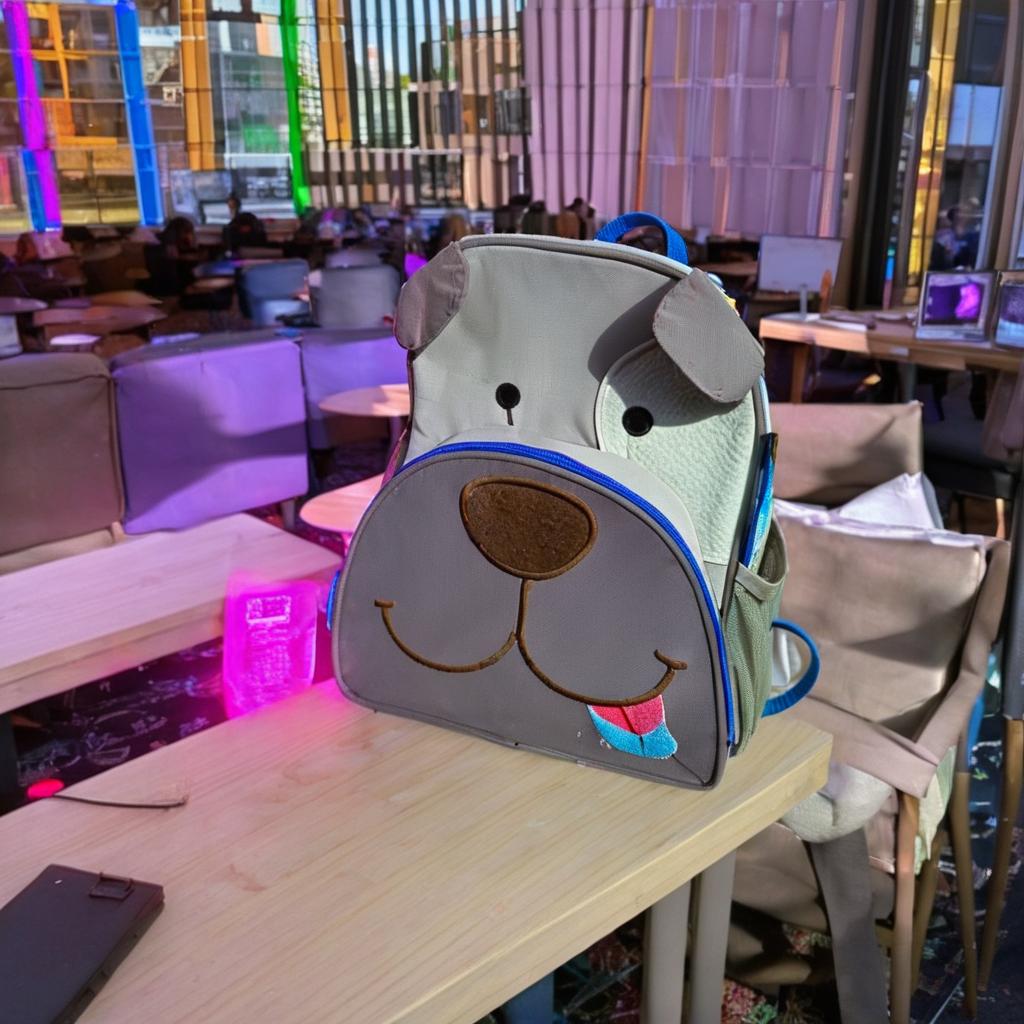}
\\
\end{tabular}%
}
\caption{Images generated using KOALA-700m backbone for the subject ``backpack dog". The original subject is present on the top left. 
}
\label{fig:qualitative_comparison_KOALA}
\end{figure}

Figure \ref{fig:qualitative_comparison_SDXL} and \ref{fig:qualitative_comparison_KOALA} show images generated with finetuned SDXL and KOALA-700m backbones, respectively. The generated images confirm that low ranks are unable to faithfully reproduce the subject: both the yellow clock and the backpack are often generated with the wrong color at ranks 8 and 64. At rank 512, LoRA finetuning struggles to follow the finer details of the prompt, such as ignoring the requested background. 
For the clock, rank 512 remains suboptimal for faithful reconstruction, with \method being the only approach to fully reproduce the content at high fidelity. Notably, the numeral ``3" on the clock face is preserved exclusively in our result; rank 512 fails to render it in both second and fifth prompts.
The same observation applies to the backpack: the patch eye on the right side is missing in the first and fourth prompts (and also the tongue). This suggests that subject fidelity does not necessarily improve with higher rank, likely because the model tends to overfit the background instead.
Per-class scores are provided in \cref{fig:per_class_SDXL}.
Finally, in some cases, the subject is not properly integrated with the background, exhibiting incorrect shadows or appearing to float above the ground. In contrast, images generated by \method remain consistent with both the subject and the prompt.

\subsection{Aggregated Results}
\begin{figure}[t]
    \centering
    \includegraphics[width=\linewidth]{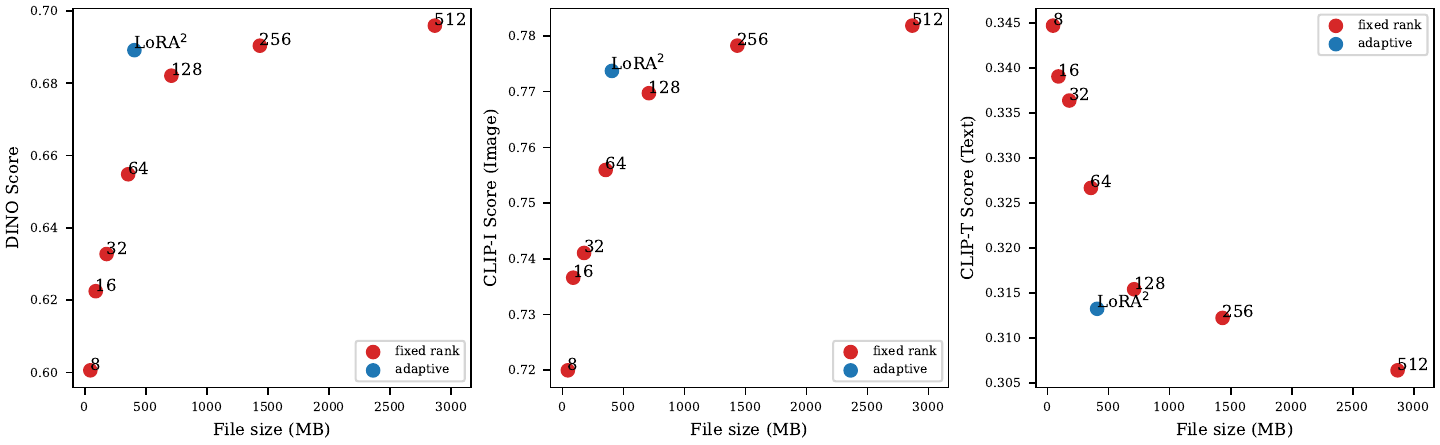}
    \caption{SDXL backbone. Aggregated results (average of all subjects). }
    \label{fig:all_metrics_scatter_SDXL}
\end{figure}

\begin{figure}[t]
    \centering
    \includegraphics[width=\linewidth]{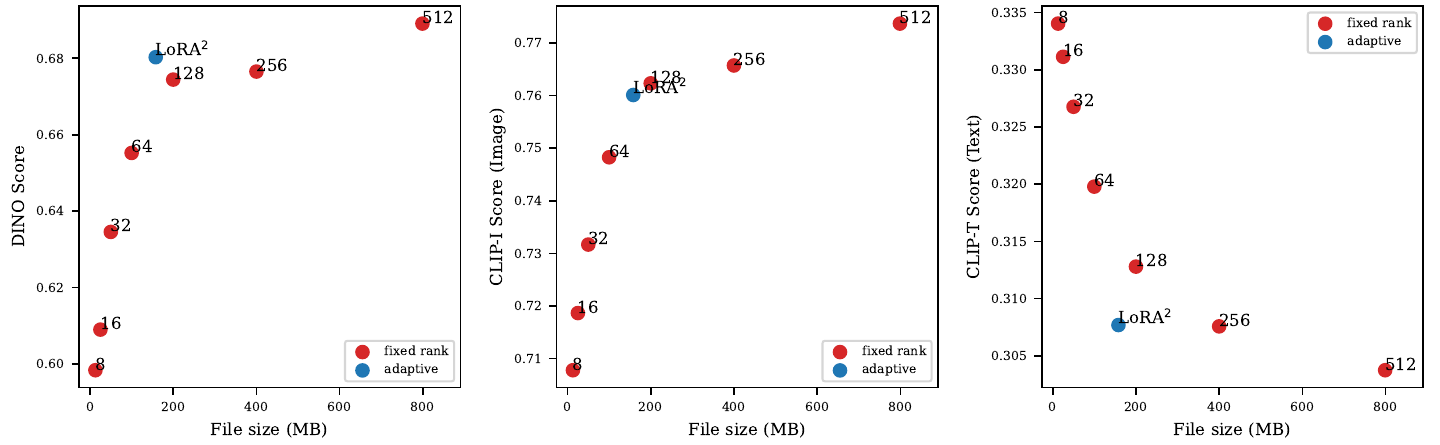}
    \caption{KOALA-700m backbone. Aggregated results (average of all subjects).}
    \label{fig:all_metrics_scatter_koala_700m}
\end{figure}

To quantitatively evaluate subject and prompt alignment in generated images, we use DINO, CLIP-I, and CLIP-T scores \cite{gal2022image,ruiz2023dreambooth}.
Figure \ref{fig:all_metrics_scatter_SDXL} and \ref{fig:all_metrics_scatter_koala_700m} report the average 
scores as a function of memory occupation for each trained model. Standard LoRA models exhibit a clear trend when trained with different ranks, where increasing the rank improves subject fidelity (higher DINO and CLIP-I) and decreases text alignment (lower CLIP-T). Low-rank models fail to consistently reproduce the target subject, frequently omitting distinctive attributes (e.g., incorrect colors or textures). High-rank models generate a stable and recognizable subject, but the surrounding scene and attributes increasingly deviate from the textual description. This indicates a tradeoff between subject consistency and text alignment as model capacity during finetuning grows, consistent with previous work~\cite{bidermanLoRALearnsLess2024}. \method achieves a more favorable tradeoff between these objectives.


\subsection{Per-Subject Performance}
To empirically support the need for adaptive ranks, we computed per-subject scores showing how there is no single rank that fits all. Figure \ref{fig:per_class_SDXL} shows per-subject scores for SDXL, while results on KOALA are in the supplementary material. We highlight with a grey band rank 64, the default value commonly used in previous works  \cite{shah2024ziplora,frenkel2024implicit, shenaj2025lora,soboleva2025tlorasingleimagediffusion,Roy_2025_ICCV, liu2025unziplora}. We also highlight in red the best value for each subject. First, we notice that rank 64 is never optimal in any of the metrics for SDXL. However, it achieves a good tradeoff considering subject alignment, text alignment, and model size. The best models on DINO and CLIP-I scores are either the high rank models or our \method. Instead, text alignment is consistently the best at lower ranks. Our \method has a model size comparable to the fixed rank 64. However, compared to the rank 64 baseline, our method achieves much higher DINO and CLIP-I scores, at the price of slightly lower CLIP-T. Instead, compared to the rank 512 model, \method has similar scores with a much lower memory occupation (0.40 GB for \method against 2.80 GB for rank 512). In conclusion, we observe that by using fixed ranks it is not possible to find an optimal solution for all the subjects, whereas \method{} provides better control by tuning the regularization hyper-parameters, which is more efficient than testing a huge number of configurations (as discussed in Section \ref{subsec:adaptive-lora}).

\vspace{-0.5cm}
\begin{figure}[!h]
    \centering
\includegraphics[width=\linewidth]{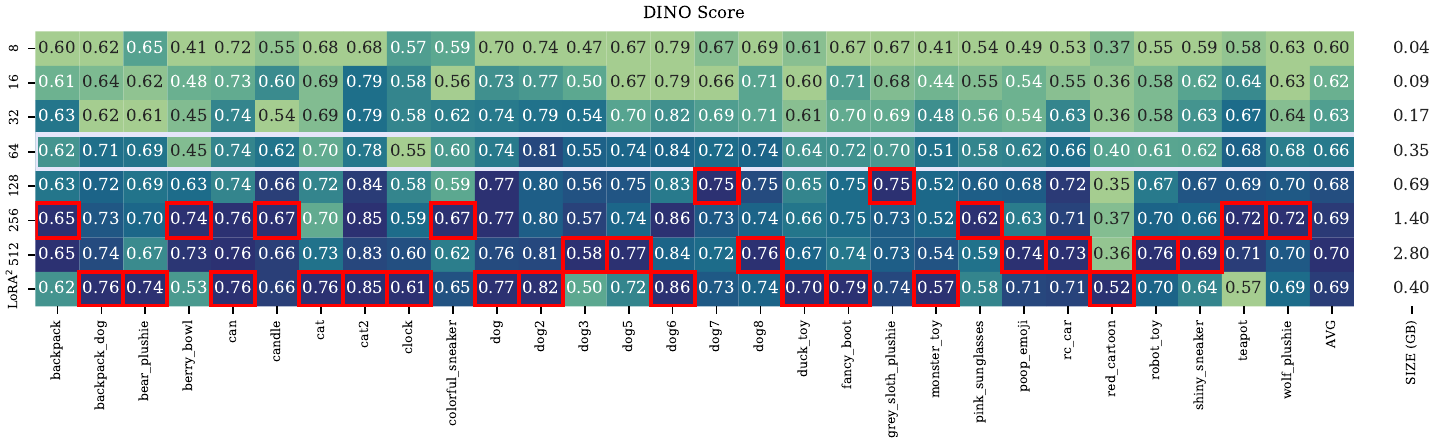}
    \includegraphics[width=\linewidth]{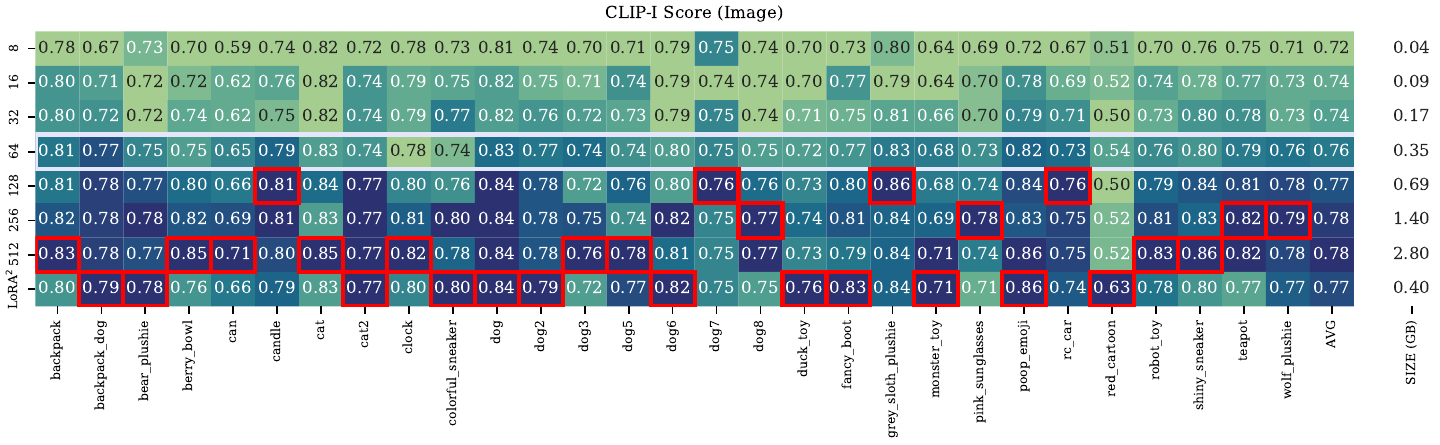}
    \includegraphics[width=\linewidth]{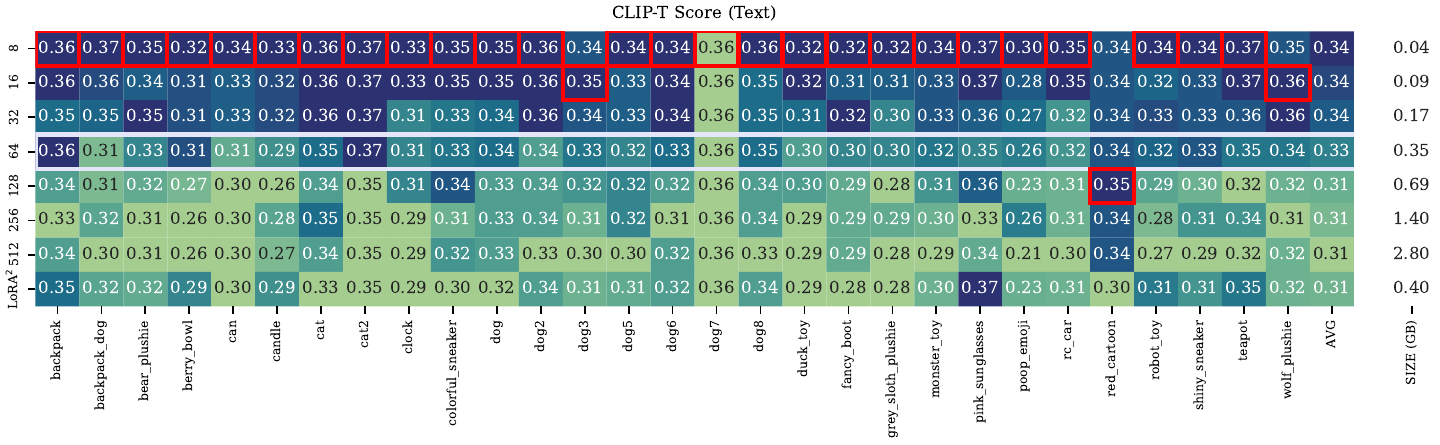}
    \caption{SDXL backbone, per-subject scores. We highlight with a grey band rank 64, the default value commonly used in previous work. We also highlight in red the best value for each subject. On the side, we also add the model size in GB. }
    \label{fig:per_class_SDXL}
\end{figure}

\subsection{LoRA Rank Analysis}
\begin{figure}[t]
    \centering
    \includegraphics[width=\linewidth]{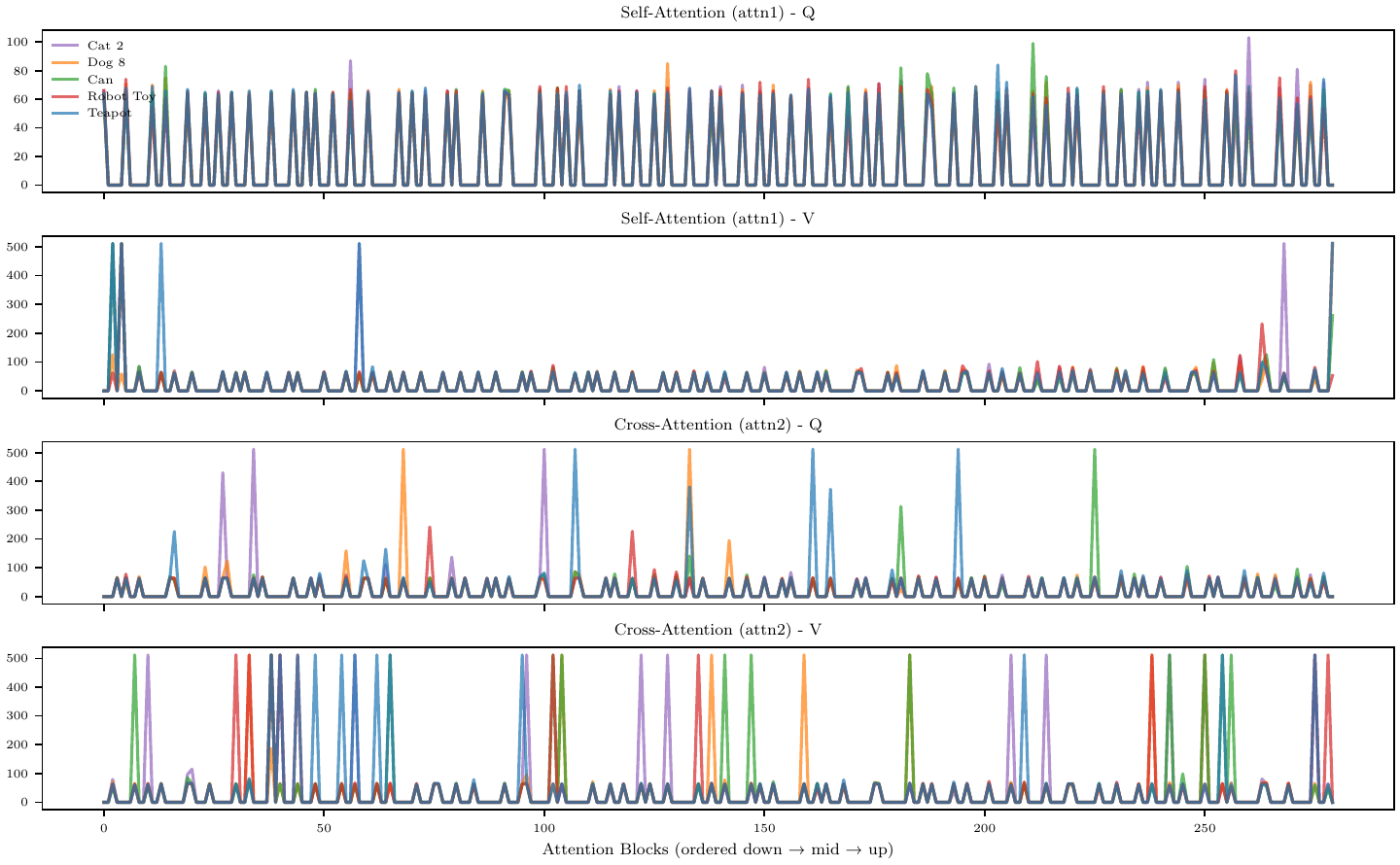}
    \caption{SDXL Self-Attention and Cross-Attention ranks, for five distinct subjects.}
    \label{fig:self_attn_ranks}
\end{figure}

One of the goals of \method is to allow the finetuning strategy to detect LoRA components that do not need adaptation, lowering their rank, and use higher capacity when necessary. To demonstrate that \method learns an ad-hoc solution for different subjects, Figure \ref{fig:self_attn_ranks} shows the ranks of self-attention and cross-attention layers (Query and Value matrices) for 5 randomly selected subjects: ``Cat 2", ``Dog 8", ``Can", ``Robot Toy", and ``Teapot".
While the figure shows the results for SDXL, and they are limited to the Query and Value matrices, we report full plots in the supplementary material. First, we notice that self-attention and cross-attention have different tendencies. Cross-attention has a higher prevalence of max rank (512) LoRAs, while self-attention layers tend to have lower ranks. A large number of components collapse to rank 1, confirming the ability of \method to save memory by reducing the rank of unnecessary components. We also notice that different subjects share most of the ranks, but they also have some differences, meaning \method adapts to different subjects though they might share some similarity. Overall, \method shows a high degree of diversity across layers and a moderate diversity across subjects and layer types, which is what we would expect from an adaptive rank method.

\vspace{-0.5cm}

\subsection{Ablation}
\begin{figure}[!h]
    \centering
    \includegraphics[width=\linewidth]{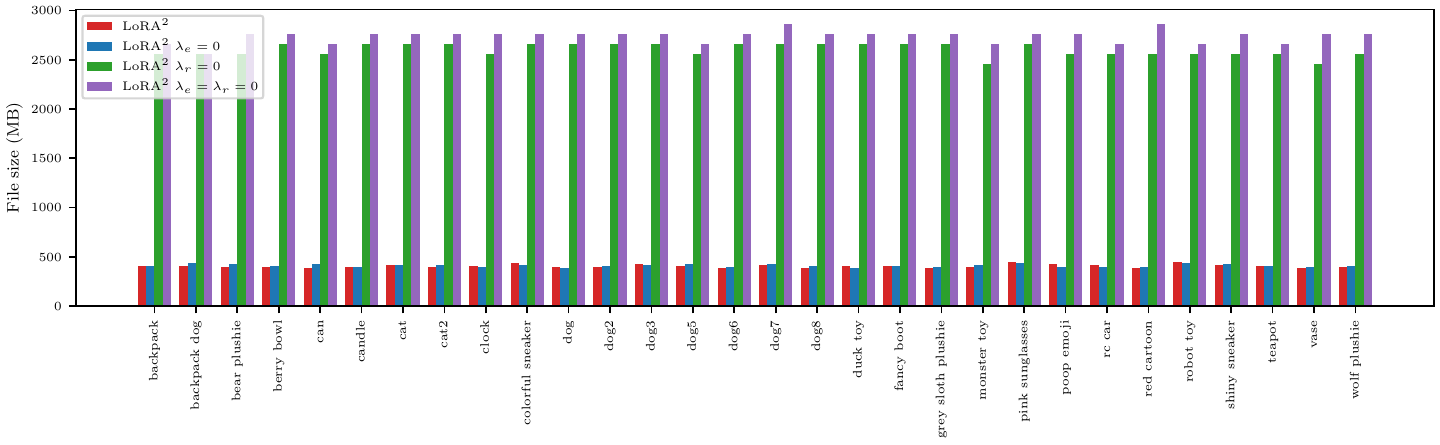}
    \caption{File size of \method for SDXL.}
    \label{fig:filesize}
\end{figure}

The MSE loss is a good proxy for subject fidelity (DINO and CLIP-I scores). Therefore, \method uses a regularization loss on the ranks and an additional entropy loss to better control the subject-text-memory tradeoff. Figure \ref{fig:filesize} shows the file size of different configurations of \method for each subject, while Table \ref{tab:ablation} shows the aggregated file size and image scores. Removing the rank regularization increases the file size from an average of 406 MB to 2.7 GB. This is a consequence of the MSE loss and its strong bias towards better subject fidelity. As a result, the resulting model obtains marginally better DINO and CLIP-I scores. Removing the entropy loss while keeping the rank regularization results in a similar file size compared to the full \method. However, the model trained with entropy regularization has a higher CLIP-T. The full \method with both regularization losses is needed to obtain a good tradeoff between subject fidelity, textual alignment, and model size.

\vspace{-1cm}
\begin{table}[!h]
    \centering
        \caption{Ablation of regularization losses, the scores are averaged across all subjects.}
    \label{tab:ablation}
    \begin{tabular}{c l c c c c}
    \hline
        Backbone & Method & DINO & CLIP-I & CLIP-T & File size \\
        \hline
       \multirow{3}*{SDXL} & \method & 0.689  & 0.773 & 0.313 & 406 MB  \\
                           & \method  ($\lambda_e=0$) & 0.696 & 0.780 & 0.303 & 410 MB\\
                           & \method  ($\lambda_r=\lambda_e=0$) & 0.699  & 0.782 & 0.299 & 2.7 GB\\
         \hline
       \multirow{3}*{KOALA-700m} & \method &  0.680 & 0.760 &  0.308 & 158 MB\\
                           & \method  ($\lambda_e=0$) & 0.681 & 0.762 & 0.304  & 160 MB\\
                           & \method  ($\lambda_r=\lambda_e=0$) & 0.693  & 0.768  & 0.302  & 734 MB \\
         \hline
    \end{tabular}

\end{table}
\vspace{-1cm}

\section{Conclusions}
\label{sec:conclusions}
We introduced \method, an easy-to-implement, fully differentiable, and model-agnostic modification of LoRA to learn a proper rank for each LoRA component in deep learning models for personalized image generation. \method{} encourages an ordering of importance across rank indices, allowing us to dynamically introduce or reduce the rank of each LoRA component depending on the specific subject at hand. 
Thanks to this approach, we do not need to manually select the rank for each LoRA component, which would have a combinatorial cost, nor to fix the same rank for all components, which we empirically show is not the best strategy. Across 29 subjects, \method achieves a very good trade-off between DINO, CLIP-I and CLIP-T scores while requiring lower memory consumption. In the future, we will investigate the role of adaptive rank learning in the multi-subject and model-merging settings and its performance on larger diffusion models.

\section*{Acknowledgments}
This paper has been partially supported by the CoEvolution project, funded by EU Horizon 2020 under GA n 101168559. We acknowledge ISCRA for awarding this project access to the LEONARDO supercomputer, owned by the EuroHPC Joint Undertaking, hosted by CINECA (Italy).

\bibliographystyle{splncs04}
\bibliography{references}

\clearpage

\renewcommand{\thefigure}{A\arabic{figure}}
\renewcommand{\theHfigure}{A\arabic{figure}}
\renewcommand{\thesection}{A\arabic{section}}
\renewcommand{\theHsection}{A\arabic{section}}
\renewcommand{\theequation}{A\arabic{equation}}
\renewcommand{\theHequation}{A\arabic{equation}}
\renewcommand{\thetable}{A\arabic{table}}
\renewcommand{\theHtable}{A\arabic{table}}

\setcounter{equation}{0}
\setcounter{figure}{0}
\setcounter{table}{0}
\setcounter{section}{0}

\title{Not All Layers Are Created Equal: Adaptive LoRA Ranks for Personalized Image Generation \\ 
\textit{Supplementary Material}}
\titlerunning{Supplementary Material}
\authorrunning{D. Shenaj, F. Errica, A. Carta}

\author{}
\institute{}

\maketitle
\section{Additional Implementation Details}

All models were trained with a resolution of $1024 \times 1024$, a batch size of 1, and a learning rate of $5 \times 10^{-5}$. We used mixed precision training (\texttt{fp16}), gradient checkpointing, and 8-bit Adam optimization. Experiments were conducted on NVIDIA Ampere A100 GPUs (64GB RAM).

\section{Prompts}

\begin{longtable}{>{\bfseries\small}p{3.5cm} >{\small}p{8cm}}
\caption{Full prompts used for evaluation}\label{tab:prompts}\\
\rowcolor{headerblue}
\color{white}\textbf{Subject} & \color{white}\textbf{Prompt} \\
\endfirsthead
\rowcolor{headerblue}
\color{white}\textbf{Subject} & \color{white}\textbf{Prompt} \\
\endhead

\rowcolor{rowgray} backpack & a <c> backpack on a wooden shelf surrounded by books \\
  & a modern minimalistic <c> backpack on a white surface \\
\rowcolor{rowgray}  & a <c> backpack in the snow under warm sunlight \\
  & a <c> backpack on a cobblestone street after rain \\
\rowcolor{rowgray}  & a vintage <c> backpack on an antique table \\
  & a <c> backpack placed on pink silk fabric \\
\rowcolor{rowgray}  & a <c> backpack on a mossy rock in a forest \\
  & a glowing <c> backpack in the dark \\
\rowcolor{rowgray}  & a <c> backpack on a glass table with reflections \\
  & a <c> backpack on a sandy beach at sunset \\
\midrule
\rowcolor{rowgray} backpack\_dog & a <c> backpack on a cobblestone street after rain \\
  & a <c> backpack with a city skyline in the background \\
\rowcolor{rowgray}  & a <c> backpack in the snow under warm sunlight \\
  & a <c> backpack surrounded by neon lights \\
\rowcolor{rowgray}  & a vintage <c> backpack on an antique table \\
  & a <c> backpack on a glass table with reflections \\
\rowcolor{rowgray}  & a <c> backpack on a wooden shelf surrounded by books \\
  & a <c> backpack with mountains and mist in the background \\
\rowcolor{rowgray}  & a <c> backpack floating in crystal clear water \\
  & a <c> backpack placed on pink silk fabric \\
\midrule
\rowcolor{rowgray} bear\_plushie & a <c> stuffed animal in the jungle \\
  & a wet <c> stuffed animal \\
\rowcolor{rowgray}  & a <c> stuffed animal in the snow \\
  & a <c> stuffed animal in a chef outfit \\
\rowcolor{rowgray}  & a <c> stuffed animal in a police uniform \\
  & a <c> stuffed animal wearing a rainbow scarf \\
\rowcolor{rowgray}  & a <c> stuffed animal in a city park surrounded by flowers \\
  & a <c> stuffed animal wearing a black top hat and a monocle \\
\rowcolor{rowgray}  & a <c> stuffed animal in a forest clearing with sunlight rays \\
  & a <c> stuffed animal with the Eiffel Tower in the background \\
\midrule
\rowcolor{rowgray} berry\_bowl & a <c> bowl in the snow under warm sunlight \\
  & a <c> bowl on a cobblestone street after rain \\
\rowcolor{rowgray}  & a vintage <c> bowl on an antique table \\
  & a <c> bowl with a city skyline in the background \\
\rowcolor{rowgray}  & a modern minimalistic <c> bowl on a white surface \\
  & a <c> bowl on a glass table with reflections \\
\rowcolor{rowgray}  & a <c> bowl in a minimalist art gallery \\
  & a <c> bowl on a sandy beach at sunset \\
\rowcolor{rowgray}  & a glowing <c> bowl in the dark \\
  & a <c> bowl floating in crystal clear water \\
\midrule
\rowcolor{rowgray} can & a glowing <c> can in the dark \\
  & a <c> can on a mossy rock in a forest \\
\rowcolor{rowgray}  & a <c> can with mountains and mist in the background \\
  & a <c> can on a wooden shelf surrounded by books \\
\rowcolor{rowgray}  & a <c> can placed on pink silk fabric \\
  & a <c> can on a sandy beach at sunset \\
\rowcolor{rowgray}  & a vintage <c> can on an antique table \\
  & a <c> can in the snow under warm sunlight \\
\rowcolor{rowgray}  & a modern minimalistic <c> can on a white surface \\
  & a <c> can on a marble table, studio lighting \\
\midrule
\rowcolor{rowgray} candle & a <c> candle on a cobblestone street after rain \\
  & a <c> candle on a sandy beach at sunset \\
\rowcolor{rowgray}  & a <c> candle on a reflective mirror surface \\
  & a <c> candle placed on pink silk fabric \\
\rowcolor{rowgray}  & a <c> candle with a city skyline in the background \\
  & a <c> candle in a minimalist art gallery \\
\rowcolor{rowgray}  & a <c> candle in the snow under warm sunlight \\
  & a <c> candle next to a cup of coffee on a kitchen counter \\
\rowcolor{rowgray}  & a glowing <c> candle in the dark \\
  & a <c> candle on a wooden shelf surrounded by books \\
\midrule
\rowcolor{rowgray} cat & a <c> cat in a forest clearing with sunlight rays \\
  & a <c> cat in a police uniform \\
\rowcolor{rowgray}  & a <c> cat in the jungle \\
  & a <c> cat in a chef outfit \\
\rowcolor{rowgray}  & a <c> cat on the beach during sunset \\
  & a <c> cat in the snow \\
\rowcolor{rowgray}  & a <c> cat wearing a rainbow scarf \\
  & a <c> cat driving a tiny car \\
\rowcolor{rowgray}  & a shiny <c> cat \\
  & a <c> cat with the Eiffel Tower in the background \\
\midrule
\rowcolor{rowgray} cat2 & a <c> cat with the Eiffel Tower in the background \\
  & a <c> cat in the jungle \\
\rowcolor{rowgray}  & a shiny <c> cat \\
  & a <c> cat on the beach during sunset \\
\rowcolor{rowgray}  & a <c> cat in a chef outfit \\
  & a <c> cat in a city park surrounded by flowers \\
\rowcolor{rowgray}  & a <c> cat floating in outer space \\
  & a <c> cat wearing a rainbow scarf \\
\rowcolor{rowgray}  & a <c> cat in a forest clearing with sunlight rays \\
  & a <c> cat sitting on a red couch indoors \\
\midrule
\rowcolor{rowgray} clock & a <c> clock surrounded by neon lights \\
  & a <c> clock next to a cup of coffee on a kitchen counter \\
\rowcolor{rowgray}  & a <c> clock on a reflective mirror surface \\
  & a <c> clock on a glass table with reflections \\
\rowcolor{rowgray}  & a <c> clock on a cobblestone street after rain \\
  & a <c> clock placed on pink silk fabric \\
\rowcolor{rowgray}  & a <c> clock on a marble table, studio lighting \\
  & a <c> clock on a mossy rock in a forest \\
\rowcolor{rowgray}  & a <c> clock with a city skyline in the background \\
  & a <c> clock in the snow under warm sunlight \\
\midrule
\rowcolor{rowgray} colorful\_sneaker & a <c> sneaker with a city skyline in the background \\
  & a <c> sneaker placed on pink silk fabric \\
\rowcolor{rowgray}  & a <c> sneaker on a glass table with reflections \\
  & a <c> sneaker surrounded by neon lights \\
\rowcolor{rowgray}  & a <c> sneaker in a minimalist art gallery \\
  & a <c> sneaker on a marble table, studio lighting \\
\rowcolor{rowgray}  & a modern minimalistic <c> sneaker on a white surface \\
  & a <c> sneaker on a reflective mirror surface \\
\rowcolor{rowgray}  & a <c> sneaker on a mossy rock in a forest \\
  & a <c> sneaker with mountains and mist in the background \\
\midrule
\rowcolor{rowgray} dog & a <c> dog with mountains in the background \\
  & a cube-shaped <c> dog \\
\rowcolor{rowgray}  & a <c> dog wearing a black top hat and a monocle \\
  & a <c> dog in a chef outfit \\
\rowcolor{rowgray}  & a <c> dog in the jungle \\
  & a <c> dog in a city park surrounded by flowers \\
\rowcolor{rowgray}  & a <c> dog floating in outer space \\
  & a <c> dog with the Eiffel Tower in the background \\
\rowcolor{rowgray}  & a <c> dog wearing sunglasses \\
  & a wet <c> dog \\
\midrule
\rowcolor{rowgray} dog2 & a <c> dog in the snow \\
  & a <c> dog wearing a black top hat and a monocle \\
\rowcolor{rowgray}  & a <c> dog in a chef outfit \\
  & a <c> dog sitting on a red couch indoors \\
\rowcolor{rowgray}  & a <c> dog in a forest clearing with sunlight rays \\
  & a <c> dog in a city park surrounded by flowers \\
\rowcolor{rowgray}  & a <c> dog on the beach during sunset \\
  & a <c> dog with the Eiffel Tower in the background \\
\rowcolor{rowgray}  & a <c> dog floating in outer space \\
  & a <c> dog driving a tiny car \\
\midrule
\rowcolor{rowgray} dog3 & a cube-shaped <c> dog \\
  & a <c> dog in the jungle \\
\rowcolor{rowgray}  & a <c> dog in a wizard robe holding a staff \\
  & a <c> dog wearing a rainbow scarf \\
\rowcolor{rowgray}  & a <c> dog wearing sunglasses \\
  & a <c> dog in a police uniform \\
\rowcolor{rowgray}  & a <c> dog in the snow \\
  & a <c> dog sitting on a red couch indoors \\
\rowcolor{rowgray}  & a <c> dog in a forest clearing with sunlight rays \\
  & a <c> dog in a chef outfit \\
\midrule
\rowcolor{rowgray} dog5 & a <c> dog wearing a red hat \\
  & a shiny <c> dog \\
\rowcolor{rowgray}  & a <c> dog wearing a black top hat and a monocle \\
  & a <c> dog in a chef outfit \\
\rowcolor{rowgray}  & a <c> dog floating in outer space \\
  & a <c> dog with mountains in the background \\
\rowcolor{rowgray}  & a <c> dog in a forest clearing with sunlight rays \\
  & a wet <c> dog \\
\rowcolor{rowgray}  & a <c> dog in a wizard robe holding a staff \\
  & a <c> dog in the snow \\
\midrule
\rowcolor{rowgray} dog6 & a wet <c> dog \\
  & a shiny <c> dog \\
\rowcolor{rowgray}  & a <c> dog driving a tiny car \\
  & a <c> dog wearing a red hat \\
\rowcolor{rowgray}  & a <c> dog with mountains in the background \\
  & a <c> dog in a forest clearing with sunlight rays \\
\rowcolor{rowgray}  & a <c> dog in the jungle \\
  & a <c> dog in a police uniform \\
\rowcolor{rowgray}  & a cube-shaped <c> dog \\
  & a <c> dog floating in outer space \\
\midrule
\rowcolor{rowgray} dog7 & a <c> dog in the snow \\
  & a <c> dog wearing a black top hat and a monocle \\
\rowcolor{rowgray}  & a <c> dog in a chef outfit \\
  & a <c> dog wearing a red hat \\
\rowcolor{rowgray}  & a <c> dog on the beach during sunset \\
  & a <c> dog wearing a rainbow scarf \\
\rowcolor{rowgray}  & a <c> dog with the Eiffel Tower in the background \\
  & a <c> dog in the jungle \\
\rowcolor{rowgray}  & a <c> dog wearing sunglasses \\
  & a <c> dog in a forest clearing with sunlight rays \\
\midrule
\rowcolor{rowgray} dog8 & a shiny <c> dog \\
  & a <c> dog in a city park surrounded by flowers \\
\rowcolor{rowgray}  & a <c> dog in a wizard robe holding a staff \\
  & a <c> dog wearing sunglasses \\
\rowcolor{rowgray}  & a <c> dog wearing a red hat \\
  & a <c> dog in a forest clearing with sunlight rays \\
\rowcolor{rowgray}  & a <c> dog wearing a black top hat and a monocle \\
  & a wet <c> dog \\
\rowcolor{rowgray}  & a <c> dog on the beach during sunset \\
  & a <c> dog floating in outer space \\
\midrule
\rowcolor{rowgray} duck\_toy & a <c> toy sitting on a red couch indoors \\
  & a <c> toy on the beach during sunset \\
\rowcolor{rowgray}  & a <c> toy in a police uniform \\
  & a <c> toy with mountains in the background \\
\rowcolor{rowgray}  & a <c> toy floating in outer space \\
  & a <c> toy wearing a red hat \\
\rowcolor{rowgray}  & a shiny <c> toy \\
  & a <c> toy in a forest clearing with sunlight rays \\
\rowcolor{rowgray}  & a <c> toy wearing a black top hat and a monocle \\
  & a wet <c> toy \\
\midrule
\rowcolor{rowgray} fancy\_boot & a <c> boot floating in crystal clear water \\
  & a <c> boot with a city skyline in the background \\
\rowcolor{rowgray}  & a <c> boot on a cobblestone street after rain \\
  & a <c> boot placed on pink silk fabric \\
\rowcolor{rowgray}  & a vintage <c> boot on an antique table \\
  & a <c> boot on a sandy beach at sunset \\
\rowcolor{rowgray}  & a <c> boot on a marble table, studio lighting \\
  & a <c> boot on a mossy rock in a forest \\
\rowcolor{rowgray}  & a glowing <c> boot in the dark \\
  & a <c> boot on a wooden shelf surrounded by books \\
\midrule
\rowcolor{rowgray} grey\_sloth\_plushie & a <c> stuffed animal in the snow \\
  & a <c> stuffed animal floating in outer space \\
\rowcolor{rowgray}  & a <c> stuffed animal sitting on a red couch indoors \\
  & a <c> stuffed animal driving a tiny car \\
\rowcolor{rowgray}  & a shiny <c> stuffed animal \\
  & a wet <c> stuffed animal \\
\rowcolor{rowgray}  & a <c> stuffed animal in a forest clearing with sunlight rays \\
  & a <c> stuffed animal with mountains in the background \\
\rowcolor{rowgray}  & a <c> stuffed animal on the beach during sunset \\
  & a cube-shaped <c> stuffed animal \\
\midrule
\rowcolor{rowgray} monster\_toy & a <c> toy in a wizard robe holding a staff \\
  & a <c> toy on the beach during sunset \\
\rowcolor{rowgray}  & a shiny <c> toy \\
  & a <c> toy wearing a black top hat and a monocle \\
\rowcolor{rowgray}  & a cube-shaped <c> toy \\
  & a <c> toy sitting on a red couch indoors \\
\rowcolor{rowgray}  & a <c> toy in a city park surrounded by flowers \\
  & a <c> toy driving a tiny car \\
\rowcolor{rowgray}  & a <c> toy wearing a rainbow scarf \\
  & a <c> toy wearing sunglasses \\
\midrule
\rowcolor{rowgray} pink\_sunglasses & a <c> glasses next to a cup of coffee on a kitchen counter \\
  & a <c> glasses on a wooden shelf surrounded by books \\
\rowcolor{rowgray}  & a vintage <c> glasses on an antique table \\
  & a <c> glasses with a city skyline in the background \\
\rowcolor{rowgray}  & a <c> glasses with mountains and mist in the background \\
  & a glowing <c> glasses in the dark \\
\rowcolor{rowgray}  & a <c> glasses on a cobblestone street after rain \\
  & a modern minimalistic <c> glasses on a white surface \\
\rowcolor{rowgray}  & a <c> glasses on a marble table, studio lighting \\
  & a <c> glasses placed on pink silk fabric \\
\midrule
\rowcolor{rowgray} poop\_emoji & a <c> toy with the Eiffel Tower in the background \\
  & a <c> toy in the snow \\
\rowcolor{rowgray}  & a <c> toy driving a tiny car \\
  & a <c> toy on the beach during sunset \\
\rowcolor{rowgray}  & a <c> toy in a wizard robe holding a staff \\
  & a <c> toy wearing a rainbow scarf \\
\rowcolor{rowgray}  & a <c> toy floating in outer space \\
  & a cube-shaped <c> toy \\
\rowcolor{rowgray}  & a <c> toy in a police uniform \\
  & a shiny <c> toy \\
\midrule
\rowcolor{rowgray} rc\_car & a <c> toy wearing sunglasses \\
  & a <c> toy wearing a rainbow scarf \\
\rowcolor{rowgray}  & a shiny <c> toy \\
  & a <c> toy in the jungle \\
\rowcolor{rowgray}  & a <c> toy driving a tiny car \\
  & a <c> toy floating in outer space \\
\rowcolor{rowgray}  & a <c> toy in a police uniform \\
  & a <c> toy in a chef outfit \\
\rowcolor{rowgray}  & a <c> toy wearing a black top hat and a monocle \\
  & a <c> toy in the snow \\
\midrule
\rowcolor{rowgray} red\_cartoon & a shiny <c> cartoon \\
  & a <c> cartoon wearing a black top hat and a monocle \\
\rowcolor{rowgray}  & a wet <c> cartoon \\
  & a <c> cartoon with the Eiffel Tower in the background \\
\rowcolor{rowgray}  & a <c> cartoon sitting on a red couch indoors \\
  & a <c> cartoon on the beach during sunset \\
\rowcolor{rowgray}  & a <c> cartoon floating in outer space \\
  & a <c> cartoon wearing a rainbow scarf \\
\rowcolor{rowgray}  & a <c> cartoon in the jungle \\
  & a <c> cartoon with mountains in the background \\
\midrule
\rowcolor{rowgray} robot\_toy & a <c> toy in a police uniform \\
  & a <c> toy in a chef outfit \\
\rowcolor{rowgray}  & a <c> toy in a forest clearing with sunlight rays \\
  & a <c> toy driving a tiny car \\
\rowcolor{rowgray}  & a <c> toy sitting on a red couch indoors \\
  & a <c> toy on the beach during sunset \\
\rowcolor{rowgray}  & a <c> toy with mountains in the background \\
  & a shiny <c> toy \\
\rowcolor{rowgray}  & a cube-shaped <c> toy \\
  & a <c> toy in a city park surrounded by flowers \\
\midrule
\rowcolor{rowgray} shiny\_sneaker & a <c> sneaker on a glass table with reflections \\
  & a <c> sneaker on a sandy beach at sunset \\
\rowcolor{rowgray}  & a modern minimalistic <c> sneaker on a white surface \\
  & a <c> sneaker on a cobblestone street after rain \\
\rowcolor{rowgray}  & a <c> sneaker in the snow under warm sunlight \\
  & a <c> sneaker on a marble table, studio lighting \\
\rowcolor{rowgray}  & a <c> sneaker with a city skyline in the background \\
  & a vintage <c> sneaker on an antique table \\
\rowcolor{rowgray}  & a <c> sneaker placed on pink silk fabric \\
  & a <c> sneaker in a minimalist art gallery \\
\midrule
\rowcolor{rowgray} teapot & a modern minimalistic <c> teapot on a white surface \\
  & a glowing <c> teapot in the dark \\
\rowcolor{rowgray}  & a <c> teapot floating in crystal clear water \\
  & a <c> teapot placed on pink silk fabric \\
\rowcolor{rowgray}  & a <c> teapot on a sandy beach at sunset \\
  & a <c> teapot on a mossy rock in a forest \\
\rowcolor{rowgray}  & a <c> teapot with mountains and mist in the background \\
  & a vintage <c> teapot on an antique table \\
\rowcolor{rowgray}  & a <c> teapot on a glass table with reflections \\
  & a <c> teapot next to a cup of coffee on a kitchen counter \\
\midrule
\rowcolor{rowgray} vase & a <c> vase on a mossy rock in a forest \\
  & a <c> vase next to a cup of coffee on a kitchen counter \\
\rowcolor{rowgray}  & a <c> vase with a city skyline in the background \\
  & a <c> vase on a sandy beach at sunset \\
\rowcolor{rowgray}  & a glowing <c> vase in the dark \\
  & a <c> vase floating in crystal clear water \\
\rowcolor{rowgray}  & a <c> vase on a wooden shelf surrounded by books \\
  & a <c> vase on a reflective mirror surface \\
\rowcolor{rowgray}  & a <c> vase in a minimalist art gallery \\
  & a <c> vase with mountains and mist in the background \\
\midrule
\rowcolor{rowgray} wolf\_plushie & a wet <c> stuffed animal \\
  & a <c> stuffed animal driving a tiny car \\
\rowcolor{rowgray}  & a <c> stuffed animal wearing a black top hat and a monocle \\
  & a <c> stuffed animal wearing a red hat \\
\rowcolor{rowgray}  & a <c> stuffed animal in a chef outfit \\
  & a <c> stuffed animal wearing a rainbow scarf \\
\rowcolor{rowgray}  & a <c> stuffed animal in a city park surrounded by flowers \\
  & a <c> stuffed animal floating in outer space \\
\rowcolor{rowgray}  & a <c> stuffed animal in a forest clearing with sunlight rays \\
  & a <c> stuffed animal on the beach during sunset \\
\midrule

\end{longtable}

\clearpage
\section{Full  Self-Attention and Cross-Attention Ranks}

\begin{figure}[!h]
  \centering
  \begin{subfigure}{\textwidth}
    \centering
    \includegraphics[width=\linewidth]{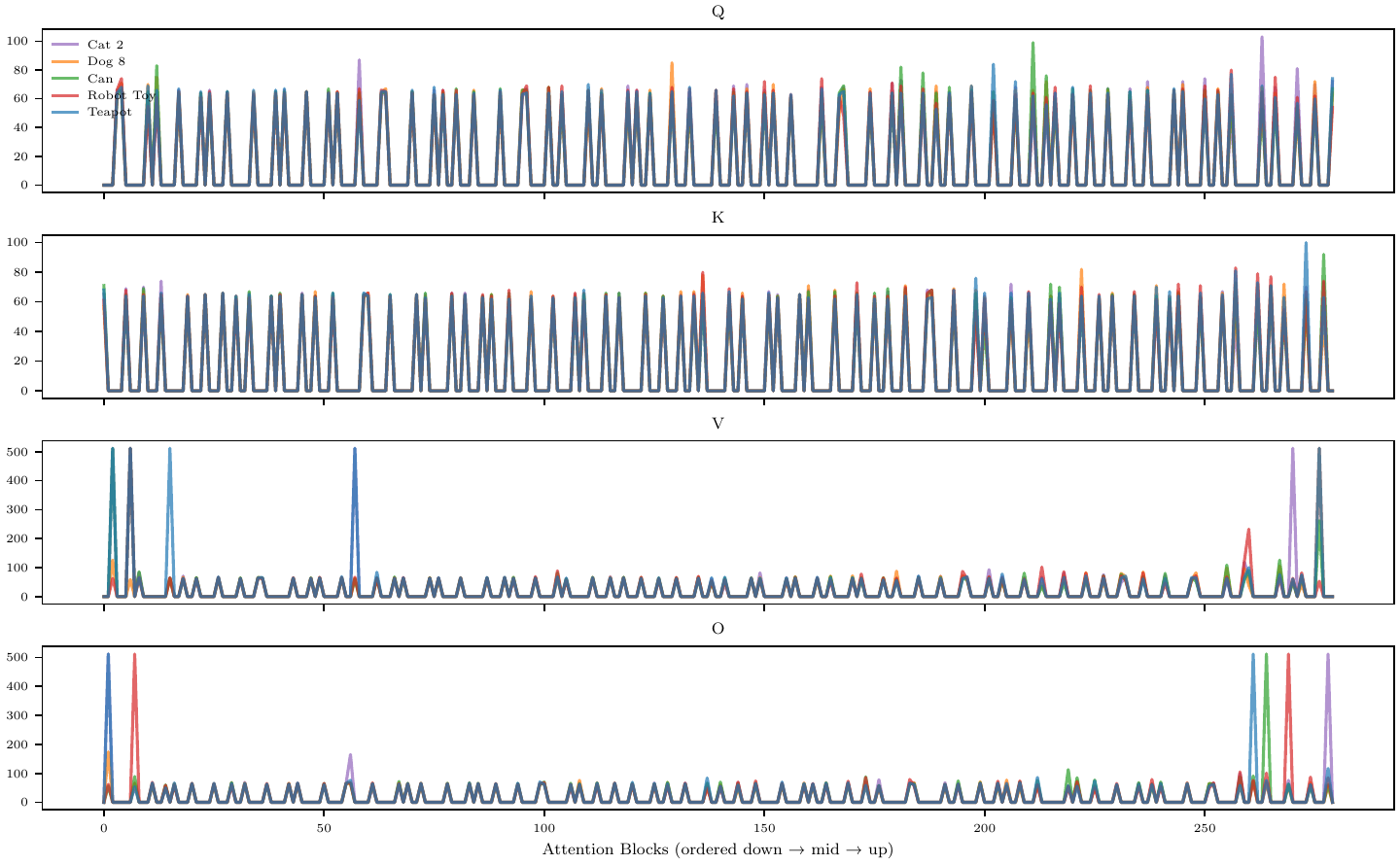}
    \caption{SDXL Self-attention ranks, for five distinct subjects.}
    \label{fig:self_attn_ranks_full}
  \end{subfigure}
  \begin{subfigure}{\textwidth}
    \centering
    \includegraphics[width=\linewidth]{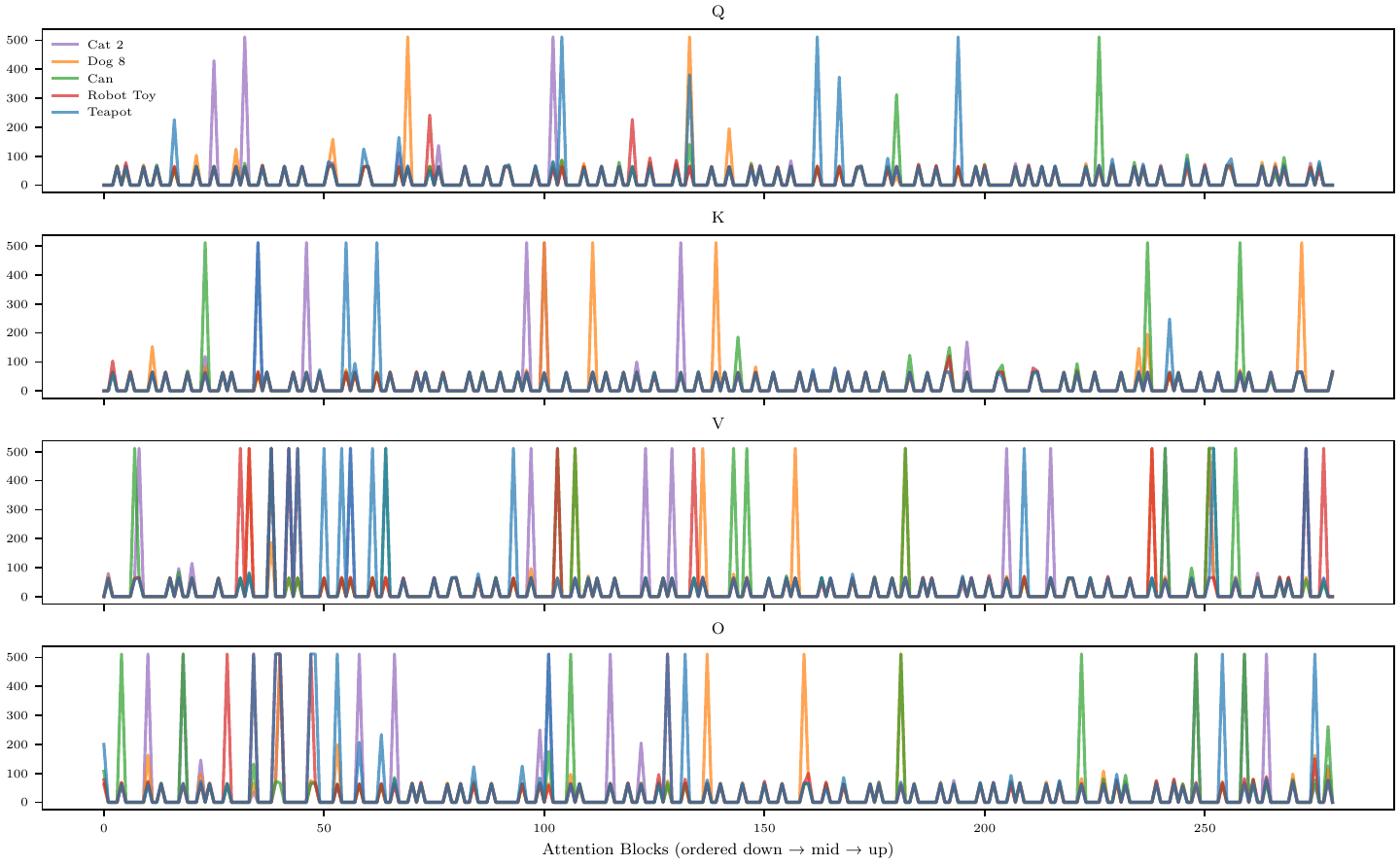}
    \caption{SDXL Cross-attention ranks, for five distinct subjects.}
    \label{fig:cross_attn_ranks_full}
  \end{subfigure}
\end{figure}

\begin{figure}[!h]
  \centering
  \begin{subfigure}{\textwidth}
    \centering
    \includegraphics[width=\linewidth]{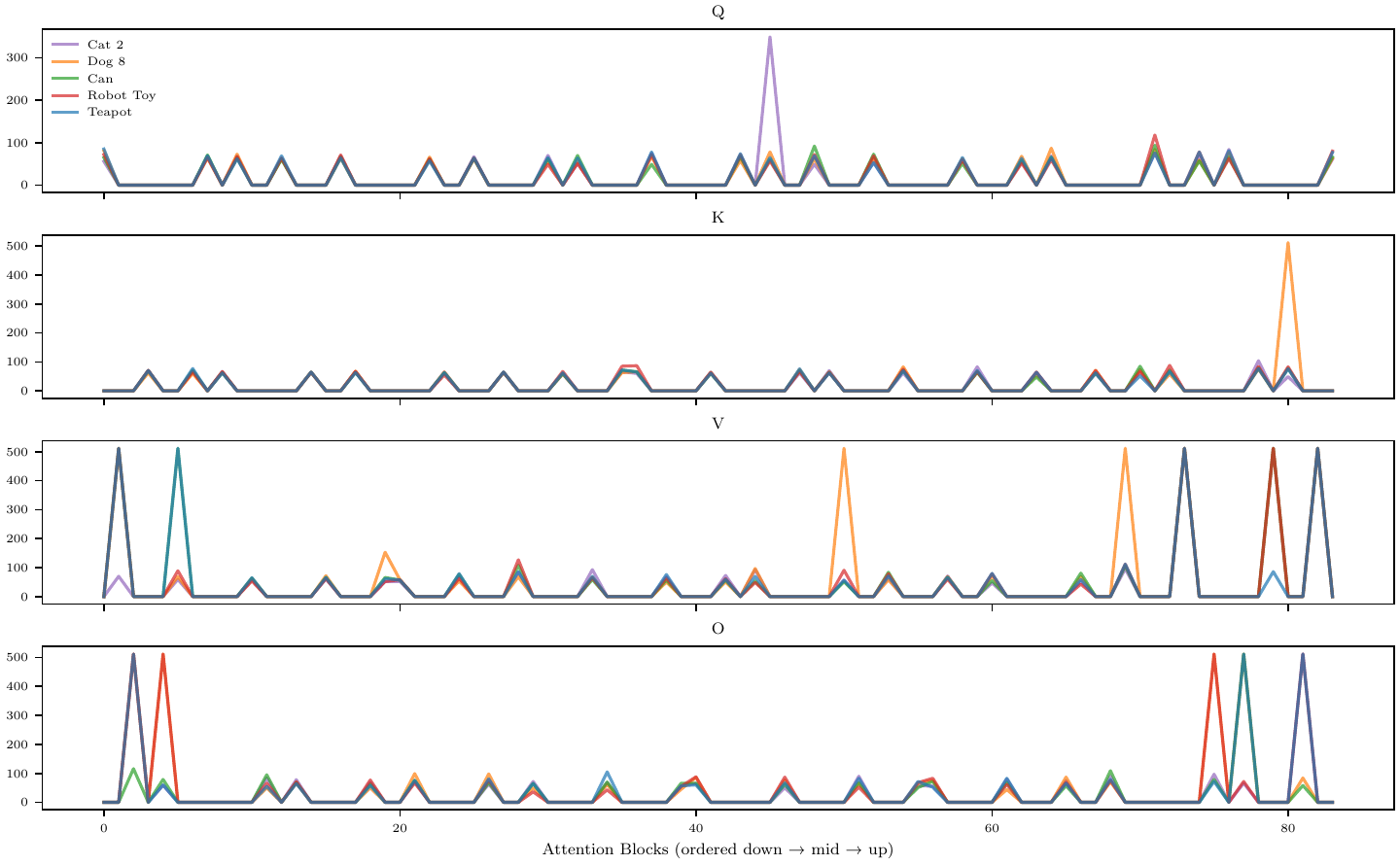}
    \caption{KOALA-700m Self-attention ranks, for five distinct subjects.}
    \label{fig:self_attn_ranks_koala}
  \end{subfigure}
  \begin{subfigure}{\textwidth}
    \centering
    \includegraphics[width=\linewidth]{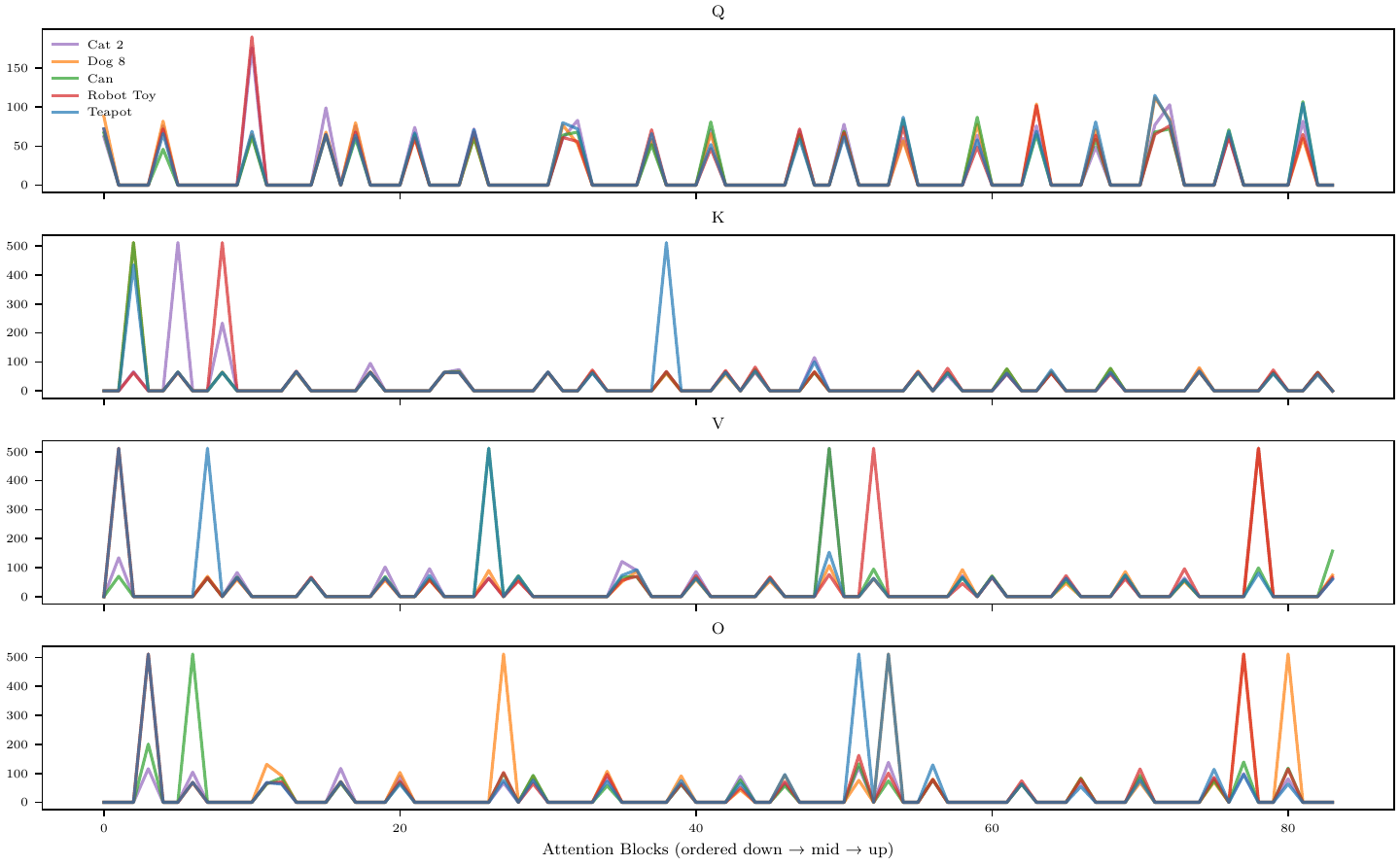}
    \caption{KOALA-700m Cross-attention ranks, for five distinct subjects.}
    \label{fig:cross_attn_ranks_koala}
  \end{subfigure}
\end{figure}

\clearpage
\section{KOALA Per-Class Scores}
\begin{figure}[!h]
    \centering
\includegraphics[width=\linewidth]{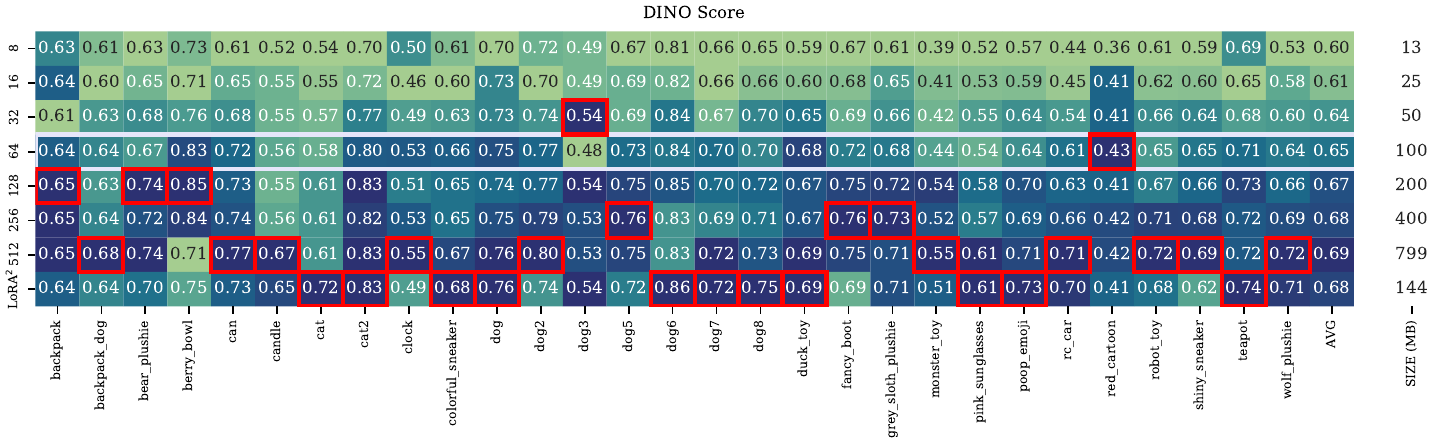}
    \includegraphics[width=\linewidth]{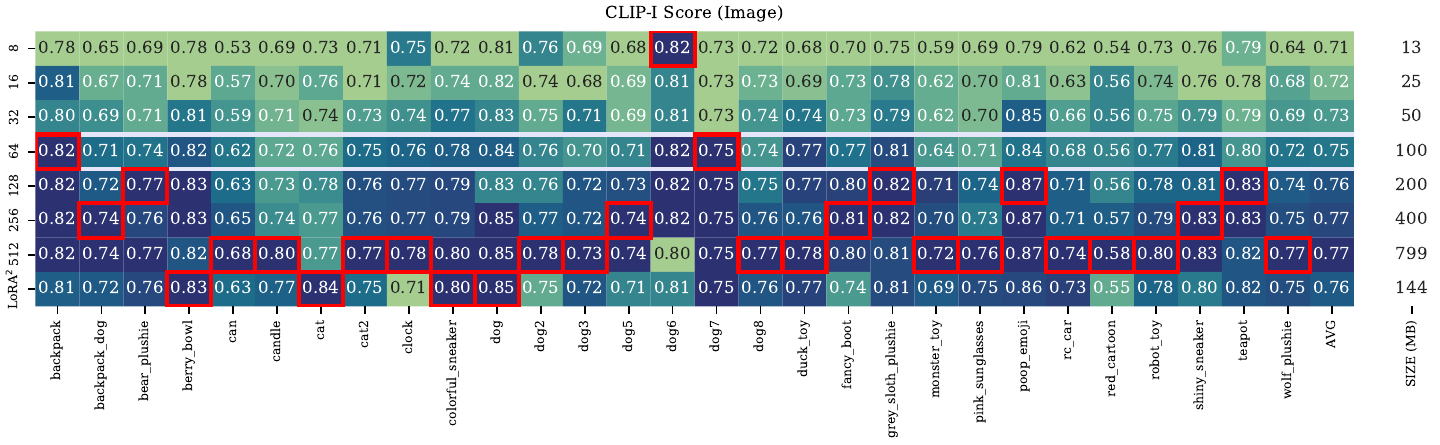}
    \includegraphics[width=\linewidth]{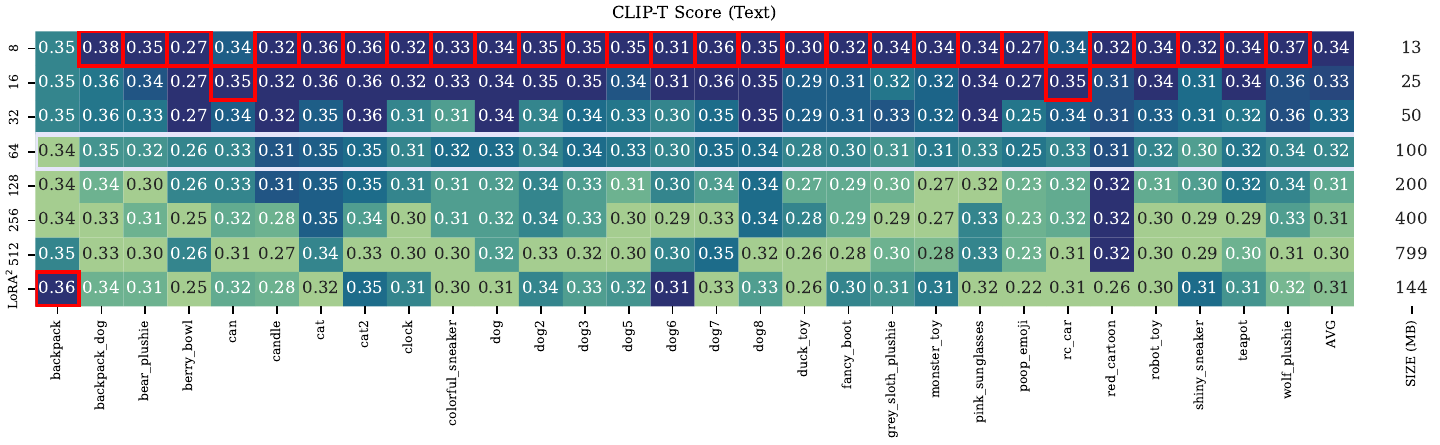}
    \caption{KOALA-700m backbone, per-subject scores. We highlight with a grey band rank 64, the default value commonly used in previous work. We also highlight in red the best value for each subject. On the side, we also add the model size in MB. }
    \label{fig:per_class_koala}
\end{figure}

In \Cref{fig:per_class_koala} we report the per-subject scores for KOALA-700m.
Similar to the SDXL in the main paper, we note that the optimal rank changes depending on the subject. We note here more variability in the best subject rank selection.
 
\section{Additional Qualitative Results}
\begin{figure}[!h]
\centering
\renewcommand{\arraystretch}{1.2}
\setlength{\tabcolsep}{2pt}
\resizebox{\linewidth}{!}{%
\begin{tabular}{
    >{\centering\arraybackslash}m{2cm}
    >{\centering\arraybackslash}m{2.5cm} 
    >{\centering\arraybackslash}m{2.5cm} 
    >{\centering\arraybackslash}m{2.5cm} 
    >{\centering\arraybackslash}m{2.5cm} 
    >{\centering\arraybackslash}m{2.5cm} 
}
\includegraphics[width=2cm]{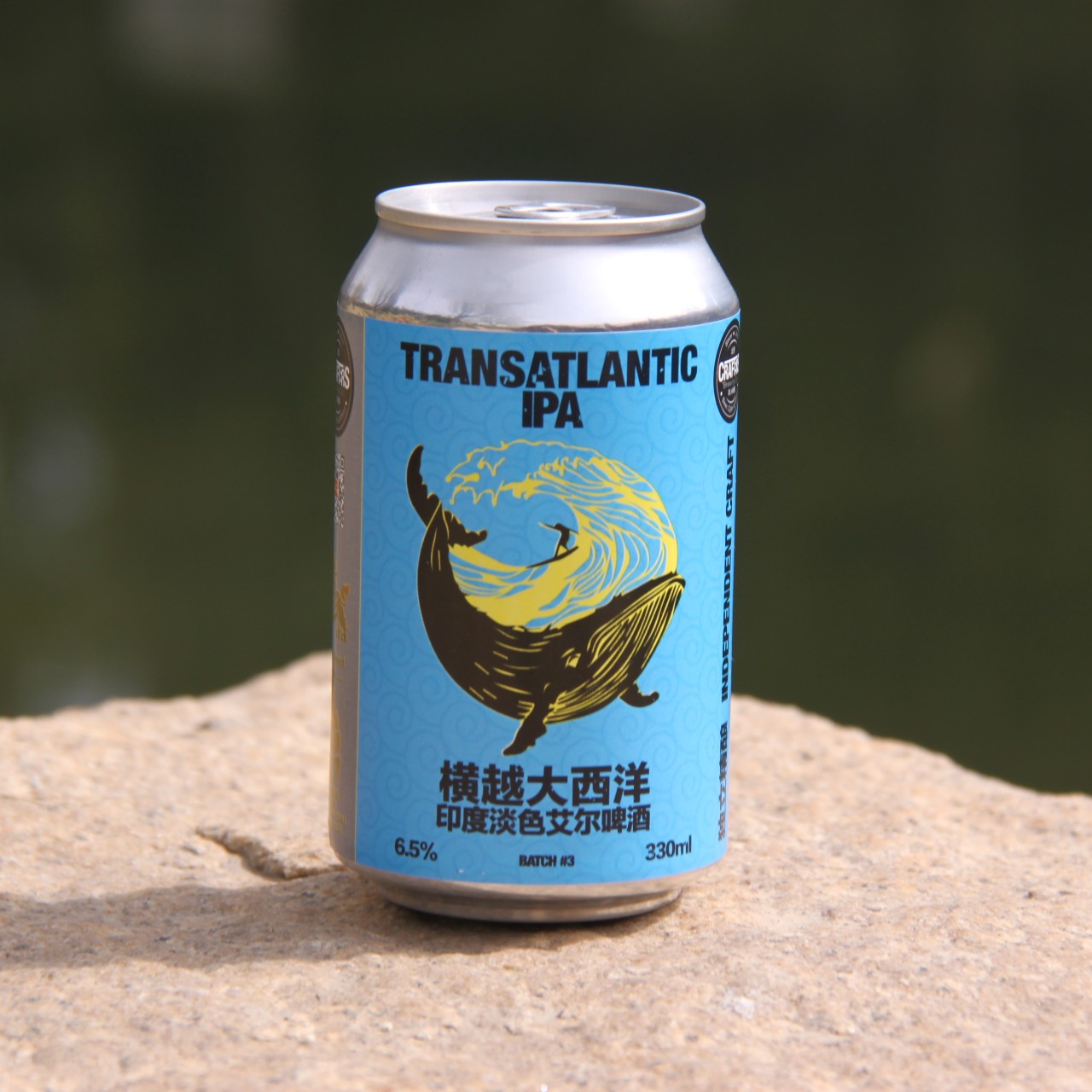}
& \textbf{``a \textcolor{red}{k} can on a mossy rock in a forest''} 
& \textbf{``a \textcolor{red}{k} can with mountains and mist in the background''} 
& \textbf{``a \textcolor{red}{k} can placed on pink silk fabric''} 
& \textbf{``a \textcolor{red}{k} can on the snow under warm sunlight'} 
& \textbf{``a \textcolor{red}{k} on a sandy beach at sunset''} \\[2mm]

\scriptsize Rank 8
& \includegraphics[width=2.3cm]{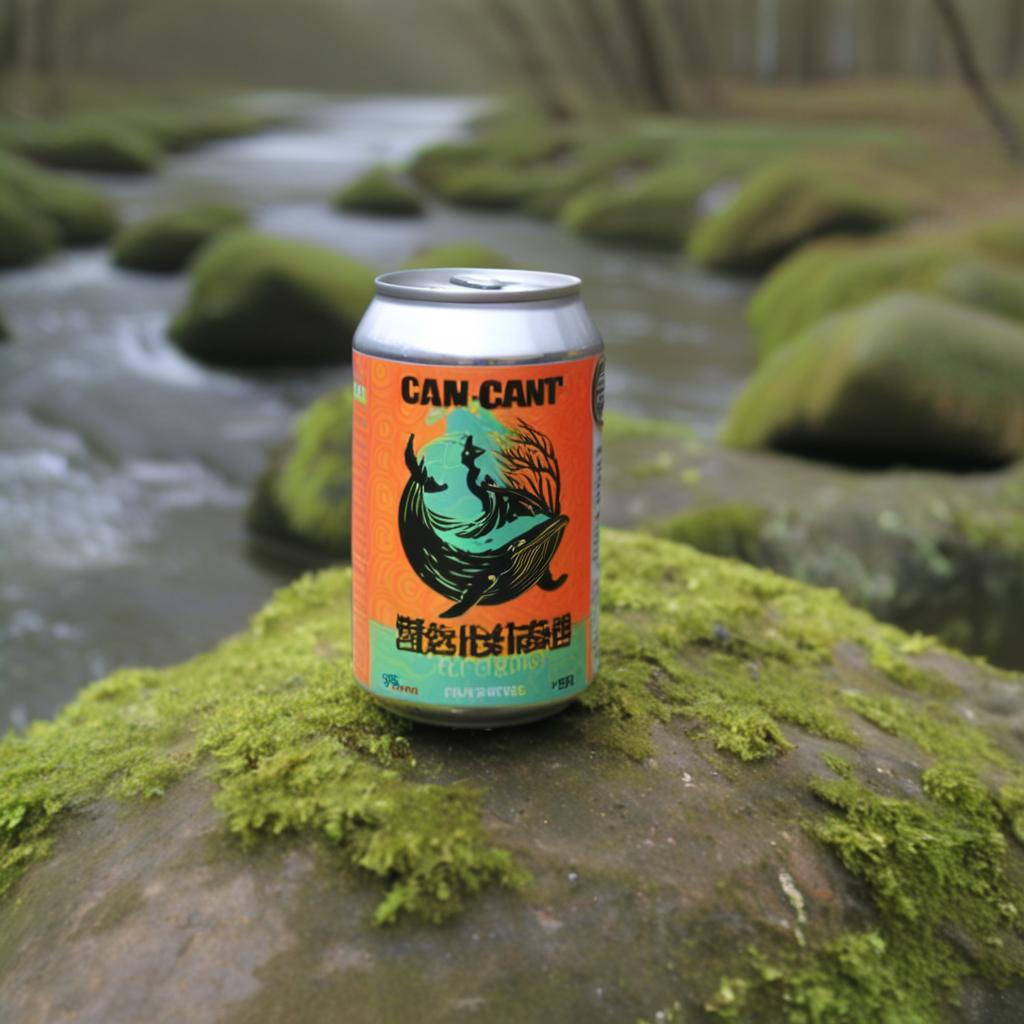}
& \includegraphics[width=2.3cm]{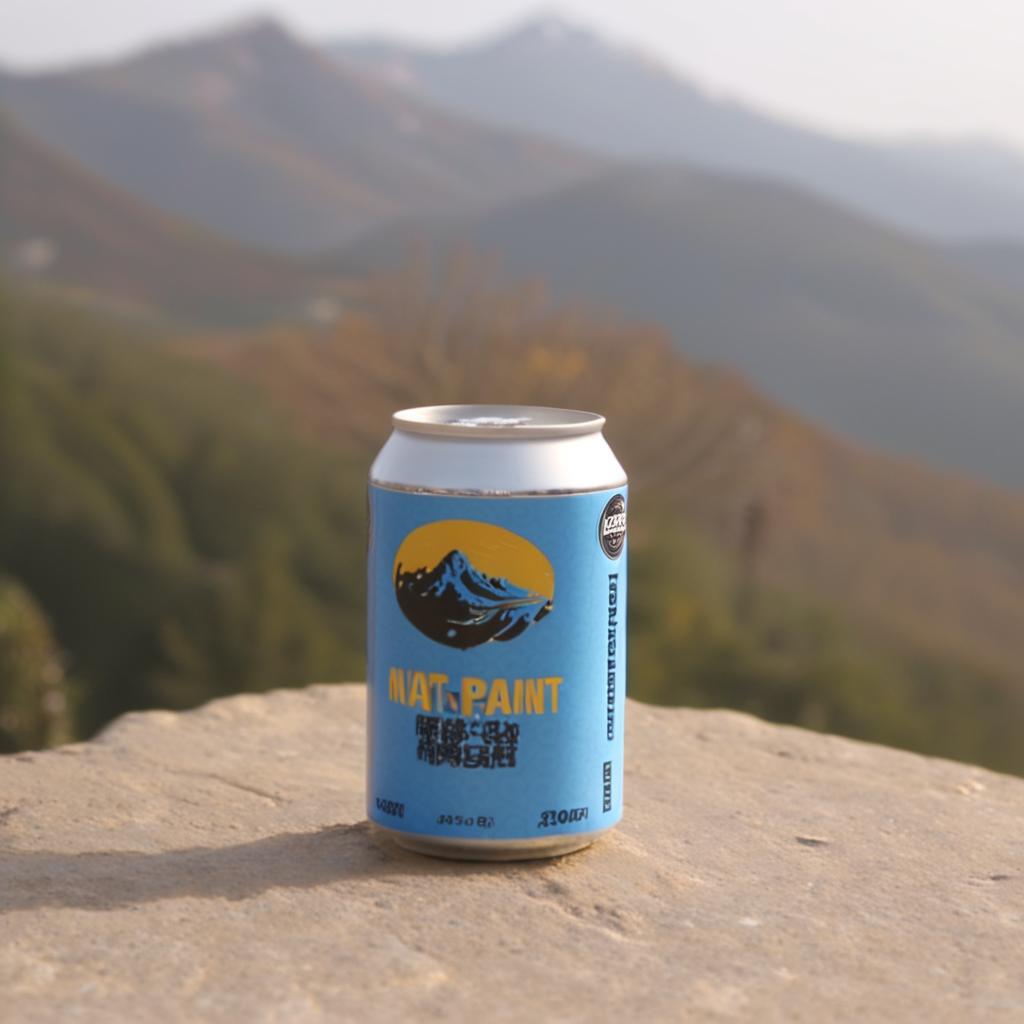}
& \includegraphics[width=2.3cm]{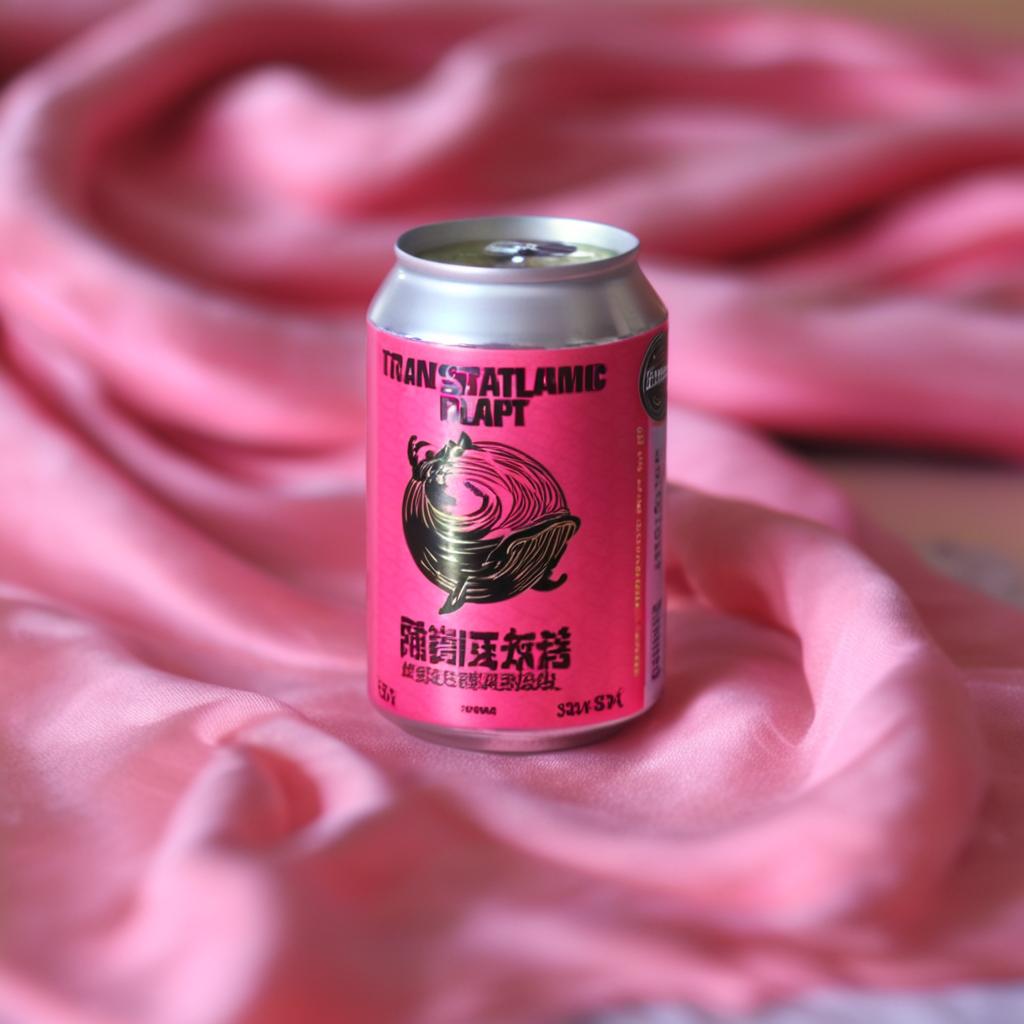}
& \includegraphics[width=2.3cm]{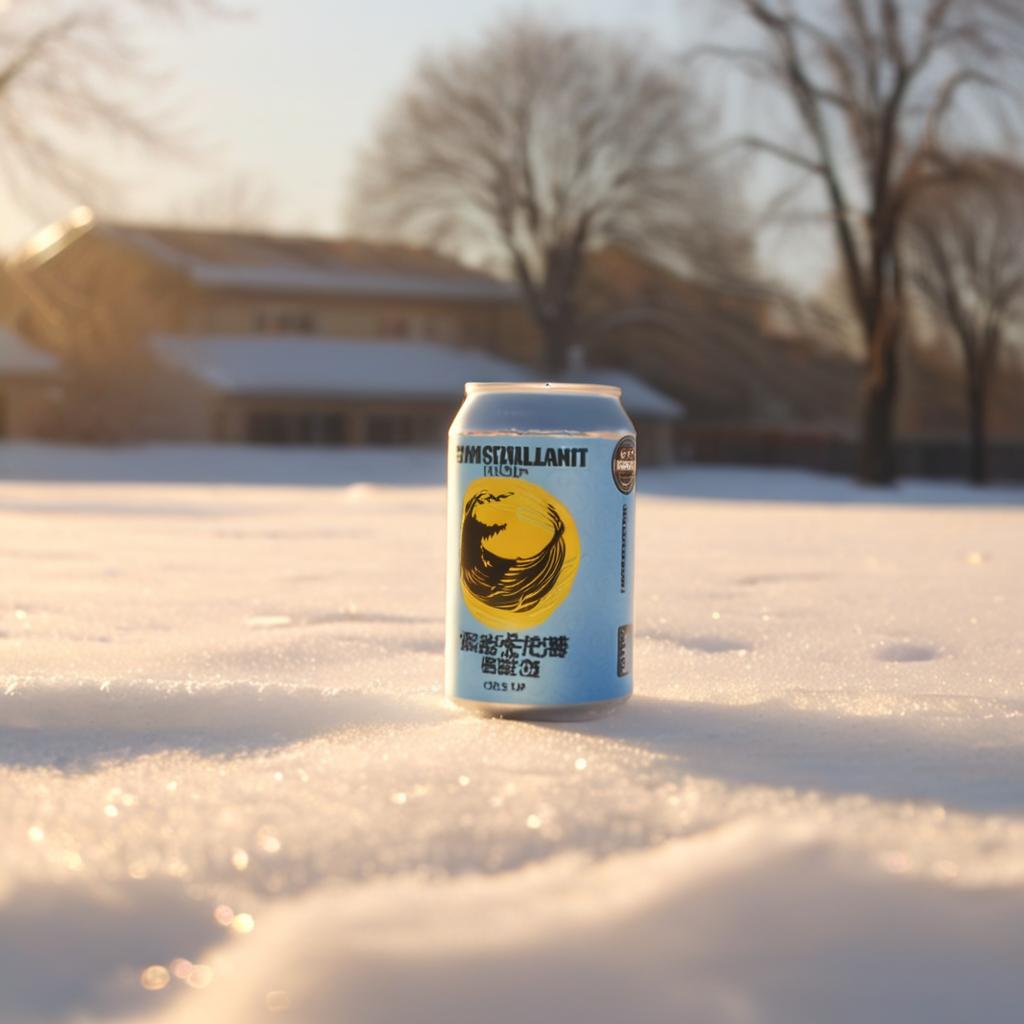}
& \includegraphics[width=2.3cm]{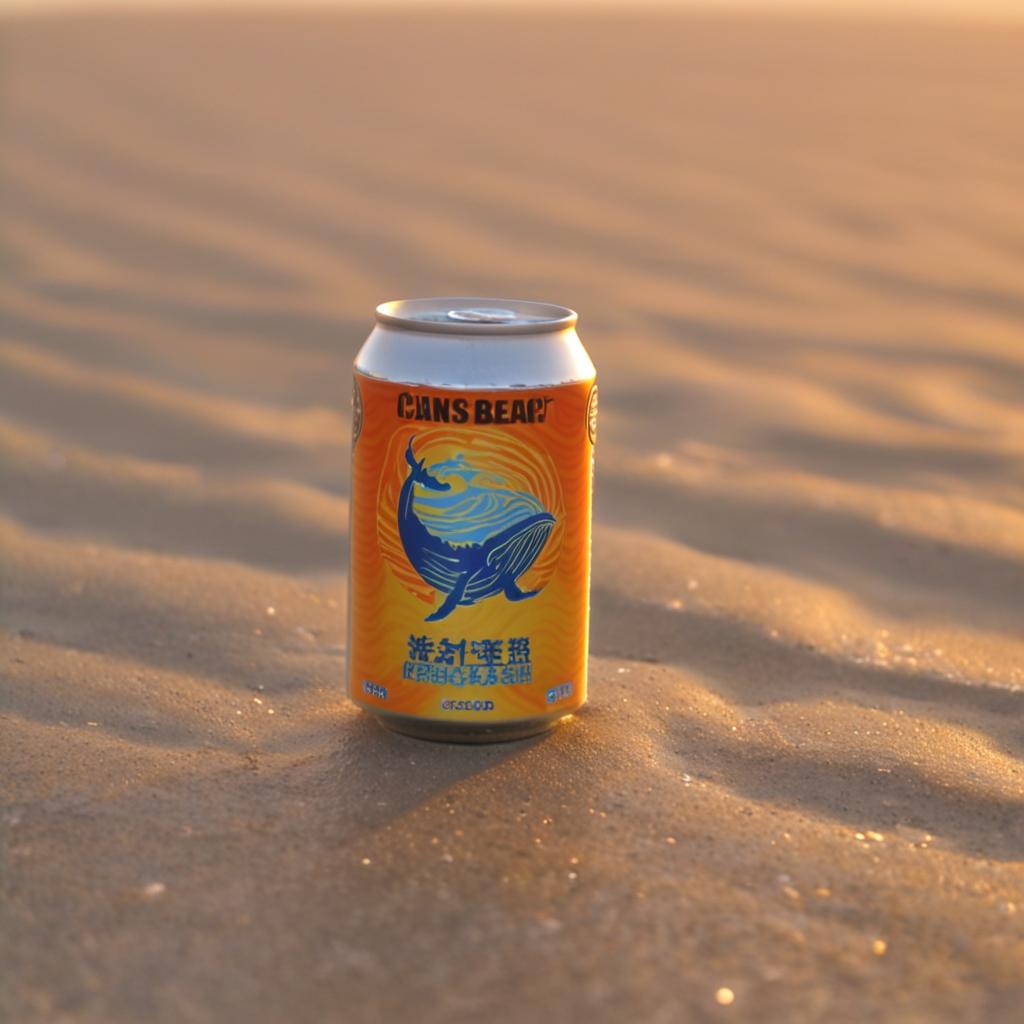}
\\[1mm]

\scriptsize Rank 64
& \includegraphics[width=2.3cm]{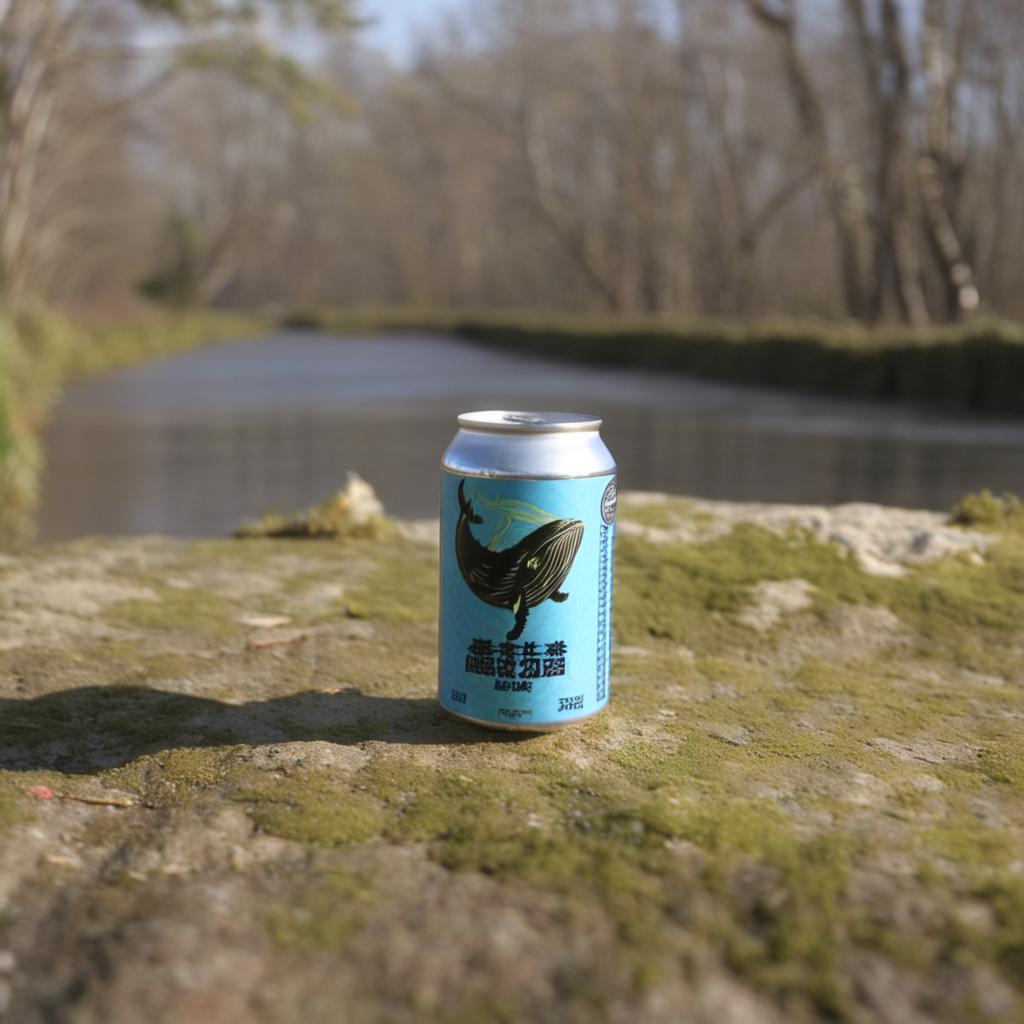}
& \includegraphics[width=2.3cm]{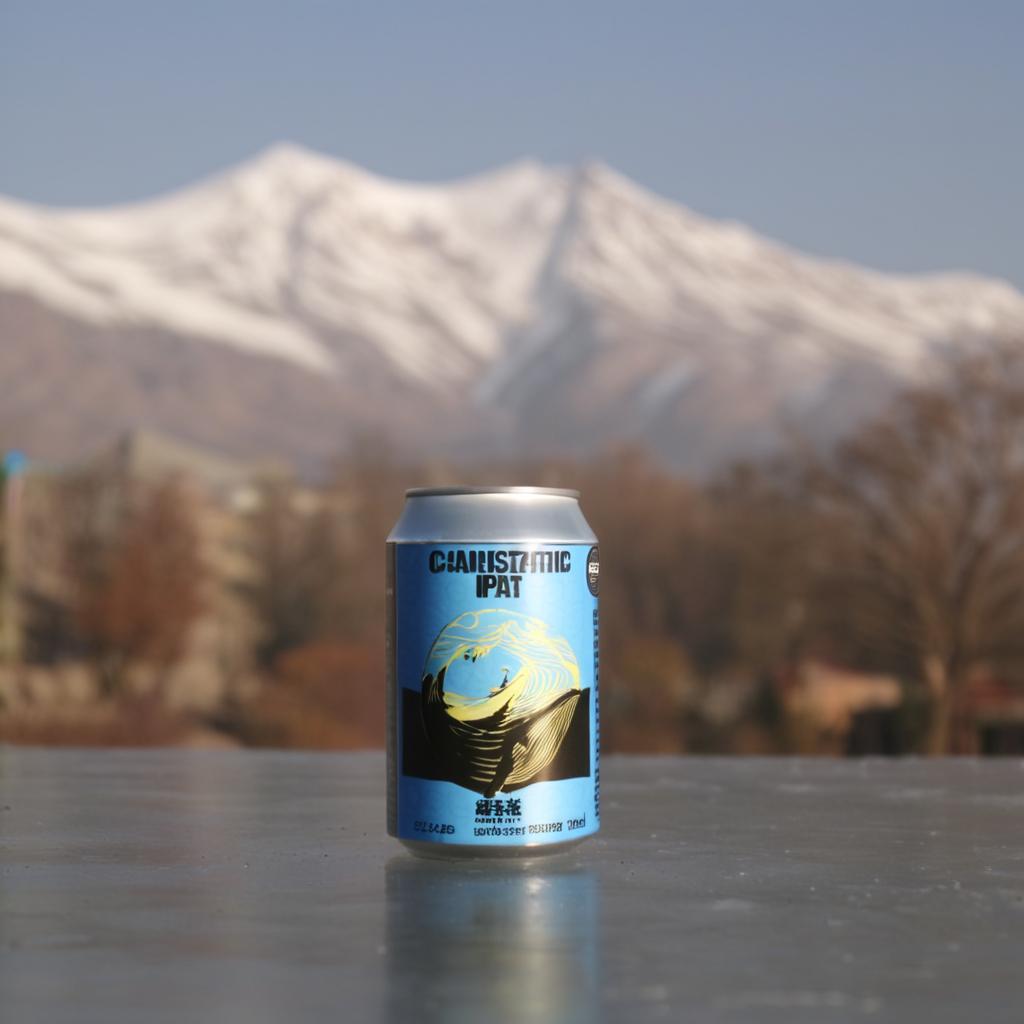}
& \includegraphics[width=2.3cm]{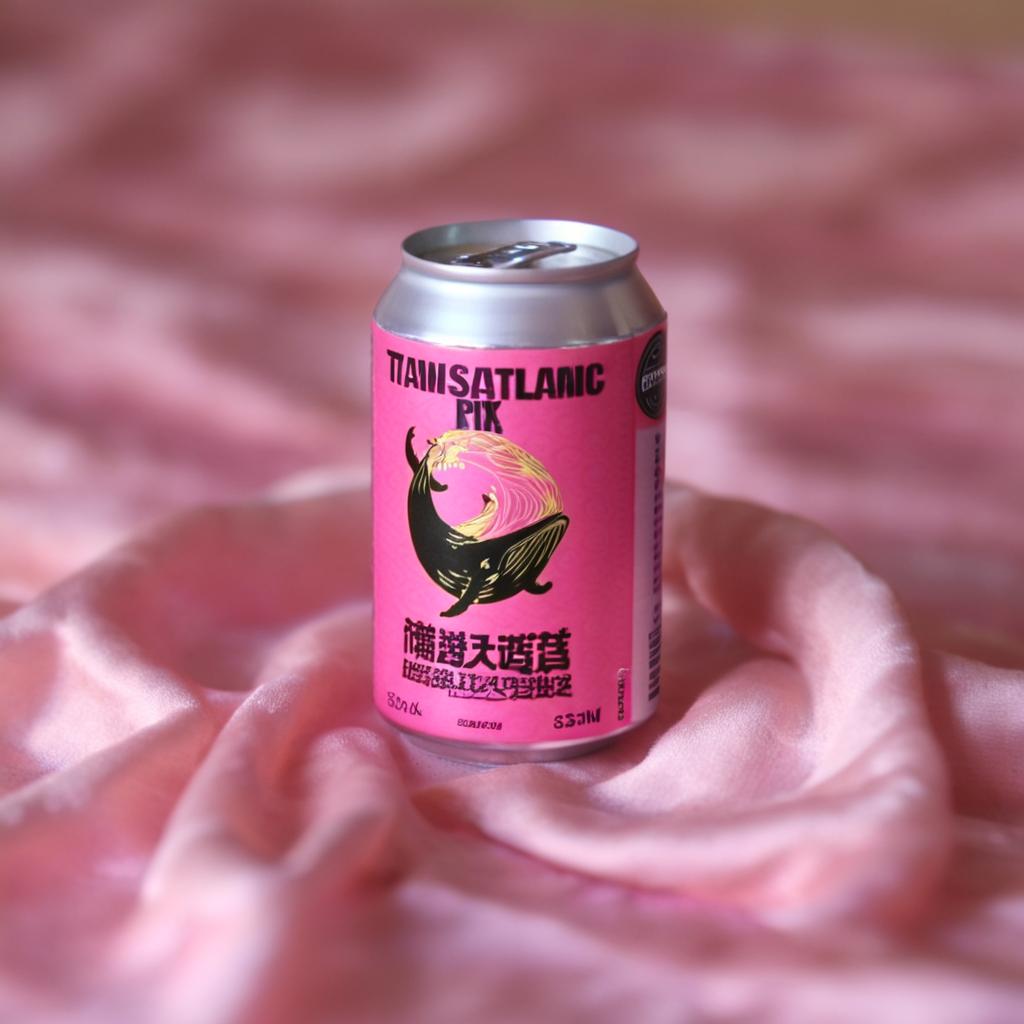}
& \includegraphics[width=2.3cm]{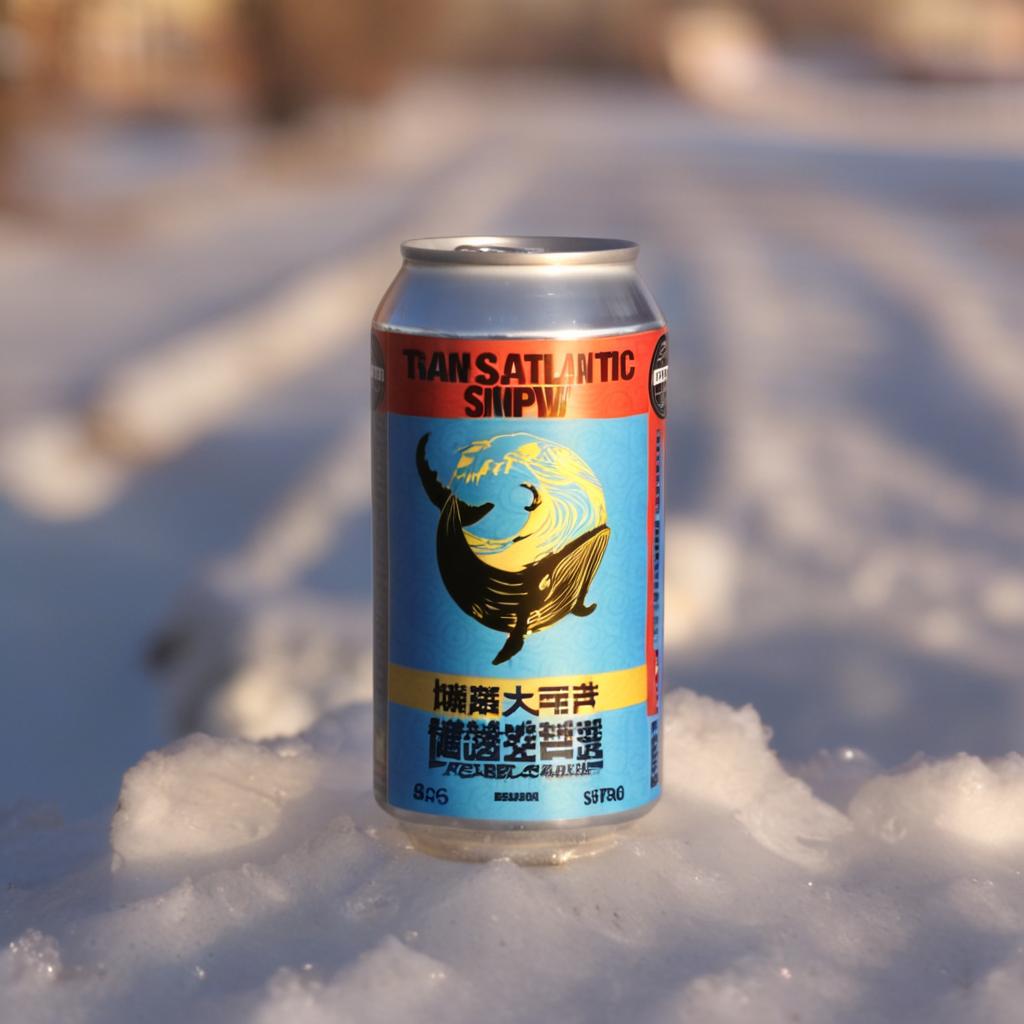}
& \includegraphics[width=2.3cm]{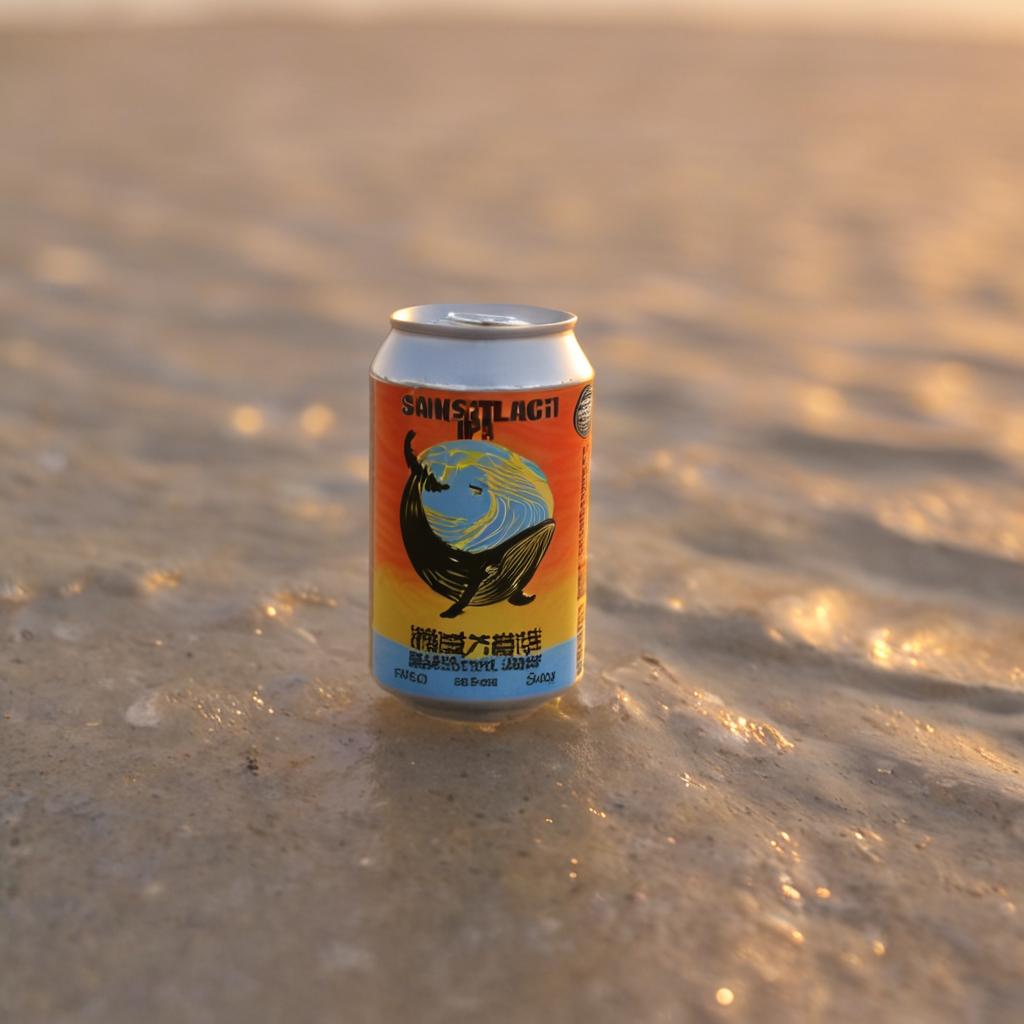}
\\[1mm]

\scriptsize Rank 512
& \includegraphics[width=2.3cm]{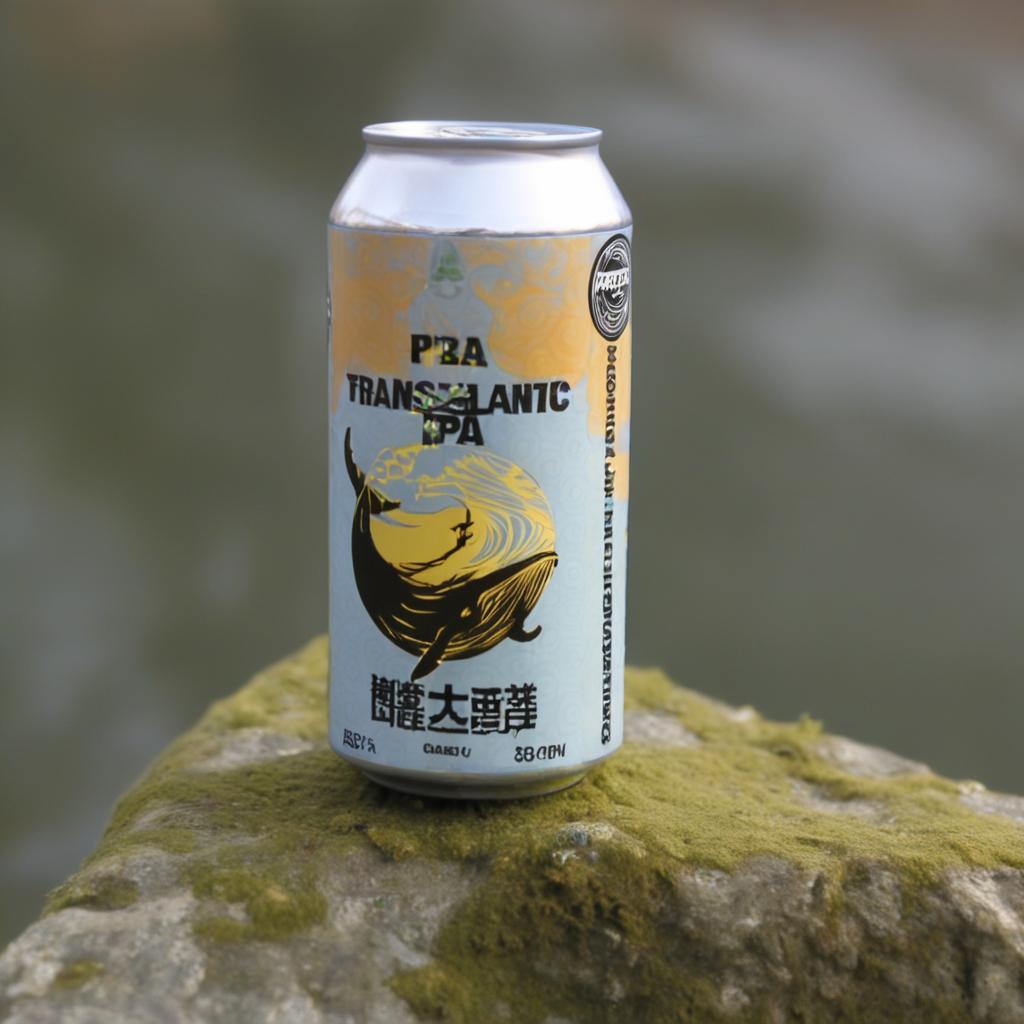}
& \includegraphics[width=2.3cm]{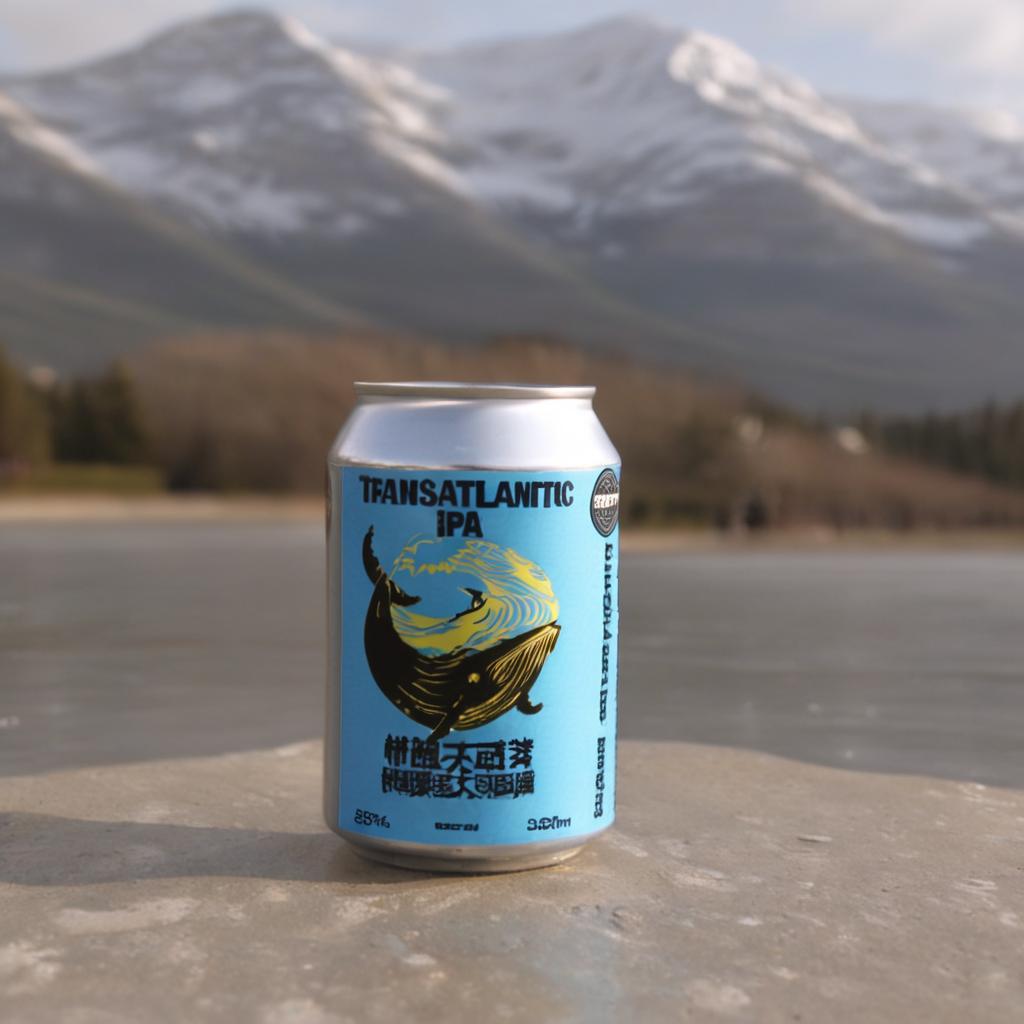}
& \includegraphics[width=2.3cm]{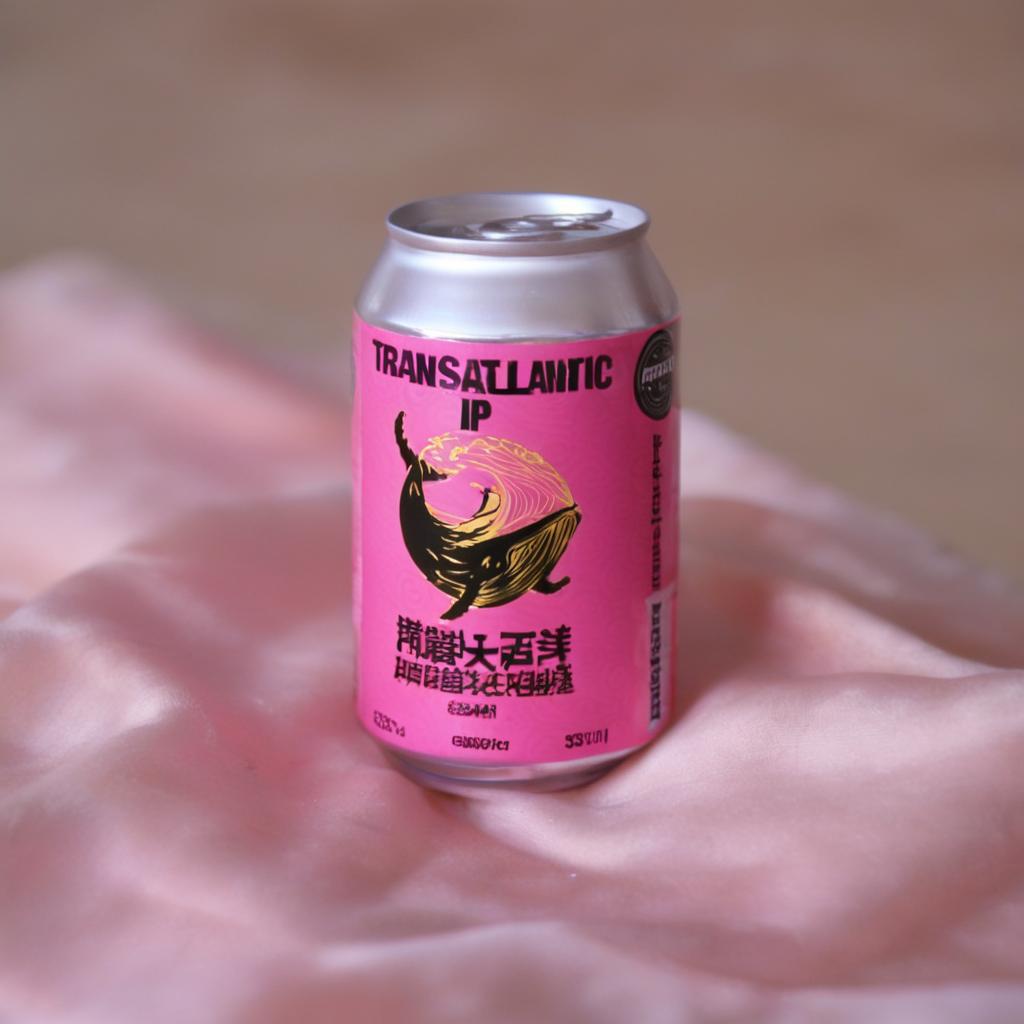}
& \includegraphics[width=2.3cm]{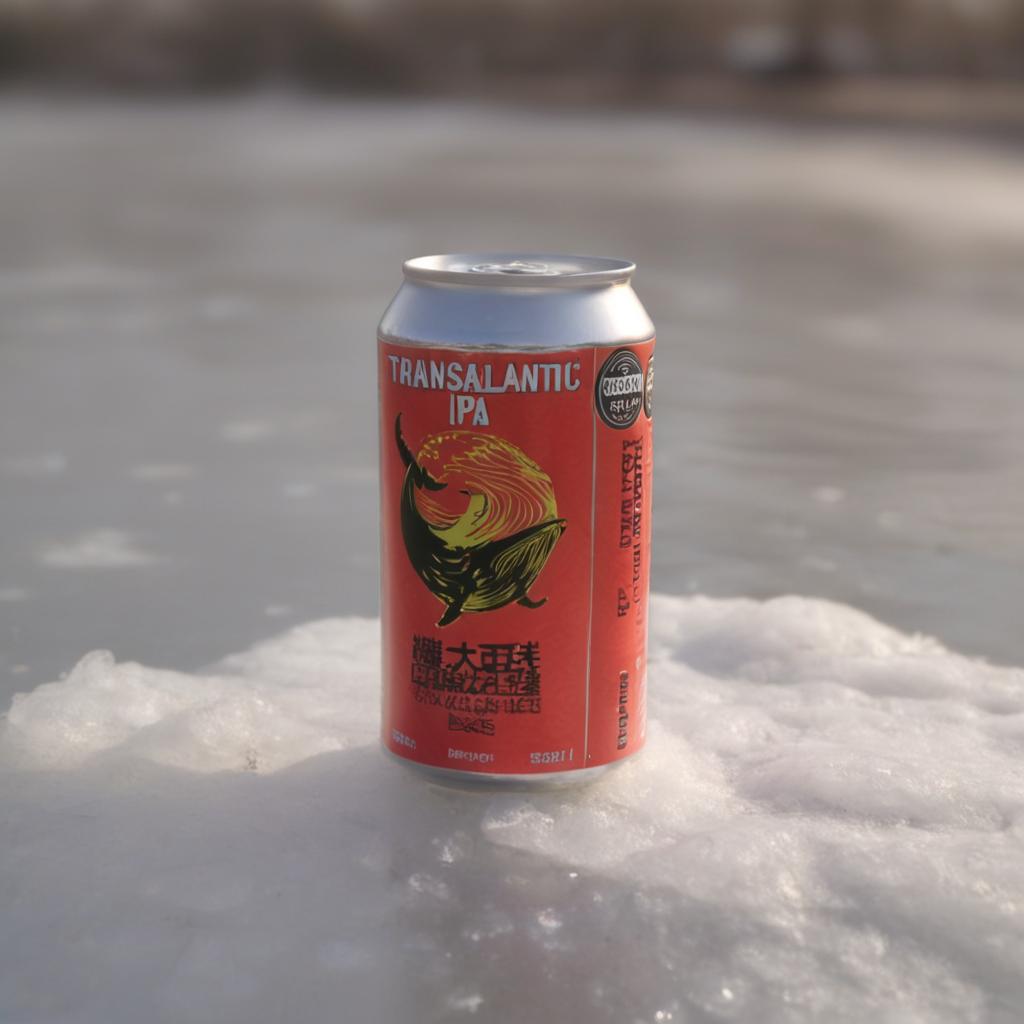}
& \includegraphics[width=2.3cm]{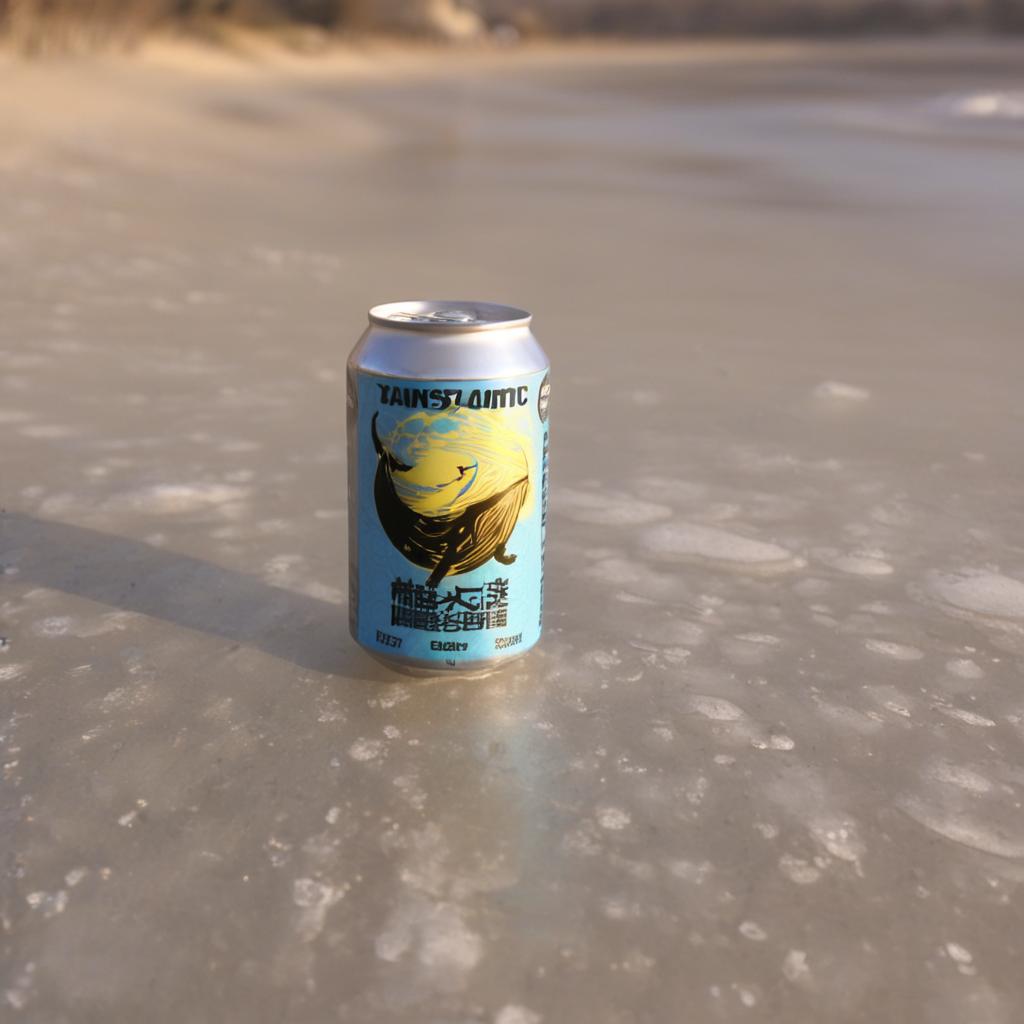}
\\[1mm]

\scriptsize \method
& \includegraphics[width=2.3cm]{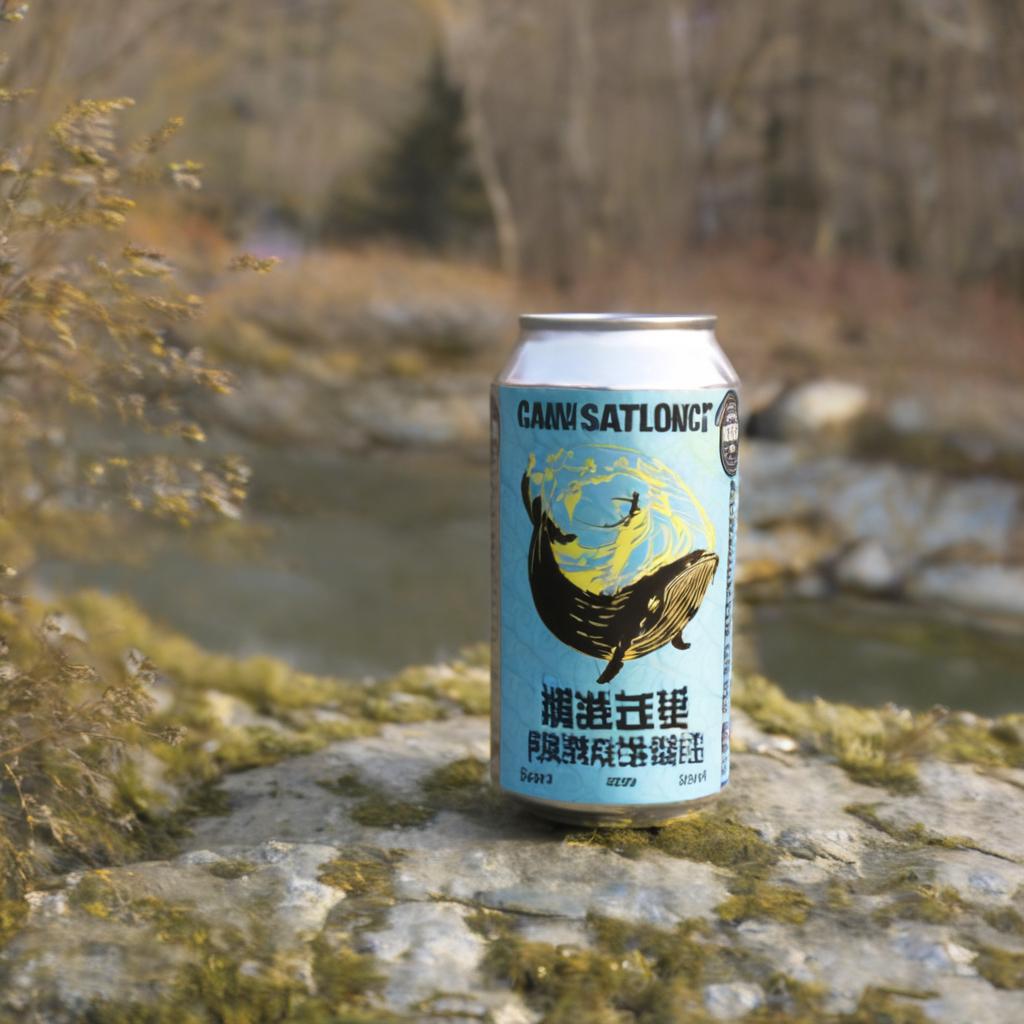}
& \includegraphics[width=2.3cm]{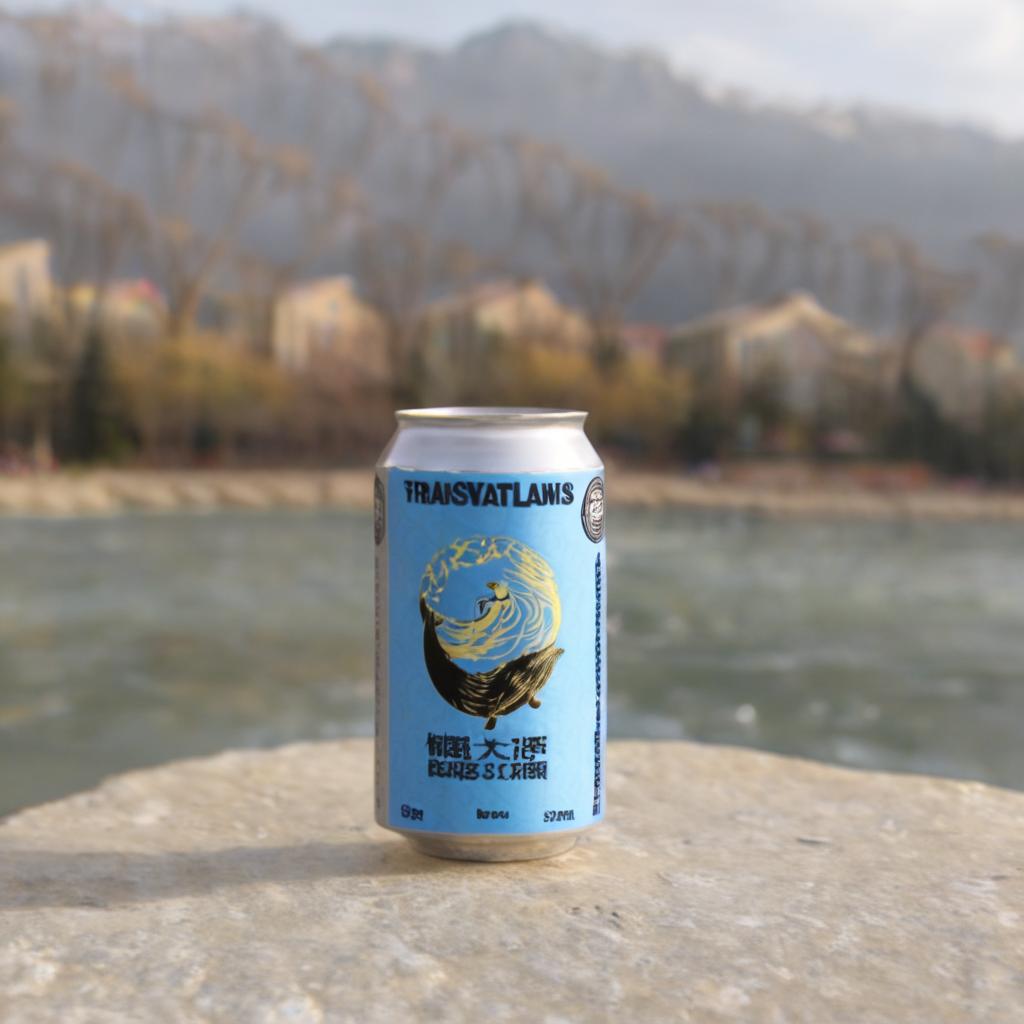}
& \includegraphics[width=2.3cm]{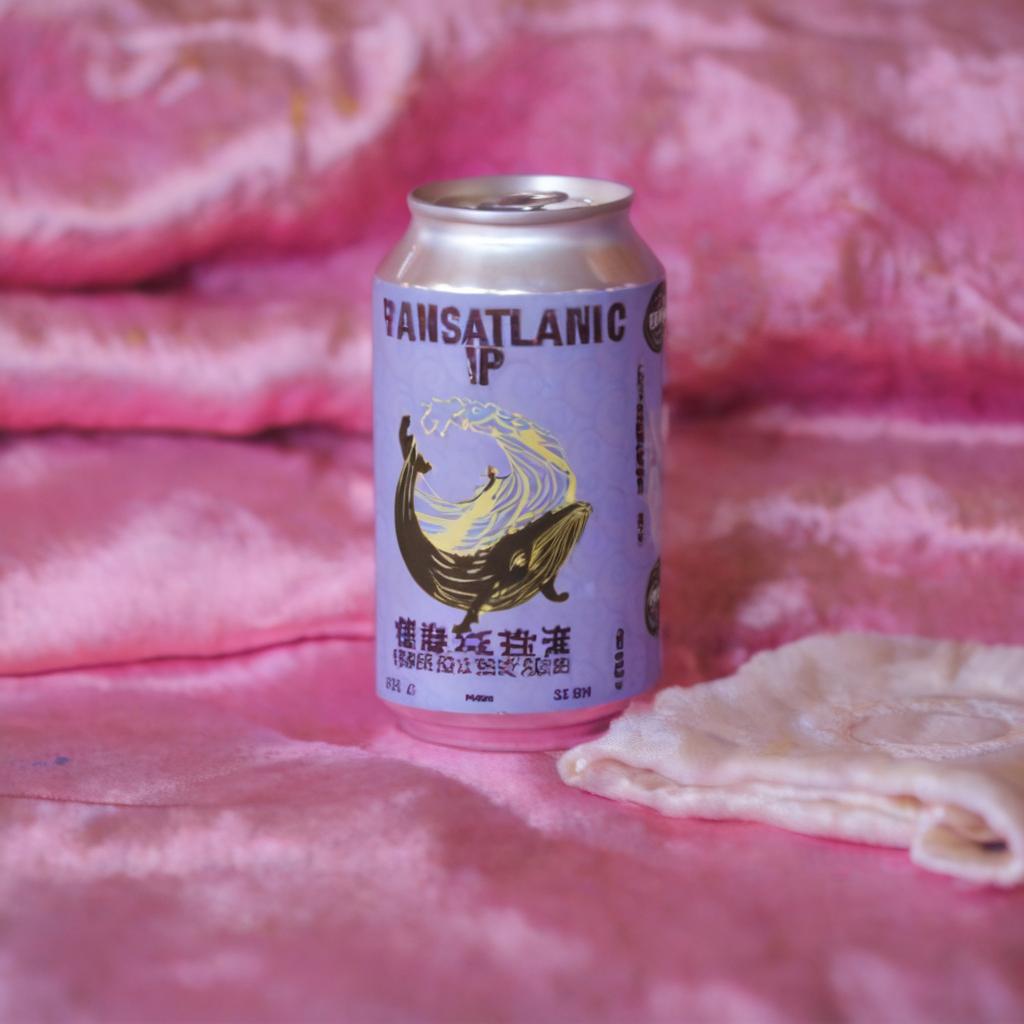}
& \includegraphics[width=2.3cm]{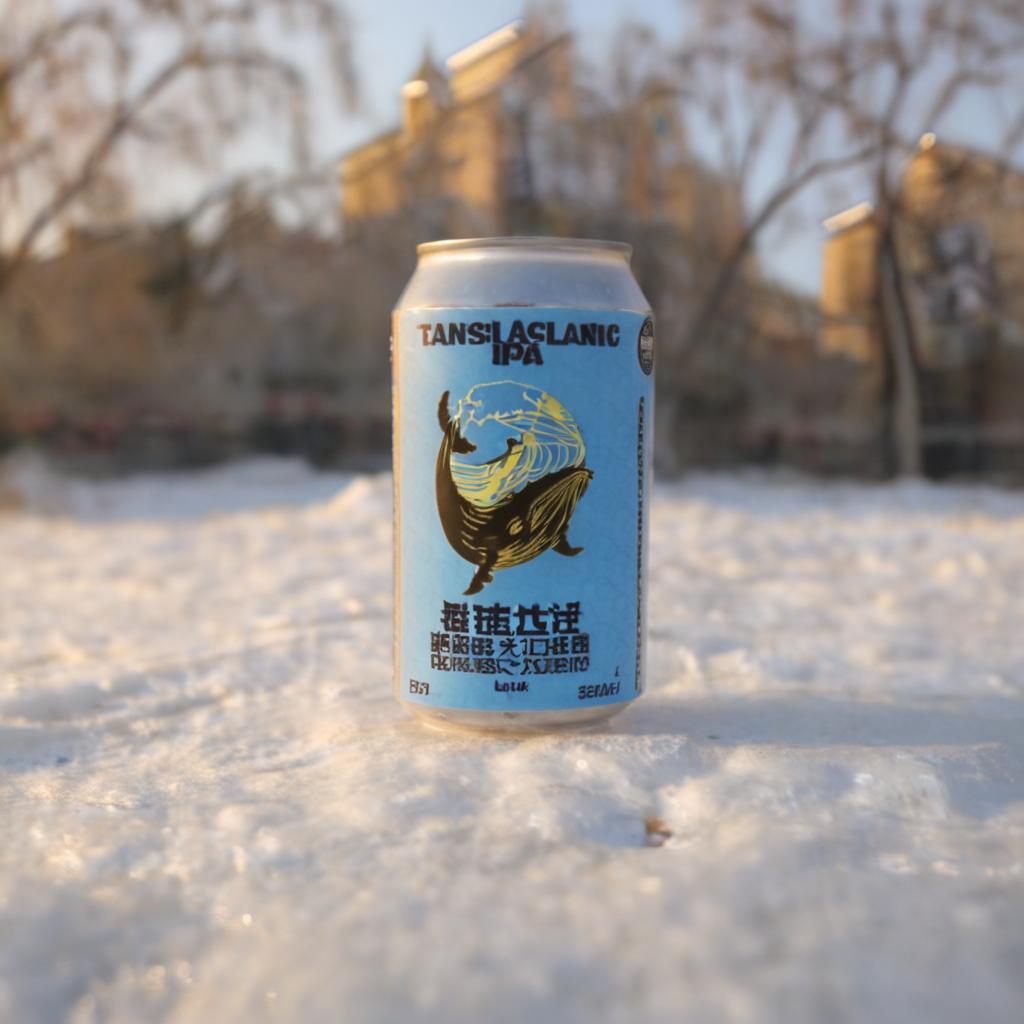}
& \includegraphics[width=2.3cm]{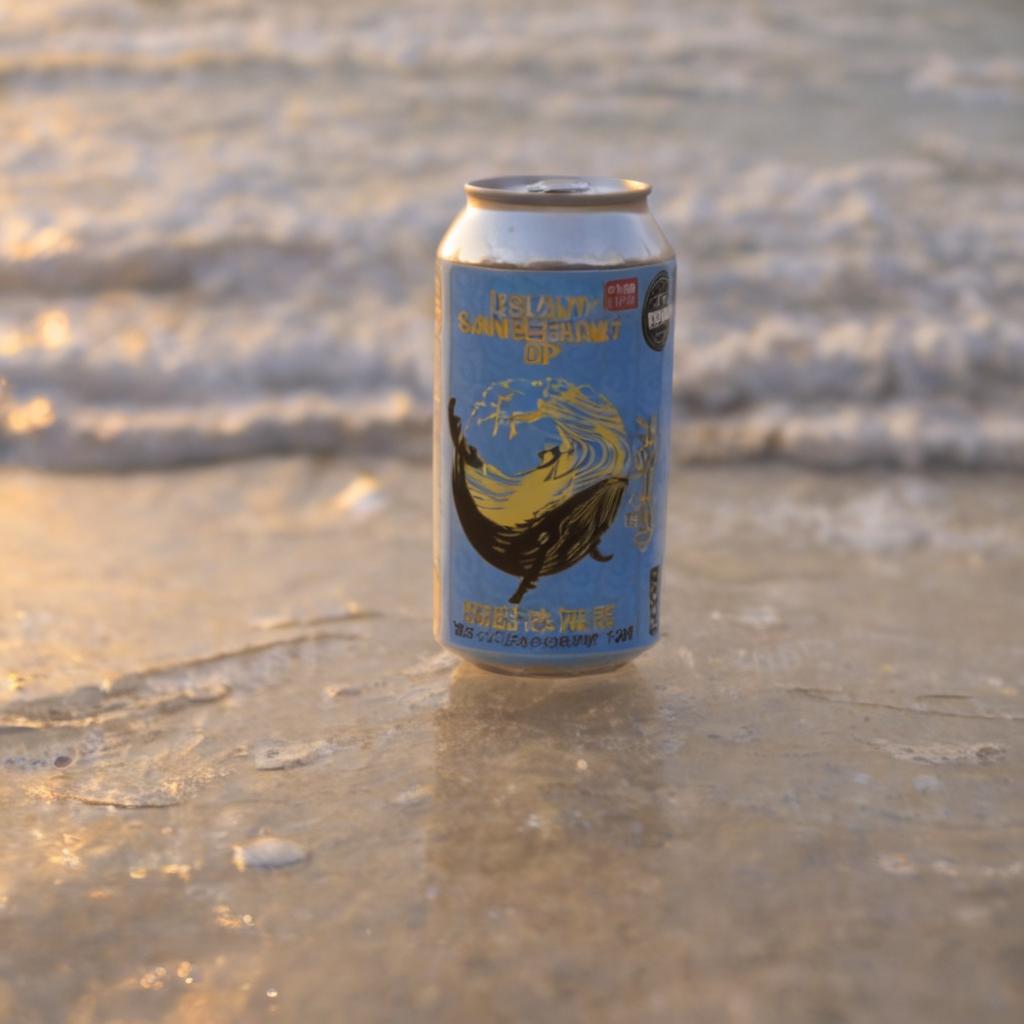}
\\
\end{tabular}%
}
\caption{Images generated using SDXL backbone for the ``can" subject.  The original subject is present on the top left. 
}
\label{fig:qualitative_comparison_SDXL_can}
\end{figure}

\begin{figure}[!h]
\centering
\renewcommand{\arraystretch}{1.2}
\setlength{\tabcolsep}{2pt}
\resizebox{\linewidth}{!}{%
\begin{tabular}{
    >{\centering\arraybackslash}m{2cm}
    >{\centering\arraybackslash}m{2.5cm} 
    >{\centering\arraybackslash}m{2.5cm} 
    >{\centering\arraybackslash}m{2.5cm} 
    >{\centering\arraybackslash}m{2.5cm} 
    >{\centering\arraybackslash}m{2.5cm} 
}
\includegraphics[width=2cm]{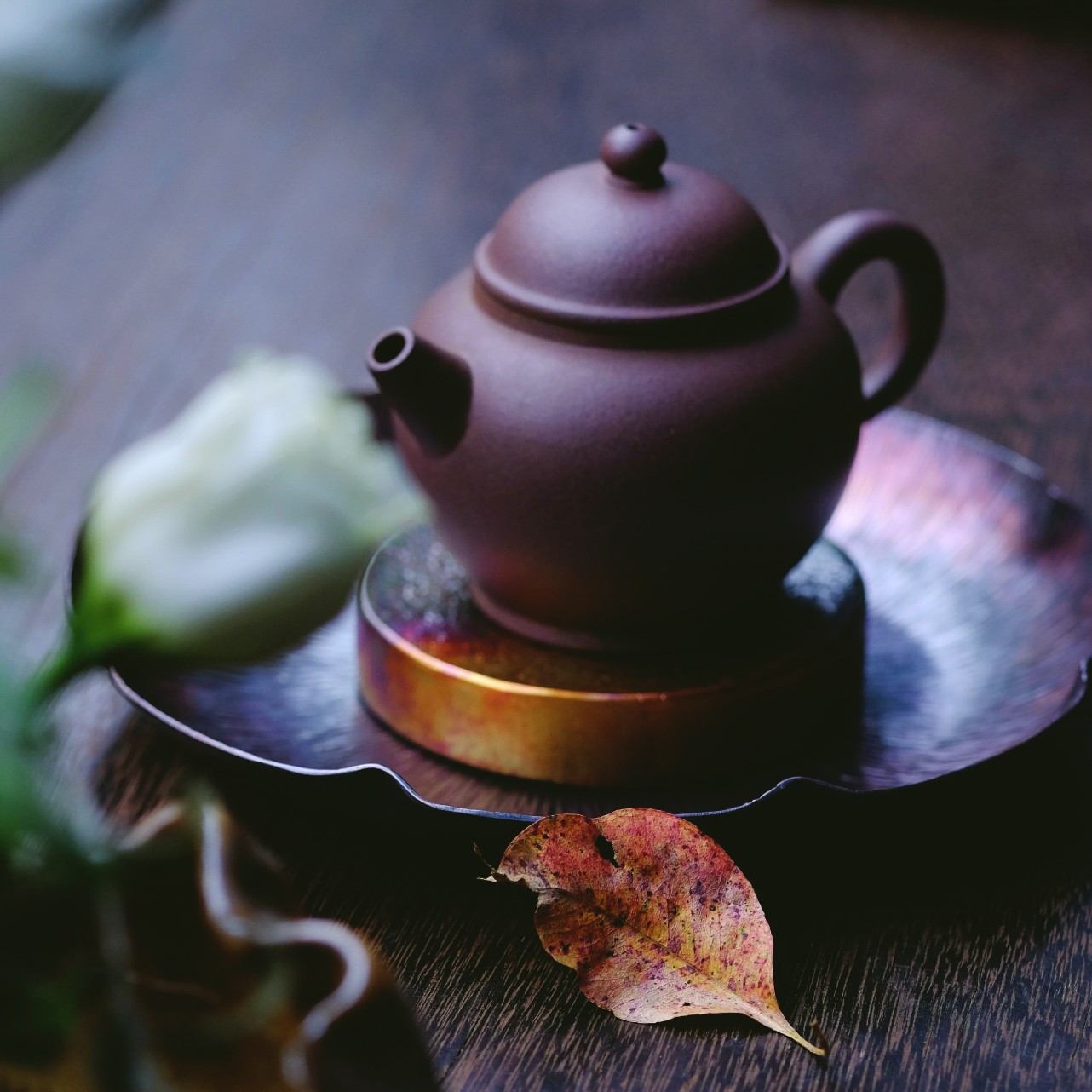}
& \textbf{``a modern minimalistic \textcolor{red}{k} teapot  on a white surface''} 
& \textbf{``a glowing \textcolor{red}{k} teapot in the dark''} 
& \textbf{``a \textcolor{red}{k} teapot floating in crystal clear water''} 
& \textbf{``a vintage \textcolor{red}{k} teapot on an antique table  ''} 
& \textbf{``a \textcolor{red}{k} teapot on a glass table with reflections ''} \\[2mm]

\scriptsize Rank 8
& \includegraphics[width=2.3cm]{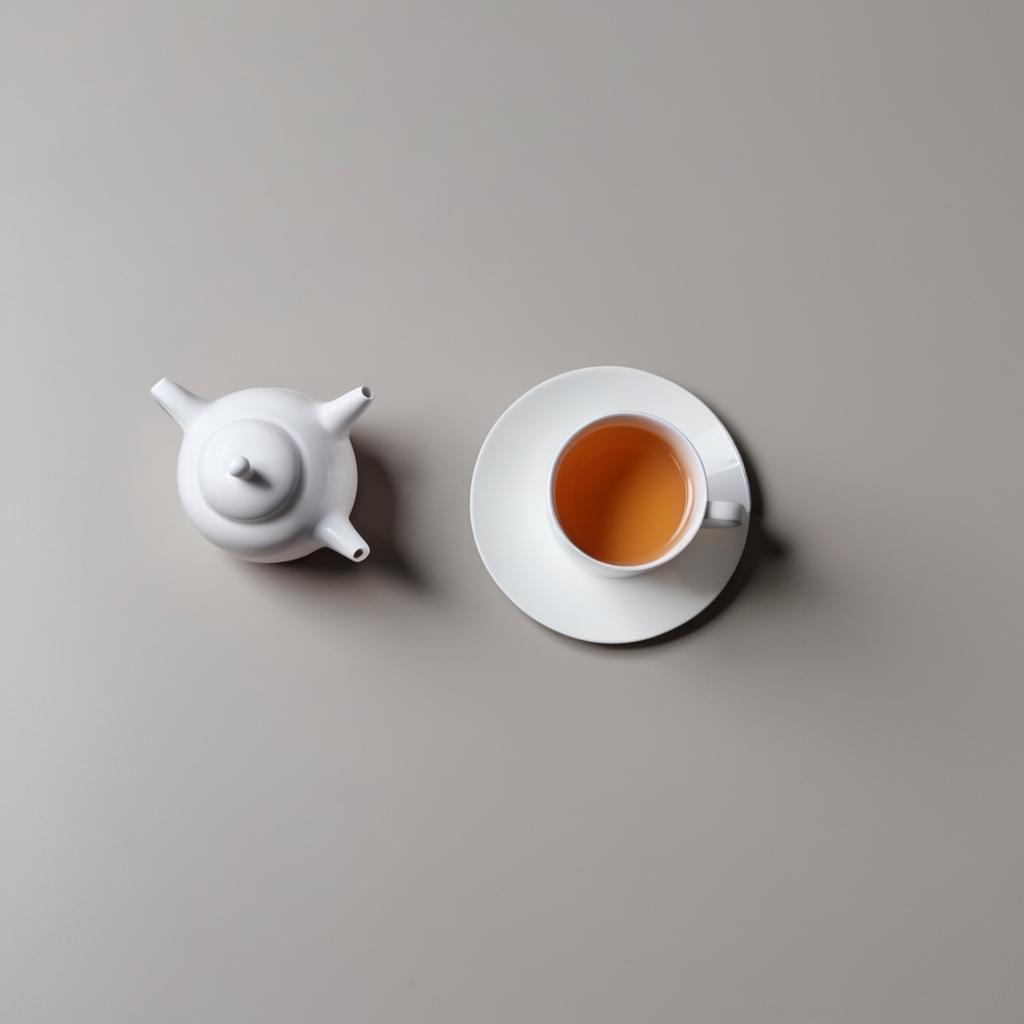}
& \includegraphics[width=2.3cm]{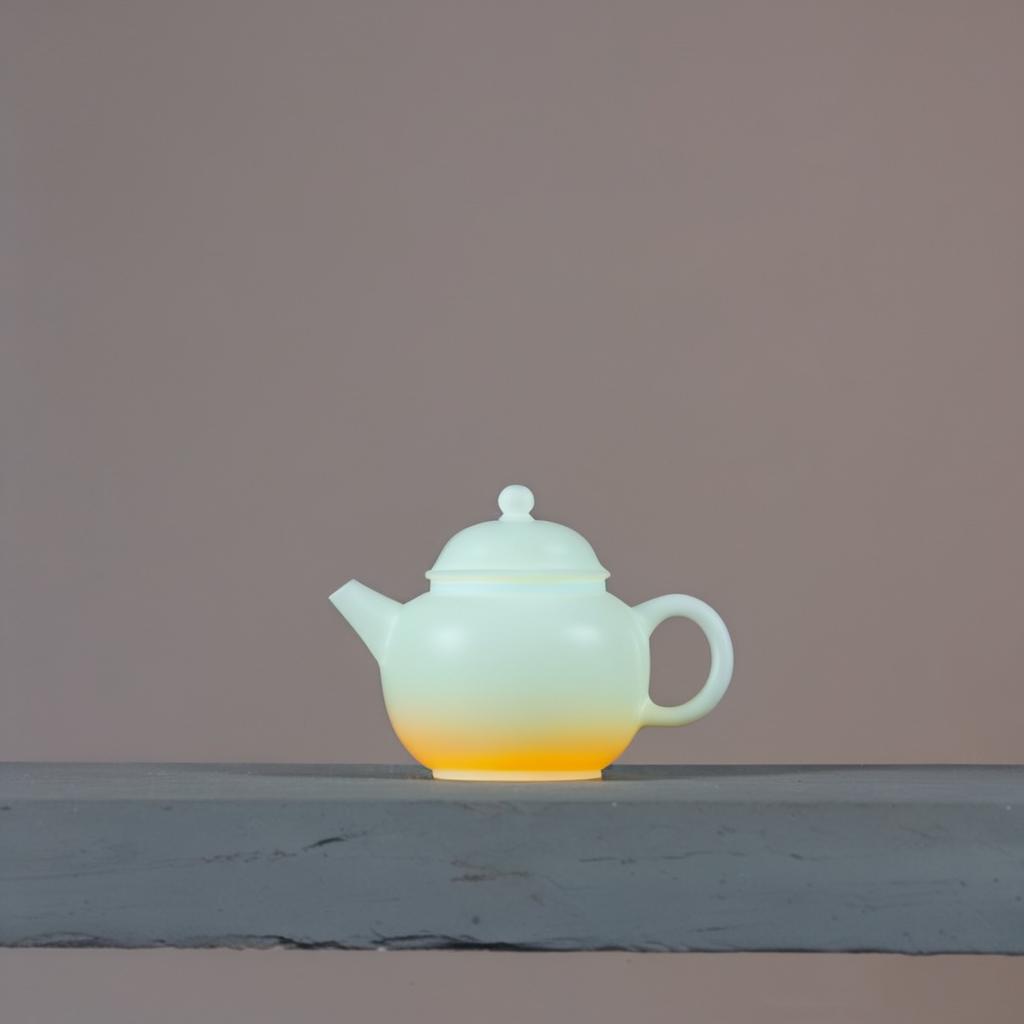}
& \includegraphics[width=2.3cm]{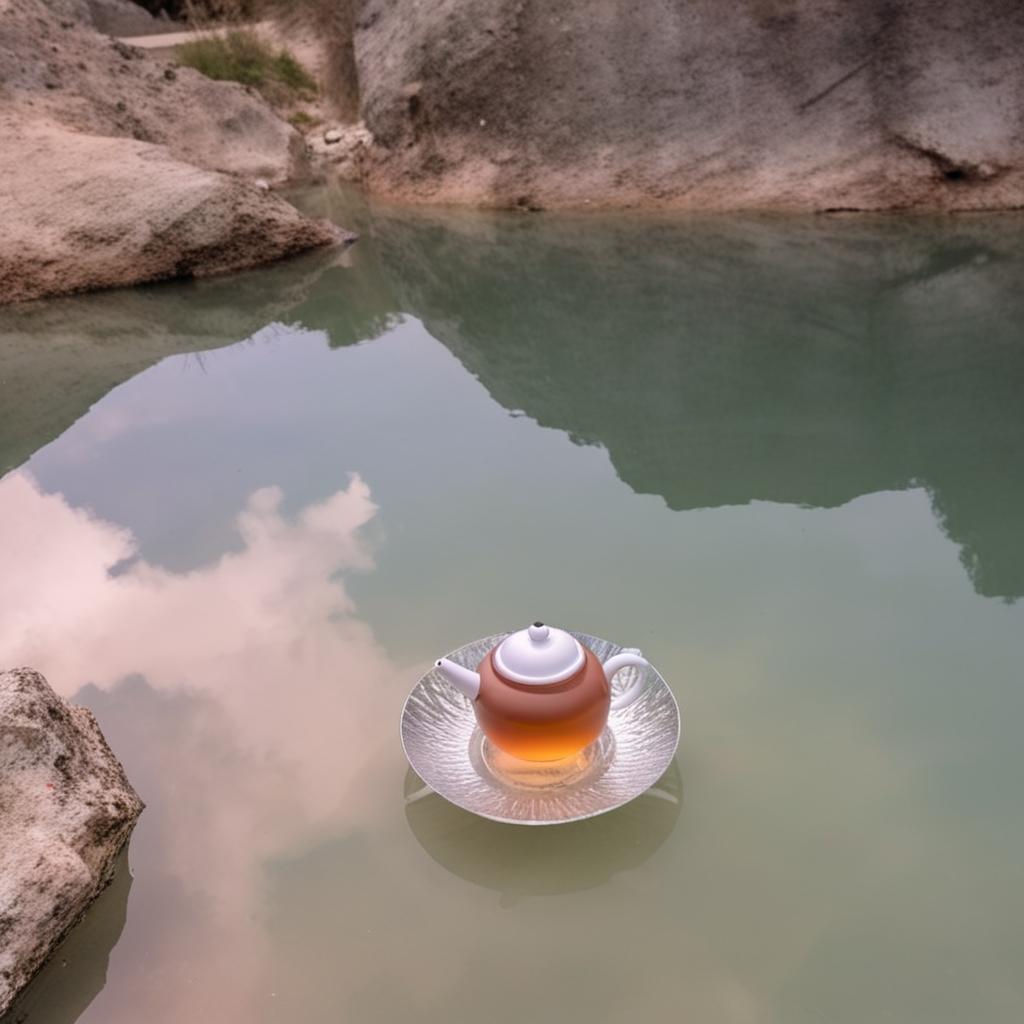}
& \includegraphics[width=2.3cm]{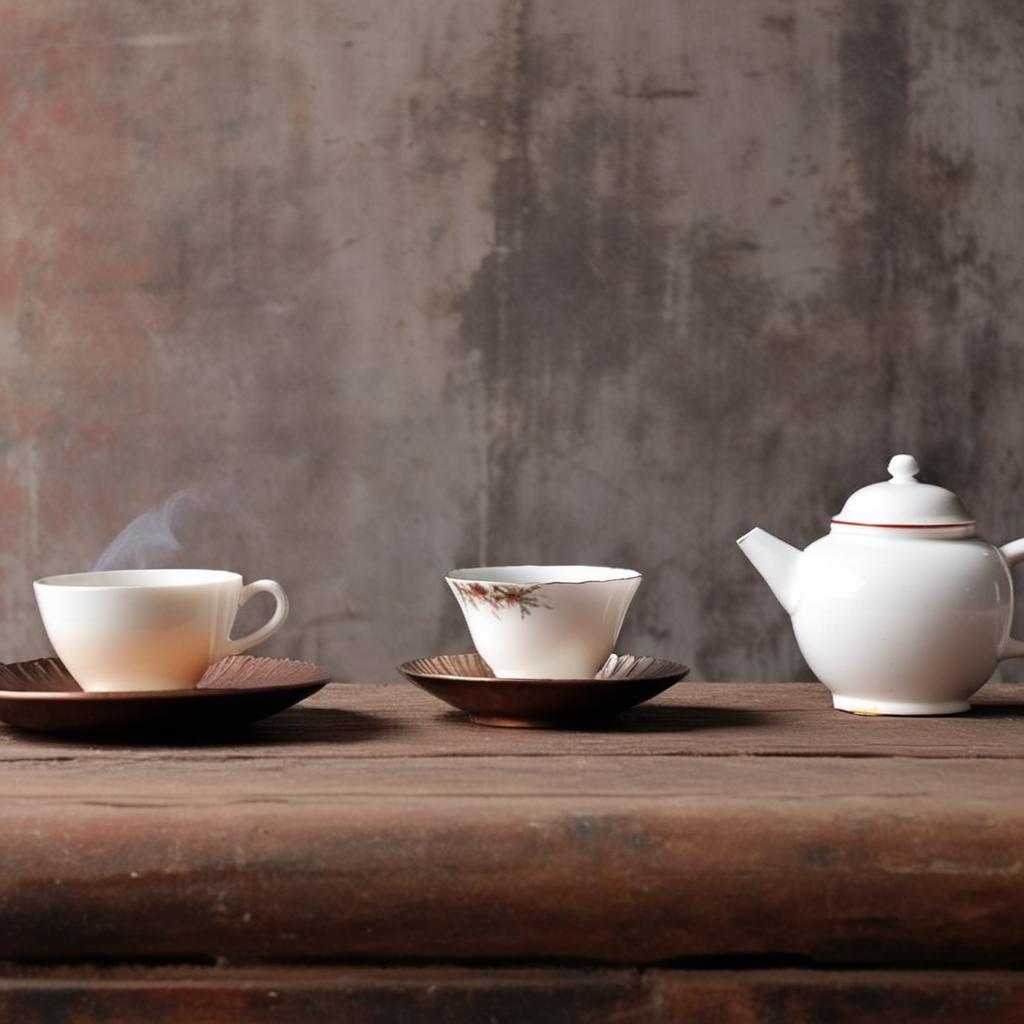}
& \includegraphics[width=2.3cm]{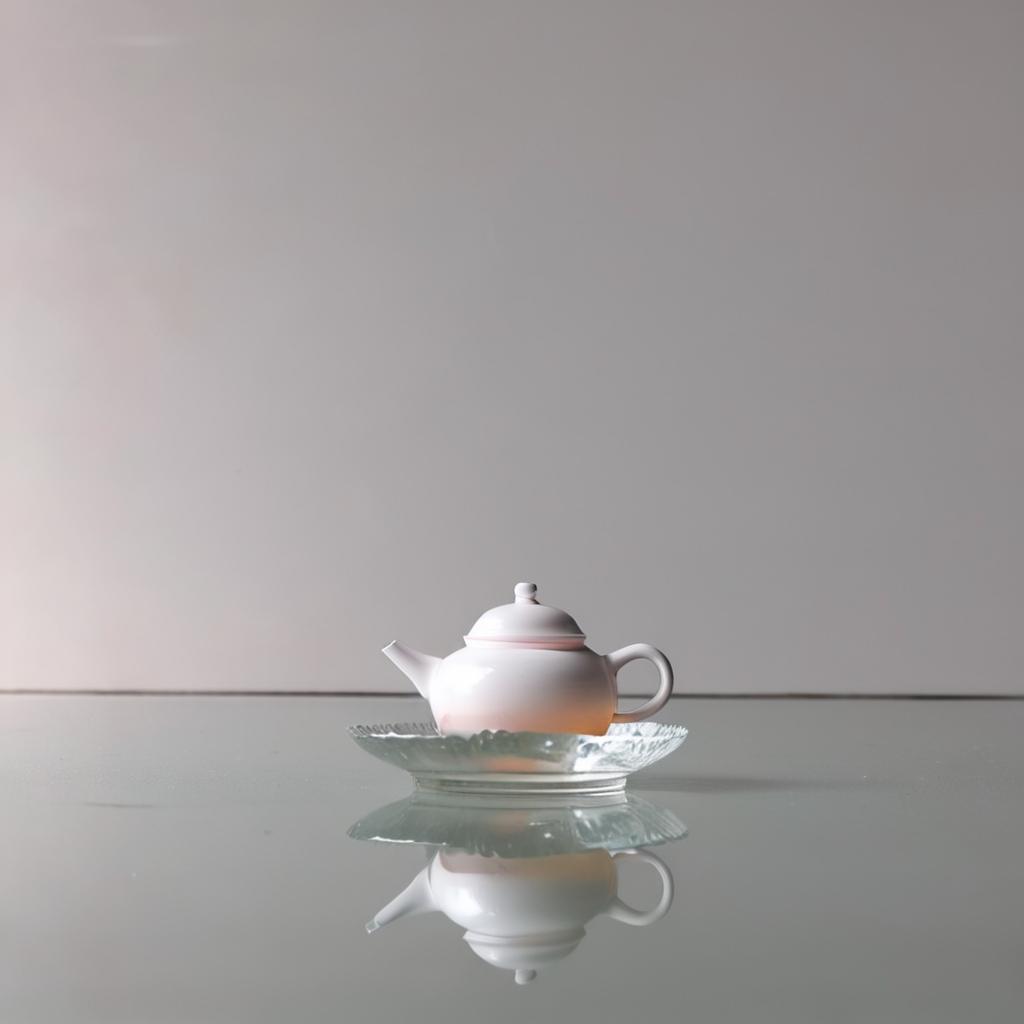}
\\[1mm]

\scriptsize Rank 64
& \includegraphics[width=2.3cm]{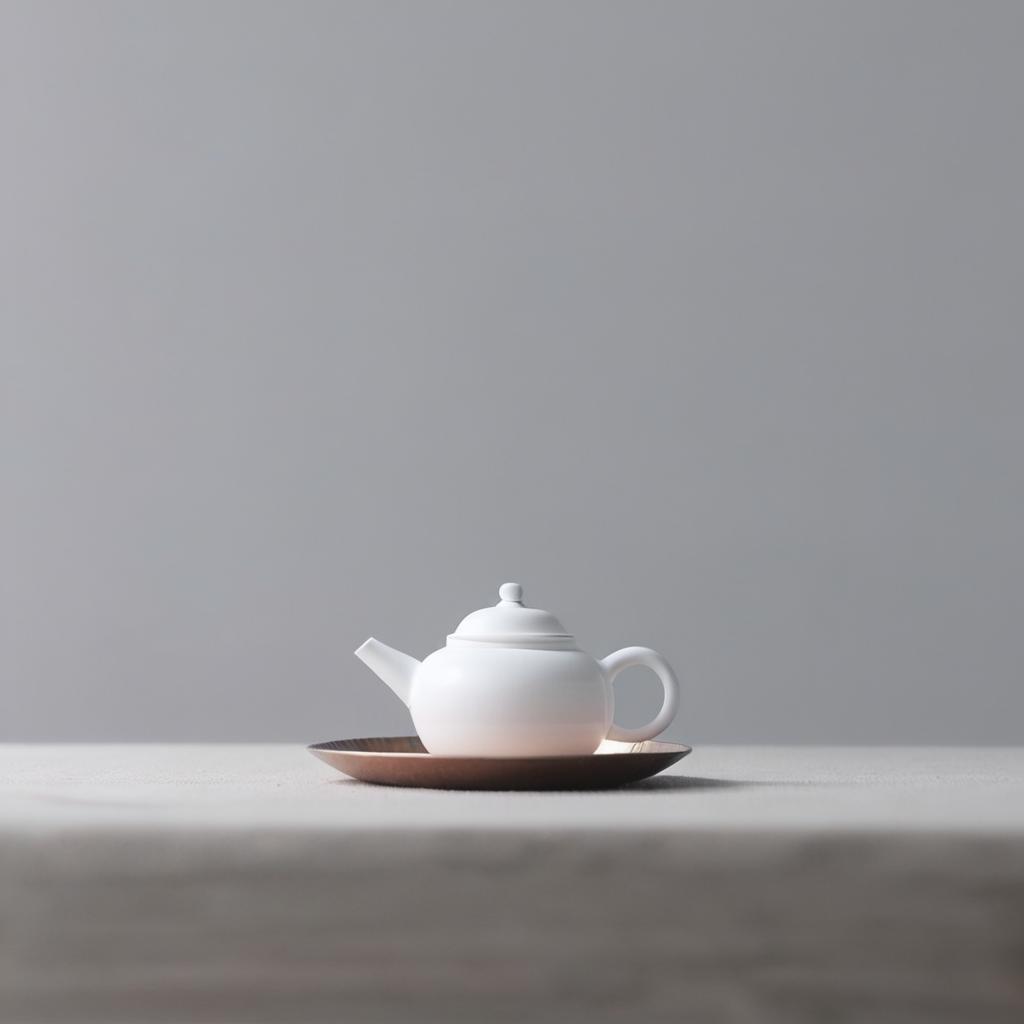}
& \includegraphics[width=2.3cm]{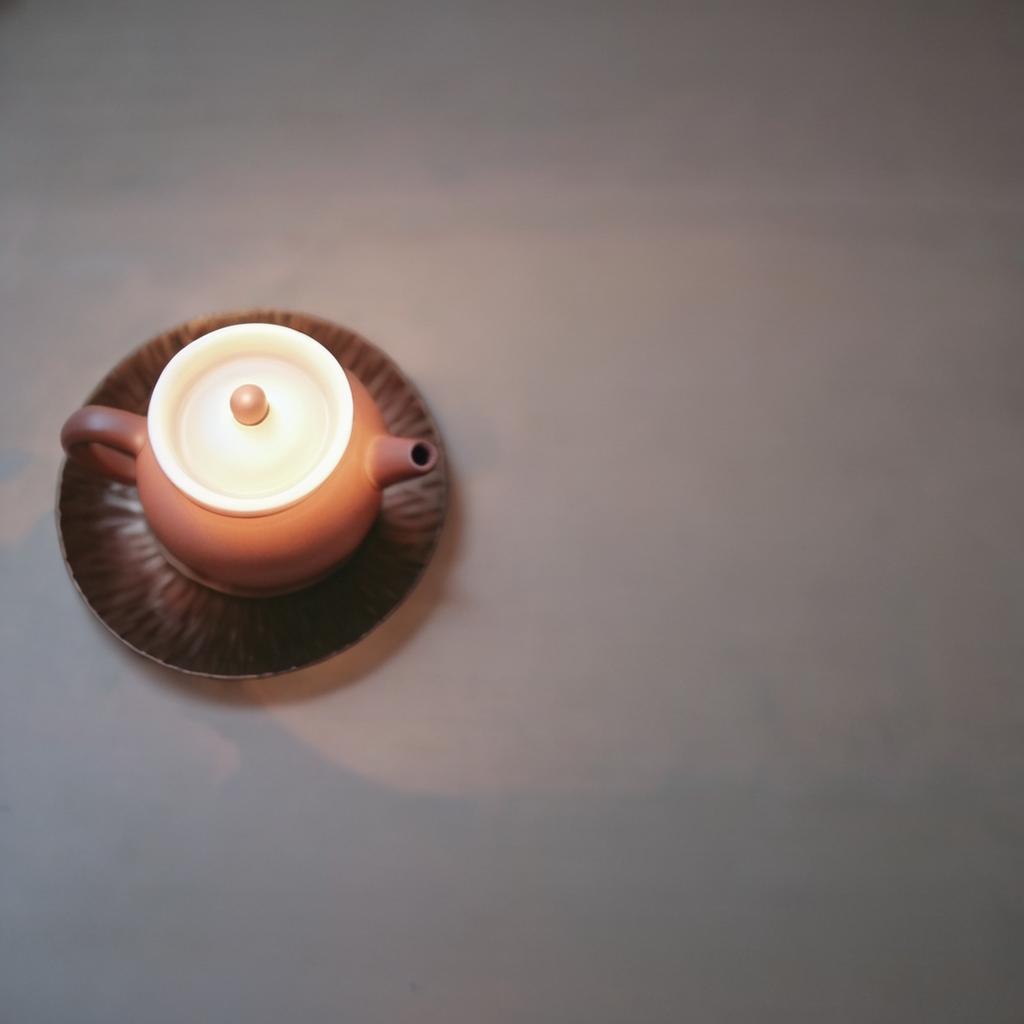}
& \includegraphics[width=2.3cm]{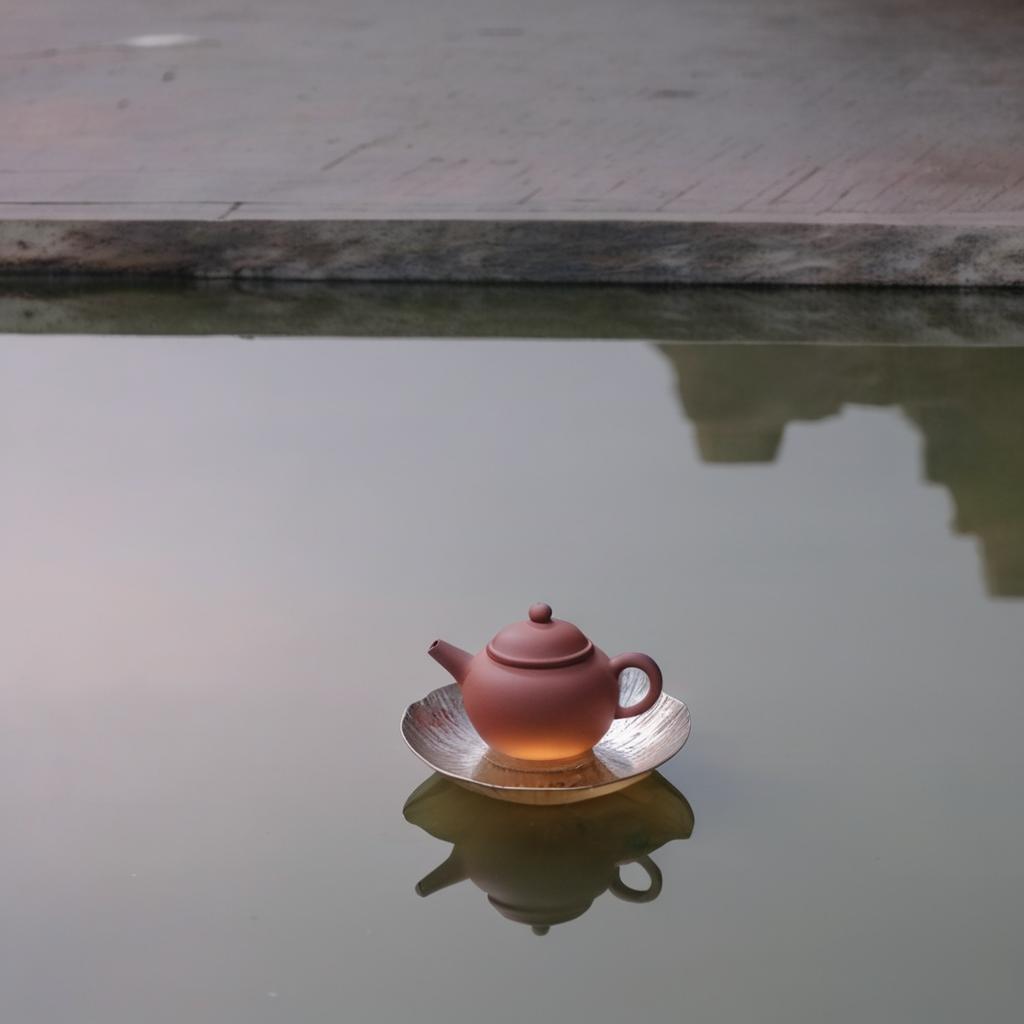}
& \includegraphics[width=2.3cm]{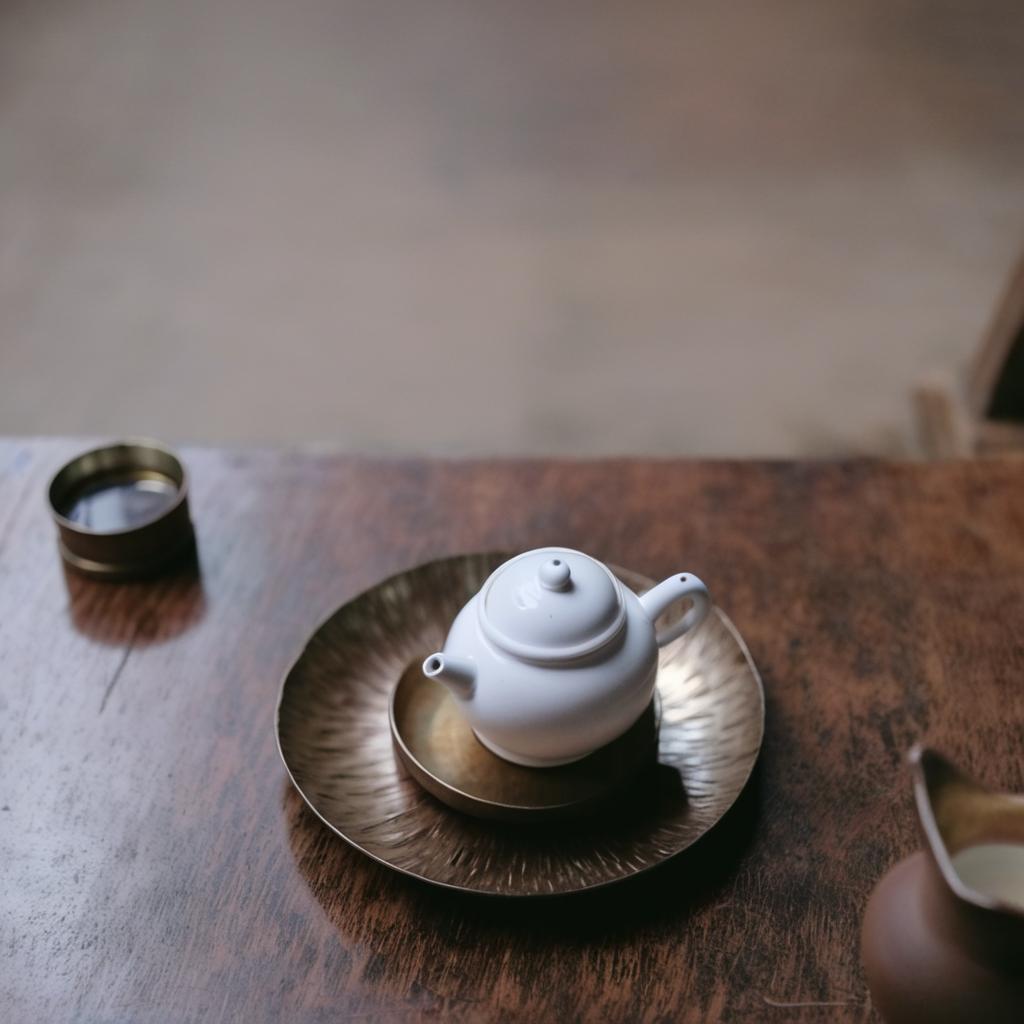}
& \includegraphics[width=2.3cm]{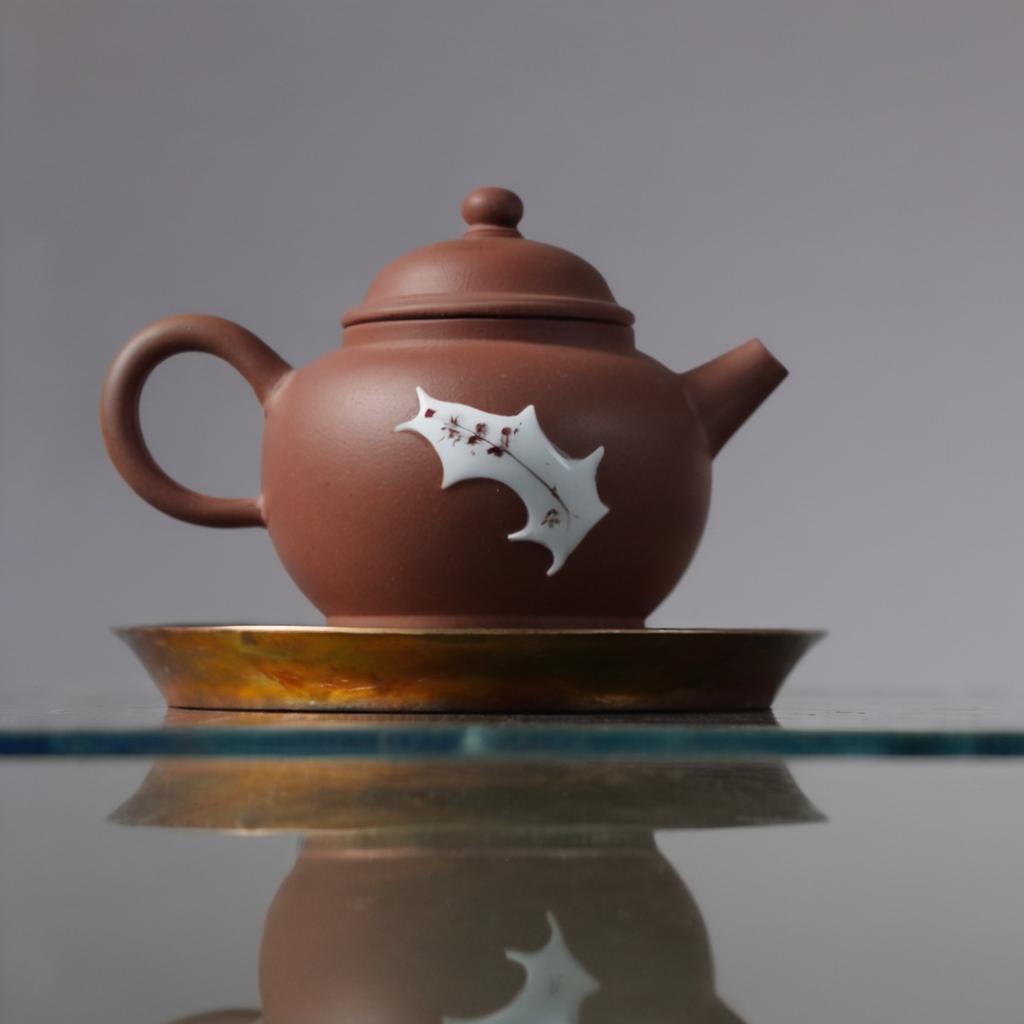}
\\[1mm]

\scriptsize Rank 512
& \includegraphics[width=2.3cm]{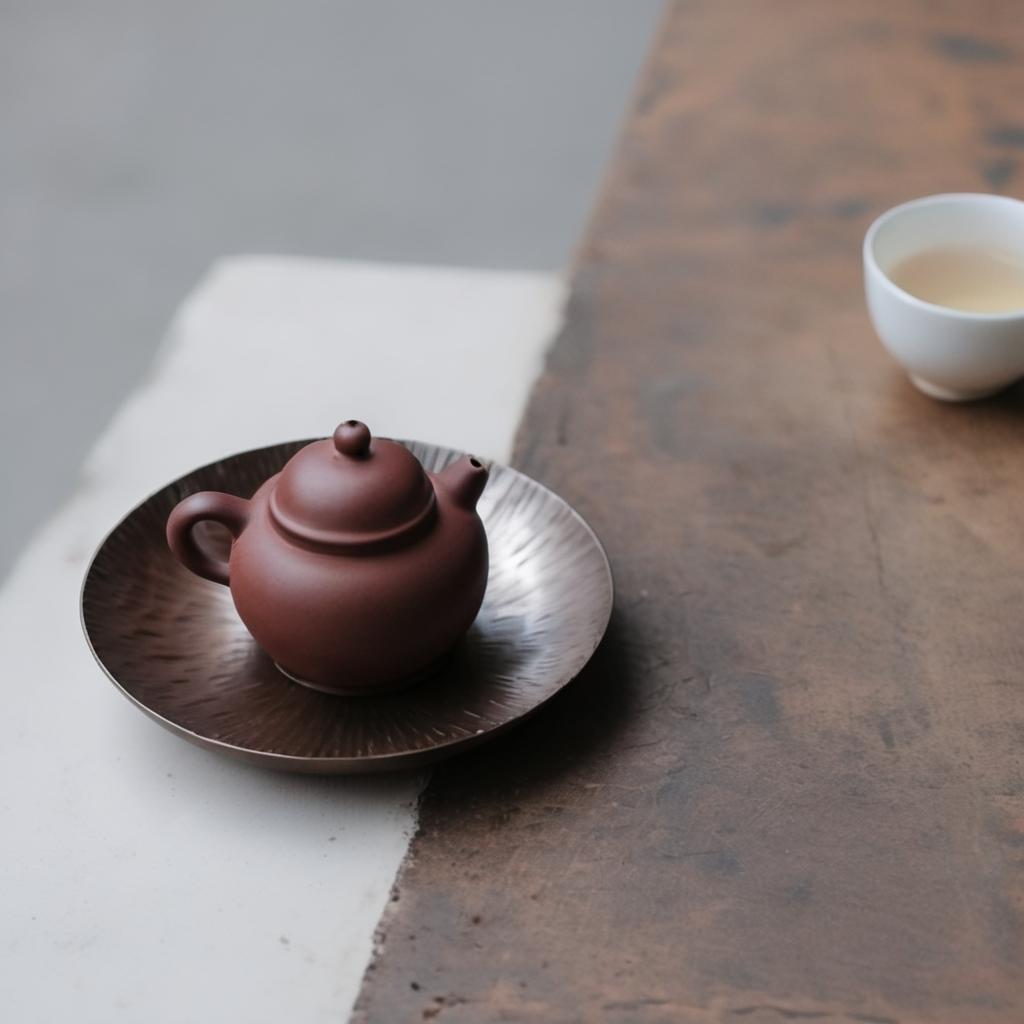}
& \includegraphics[width=2.3cm]{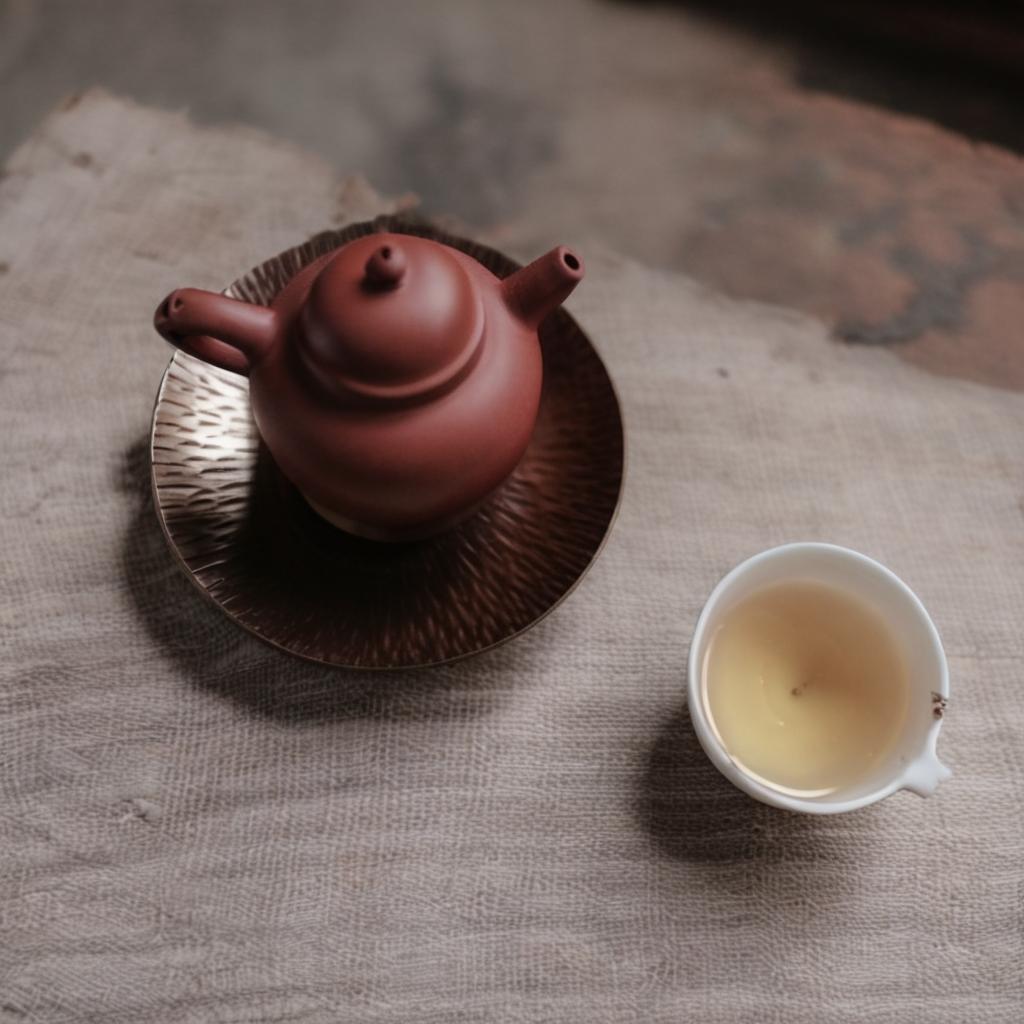}
& \includegraphics[width=2.3cm]{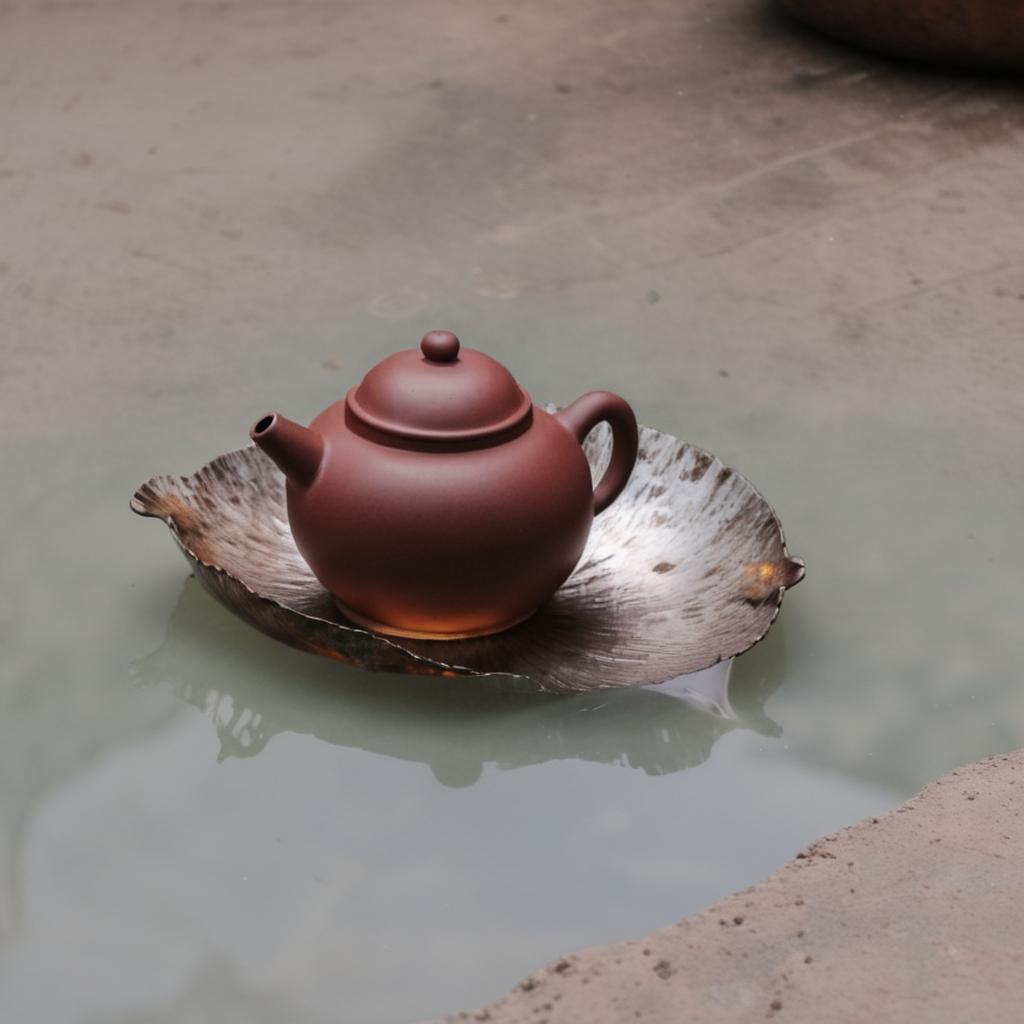}
& \includegraphics[width=2.3cm]{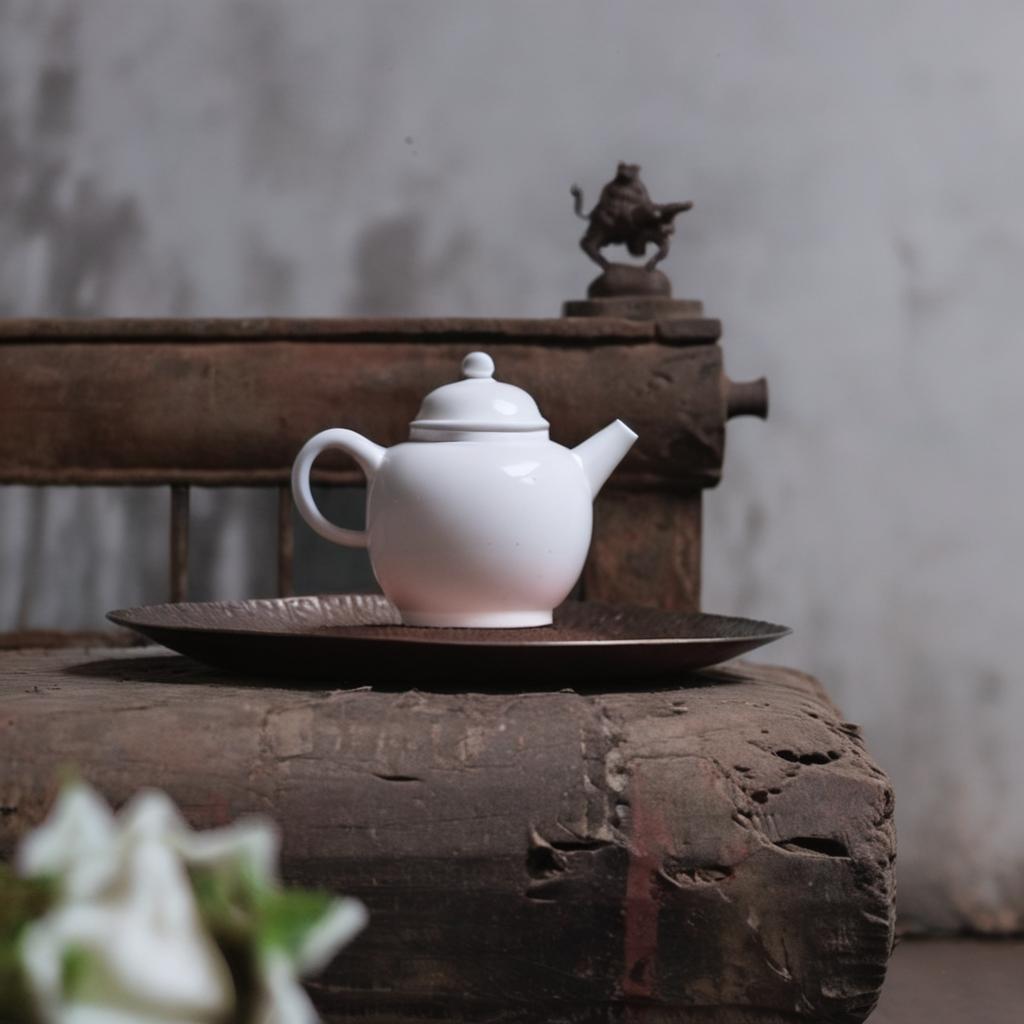}
& \includegraphics[width=2.3cm]{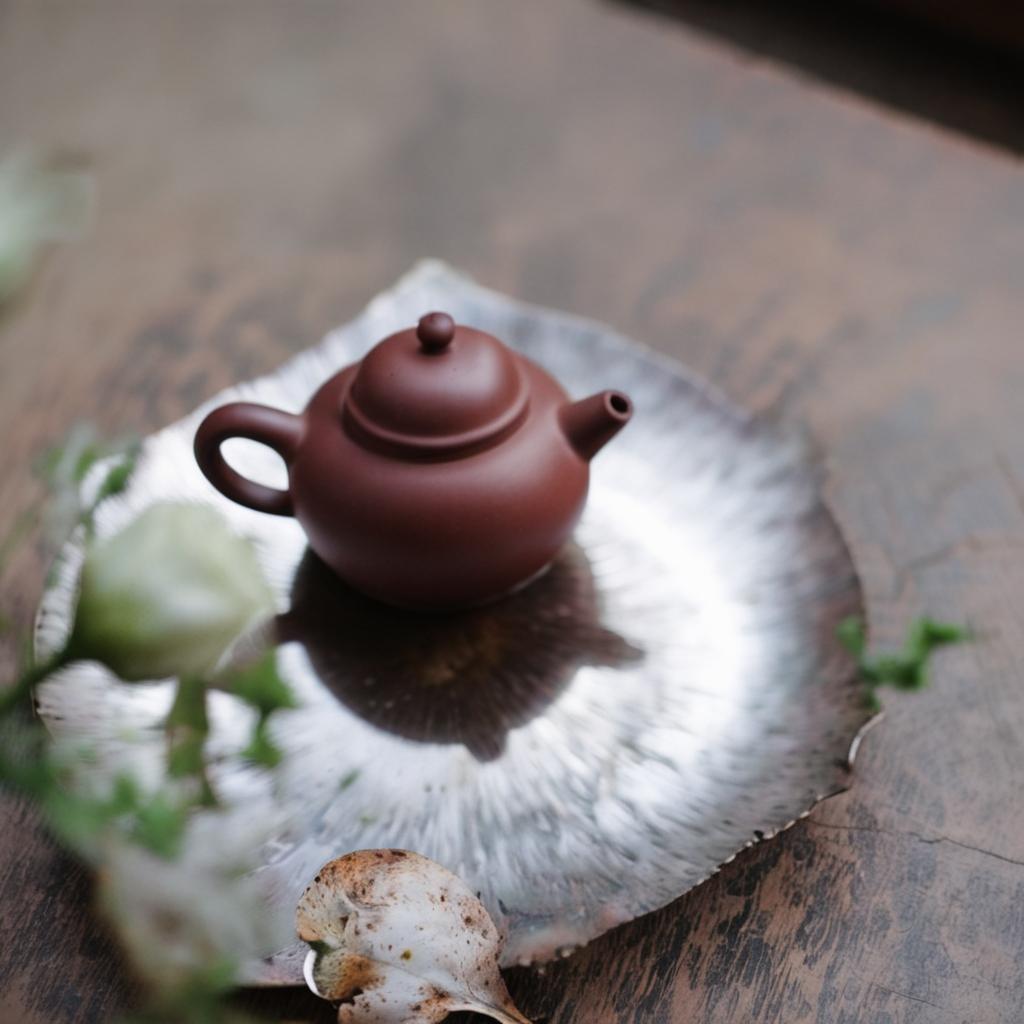}
\\[1mm]

\scriptsize \method
& \includegraphics[width=2.3cm]{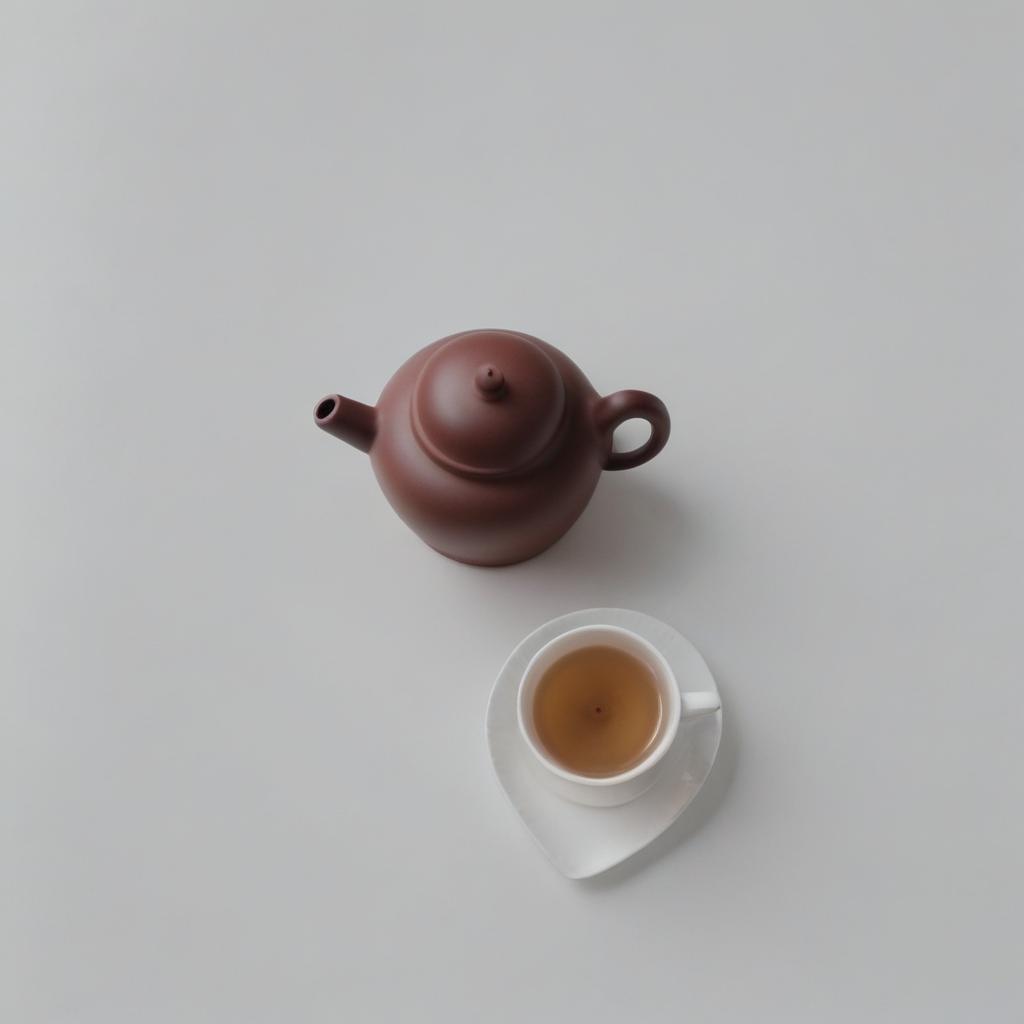}
& \includegraphics[width=2.3cm]{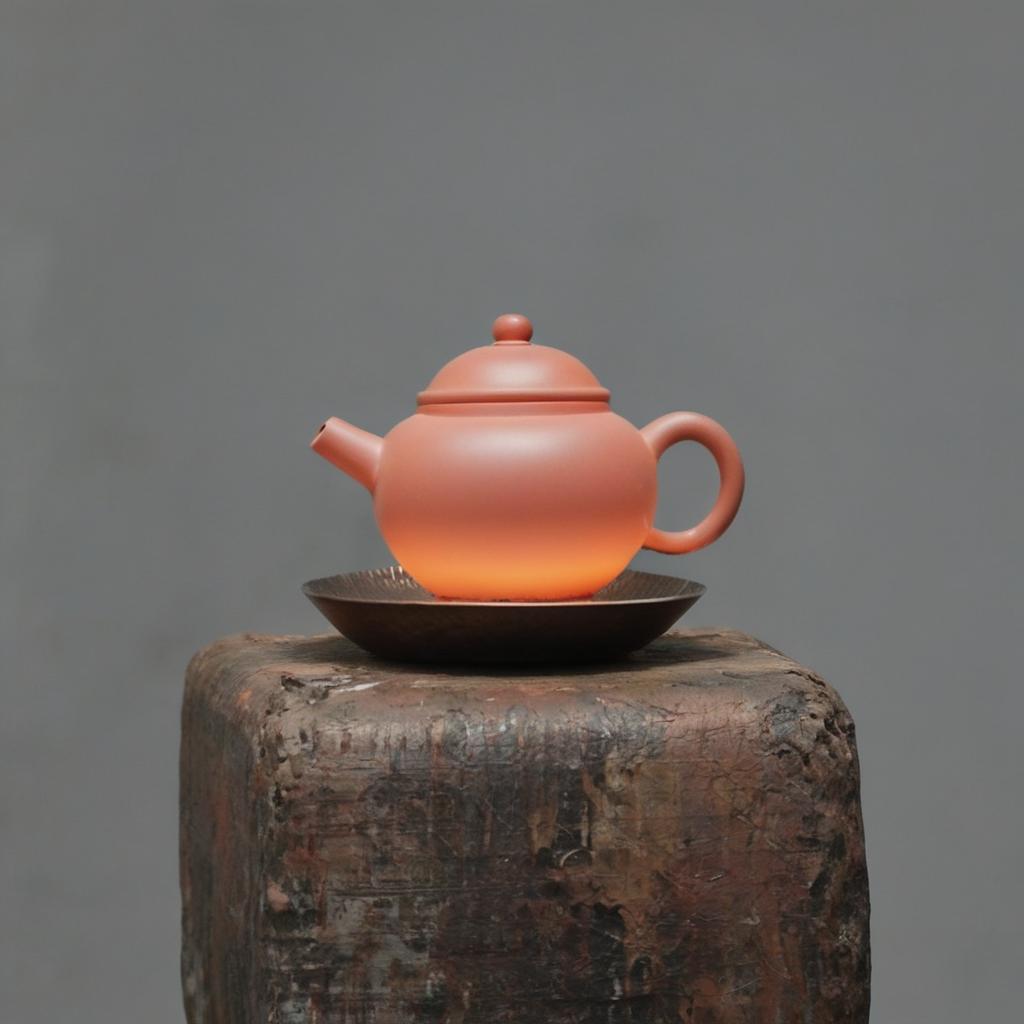}
& \includegraphics[width=2.3cm]{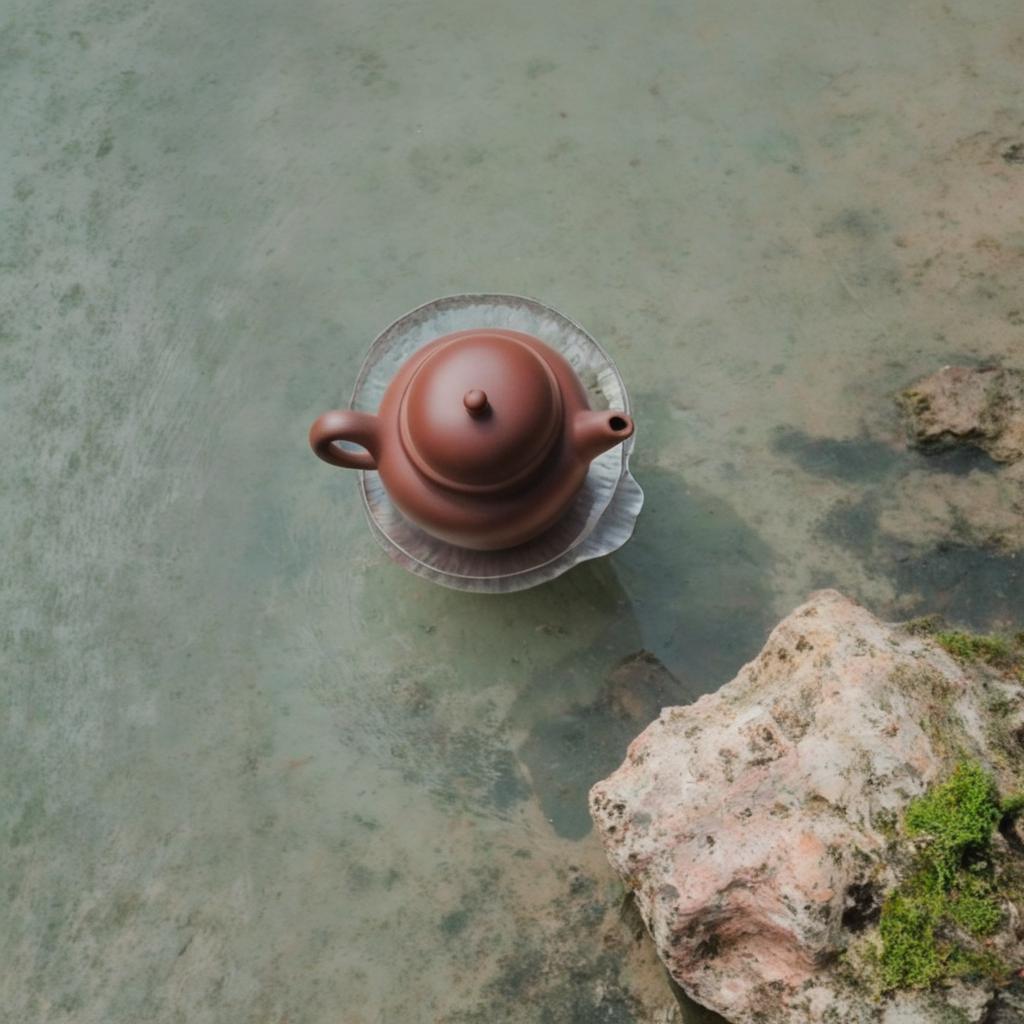}
& \includegraphics[width=2.3cm]{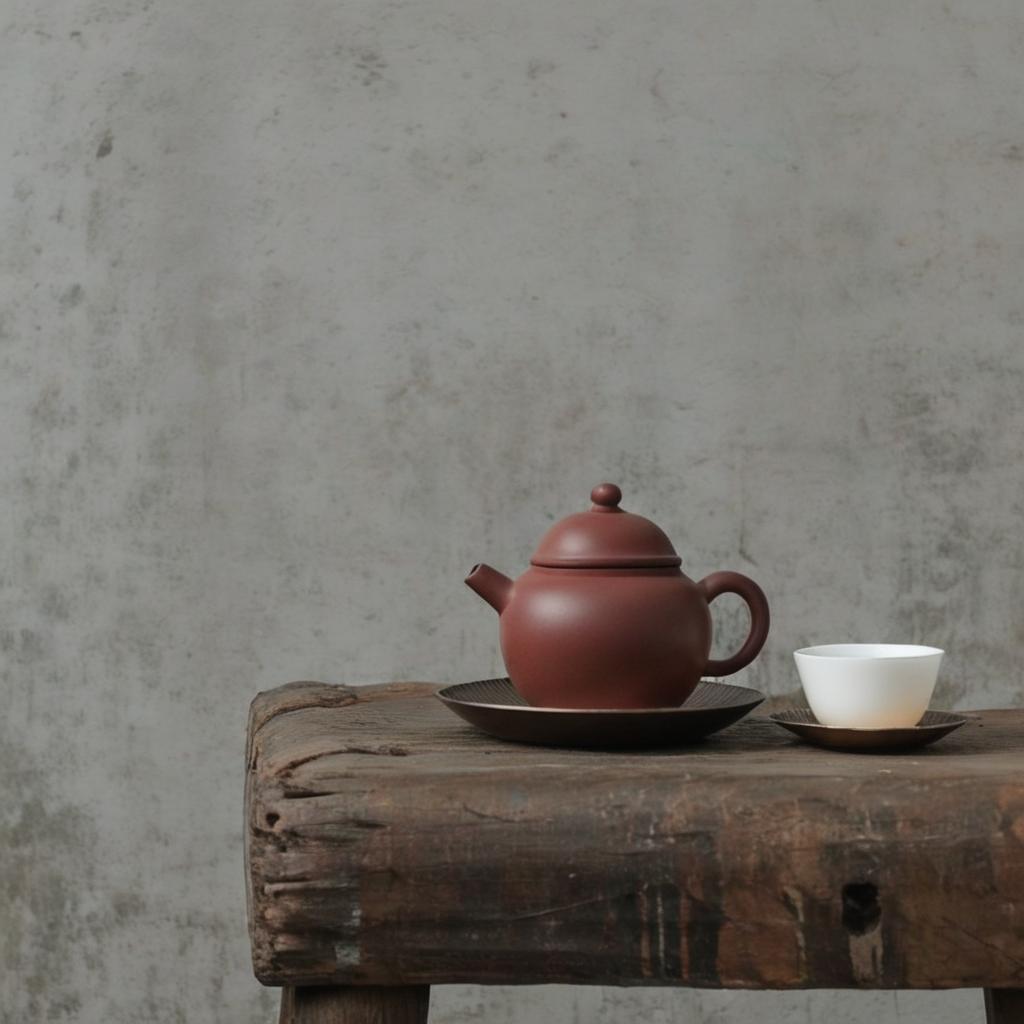}
& \includegraphics[width=2.3cm]{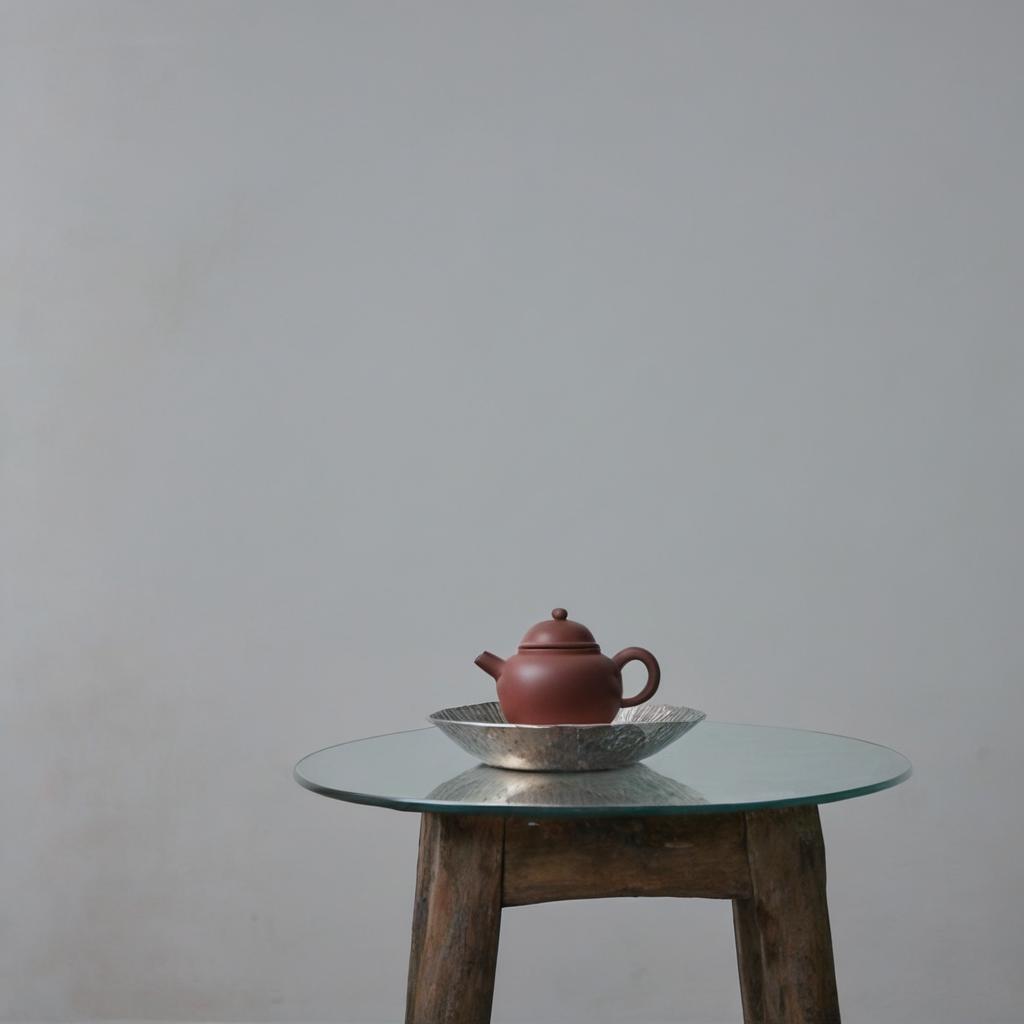}
\\
\end{tabular}%
}
\caption{Images generated using SDXL backbone for the ``teapot" subject.  The original subject is present on the top left. 
}
\label{fig:qualitative_comparison_SDXL_teapot}
\end{figure}

\Cref{fig:qualitative_comparison_SDXL_teapot,fig:qualitative_comparison_SDXL_can} present additional qualitative comparisons using SDXL for the \textit{teapot} and \textit{can} subjects. Notably, our approach is the only method that consistently reproduces the label on the can across all generated images, demonstrating superior fidelity to fine-grained subject details.

\begin{figure}[h!]
    \centering

    \begin{subfigure}[t]{0.4\textwidth}
        \centering
        \includegraphics[width=\textwidth]{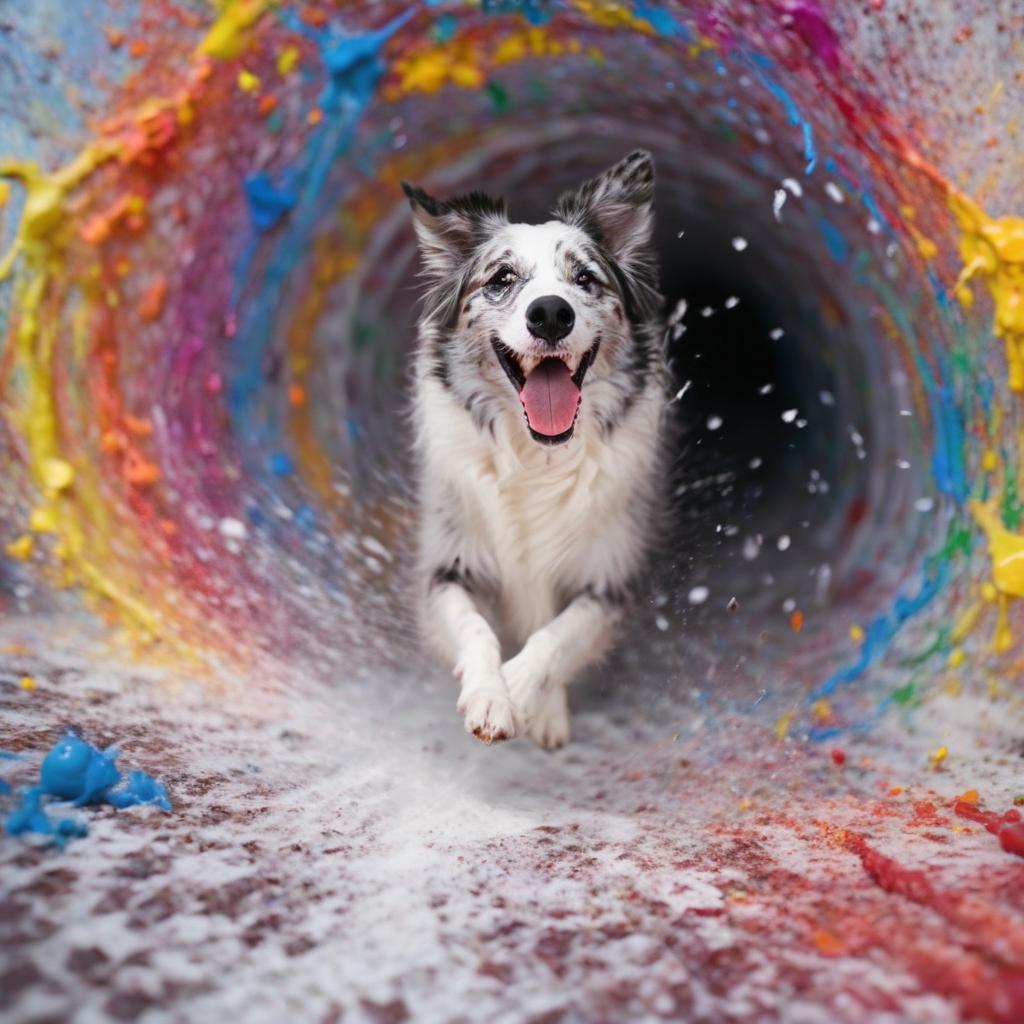}
        \caption*{\scriptsize``a  \textcolor{red}{k} dog racing through an exploding tunnel of colorful paint splashes, motion blur, frozen droplets mid-air, low angle high-speed shot.''}
    \end{subfigure}
    \begin{subfigure}[t]{0.4\textwidth}
        \centering
        \includegraphics[width=\textwidth]{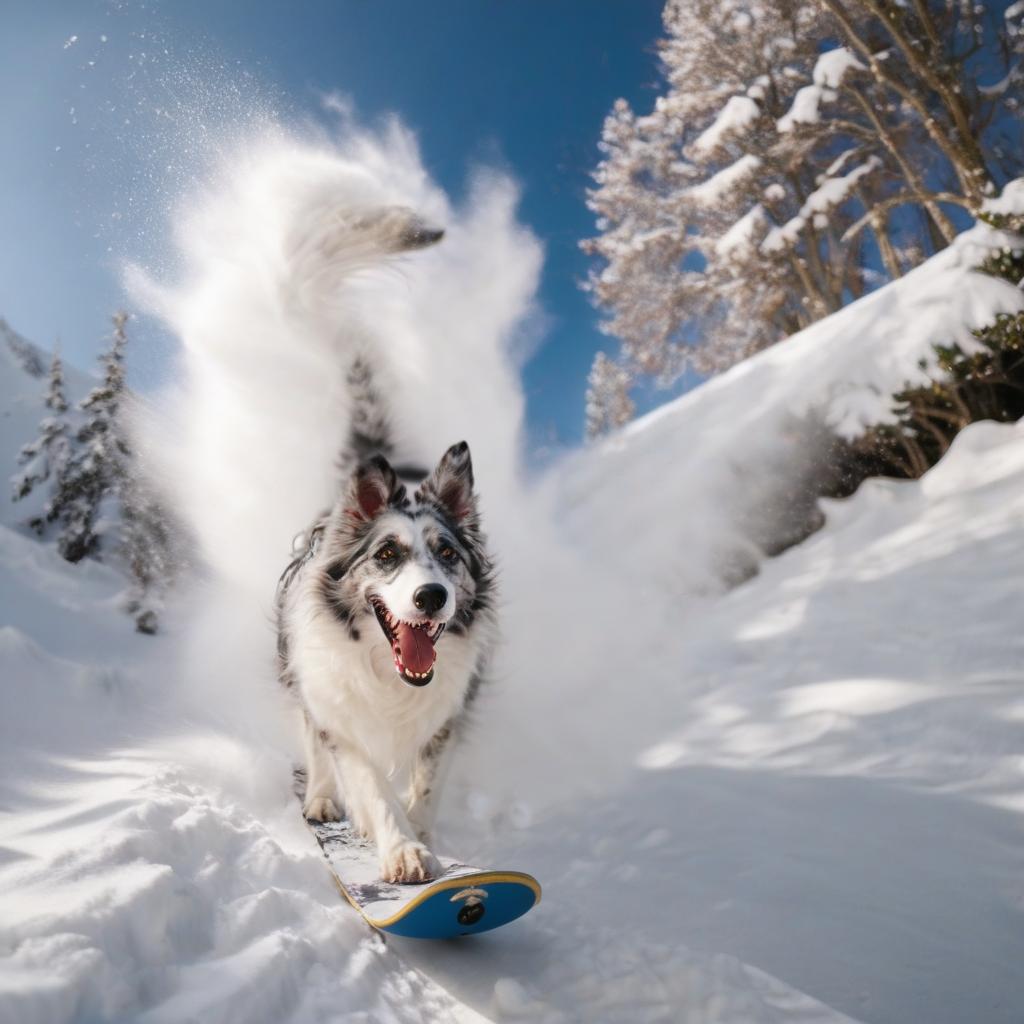}
        \caption*{\scriptsize ``a  \textcolor{red}{k} dog launching off a snowy mountain peak on a snowboard, massive powder explosion, crisp blue sky, low angle action shot.''}
    \end{subfigure}

    \vspace{1em}

    \begin{subfigure}[t]{0.4\textwidth}
        \centering
        \includegraphics[width=\textwidth]{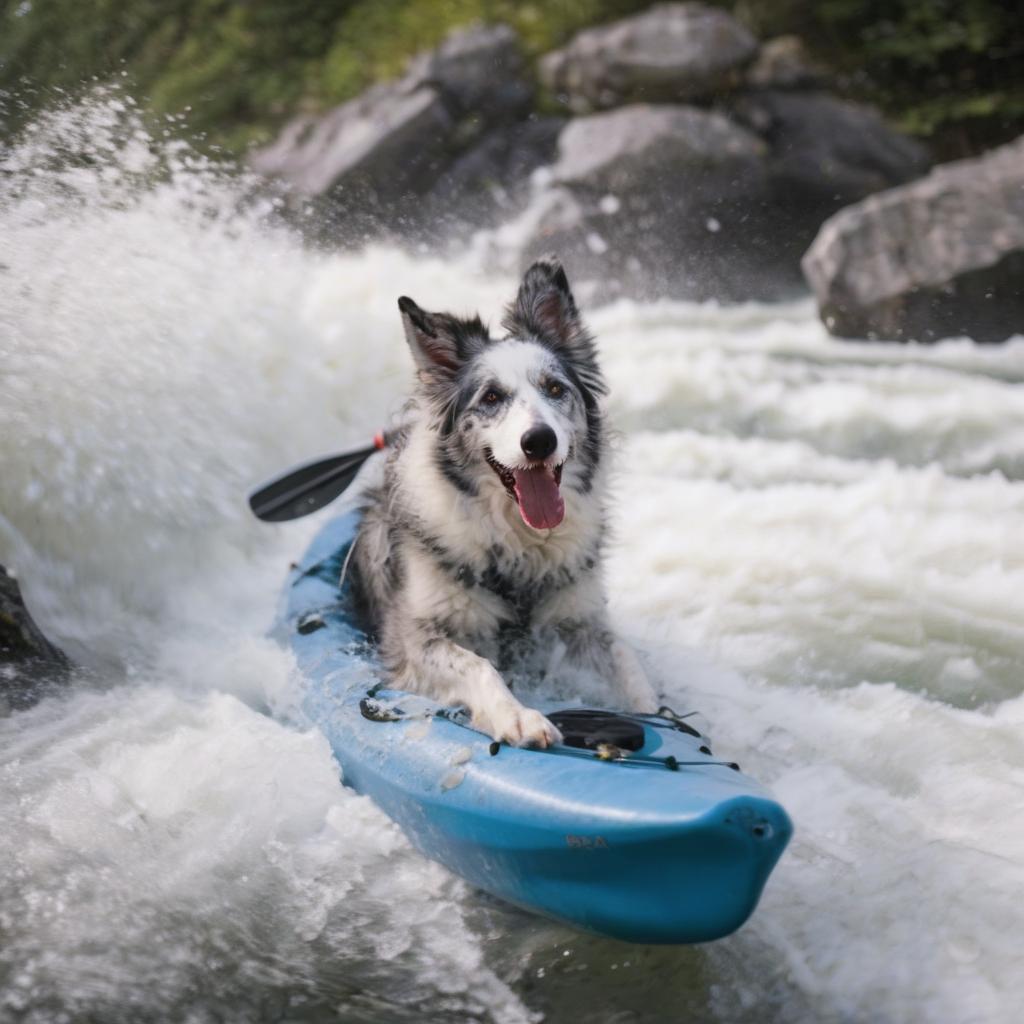}
        \caption*{\scriptsize ``a  \textcolor{red}{k} dog kayaking through a raging white-water rapid, water exploding around the boat, soaked fur, intense focus, action shot frozen mid-crash''.}
    \end{subfigure}
    \begin{subfigure}[t]{0.4\textwidth}
        \centering
        \includegraphics[width=\textwidth]{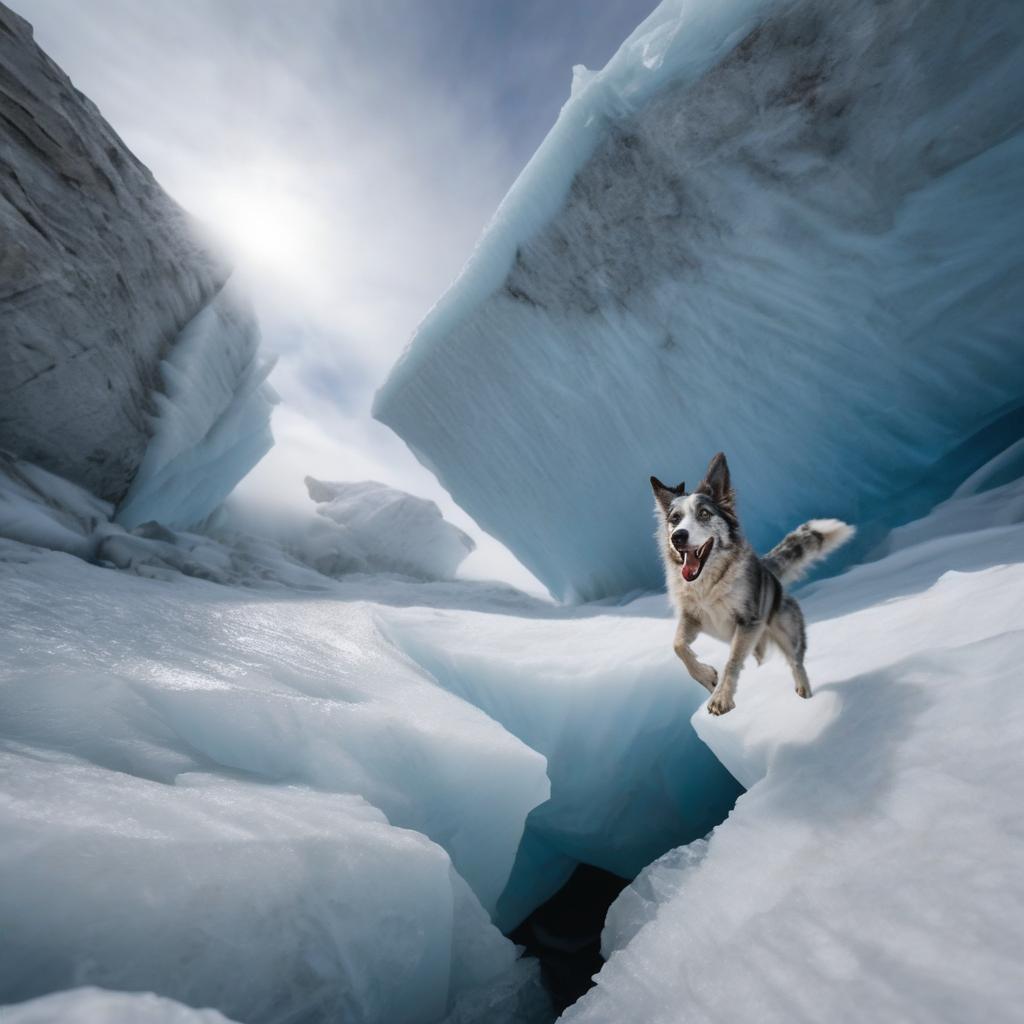}
        \caption*{\scriptsize ``a  \textcolor{red}{k} dog leaping between two glaciers over an icy blue crevasse, paws mid-air, frozen mist, dramatic arctic light, ultra-wide low angle''.}
    \end{subfigure}

    \vspace{1em}

    \begin{subfigure}[t]{0.4\textwidth}
        \centering
        \includegraphics[width=\textwidth]{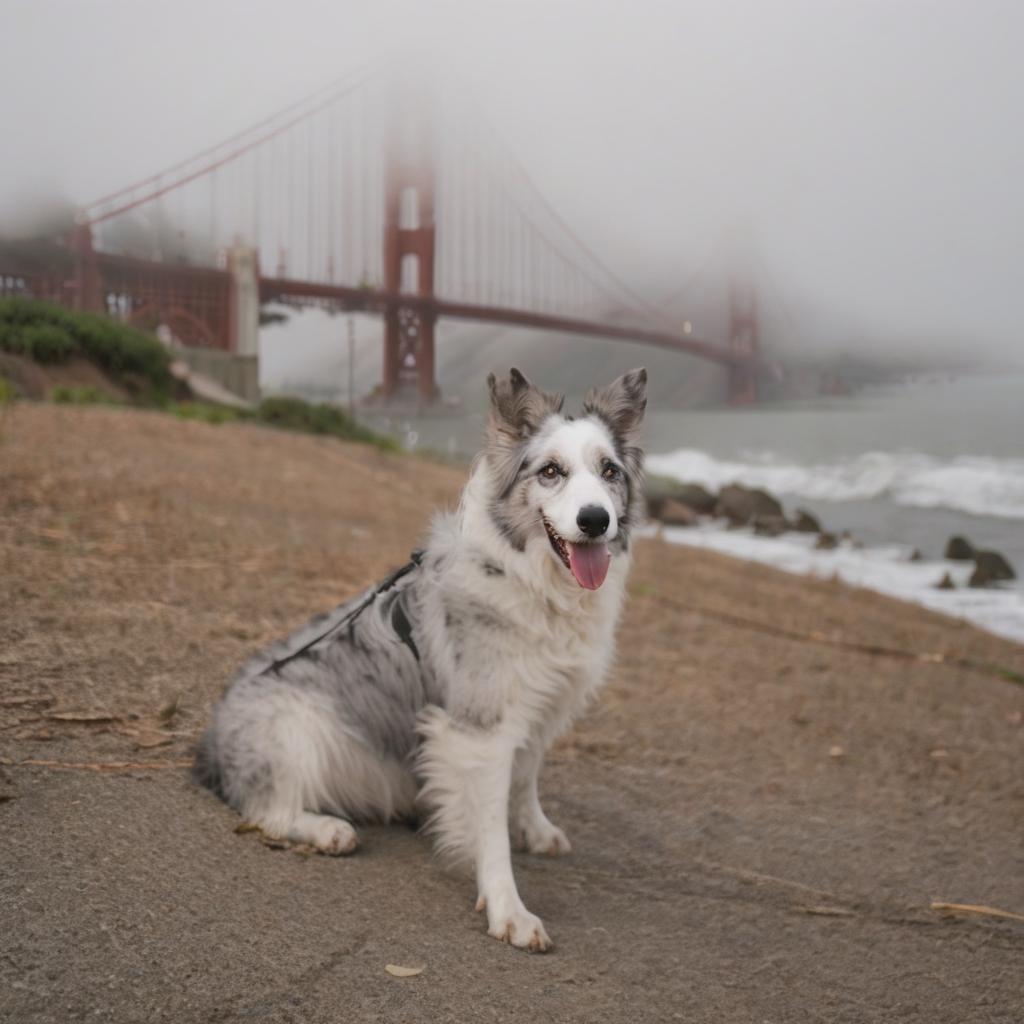}
        \caption*{\scriptsize ``a  \textcolor{red}{k} dog sitting on the waterfront, Golden Gate Bridge emerging from thick morning fog in the background, soft diffused light filtering through the mist''.}
    \end{subfigure}
    \begin{subfigure}[t]{0.4\textwidth}
        \centering
        \includegraphics[width=\textwidth]{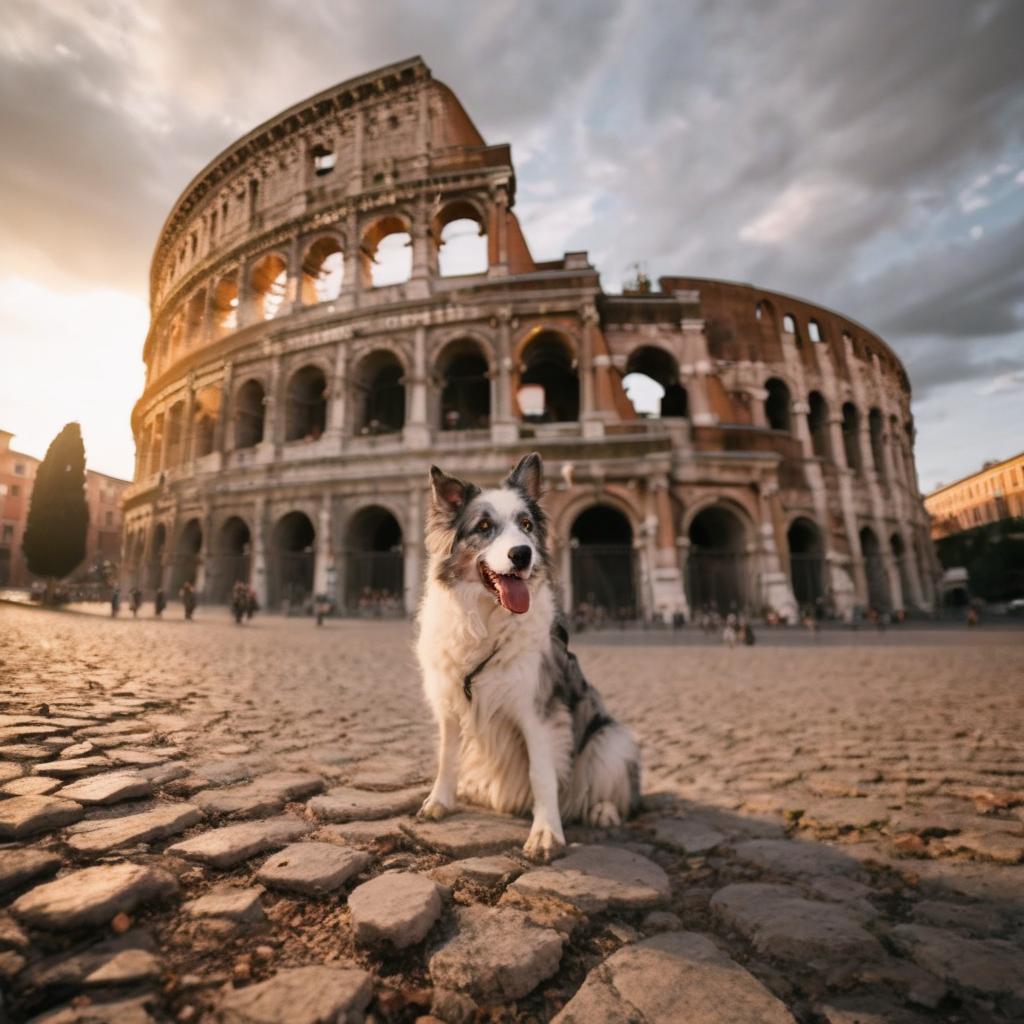}
        \caption*{\scriptsize ``a  \textcolor{red}{k} dog standing in front of the Colosseum at golden hour, warm amber light on ancient stone, dramatic clouds above, cinematic wide angle''.}
    \end{subfigure}

    \caption{\method generated images of ``dog8" across complex scenarios.}
    \label{fig:complex_dog8}
\end{figure}

\begin{figure}[h!]
    \centering

    \begin{subfigure}[t]{0.4\textwidth}
        \centering
        \includegraphics[width=\textwidth]{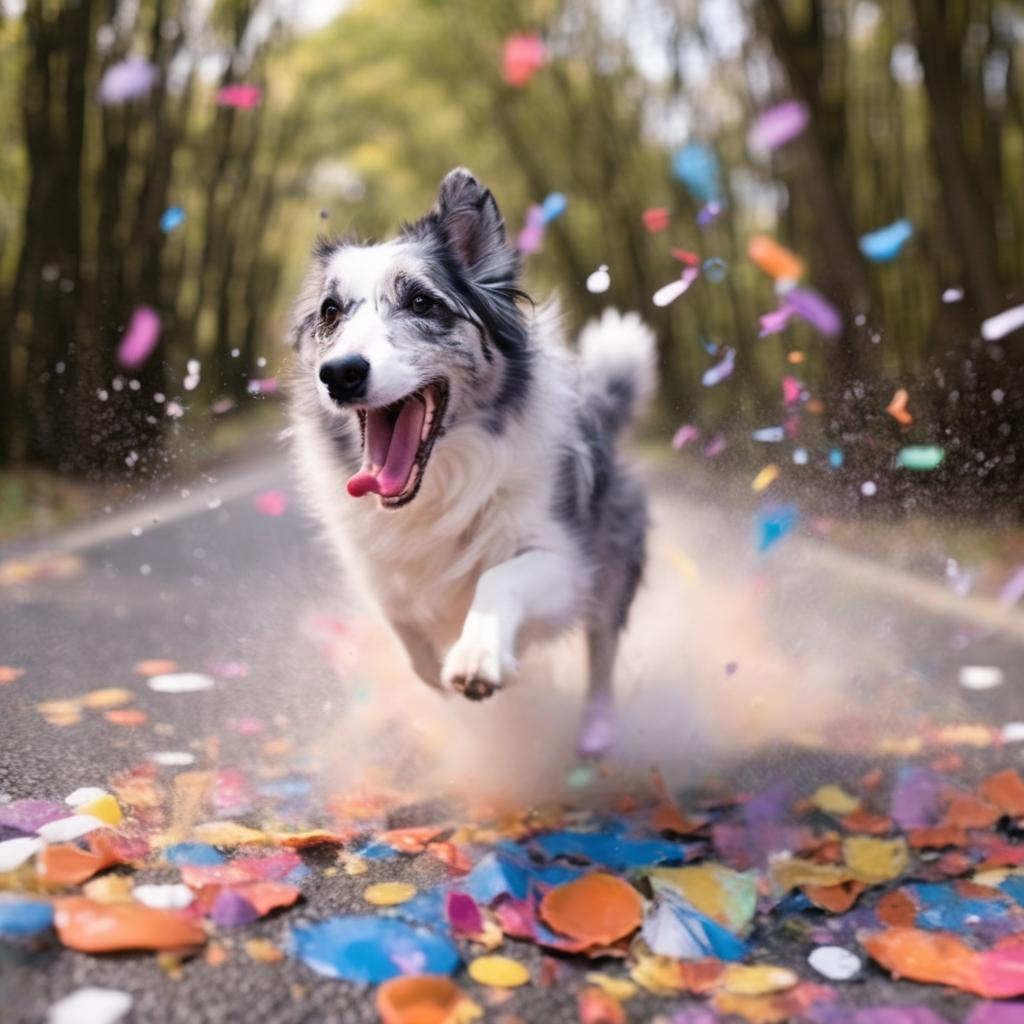}
        \caption*{\scriptsize``a  \textcolor{red}{k} dog racing through an exploding tunnel of colorful paint splashes, motion blur, frozen droplets mid-air, low angle high-speed shot.''}
    \end{subfigure}
    \begin{subfigure}[t]{0.4\textwidth}
        \centering
        \includegraphics[width=\textwidth]{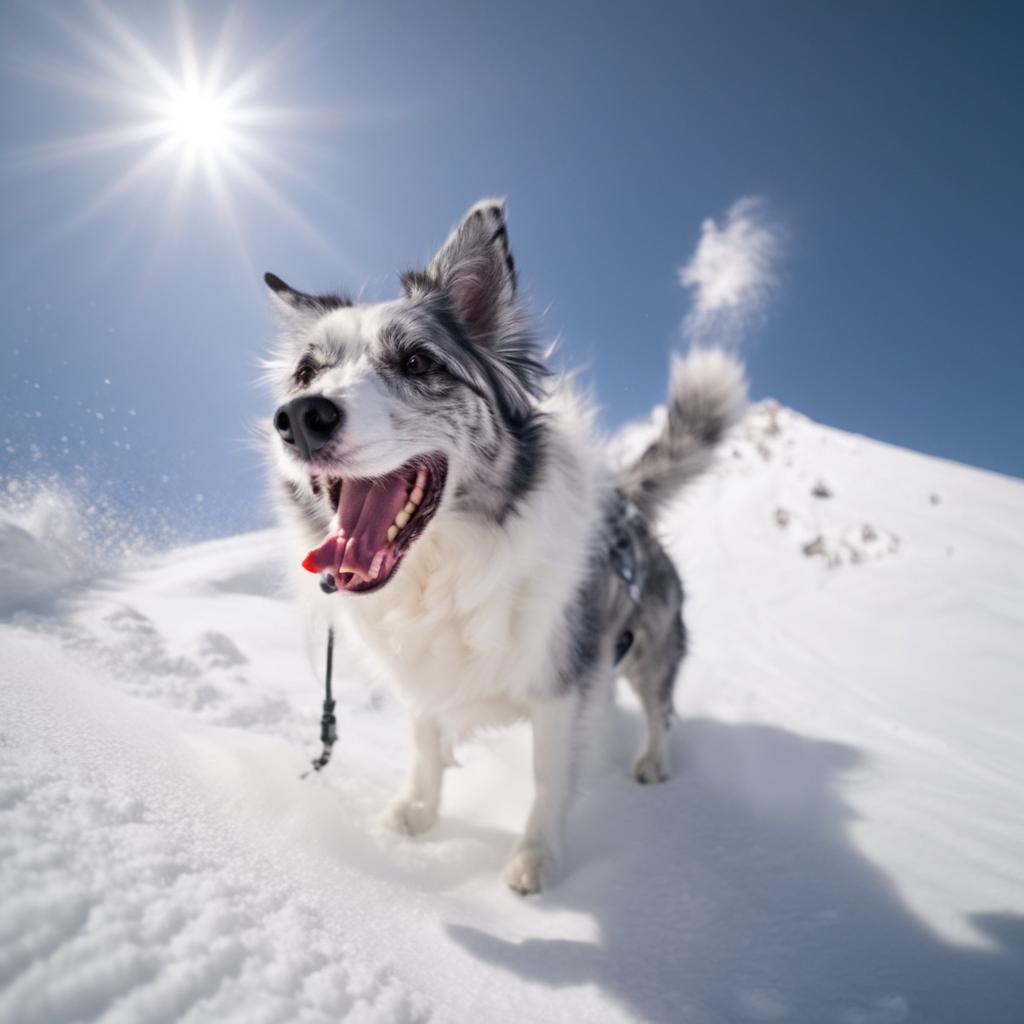}
        \caption*{\scriptsize ``a  \textcolor{red}{k} dog launching off a snowy mountain peak on a snowboard, massive powder explosion, crisp blue sky, low angle action shot.''}
    \end{subfigure}

    \vspace{1em}

    \begin{subfigure}[t]{0.4\textwidth}
        \centering
        \includegraphics[width=\textwidth]{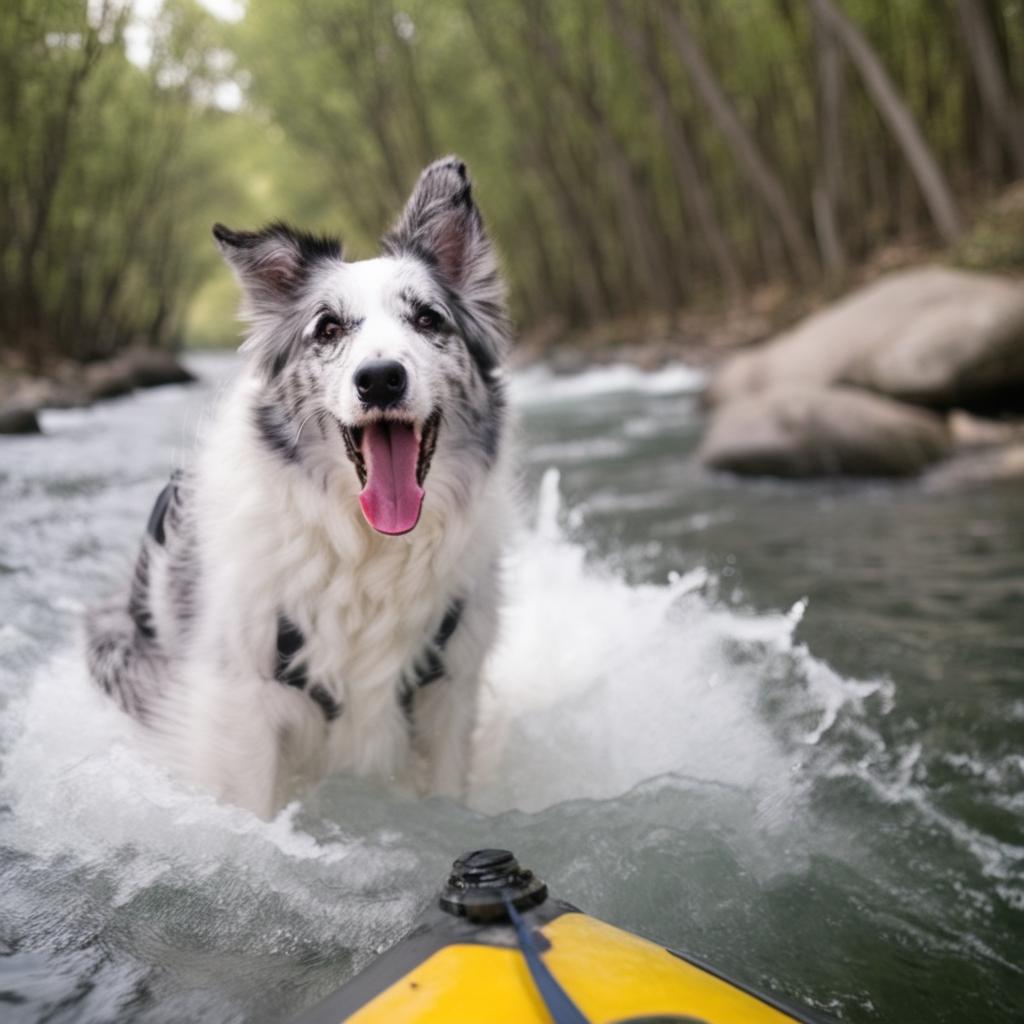}
        \caption*{\scriptsize ``a  \textcolor{red}{k} dog kayaking through a raging white-water rapid, water exploding around the boat, soaked fur, intense focus, action shot frozen mid-crash''.}
    \end{subfigure}
    \begin{subfigure}[t]{0.4\textwidth}
        \centering
        \includegraphics[width=\textwidth]{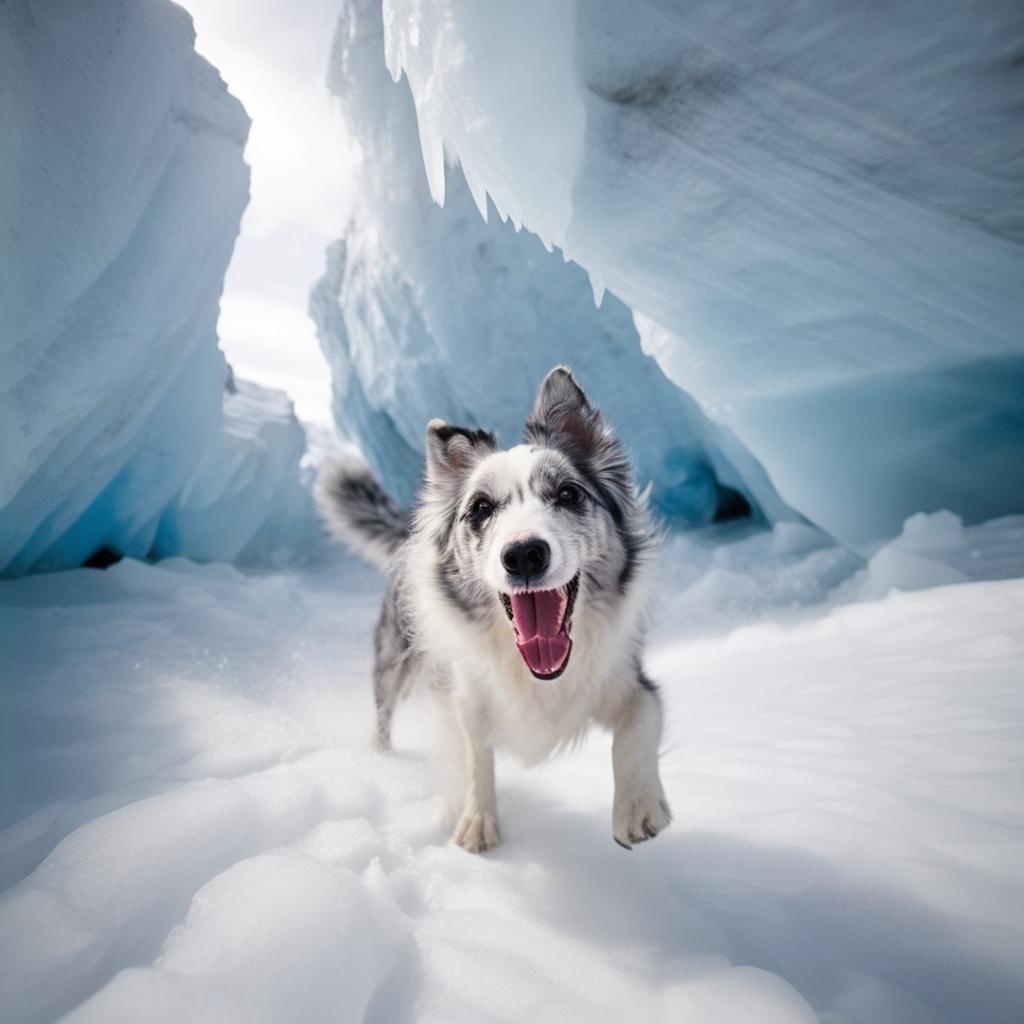}
        \caption*{\scriptsize ``a  \textcolor{red}{k} dog leaping between two glaciers over an icy blue crevasse, paws mid-air, frozen mist, dramatic arctic light, ultra-wide low angle''.}
    \end{subfigure}

    \vspace{1em}

    \begin{subfigure}[t]{0.4\textwidth}
        \centering
        \includegraphics[width=\textwidth]{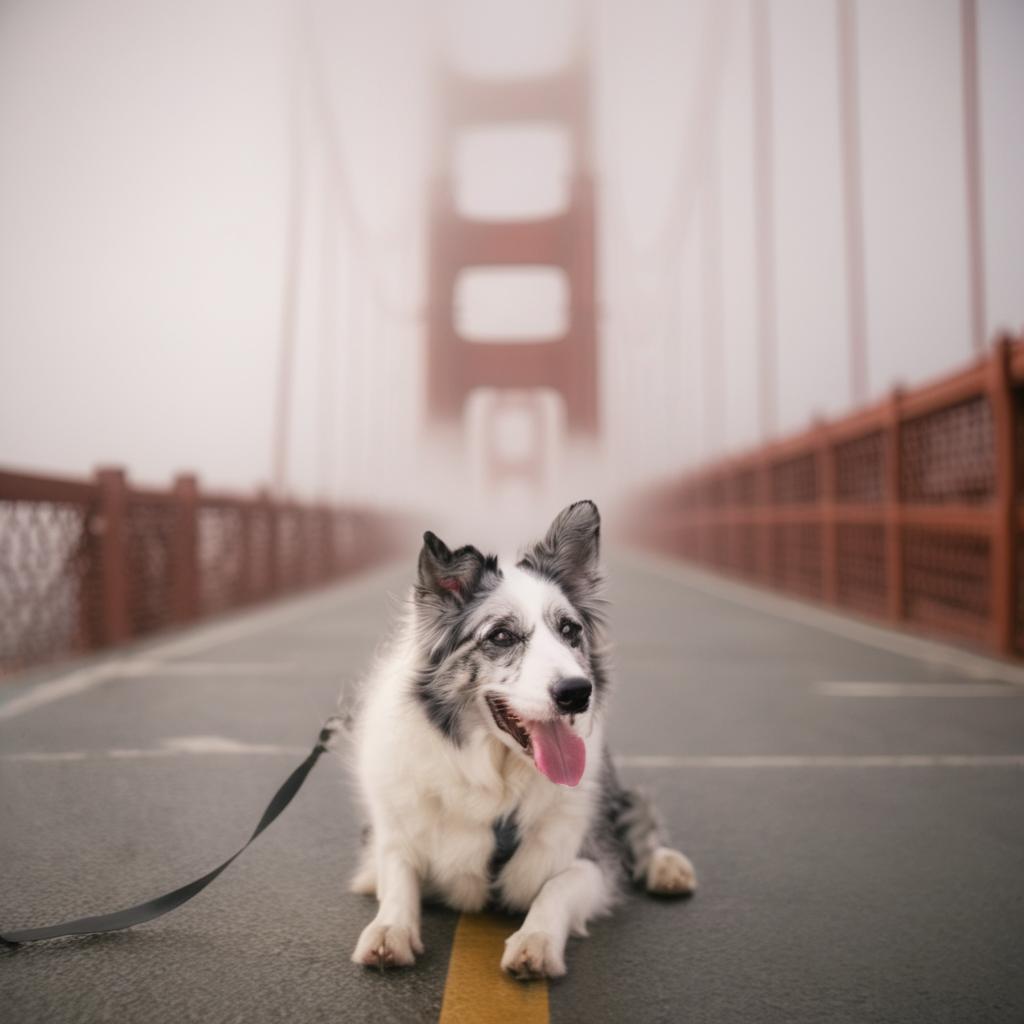}
        \caption*{\scriptsize ``a  \textcolor{red}{k} dog sitting on the waterfront, Golden Gate Bridge emerging from thick morning fog in the background, soft diffused light filtering through the mist''.}
    \end{subfigure}
    \begin{subfigure}[t]{0.4\textwidth}
        \centering
        \includegraphics[width=\textwidth]{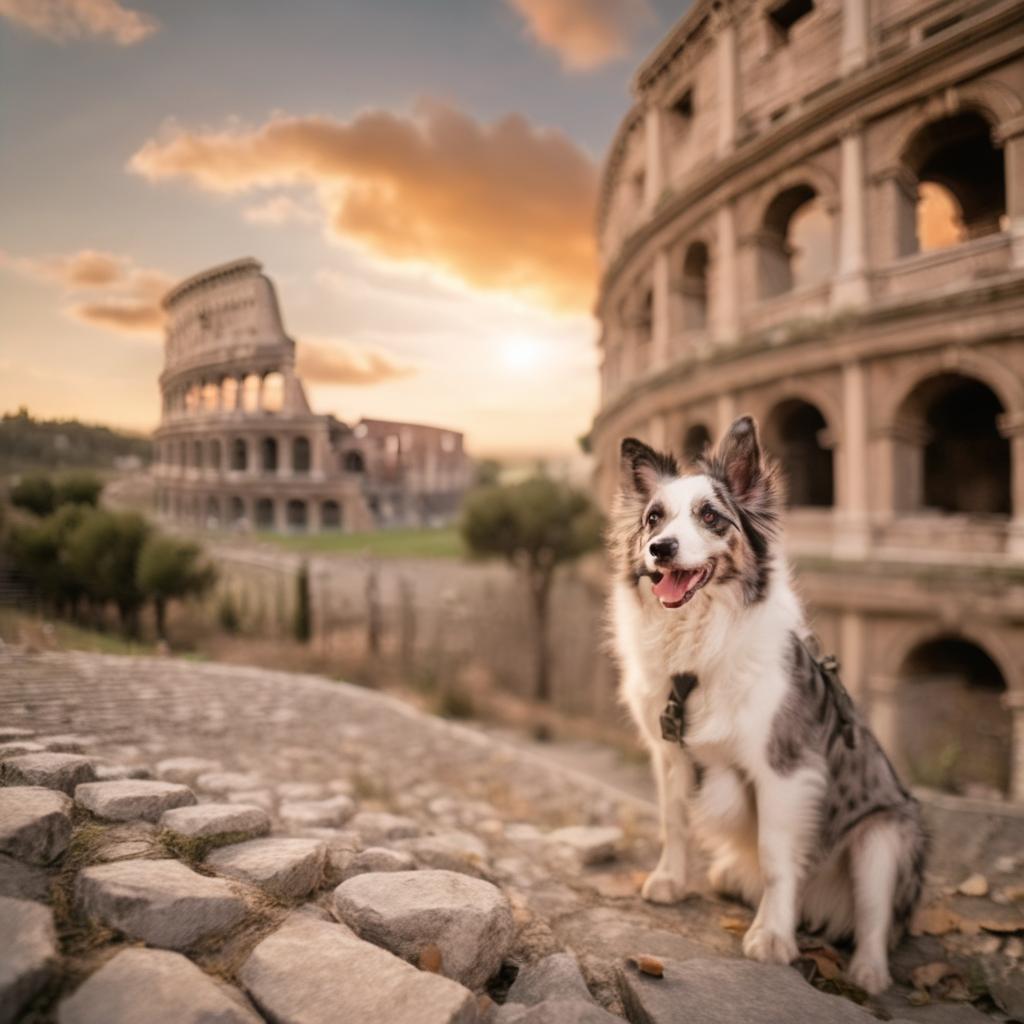}
        \caption*{\scriptsize ``a  \textcolor{red}{k} dog standing in front of the Colosseum at golden hour, warm amber light on ancient stone, dramatic clouds above, cinematic wide angle''.}
    \end{subfigure}

    \caption{LoRA (rank 512) generated images of ``dog8" across complex scenarios do not produce satisfactory results.}
    \label{fig:complex_dog8_512}
\end{figure}

\begin{figure}[h!]
    \centering

    \begin{subfigure}[t]{0.4\textwidth}
        \centering
        \includegraphics[width=\textwidth]{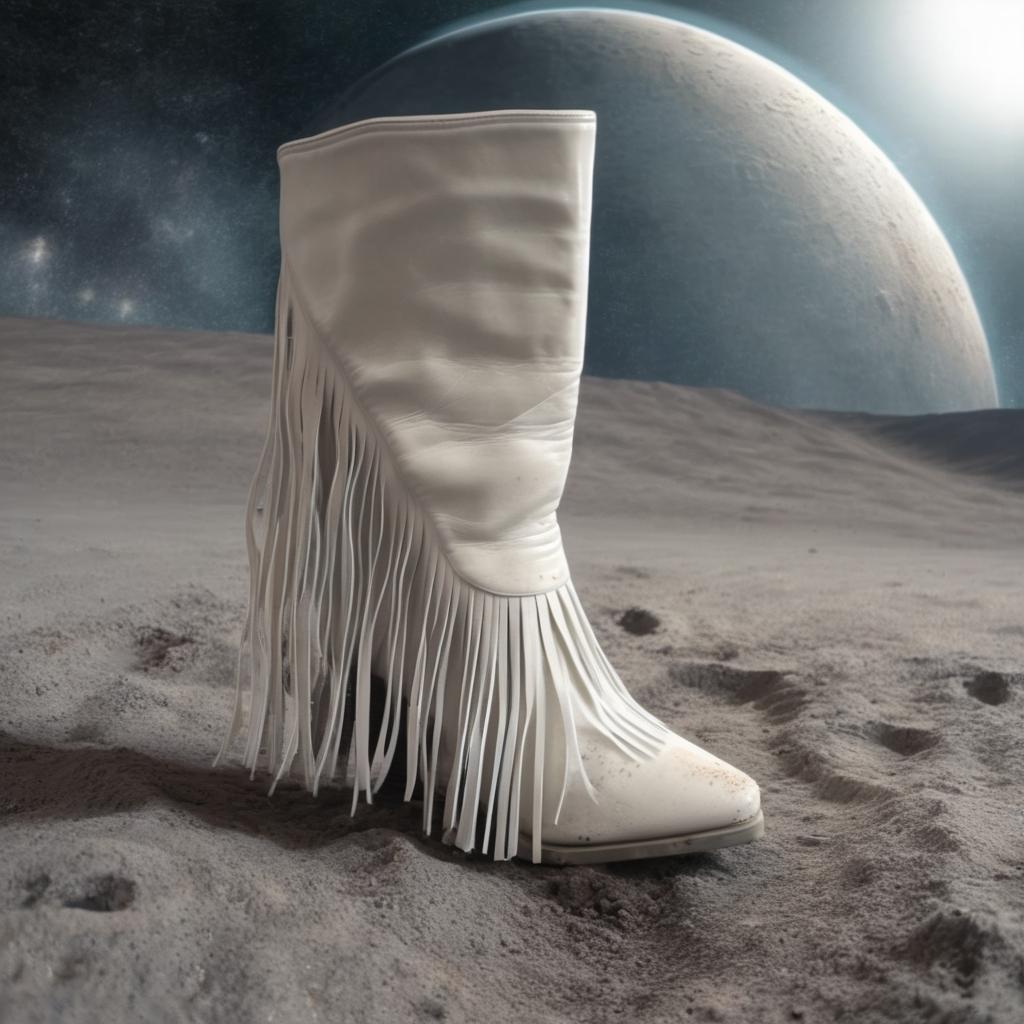}
        \caption*{\scriptsize ``a \textcolor{red}{k} boot standing on the moon surface, Earth rising on the horizon, ultra-realistic cinematic lighting''.}
    \end{subfigure}
    \begin{subfigure}[t]{0.4\textwidth}
        \centering
        \includegraphics[width=\textwidth]{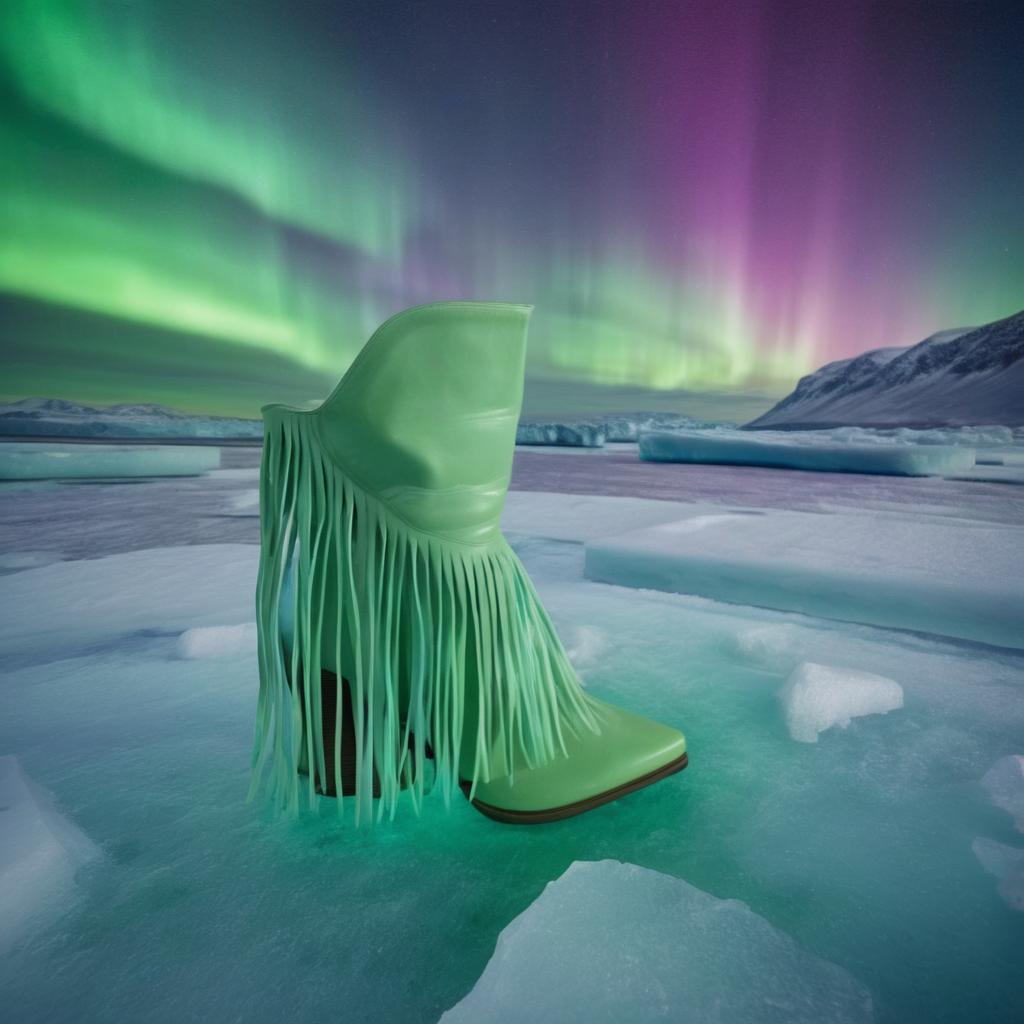}
        \caption*{\scriptsize ``a \textcolor{red}{k} boot on a giant block of ice in an arctic tundra, northern lights glowing green and purple above, cinematic blue tones, photorealistic''.}
    \end{subfigure}

    \vspace{1em}

    \begin{subfigure}[t]{0.4\textwidth}
        \centering
        \includegraphics[width=\textwidth]{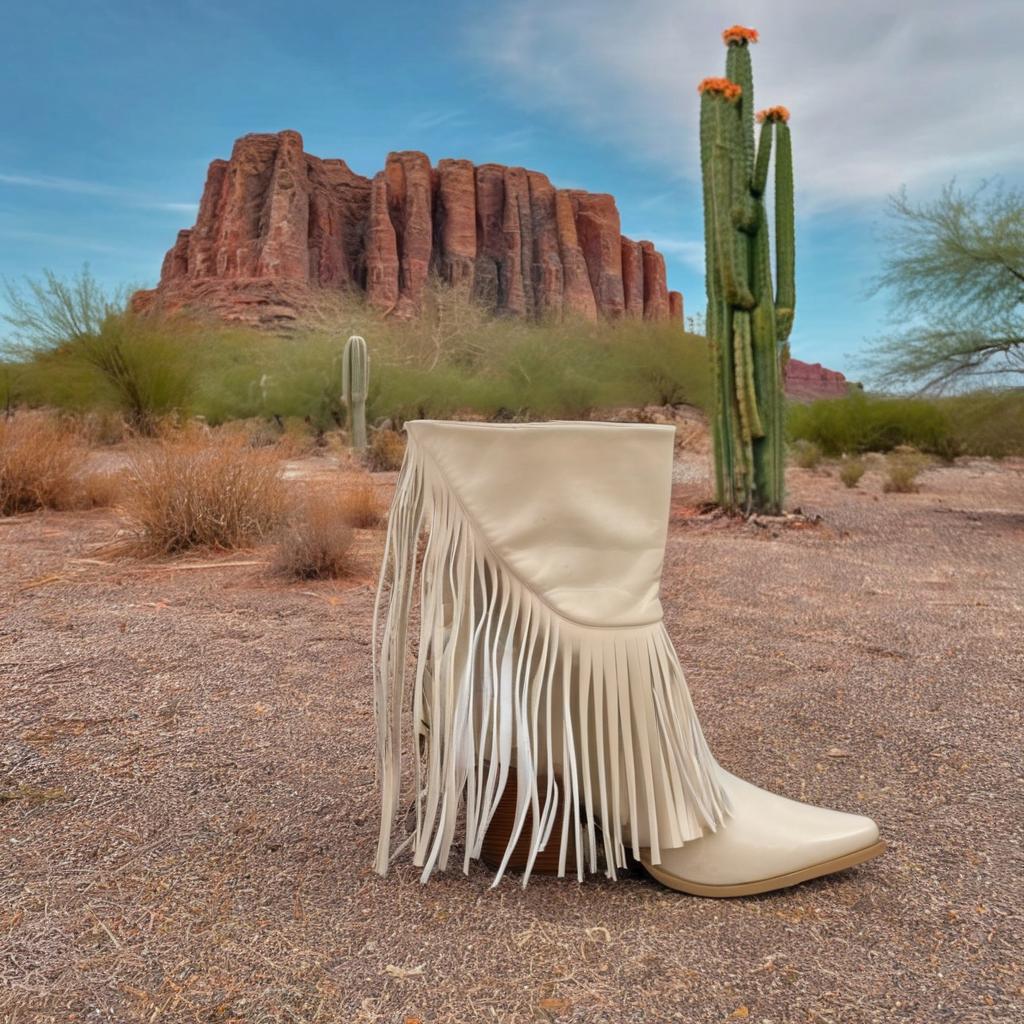}
        \caption*{\scriptsize ``a \textcolor{red}{k} boot in the Sonoran desert, cactus and red rocks behind, blue sky, warm natural light''.}
    \end{subfigure}
    \begin{subfigure}[t]{0.4\textwidth}
        \centering
        \includegraphics[width=\textwidth]{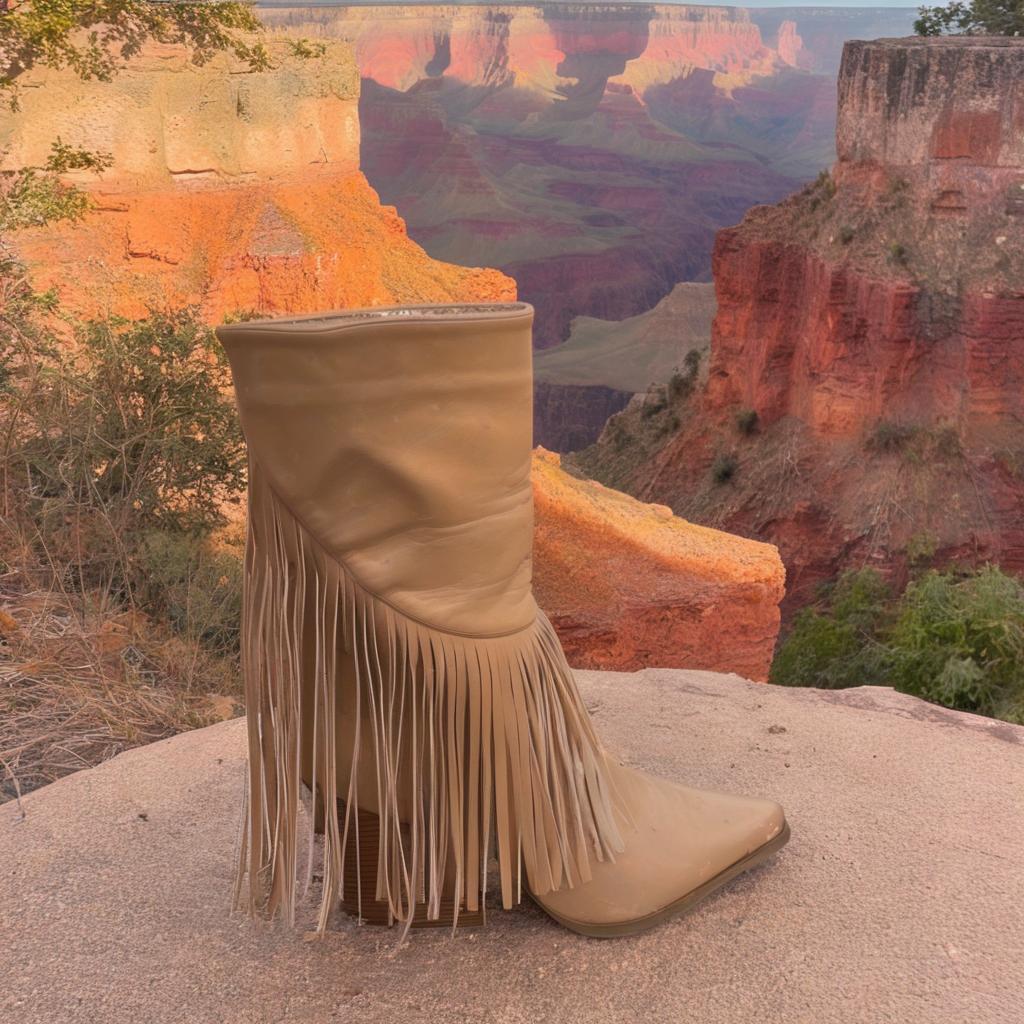}
        \caption*{\scriptsize ``a \textcolor{red}{k} boot on a Grand Canyon overlook, vast red canyon stretching behind, golden hour''.}
    \end{subfigure}

    \vspace{1em}

    \begin{subfigure}[t]{0.4\textwidth}
        \centering
        \includegraphics[width=\textwidth]{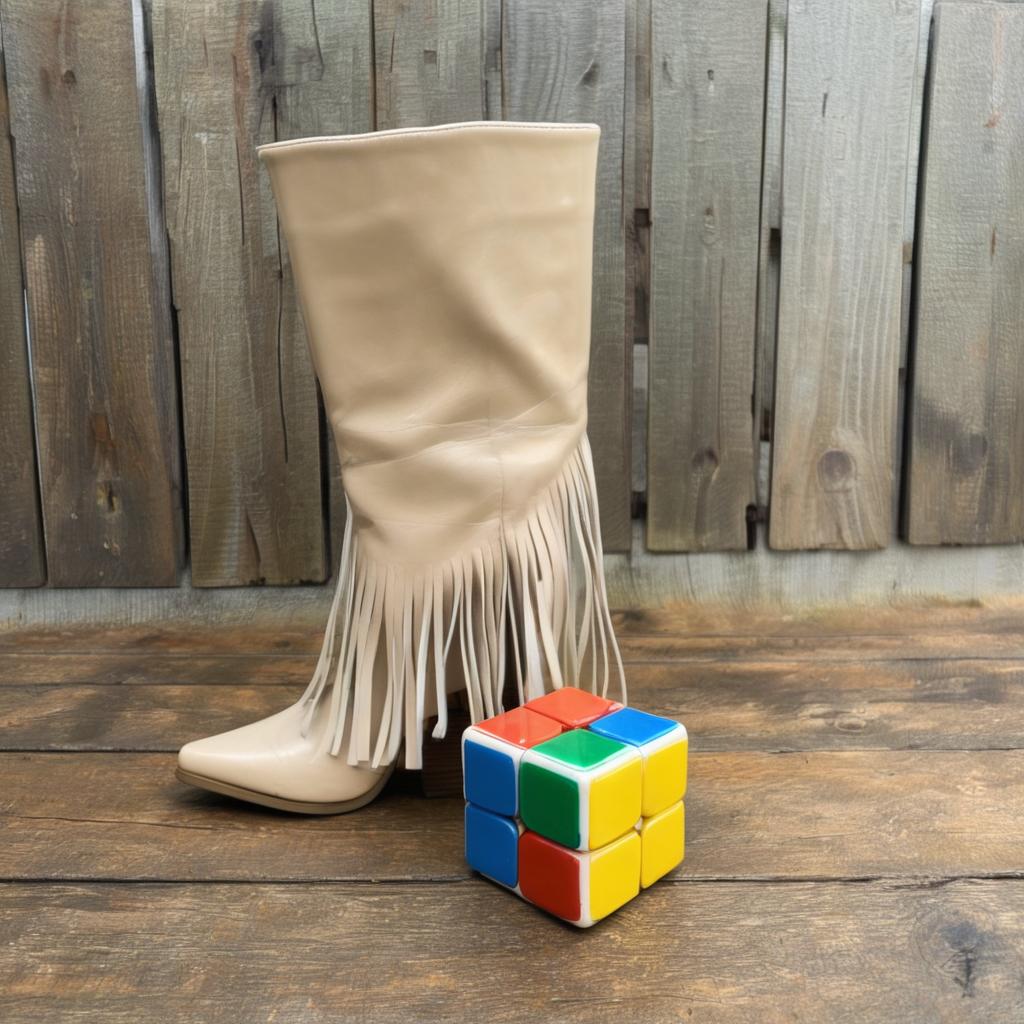}
        \caption*{\scriptsize ``a \textcolor{red}{k} boot next to a rubik's cube''.}
    \end{subfigure}
    \begin{subfigure}[t]{0.4\textwidth}
        \centering
        \includegraphics[width=\textwidth]{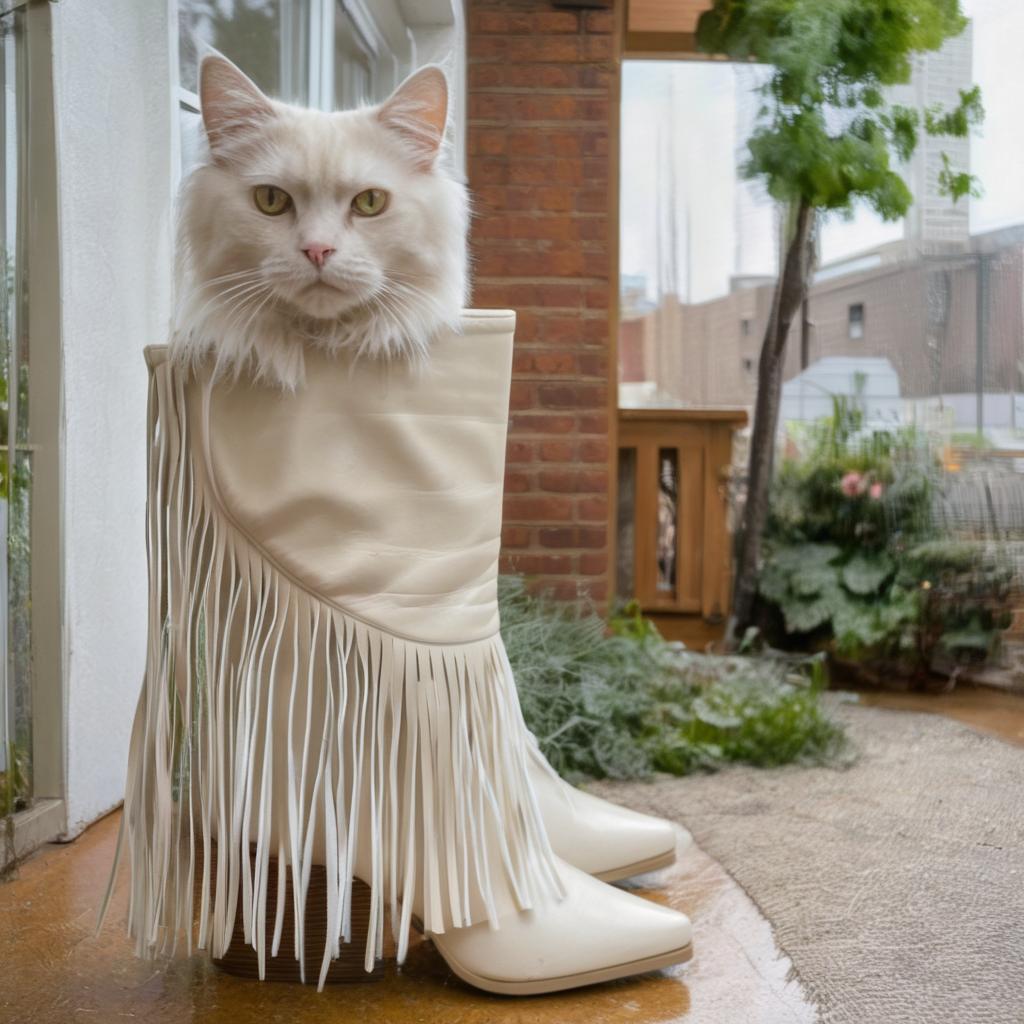}
        \caption*{\scriptsize ``a cat inside a \textcolor{red}{k} boot,  soft natural light, cozy home''.}
    \end{subfigure}

    \caption{\method generated images of ``fancy boot" across complex scenarios.}
    \label{fig:complex_fancy_boot}
\end{figure}

\begin{figure}[h!]
    \centering

    \begin{subfigure}[t]{0.4\textwidth}
        \centering
        \includegraphics[width=\textwidth]{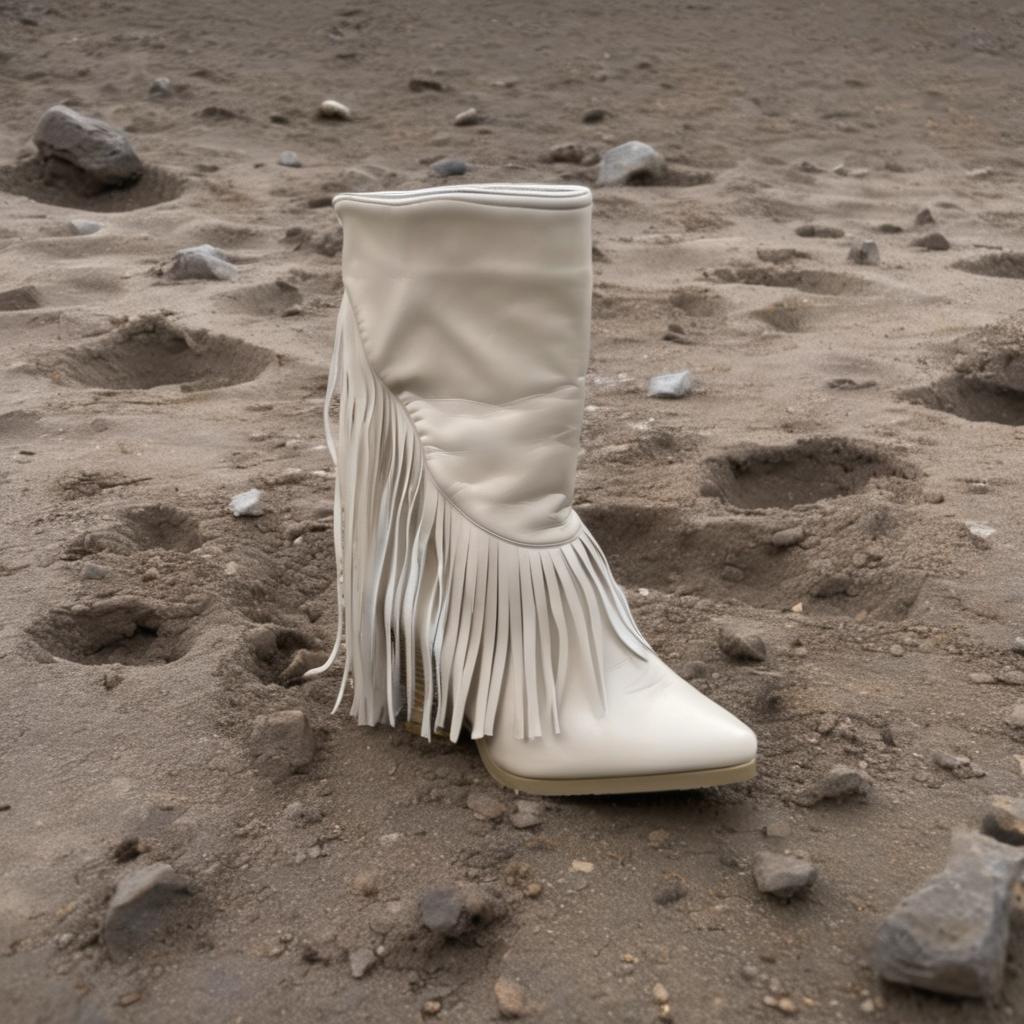}
        \caption*{\scriptsize ``a \textcolor{red}{k} boot standing on the moon surface, Earth rising on the horizon, ultra-realistic cinematic lighting''.}
    \end{subfigure}
    \begin{subfigure}[t]{0.4\textwidth}
        \centering
        \includegraphics[width=\textwidth]{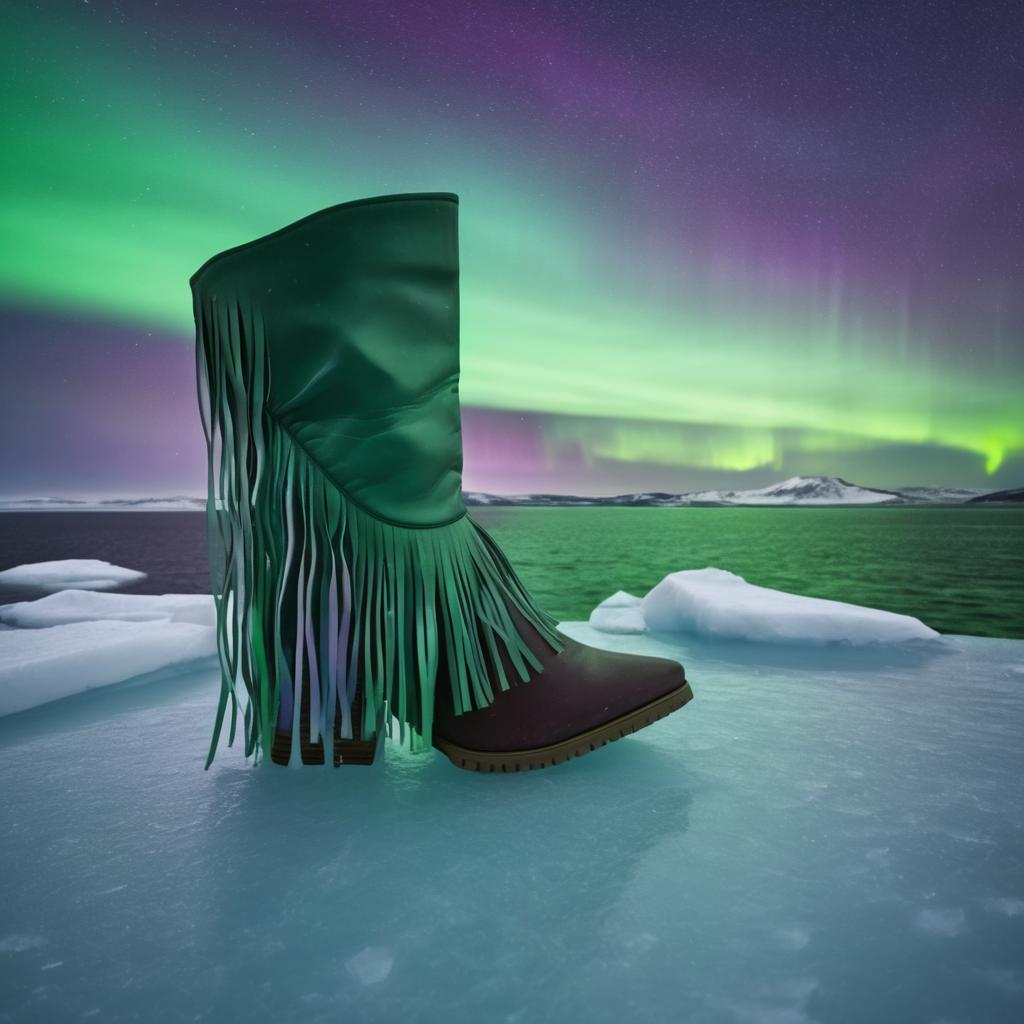}
        \caption*{\scriptsize ``a \textcolor{red}{k} boot on a giant block of ice in an arctic tundra, northern lights glowing green and purple above, cinematic blue tones, photorealistic''.}
    \end{subfigure}

    \vspace{1em}

    \begin{subfigure}[t]{0.4\textwidth}
        \centering
        \includegraphics[width=\textwidth]{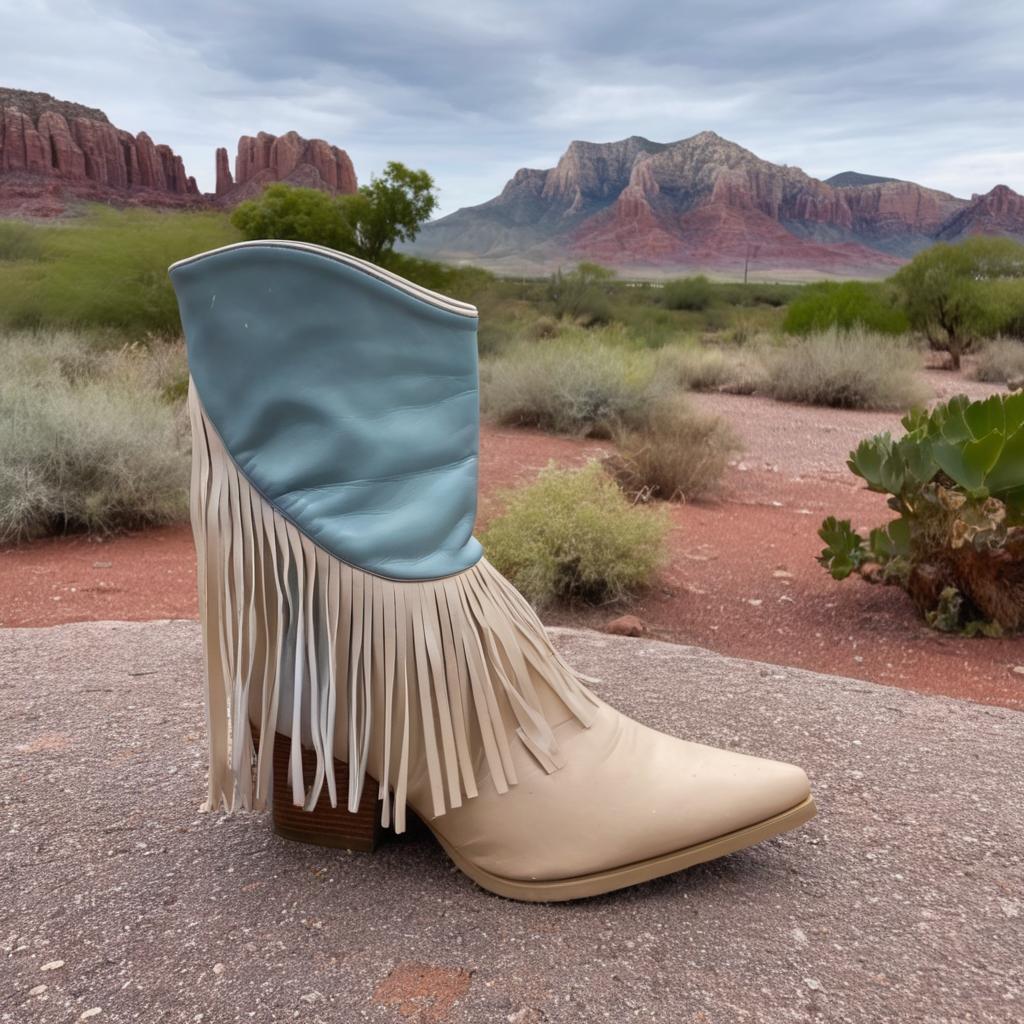}
        \caption*{\scriptsize ``a \textcolor{red}{k} boot in the Sonoran desert, cactus and red rocks behind, blue sky, warm natural light''.}
    \end{subfigure}
    \begin{subfigure}[t]{0.4\textwidth}
        \centering
        \includegraphics[width=\textwidth]{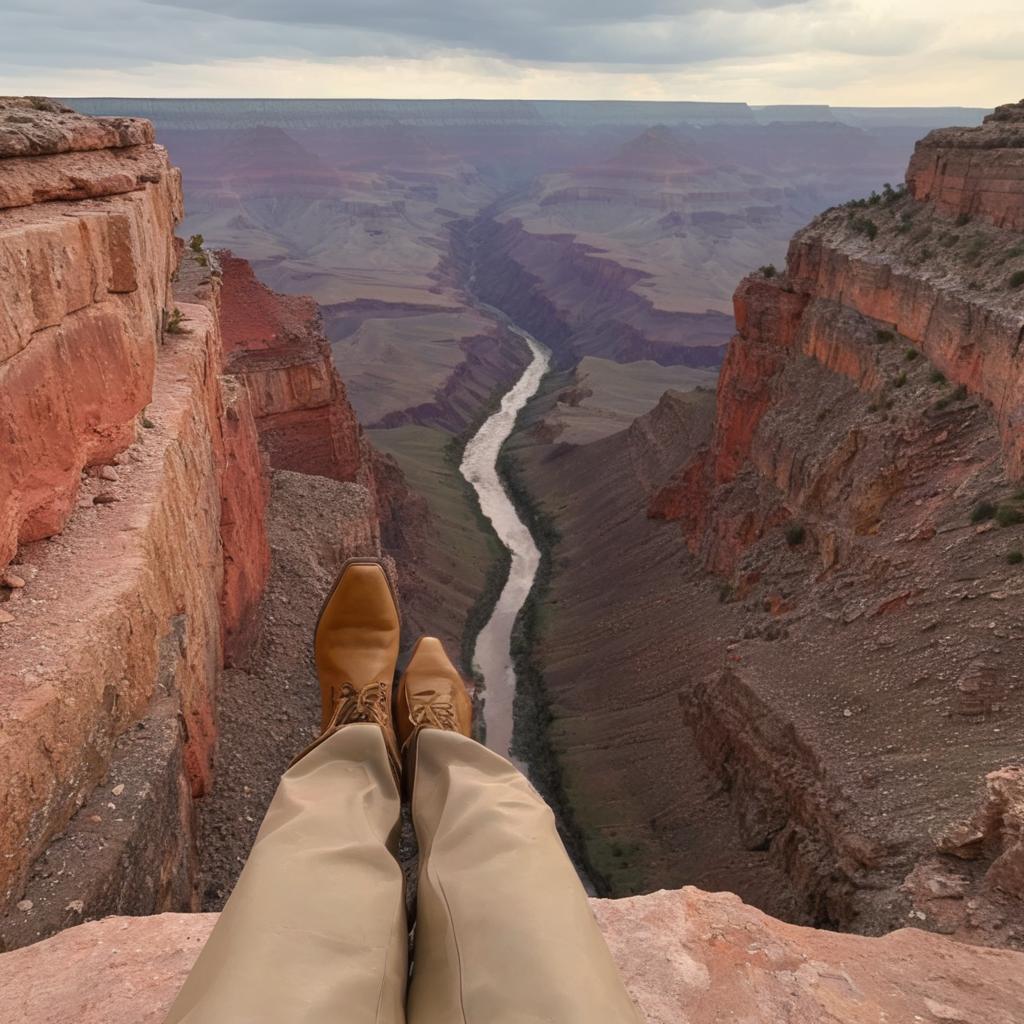}
        \caption*{\scriptsize ``a \textcolor{red}{k} boot on a Grand Canyon overlook, vast red canyon stretching behind, golden hour''.}
    \end{subfigure}

    \vspace{1em}

    \begin{subfigure}[t]{0.4\textwidth}
        \centering
        \includegraphics[width=\textwidth]{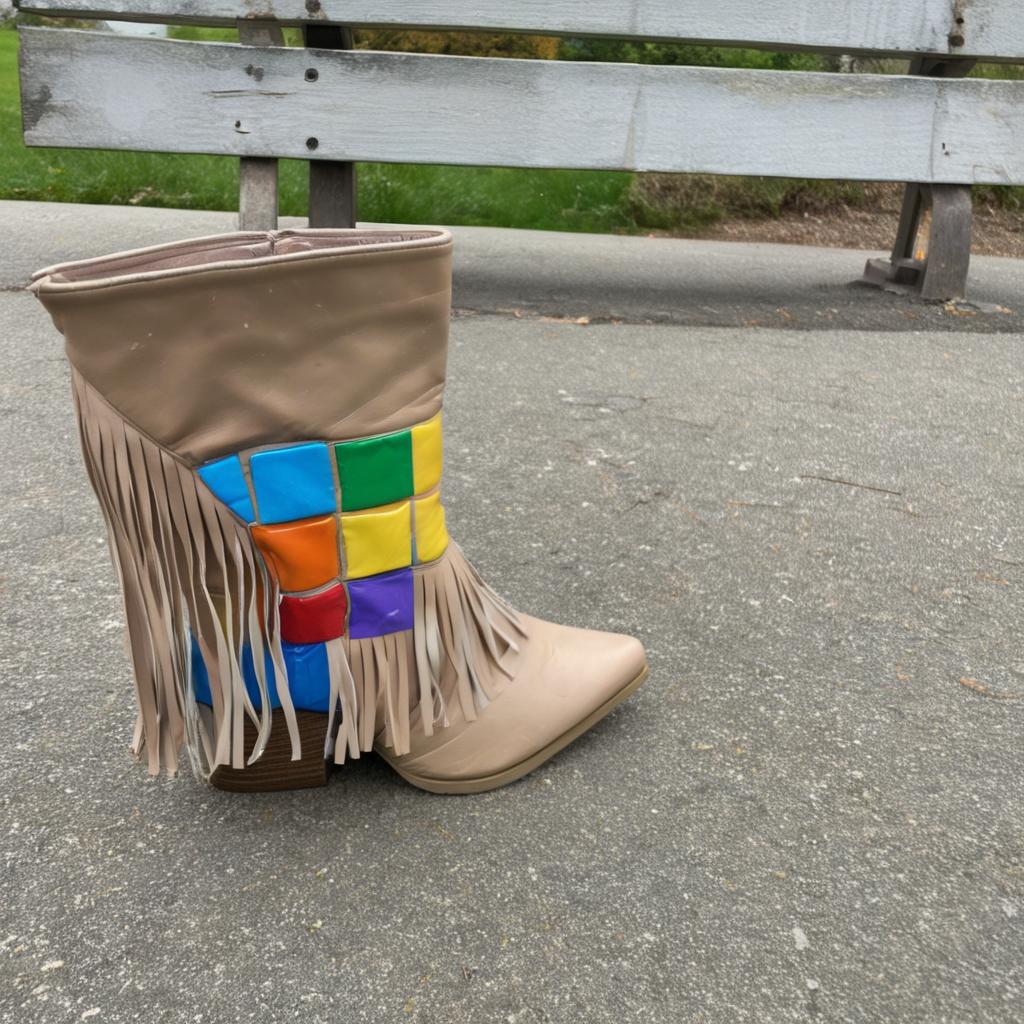}
        \caption*{\scriptsize ``a \textcolor{red}{k} boot next to a rubik's cube''.}
    \end{subfigure}
    \begin{subfigure}[t]{0.4\textwidth}
        \centering
        \includegraphics[width=\textwidth]{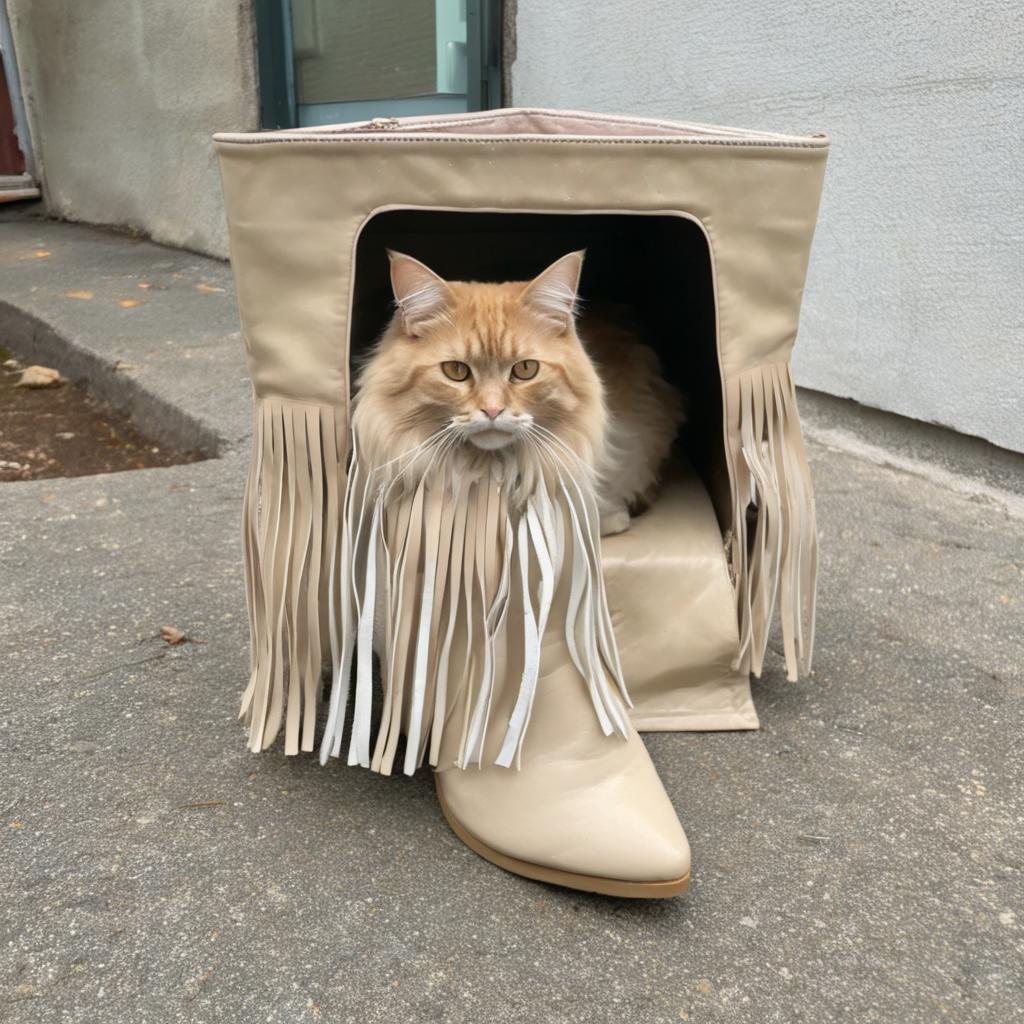}
        \caption*{\scriptsize ``a cat inside a \textcolor{red}{k} boot,  soft natural light, cozy home''.}
    \end{subfigure}

    \caption{LoRA (rank 512) generated images of ``fancy boot" across complex scenarios do not produce satisfactory results.}
    \label{fig:complex_fancy_boot_512}
\end{figure}

\Cref{fig:complex_dog8,fig:complex_fancy_boot} showcase complex prompt generations, illustrating that \method generalizes effectively to broader, more challenging generation scenarios beyond simple subject reconstruction, while LoRA with fixed rank in \Cref{fig:complex_dog8_512,fig:complex_fancy_boot_512} often fails to recontextualize properly.

\section{Limitations}
Our current evaluation of \method focuses on personalized subject learning; extending the approach to style learning remains an interesting direction for future work.

For model merging, a current limitation arises from the fact that \method produces LoRA adapters of different ranks across subjects. To merge two such adapters, the lower-rank LoRA must be expanded to match the rank of the larger one prior to merging.
Alternatively, composition-based approaches such as~\cite{meral2025contrastive} sidestep this issue entirely by combining subjects without requiring explicit adapter merging.

Finally, when generating images with complex prompts, we observe that background colors can occasionally leak into the subject, subtly shifting its appearance. However, this artifact is not unique to \method and manifests across all competing approaches. Despite this, \method consistently produces superior subject fidelity compared to existing methods, even under challenging prompt conditions.

\end{document}